\documentclass[runningheads]{llncs}
\usepackage[T1]{fontenc}
\usepackage{graphicx}
\usepackage{booktabs}
\usepackage{amsmath}
\usepackage{amssymb}
\usepackage{multirow}
\usepackage{times}
\usepackage{epsfig}
\usepackage{graphicx}
\usepackage{amsmath}
\usepackage{tabularx}
\usepackage{amssymb}
\usepackage{hyperref}
\usepackage{multirow}
\usepackage[utf8]{inputenc}
\usepackage[T1]{fontenc}
\usepackage{float}
\usepackage{arydshln} 
\usepackage{caption}
\usepackage{booktabs}
\begin{document}
\title{Granularity-Aware Transfer for Tree Instance Segmentation in Synthetic and Real Forests.}
\titlerunning{MGTD: Granularity-Aware Synthetic-to-Real Tree Segmentation}
%
\author{Pankaj Deoli\inst{1,2} \orcidID{0009-0001-0876-7367} \and
Atef Tej\inst{2} \orcidID{0009-0005-8314-090X} \and
Anmol Ashri\inst{2} \orcidID{0009-0001-9458-1461} \and
Anandatirtha JS\inst{2} \orcidID{0000-0001-8305-903X} \and
Karsten Berns\inst{1,2} \orcidID{0000-0002-9080-1404}}
\authorrunning{Deoli et al.,}
%
\institute{Robotics Research Lab, University of Kaiserslautern-Landau, Kaiserslautern, Germany \\
	\url{https://rrlab.cs.rptu.de/en} \and
University of Kaiserslautern-Landau, Kaiserslautern, Germany}
\maketitle              
\begin{abstract}
	We address the challenge of synthetic-to-real transfer in forestry perception where real data have only coarse ``Tree'' labels while synthetic data provide fine-grained trunk/crown annotations. We introduce \textbf{MGTD}, a mixed-granularity dataset with 53k synthetic and 3.6k real images, and a \textbf{four-stage protocol} isolating domain shift and granularity mismatch. Our core contribution is \textbf{granularity-aware distillation}, which transfers structural priors from fine-grained synthetic teachers to a coarse-label student via logit-space merging and mask unification. Experiments show consistent mask AP gains, especially for small/distant trees, establishing a testbed for Sim$\rightarrow$Real transfer under label granularity constraints.
	
	\noindent\textbf{Keywords:} synthetic-to-real transfer, mixed granularity, tree segmentation, knowledge distillation.
\end{abstract}

\section{Introduction}
\label{sec:intro}

Accurate tree perception, encompassing detection, segmentation, and structural analysis is critical for forestry robotics applications such as autonomous navigation and environmental monitoring. However, perception in forest environments remains challenging due to strong illumination variations, seasonal changes, occlusion from foliage, and background clutter. A fundamental bottleneck is the scarcity of large, well-annotated real-world forest datasets. Collecting fine-grained annotations (e.g., distinguishing trunks from whole trees) is logistically complex and expensive. However, synthetic data generation offers a promising alternative, enabling large-scale, fully annotated imagery under controlled conditions. Yet models trained purely on synthetic data often fail to generalize to real forest scenes due to appearance differences and texture mismatches, a challenge known as the \emph{domain gap}.

This gap is further amplified when label granularity differs between domains. In our setting, synthetic data provide separate labels for \emph{tree trunks} and \emph{whole trees} (fine granularity), while real-world images are typically annotated with a single \emph{Tree} class (coarse granularity). This label mismatch compounds the appearance shift, leading to severe performance degradation.

To systematically study this mixed-granularity setting, we introduce the \textbf{Mixed-Granularity Tree Dataset (MGTD)}, combining 53k synthetic forest images with fine-grained trunk and whole-tree annotations, and 3.6k real images annotated only with a single \emph{Tree} label under diverse illumination conditions. We further propose a \textbf{four-stage evaluation protocol} that isolates key challenges: (i) synthetic-only training, (ii) zero-shot synthetic-to-real transfer, (iii) real-only supervised training, and (iv) granularity-aware distillation from fine-grained synthetic teachers to a real-data student.

Our core contribution is adapting \textbf{knowledge distillation for granularity mismatch}, transferring structural priors from synthetic teachers (trained separately on trunk and whole-tree labels) to a student trained only on coarse real labels, without requiring any relabeling. Experiments demonstrate consistent improvements over real-only baselines, particularly for small and distant trees. 

\textbf{Our contributions are:}
\begin{itemize}
	\item \textbf{MGTD}, a large-scale mixed-granularity synthetic-to-real dataset for studying tree perception under domain and label mismatch.
	\item A \textbf{four-stage evaluation protocol} isolating domain shift and granularity effects.
	\item A \textbf{granularity-aware distillation baseline} that transfers fine-grained structural cues from synthetic-to-real data, showing consistent improvements across architectures.
\end{itemize}

Together, these contributions establish a controlled testbed for synthetic-to-real transfer under realistic label constraints in forestry and off-road robotics.

\section{Related Works}
\label{related_works}

We position our work at the intersection of three research areas: (1) forestry perception datasets, (2) synthetic-to-real domain adaptation, and (3) learning with mixed-granularity supervision. To our knowledge, no prior work addresses the \textit{combined} challenge of domain shift \textit{and} label granularity mismatch in forestry robotics.

\subsection{Forestry Perception Datasets}
Existing tree perception datasets fall into three categories. \textbf{Synthetic datasets} like SynthTree43k~\cite{grondin2022training} and SPREAD~\cite{feng2025spread} provide large-scale, fine-grained annotations (trunk vs. whole-tree) but lack paired real-world counterparts for controlled domain analysis. \textbf{Real-world ground-level datasets} such as ForTrunkDet~\cite{10944111} and FinnWoodlands~\cite{lagos2023finnwoodlands} offer authentic forest imagery but are small-scale and typically provide only coarse or single-category labels. \textbf{Aerial datasets} such as NEON crowns~\cite{weinstein2020neon} and FOR-Instance~\cite{puliti2023instance} operate at canopy level, diverging from ground-robot perception needs. Notably, none provide a \textit{paired} synthetic-real benchmark with \textit{differing} label granularity, the gap MGTD dataset addresses.

\subsection{Synthetic-to-Real Domain Adaptation}
Sim$\rightarrow$Real transfer has been extensively studied for urban scenes, with established methods for semantic segmentation (e.g., AdaptSegNet~\cite{Tsai_adaptseg_2018}, DACS~\cite{tranheden2021dacs}) and object detection (e.g., Domain-adaptive Faster R-CNN~\cite{ren2016faster}, Unbiased Mean Teacher~\cite{deng2021unbiased}). These typically employ adversarial alignment, self-training, or consistency regularization to bridge appearance gaps. In off-road environments, datasets like RUGD~\cite{wigness2019rugd} and RELLIS-3D~\cite{jiang2021rellis} reveal that urban-trained models degrade significantly due to texture complexity, illumination variance, and class imbalance. However, a critical but often implicit assumption in these works is \textit{identical class definitions} between source and target domains, an assumption violated in practical forestry settings where synthetic data have fine-grained labels while real annotations are coarse.

\subsection{Learning with Mixed-Granularity Supervision}
When label granularity differs within or across datasets, common strategies include collapsing fine categories into super-classes~\cite{finegrainedobjectclassficiation}, multi-task heads for joint coarse/fine prediction~\cite{LI2023102215}, or label-agnostic evaluation metrics. More recently, open-vocabulary detectors like ViLD~\cite{gu2021open} and Detic~\cite{zhou2022detecting} decouple localization from fixed taxonomies via vision-language alignment. In hierarchical learning, knowledge distillation has been used to transfer fine-grained knowledge to coarse predictors~\cite{dubey2018deep}. However, these approaches either assume \textit{single-domain} settings or require language supervision. The specific problem of \textit{cross-domain granularity transfer}, where source domain has fine labels and target domain has coarse labels still remains under-explored, particularly in forestry perception.

\subsection{Positioning of Our Work}
Our work uniquely addresses the \textit{dual challenge} of domain shift (Sim$\rightarrow$Real) \textit{and} label granularity mismatch (fine$\rightarrow$coarse) in forestry perception. Unlike existing Sim$\rightarrow$Real methods, we explicitly handle differing class definitions through granularity-aware distillation. Unlike hierarchical learning methods, we address cross-domain transfer without language supervision. MGTD provides the first paired benchmark for this setting, and our four-stage protocol enables systematic analysis previously impossible.
\section{Dataset}
\label{dataset}

MGTD combines a large-scale synthetic forest dataset with real ground-level 
imagery to study tree perception under domain shift and label granularity 
mismatch. The synthetic portion provides fine-grained annotations for trunks and 
whole trees, while the real portion reflects practical field constraints where 
only a single ``Tree'' label is available. This combination enables controlled 
experiments on how fine-grained synthetic supervision can benefit real-world 
performance.

\subsection{Synthetic Data}
\label{simulated_data}

We generate 53{,}521 images using Unreal Engine~5, covering four illumination 
conditions (dawn, dusk, noon, snow). Each image is annotated with two categories: 
\emph{tree trunk} and \emph{whole tree}. The dataset is split into 70\% 
training, 20\% validation, and 10\% testing, ensuring balanced coverage across 
illumination regimes. For more details, we refer readers to supplementary.

\paragraph{Annotation pipeline.}
To obtain instance-level supervision at scale, we adopt a structured annotation process. Bounding boxes for trunks and whole trees are first produced by YOLOv8-x trained on a manually validated seed set. Instance masks are then obtained through box-guided segmentation using Segment Anything Model (SAM)~\cite{kirillov2023segany}. This two-stage process yields consistent, fine-grained masks aligned with tree geometry. All annotations were manually verified for quality. Oriented bounding boxes are derived from keypoint-based fitting for improved geometric description.

\subsection{Real Data}
\label{real_data}

The real dataset contains approximately 3{,}600 frames captured using a 
ground-mounted multispectral camera (as mentioned in \cite{10597528}) in mixed forest environments. Real images 
are annotated with bounding boxes and instance masks using the same 
box-to-mask procedure as in the synthetic data, but with a \emph{single ``Tree''} 
label. This intentional granularity mismatch i.e. fine-grained labels in simulation 
vs.\ coarse labels in reality is central to MGTD and allows systematic study 
of mixed-label transfer. The same 70/20/10 split protocol is applied. For more details, we refer readers to supplementary.

\begin{table*}[ht]
	\centering
	\renewcommand{\arraystretch}{0.9}
	\setlength{\tabcolsep}{2pt}
	\caption{Examples of RGB images, Bounding Boxes (BB), Oriented Bounding Boxes (OBB), Instance Masks (IM), and Oriented Instance Masks (OIM) for simulated (row 2,3) and real (row 1) forest scenes in the MGTD dataset.}
	\begin{tabular}{ccccc}
		\toprule
		\textbf{RGB} & \textbf{BB} & \textbf{OBB} & \textbf{IM} & \textbf{OIM} \\
		\midrule
		\includegraphics[width=0.13\textwidth]{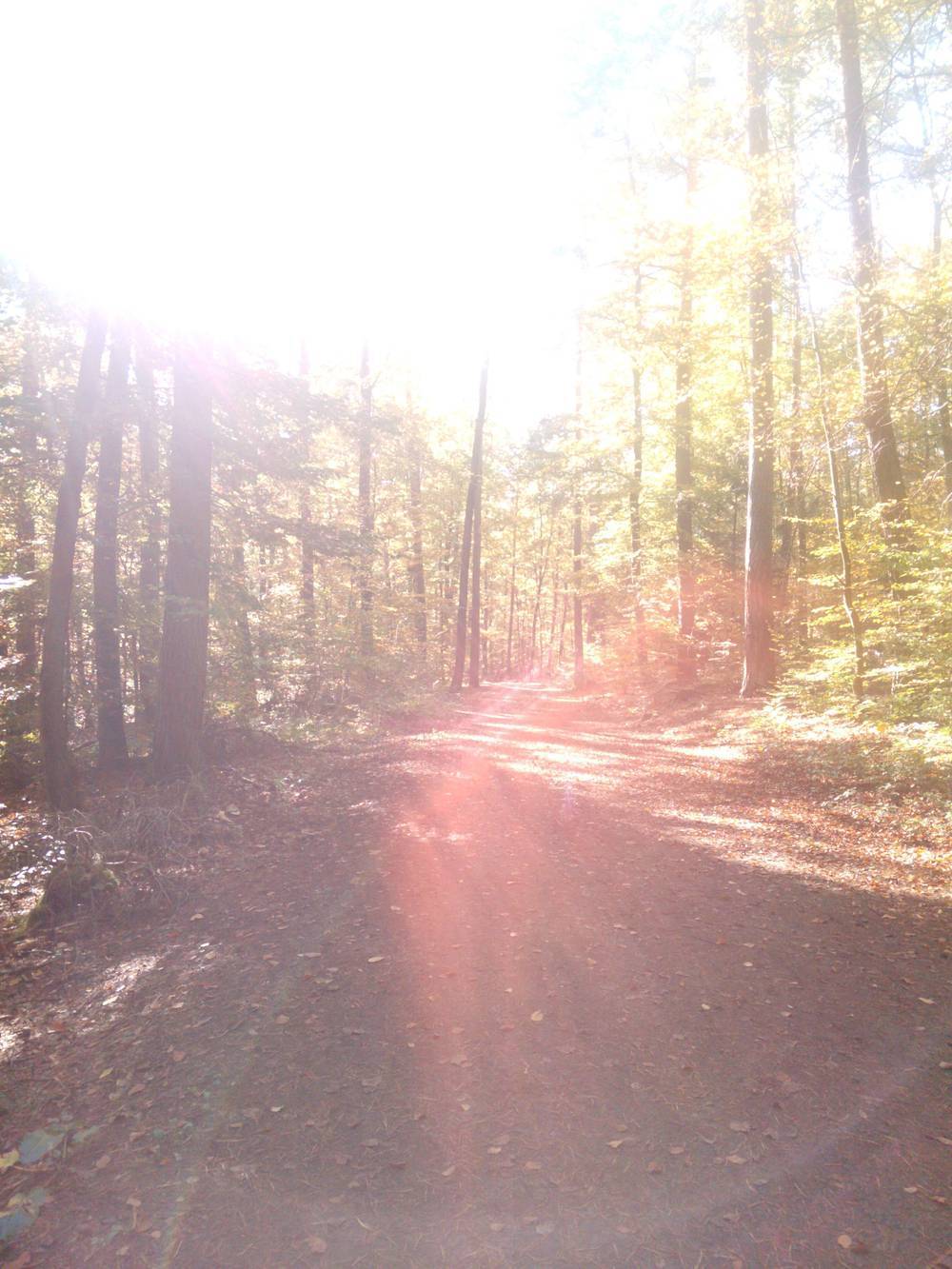} & 
		\includegraphics[width=0.13\textwidth]{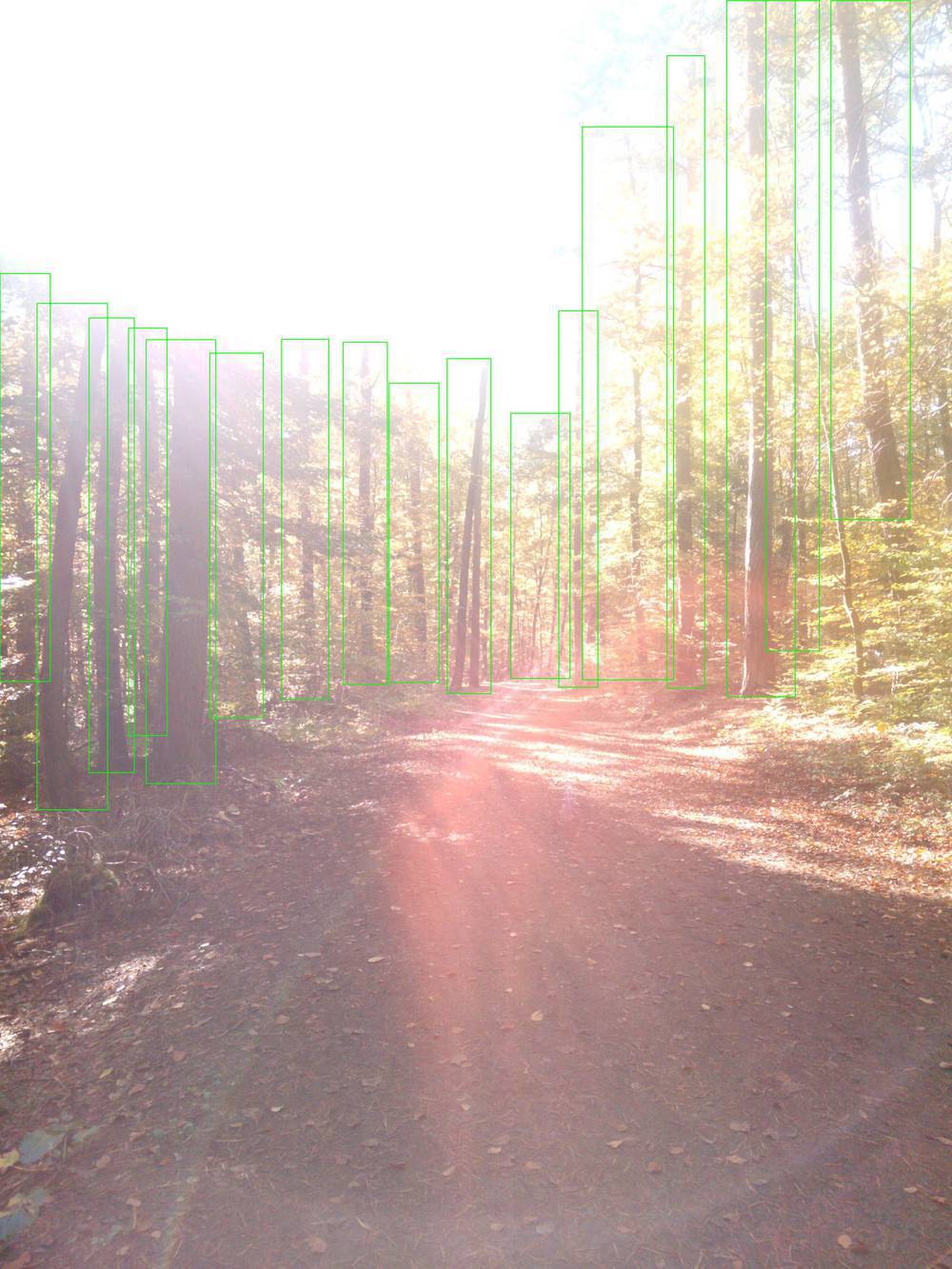} & 
		\includegraphics[width=0.13\textwidth]{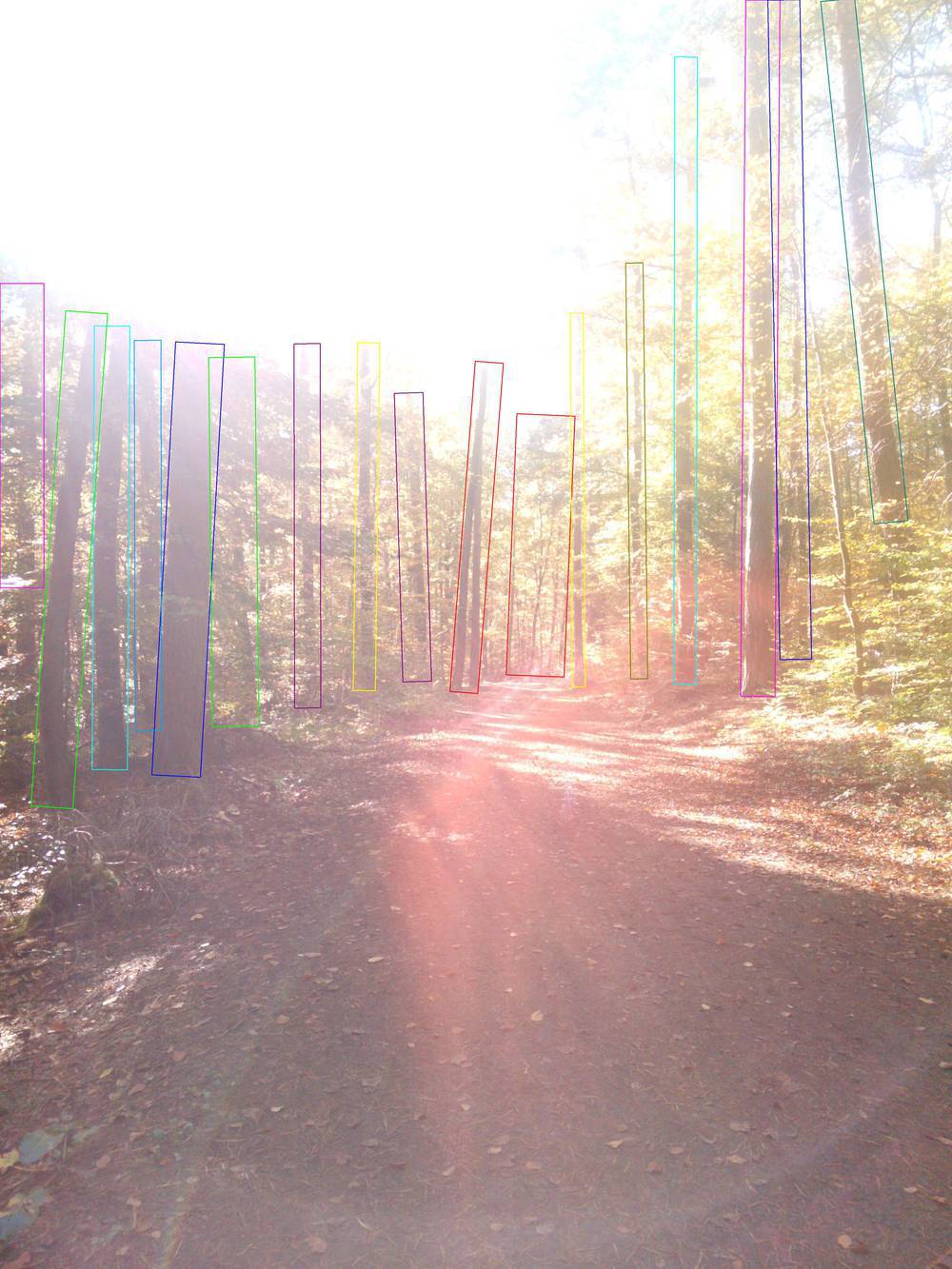}&  
		\includegraphics[width=0.13\textwidth]{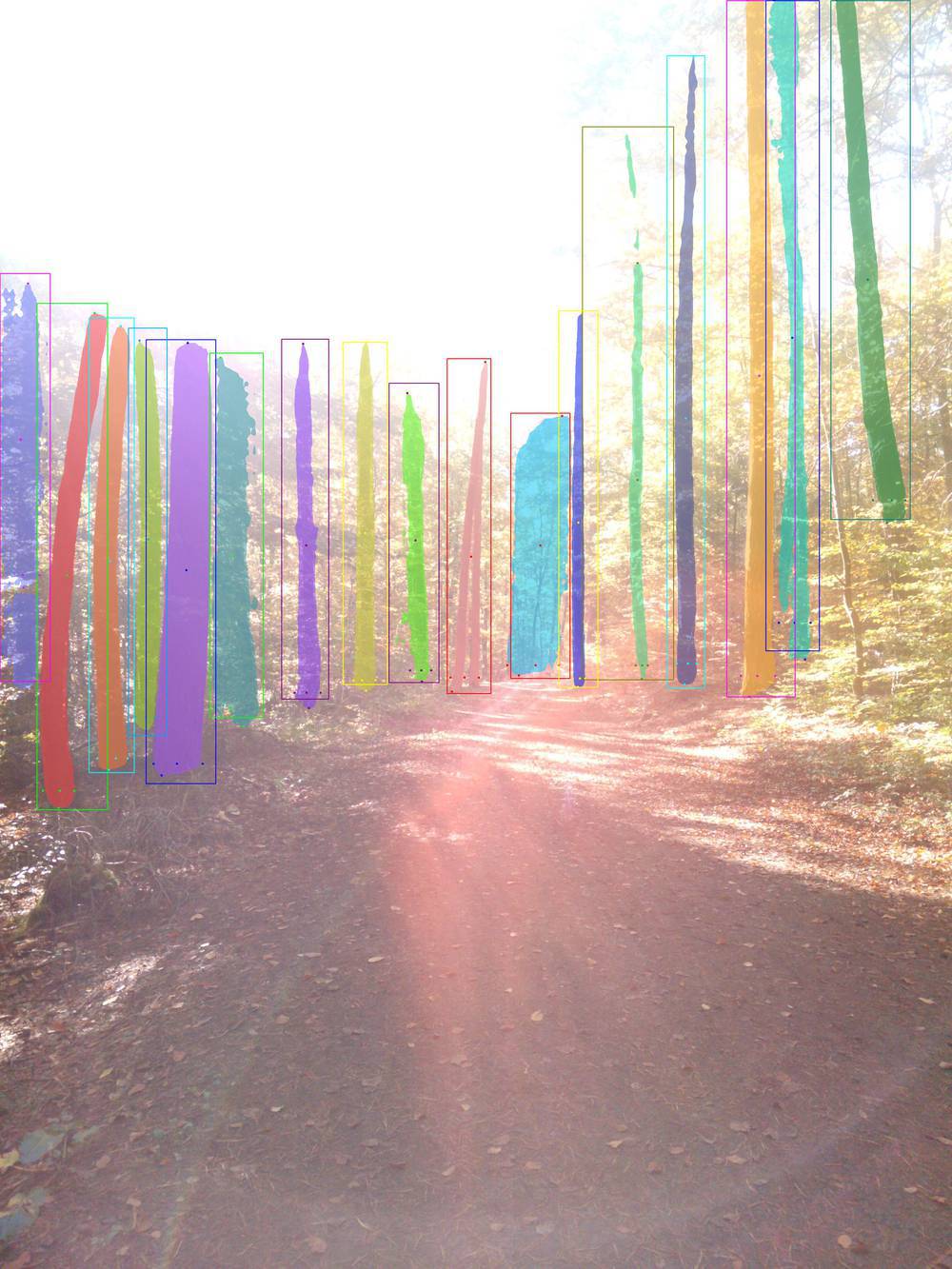}& 
		\includegraphics[width=0.13\textwidth]{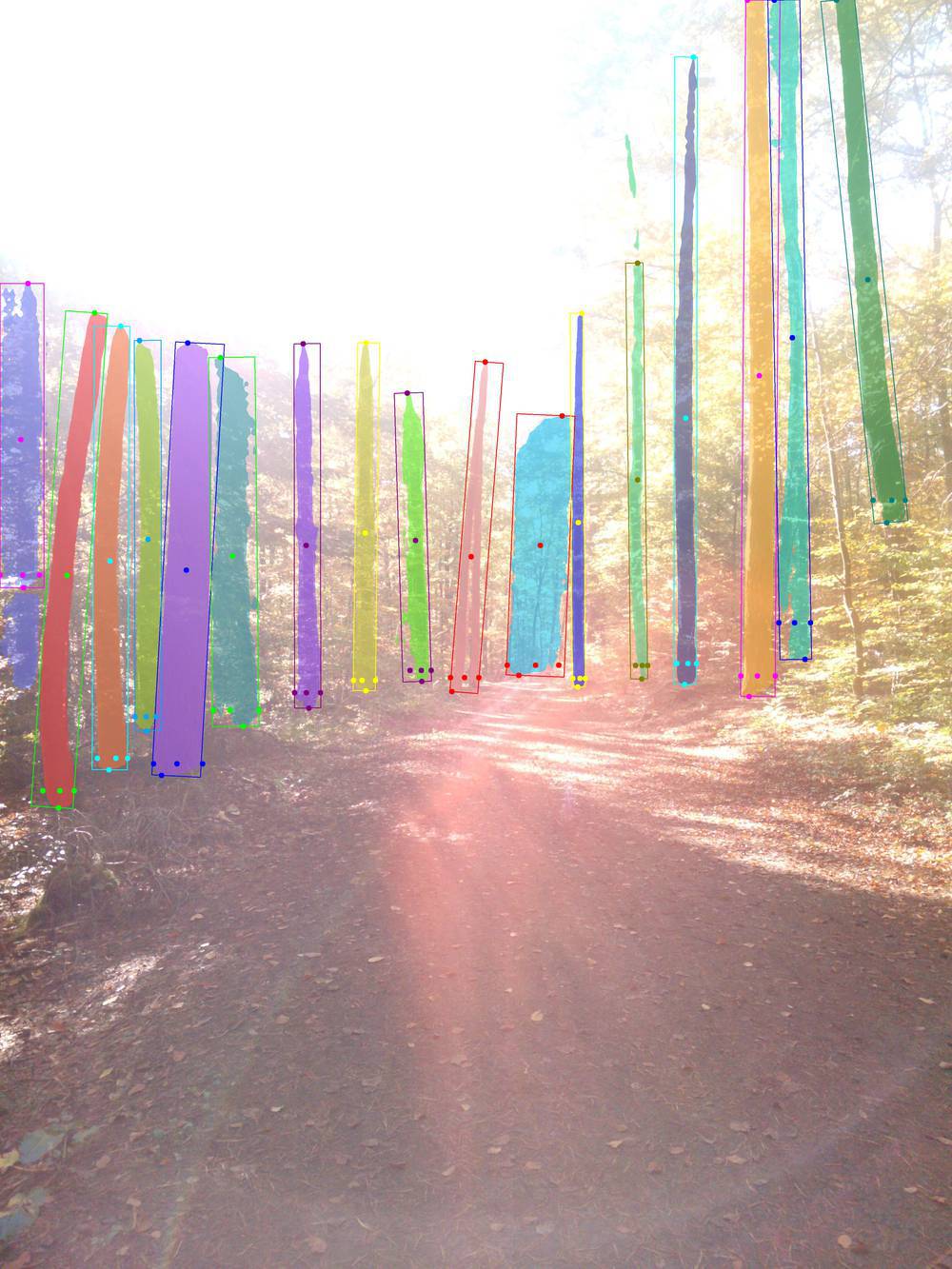} \\
		
		\includegraphics[width=0.13\textwidth]{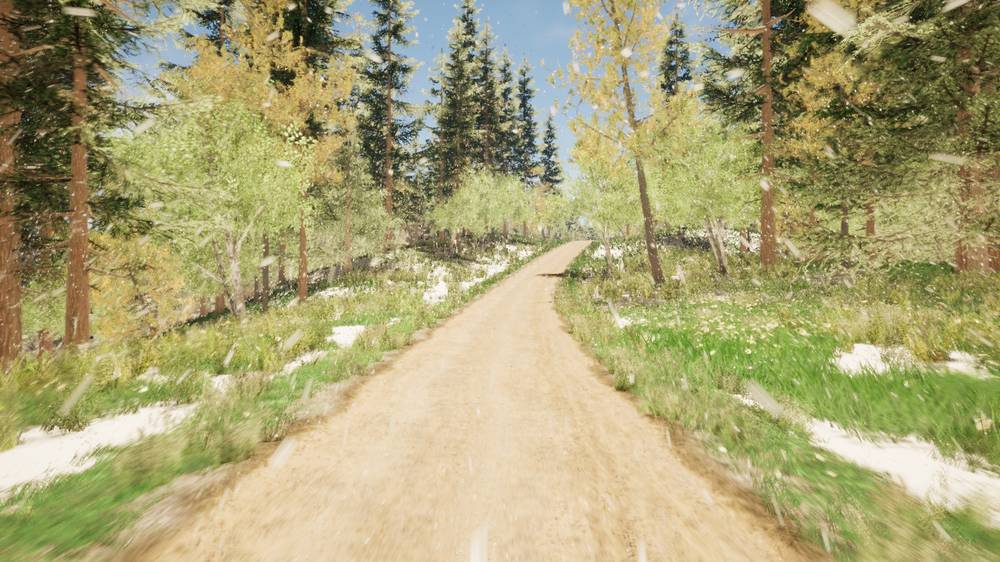} & 
		\includegraphics[width=0.13\textwidth]{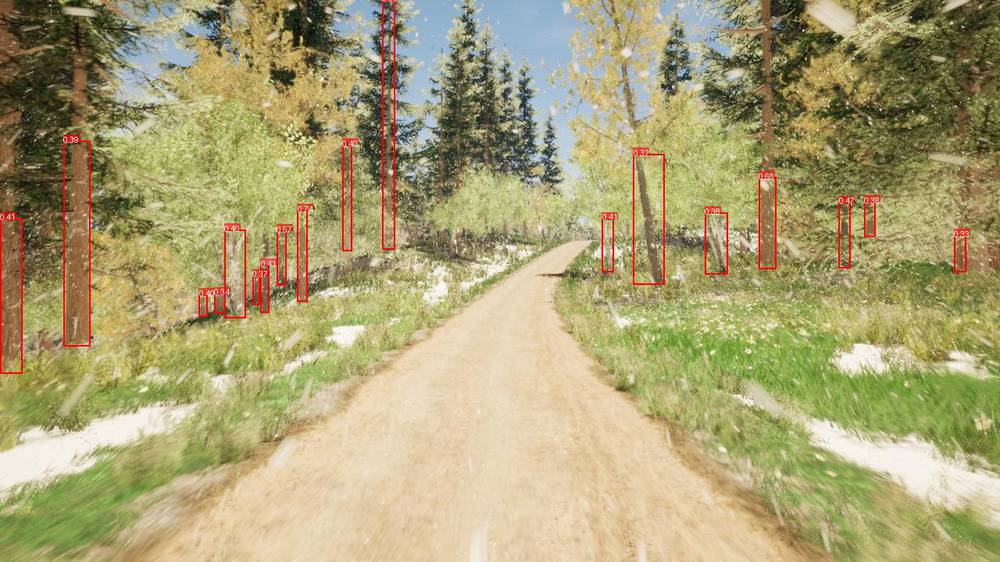} & 
		\includegraphics[width=0.13\textwidth]{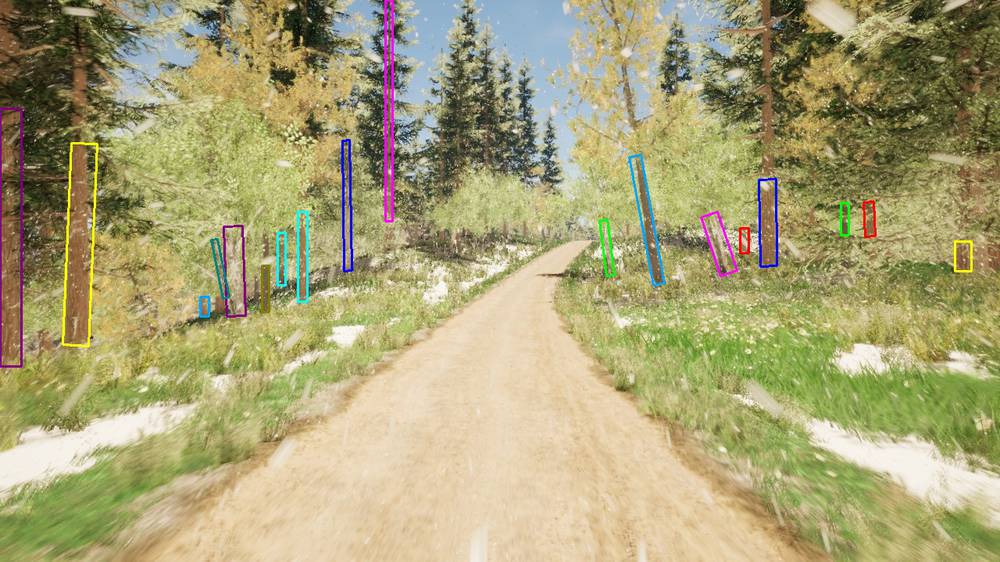}& 
		\includegraphics[width=0.13\textwidth]{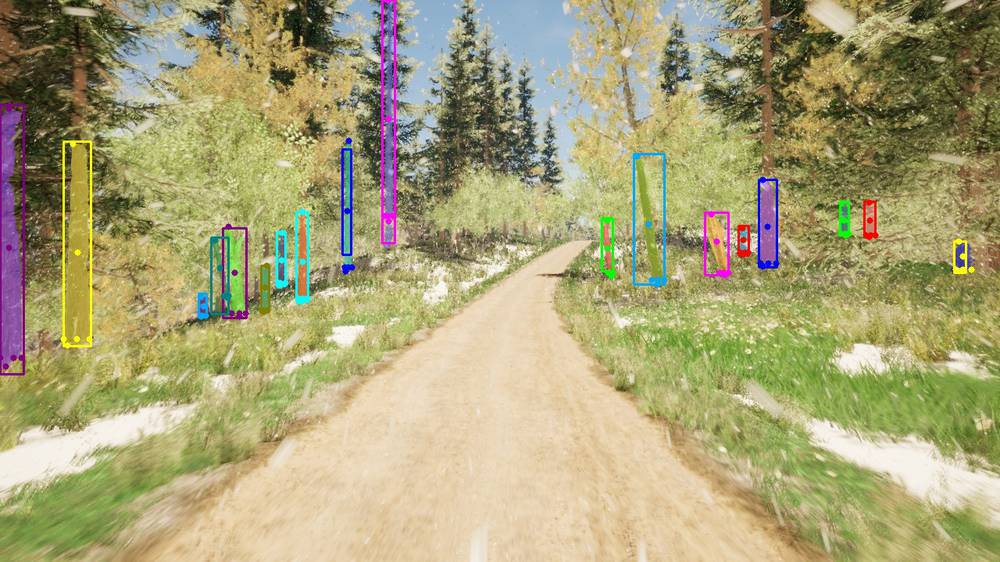} & 
		\includegraphics[width=0.13\textwidth]{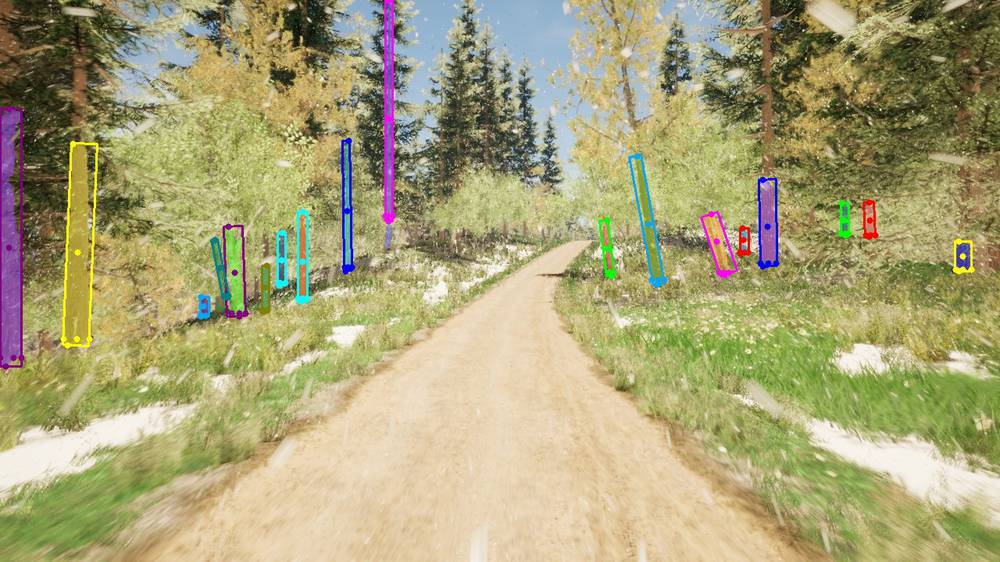} \\
		
		\includegraphics[width=0.13\textwidth]{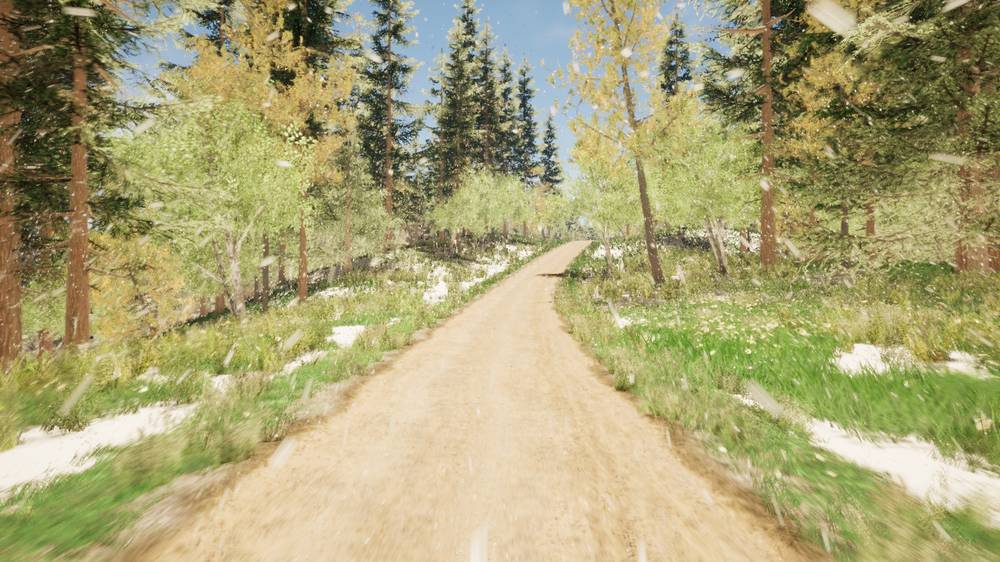} & 
		\includegraphics[width=0.13\textwidth]{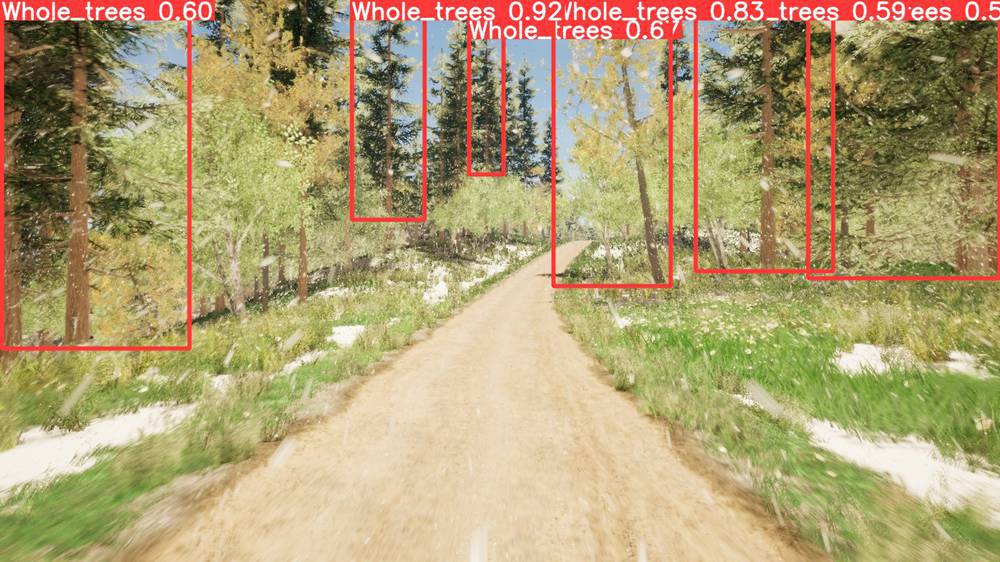} & 
		\includegraphics[width=0.13\textwidth]{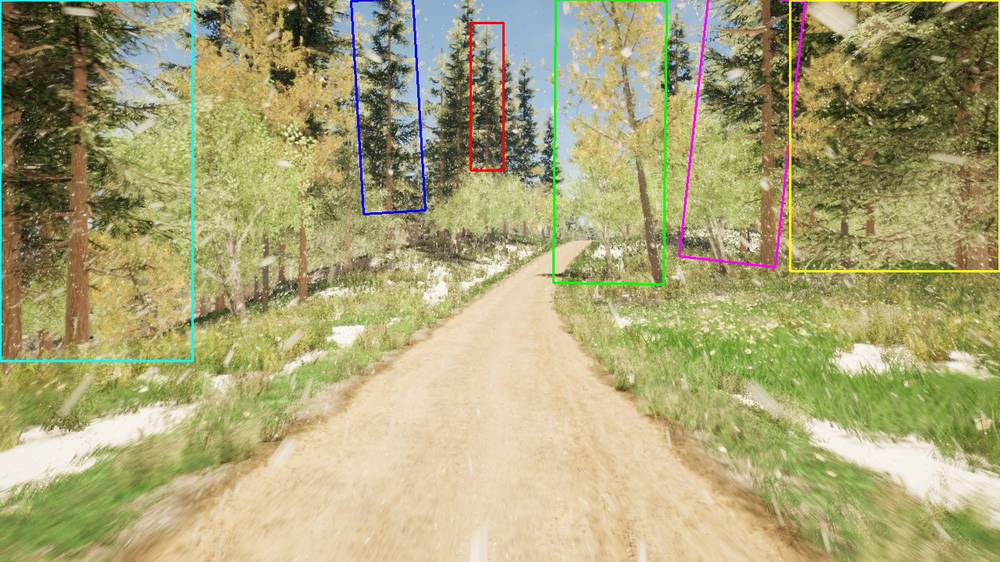} & 
		\includegraphics[width=0.13\textwidth]{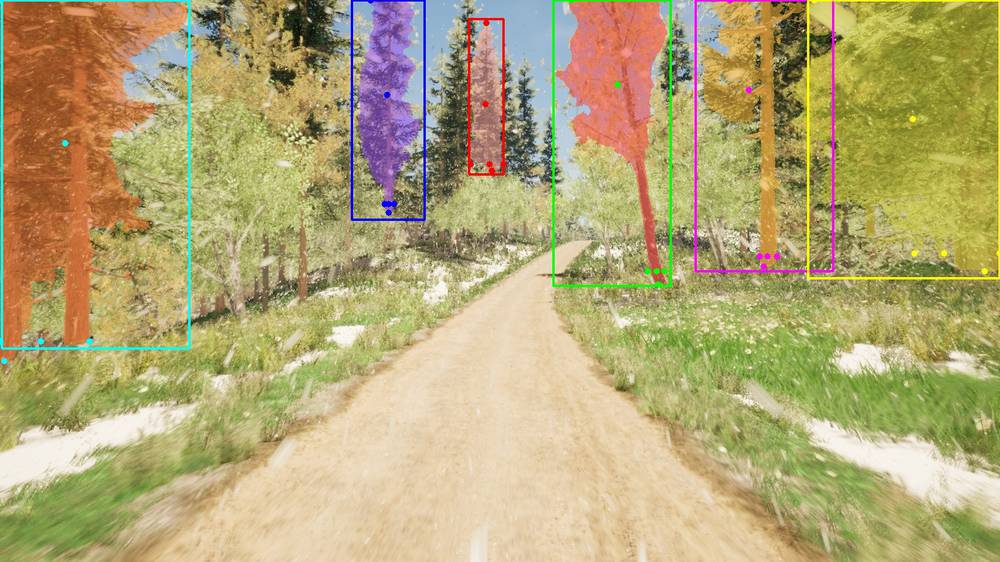} & 
		\includegraphics[width=0.13\textwidth]{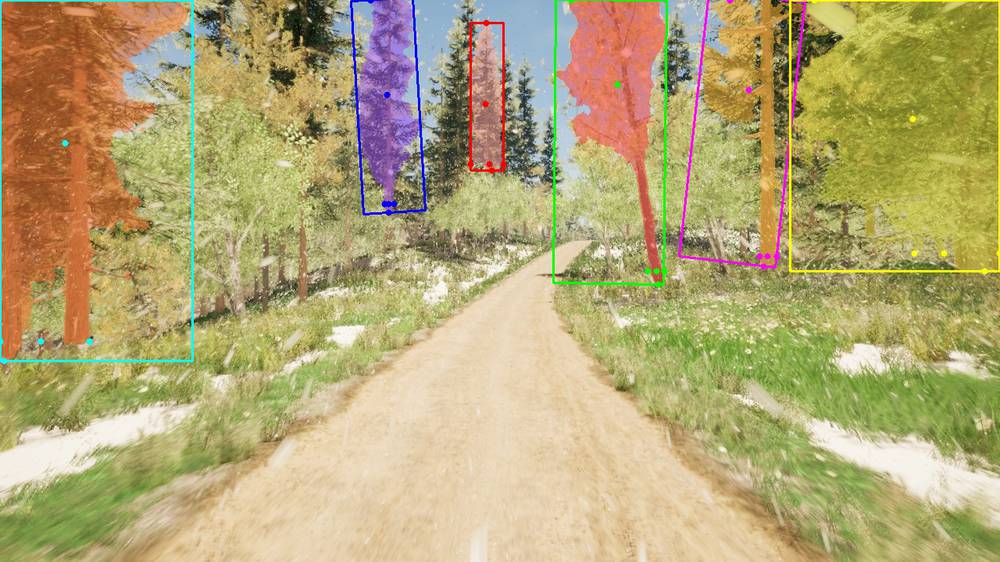} \\
		\bottomrule
	\end{tabular}
	
	\label{tab:annotations_examples}
\end{table*}

\section{Methodology}
\label{methodology}

Tree perception in forests is influenced by three factors: (i) a strong appearance
shift between synthetic and real imagery, (ii) a mismatch in label granularity
(trunk vs.\ whole tree vs.\ unified ``tree''), and (iii) pronounced scale
imbalance, as small and distant trees are disproportionately difficult to detect.
To analyze and address these effects, we adopt a four-stage evaluation protocol
that isolates each challenge before combining them into a unified distillation
baseline.

\subsection{Motivation and Design Rationale}

Our central question is:
\textit{How can a detector trained with only a single ``Tree'' label benefit from
	fine-grained synthetic supervision (trunk vs.\ whole tree) without re-annotating
	real data?}

The four stages of our protocol addresses this in a structured manner:
\begin{itemize}
	\item \textbf{Phase 1: Synthetic teachers} learn specialized trunk and whole-tree cues.
	\item \textbf{Phase 2: Zero-shot Sim$\rightarrow$Real} measures the pure domain gap.
	\item \textbf{Phase 3: Real-only student} establishes a coarse-label baseline.
	\item \textbf{Phase 4: Granularity-aware distillation} merges synthetic structure
	with real supervision.
\end{itemize}

This design tries to isolate the sources of error and provides an interpretable path from synthetic pretraining to real-world performance under mixed label granularity.

\subsection{Phase Descriptions}
\label{phase_descriptions}

\textbf{Phase 1: Synthetic Teachers.}
We train two teacher models: \(T_{\text{trunk}}\) (for tree trunks) and
\(T_{\text{whole}}\) (for whole trees), using the fine-grained synthetic labels.
Each teacher is optimized with a standard detection/segmentation objective:
\begin{equation}
	L_{T_x} = L_{\text{cls}} + L_{\text{box}} + L_{\text{mask}}, 
	\qquad x \in \{\text{trunk}, \text{whole}\}
	\label{eq:teacher_loss}
\end{equation}
Trunk and whole-tree teachers capture complementary geometric structure: dense
vertical patterns for trunks and larger silhouettes for whole trees.

\medskip
\noindent
\textbf{Phase 2: Zero-shot Synthetic$\rightarrow$Real.}
Since real images contain only the unified ``Tree'' label, we evaluate the
synthetic teachers using a label-agnostic AP metric:
\begin{equation}
	\mathrm{AP}_{\mathrm{LA}}
	= \mathrm{AP}\!\left(
	\hat{y}^{\text{trunk}} \cup \hat{y}^{\text{whole}},
	\;y^{\text{tree}}
	\right),
	\label{eq:la_ap}
\end{equation}
where a prediction is counted correct if either teacher matches a real tree
instance. This quantifies the domain gap independently of label mismatch.

\medskip
\noindent
\textbf{Phase 3: Real-only Student.}
A student model \(S\) is trained directly on real data using the single ``Tree''
label:
\begin{equation}
	L_S =
	L_{\text{cls}}^{\text{tree}}
	+ L_{\text{box}}^{\text{tree}}
	+ L_{\text{mask}}^{\text{tree}}
	\label{eq:student_loss}
\end{equation}
This stage provides a baseline aligned with real appearance statistics, but lacks
the trunk/whole-tree decomposition available in simulation.

\medskip
\noindent
\textbf{Phase 4: Granularity-Aware Knowledge Distillation.}
The final stage unifies synthetic and real supervision by transferring 
fine-grained structural cues (trunk vs.\ whole tree) from frozen synthetic 
teachers into the student, which predicts only a single ``Tree'' class. 
Our approach adapts the Mean Teacher framework~\cite{tarvainen2017mean} to handle 
granularity mismatch through three key modifications:
\begin{enumerate}
	\item Using two frozen synthetic teachers: $T_{\text{trunk}}$ and $T_{\text{whole}}$
	\item Merging their predictions via a granularity bridge into soft targets compatible with the student's single label space
	\item Adding knowledge distillation losses across classification, bounding box, and mask prediction heads
\end{enumerate}

A key requirement is to map the teachers' fine-grained outputs into the 
student's coarse label space without assuming independence between the 
fine-grained classes. Since a trunk is physically part of a whole tree, 
probability-level combinations (e.g., OR rules) would be incorrect under 
Softmax-based classification. To address this, we merge the teachers' 
outputs in \emph{logit space}, which is standard in hierarchical and 
coarse-to-fine distillation.

Let $z_{\text{trunk}}$ and $z_{\text{whole}}$ denote the classification 
logits from the two teachers. A unified ``Tree'' logit is defined using 
a smooth log-sum-exp merge:
\begin{equation}
	z_{\text{tree}}
	=
	\log\!\left(
	e^{z_{\text{trunk}}}
	+
	e^{z_{\text{whole}}}
	\right)
	\label{eq:lse_merge}
\end{equation}
This operation preserves contribution from either teacher while making no 
assumptions about independence. The resulting soft target probability is:
\begin{equation}
	p_T(\text{tree}\mid x)
	= \sigma(z_{\text{tree}}),
\end{equation}
where $\sigma(\cdot)$ is a sigmoid (for binary tree/non-tree prediction).

Spatial predictions are mapped analogously. Teacher masks are merged via set union:
\[
M_T^{\text{tree}}
=
M_T^{\text{trunk}} \cup M_T^{\text{whole}},
\]
and bounding boxes use the enclosing box of the trunk and whole-tree boxes.
This provides unified geometric signals compatible with the student's label
space while retaining the richer structure present in simulation.

The student is optimized using a joint objective:
\begin{equation}
	L 
	= 
	L_S
	+ \lambda_{\text{KD}} \big(L^{\text{cls}}_{\text{KD}} + L^{\text{box}}_{\text{KD}} + L^{\text{mask}}_{\text{KD}}\big)
	+ \lambda_{\text{cons}} L_{\text{cons}},
	\label{eq:phase4_loss}
\end{equation}
where $L_S$ is the supervised loss on real data, $L_{\text{KD}}$ distills 
classification, box, and mask predictions from the merged teacher signals, 
and $L_{\text{cons}}$ enforces augmentation-invariant consistency in the 
spirit of Mean Teacher. The distillation term injects fine-grained structural 
priors from the synthetic teachers, while consistency encourages stable 
predictions under appearance perturbations. Hyperparameters $\lambda_{\text{KD}}$ 
and $\lambda_{\text{cons}}$ balance the relative importance of distillation 
and consistency. Together, these components reconcile both the domain shift 
(Sim$\rightarrow$Real) and the granularity mismatch (trunk/whole$\rightarrow$tree).

\subsection{Implementation Details}

\textbf{Object detection.}
We train YOLOv8, YOLOv11 (n/s/m/l/x), and RT-DETR on both synthetic and real
data. Synthetic images (1280$\times$720) are trained at 1024$\times$1024
resolution for 100 epochs; real images (4608$\times$3456) are trained at
640$\times$640 for 50 epochs. Optimization uses SGD with momentum~0.937, weight
decay~$5\times10^{-4}$, and linear learning-rate decay with 3-epoch warmup.

\medskip
\noindent
\textbf{Instance segmentation.}
Mask R-CNN and Cascade Mask R-CNN are trained using MMDetection~\cite{mmdetection}
with ResNet, ResNeXt, and Swin-T backbones. Training follows the standard 24-epoch
schedule with data augmentation (resize, flip, normalization). Loss functions use
cross-entropy for classification, L1/IoU for bounding boxes, and BCE for masks.
Full hyperparameters are provided in the supplementary material.

\section{Experiments}
\label{experimentation}

We evaluate the proposed four-phase protocol on MGTD with a focus on three questions: (i) how large is the \emph{pure} Sim$\rightarrow$Real gap under mixed granularity, (ii) what is gained by training on real images alone and what is lost by discarding fine-grained structure, and (iii) to what extent granularity-aware transfer can improve real-world performance.

In this section we focus on instance segmentation in the mixed Sim$\rightarrow$Real setting, using Mask R-CNN as the canonical backbone for both teachers and student. We emphasize segmentation over detection because accurate pixel-level masks are critical for robotics tasks requiring close-proximity navigation and structural understanding of trees, especially under granularity mismatch where trunk and crown shapes must be preserved. However, a broader set of object detectors (YOLOv8/11, Cascade R-CNN, RT-DETR) is also trained; their detection results are reported in the supplementary material. 

\subsection{Experimental Setup}

\paragraph{\textbf{Teachers}.}
Two synthetic teachers are trained independently: one on the \emph{tree trunk} class and one on the \emph{whole tree} class. For each category, we evaluate Mask R-CNN with ResNeXt-101 and Swin-T backbones. Swin-T consistently achieves higher mask AP (see supplementary Table 8) and is therefore adopted as the final teacher architecture in subsequent phases. Teachers are trained on the synthetic set using the 70/20/10 (train/val/test) split and evaluated both on their synthetic validation split (upper bound) and on the real validation split using the label-agnostic metric from Eq.~\ref{eq:la_ap}.

\paragraph{\textbf{Students}.}
For Phase~3 (real-only baseline), we train multiple Mask R-CNN variants with different backbones: ResNet-50, ResNeXt-101, and Swin-T, all initialized from COCO pretrained weights. This provides a comprehensive baseline of what can be achieved with coarse real labels alone.

For Phase~4 (granularity-aware distillation), we focus on a Mask R-CNN with ResNet-50 backbone to demonstrate that effective knowledge transfer can compensate for-and even surpass-the capacity advantage of larger models. This student receives additional guidance from the frozen Swin-T synthetic teachers via the distillation framework described in Section~\ref{phase_descriptions}. In both phases, real data follow a 70/20/10 split, and all reported metrics are computed on the held-out real validation set for fair comparison.

\paragraph{\textbf{Metrics}.}
We report COCO-style mask AP:
\begin{equation}
	\text{AP}_{50:95} = \frac{1}{10} \sum_{\text{IoU}=0.50}^{0.95} 
	\text{AP}_{\text{IoU}},
\end{equation}
along with size-stratified AP$_s$, AP$_m$, and AP$_l$ in the supplementary
material. For zero-shot transfer of synthetic teachers to real data (Phase~2),
we use the \emph{label-agnostic} metric
\begin{equation}
	\text{AP}_{\text{LA}} 
	= \text{AP}\!\left(
	\hat{y}^{\text{trunk}} \cup \hat{y}^{\text{whole}},
	\, y^{\text{tree}}
	\right),
\end{equation}
where any match between either teacher prediction and a ground-truth tree
instance is counted as correct.

\subsection{Results}

\begin{figure}[ht]
	\centering
	\renewcommand{\arraystretch}{0.9}
	\setlength{\tabcolsep}{1pt}
	\begin{tabular}{c c c}
		\textbf{RGB Image} & \textbf{Ground Truth} & \textbf{Prediction} \\
		
		\includegraphics[width=0.3\linewidth]{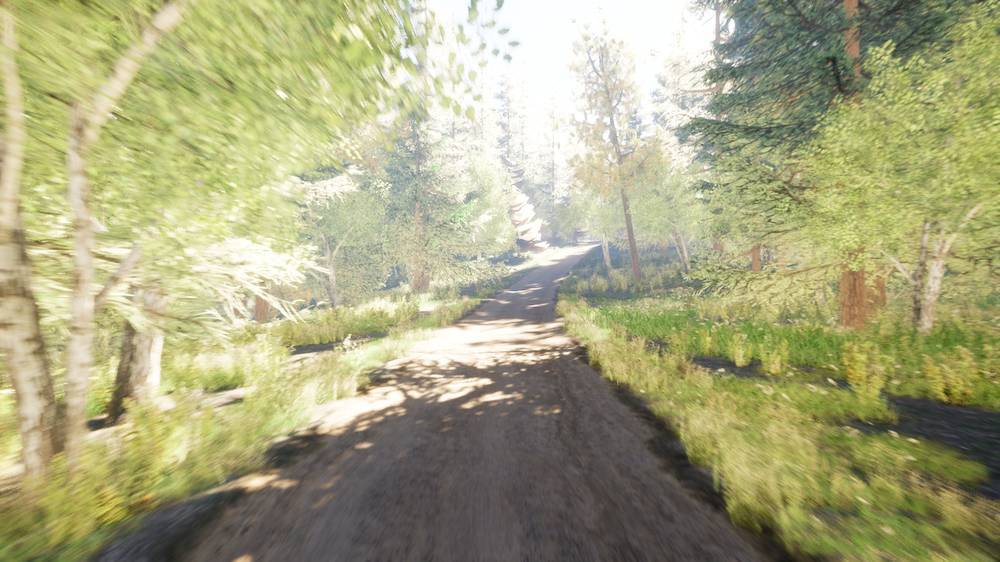} &
		\includegraphics[width=0.3\linewidth]{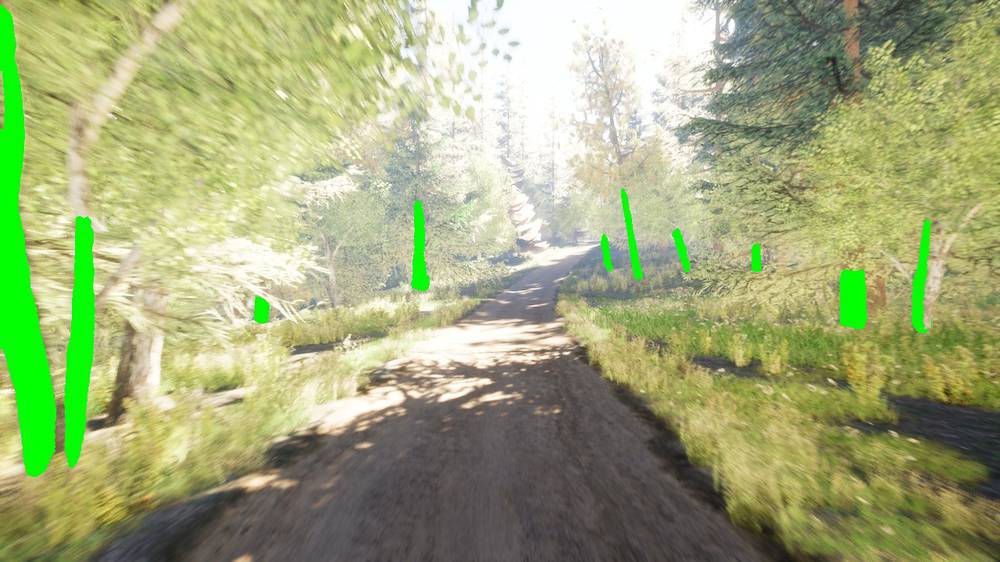} &
		\includegraphics[width=0.3\linewidth]{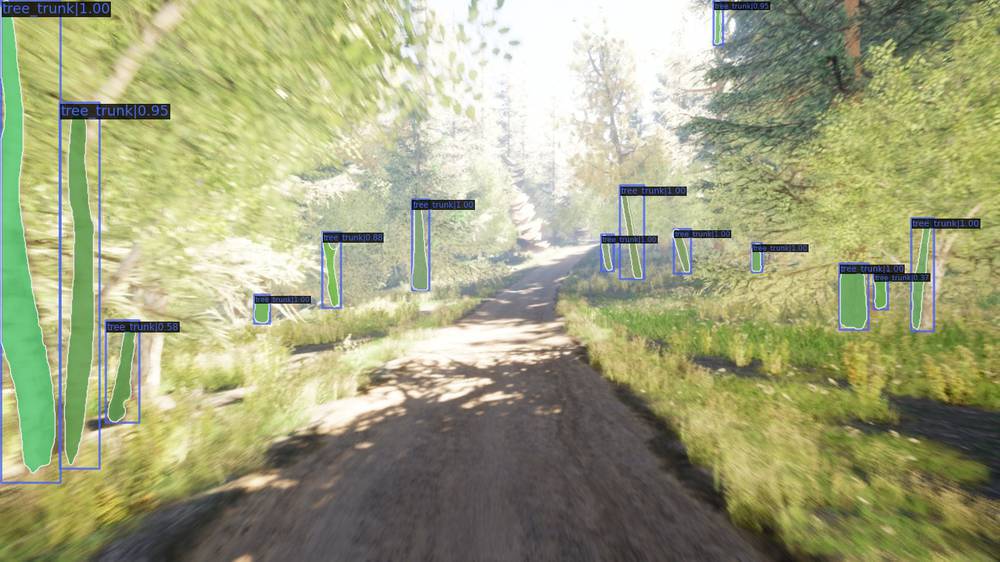} \\
	
		\includegraphics[width=0.3\linewidth]{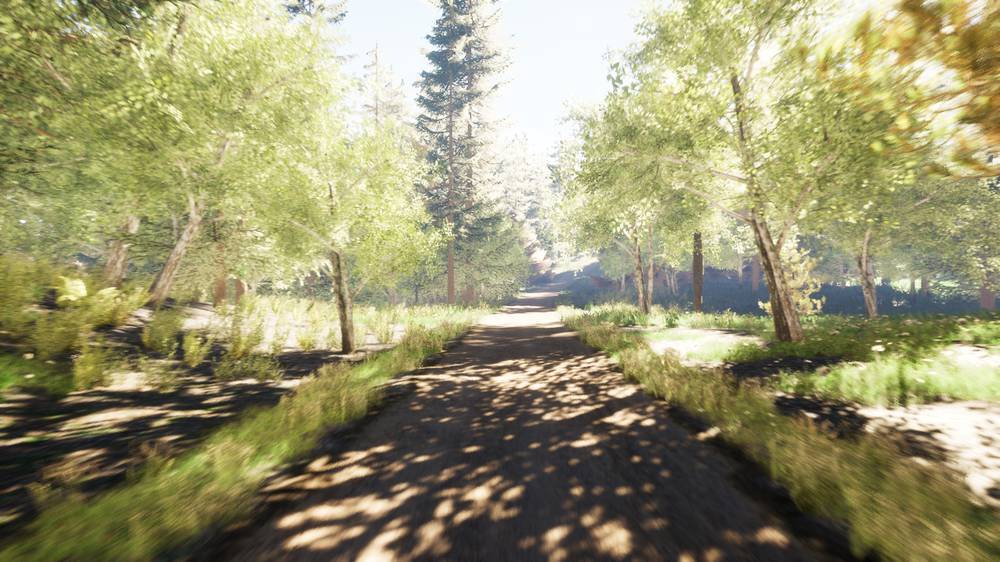} &
		\includegraphics[width=0.3\linewidth]{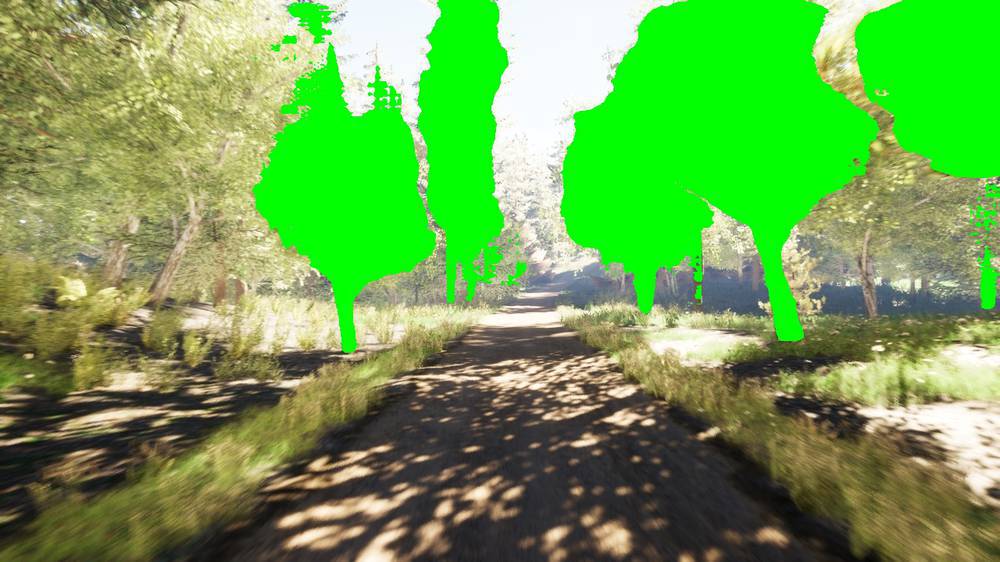} &
		\includegraphics[width=0.3\linewidth]{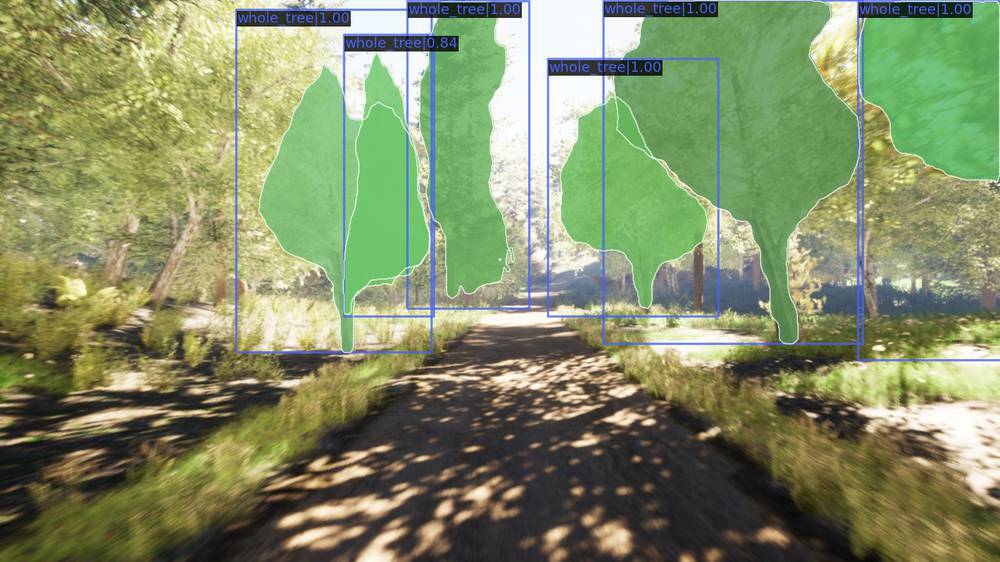} \\
		 
	\end{tabular}
	\caption{Phase~1: Instance segmentation results on synthetic validation data. Top row shows \textbf{tree trunk} example, bottom row shows \textbf{whole tree}. Columns display RGB input, ground-truth and model prediction. More results in supplementary (Fig. 2 \& Fig. 3).}
	\label{fig:phase1_examples}
\end{figure}

\paragraph{Phase 1: Synthetic upper bound.}
Table~\ref{tab:phase1} summarizes teacher performance on the synthetic
validation set. Mask R-CNN with Swin-T achieves
\textbf{0.788} mAP$_{50:95}$ (bbox) and \textbf{0.753} mAP$_{50:95}$
(segm) for tree trunks, and up to \textbf{0.880} mAP$_{50:95}$ (bbox) for
whole trees. These results define an upper bound achievable when training and
testing on the same simulated domain with fine-grained labels. Qualitative
examples in Fig.~\ref{fig:phase1_examples} show that teachers capture most visible structures, with remaining errors
concentrated on very thin or distant trunks and partially occluded crowns.

\begin{table}[t]
	\centering
	\caption{Phase~1 (synthetic only). Teacher models trained on tree trunk and 
	whole tree simulation data, evaluated on the synthetic validation set. 
	We observe that Swin-T outperforms X101 (for both Mask RCNN \& Cascade Mask RCNN) and is therefore chosen as the final teacher. Full results are provided in the supplementary (Table~8).}
	\resizebox{\linewidth}{!}{
		\begin{tabular}{|c|c|c|c|c|c|}
			\toprule
			\textbf{Model} & \textbf{Type} & \textbf{Backbone} & \textbf{mAP@0.5:0.95 (BBox)} & \textbf{mAP@0.5 (BBox)} & \textbf{mAP@0.5:0.95 (Segm)} \\
			\midrule
			Mask-RCNN & Tree trunk & X101   & 0.732 & 0.904 & 0.686 \\
			Mask-RCNN & Tree trunk & Swin-T & 0.788 & 0.936 & \textbf{0.753} \\
			\midrule
			Mask-RCNN & Whole tree & X101   & 0.859 & 0.947 & 0.670 \\
			Mask-RCNN & Whole tree & Swin-T & 0.880 & 0.948 & \textbf{0.688} \\
			\bottomrule
	\end{tabular}}

	\label{tab:phase1}
\end{table}

\begin{figure}[ht]
	\centering
	\renewcommand{\arraystretch}{1.2}
	\setlength{\tabcolsep}{1pt}
	\begin{tabular}{c c c c}
		RGB & GT & Trunk$\rightarrow$Real & Whole$\rightarrow$Real  \\
		\includegraphics[width=0.23\linewidth, angle=180, origin=c]{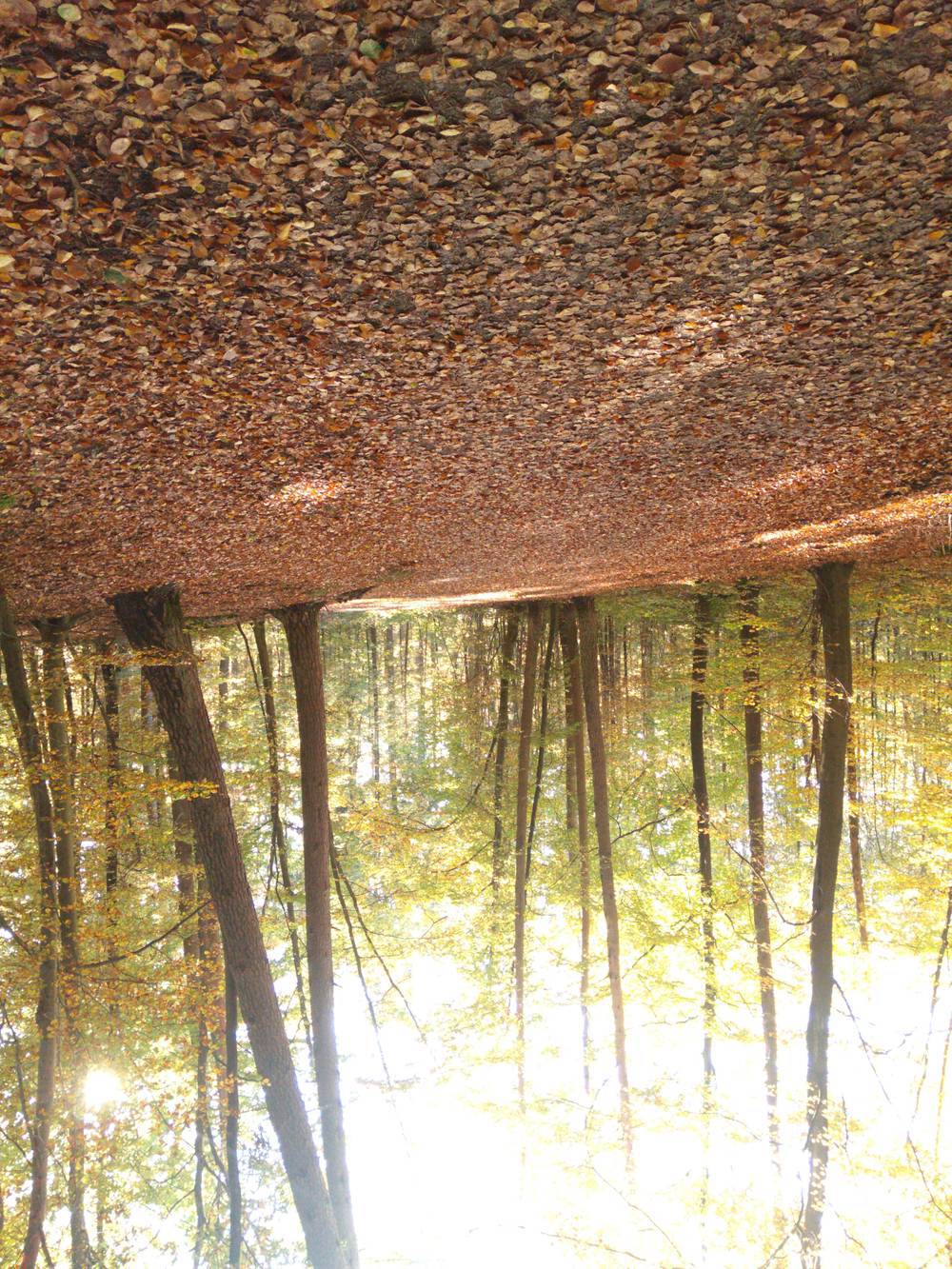} &
		\includegraphics[width=0.23\linewidth]{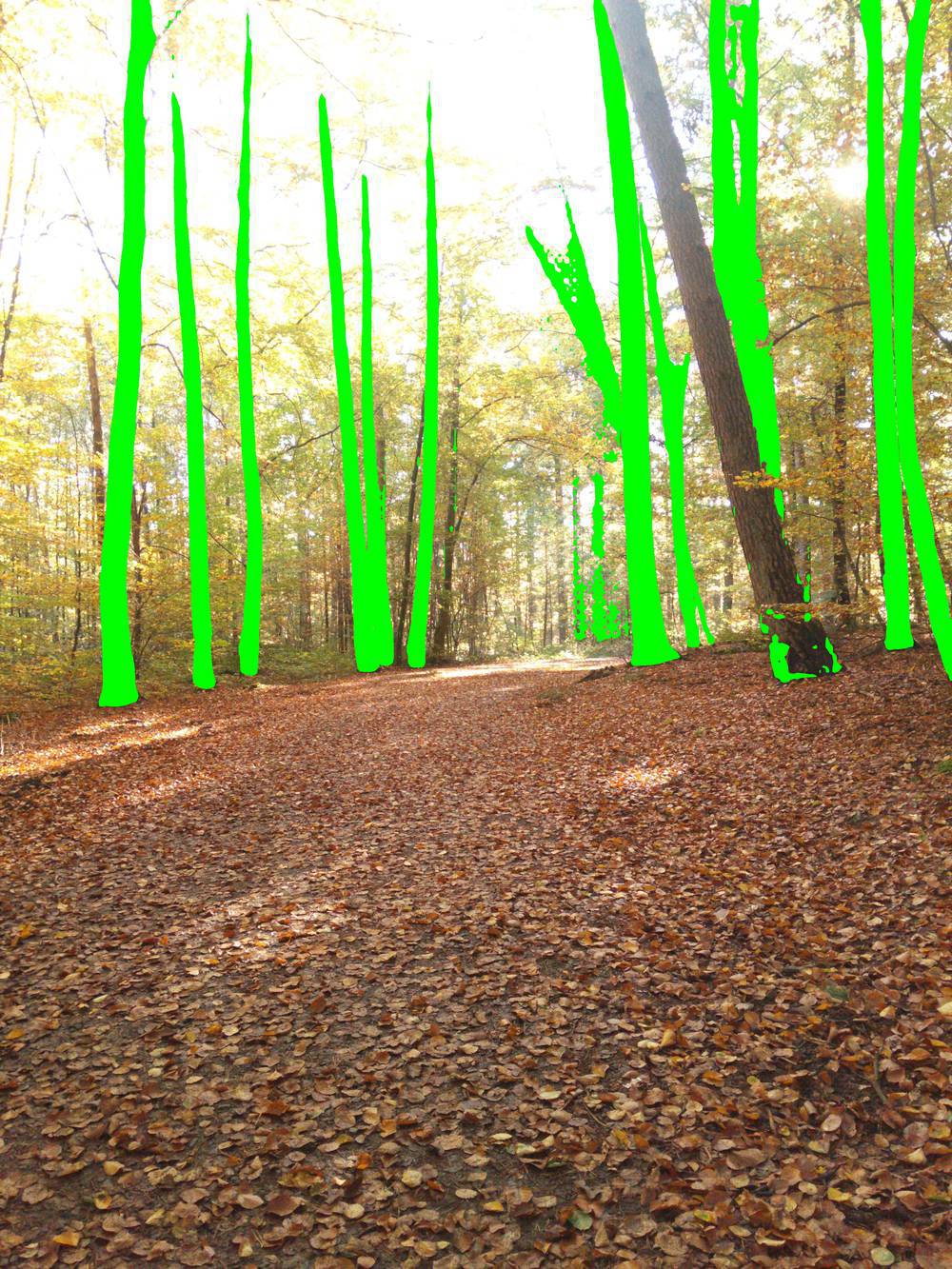} &
		\includegraphics[width=0.23\linewidth]{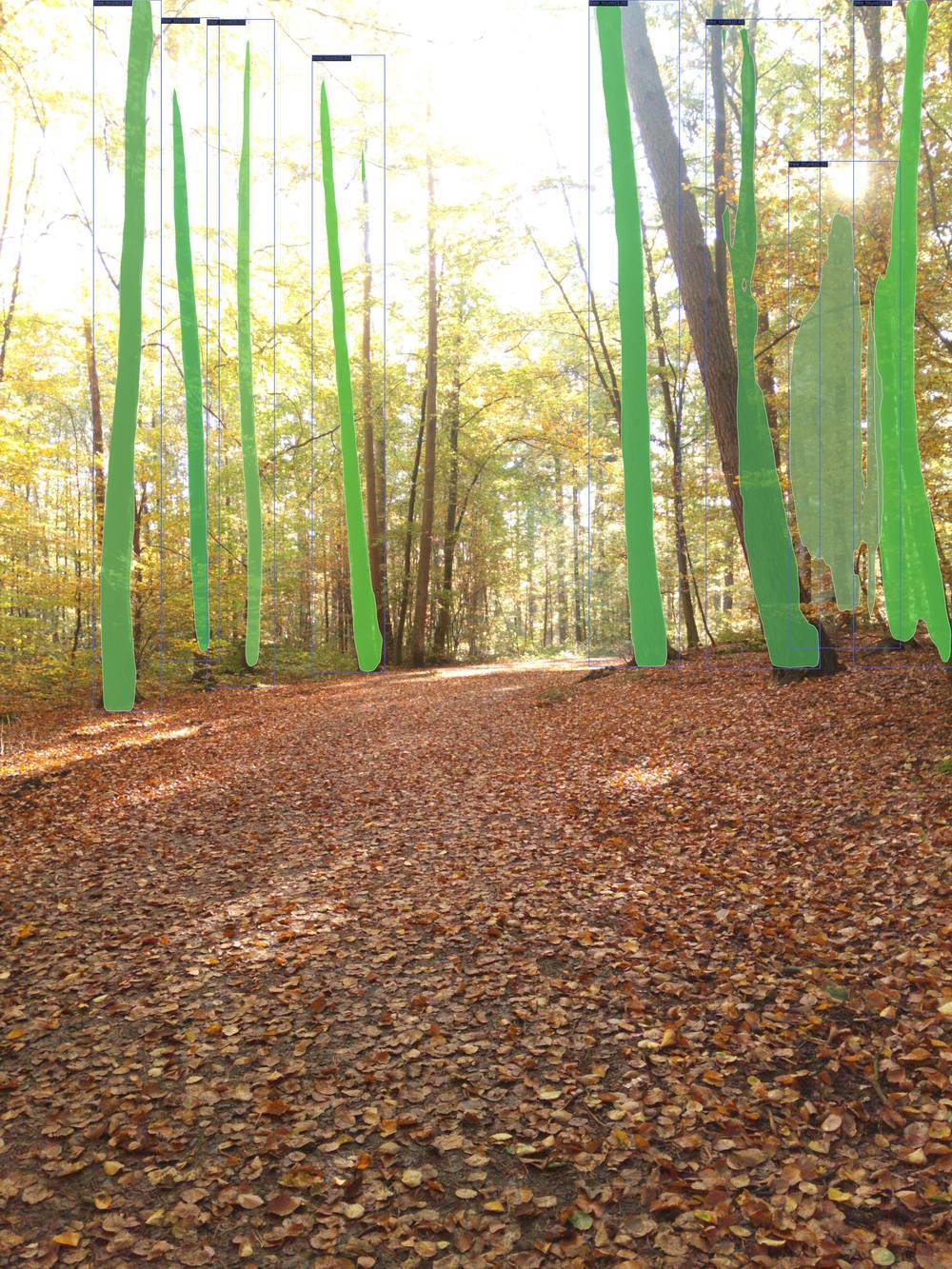} &
		\includegraphics[width=0.23\linewidth]{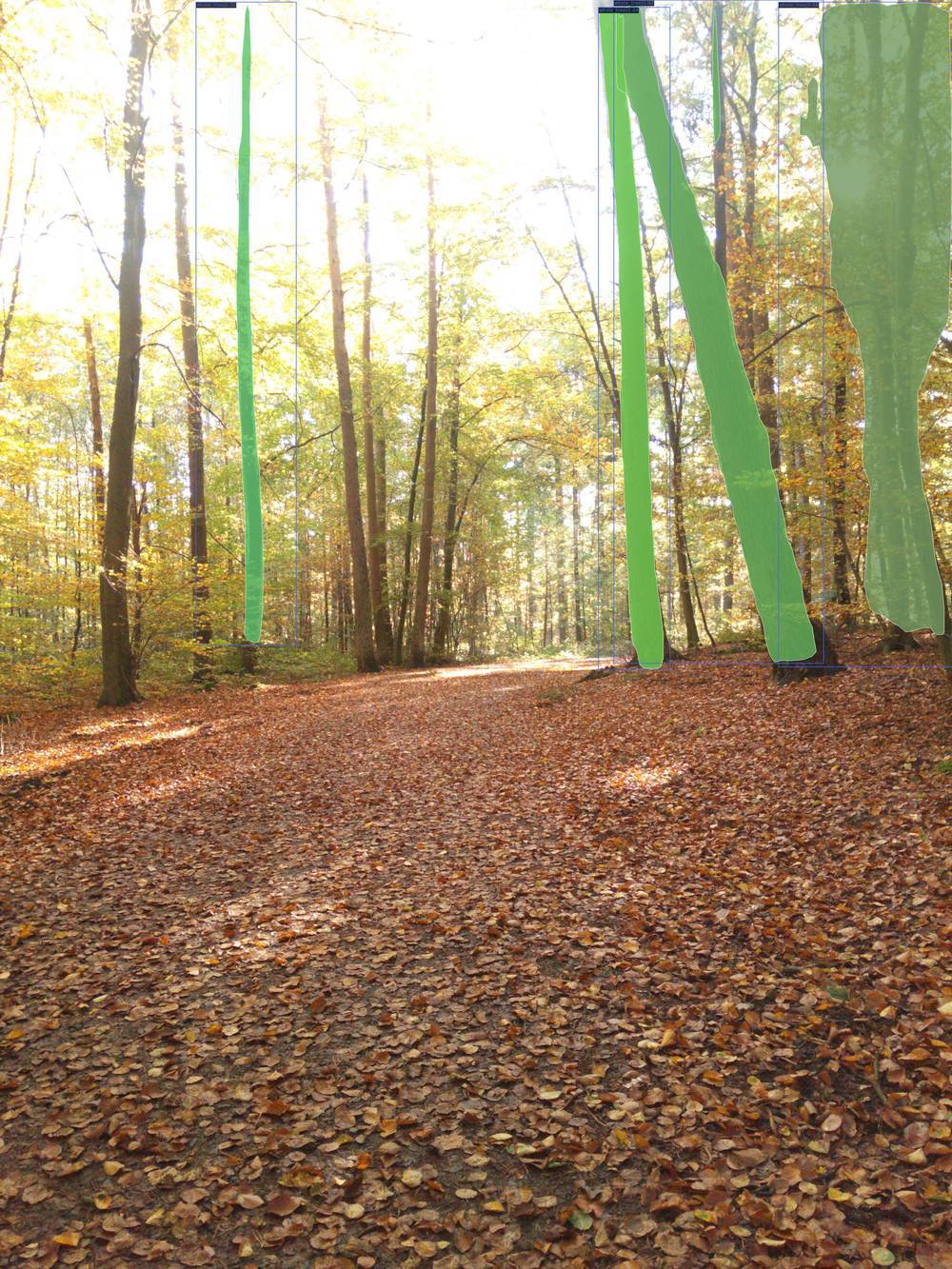} \\
		
		\includegraphics[width=0.23\linewidth, angle=180, origin=c]{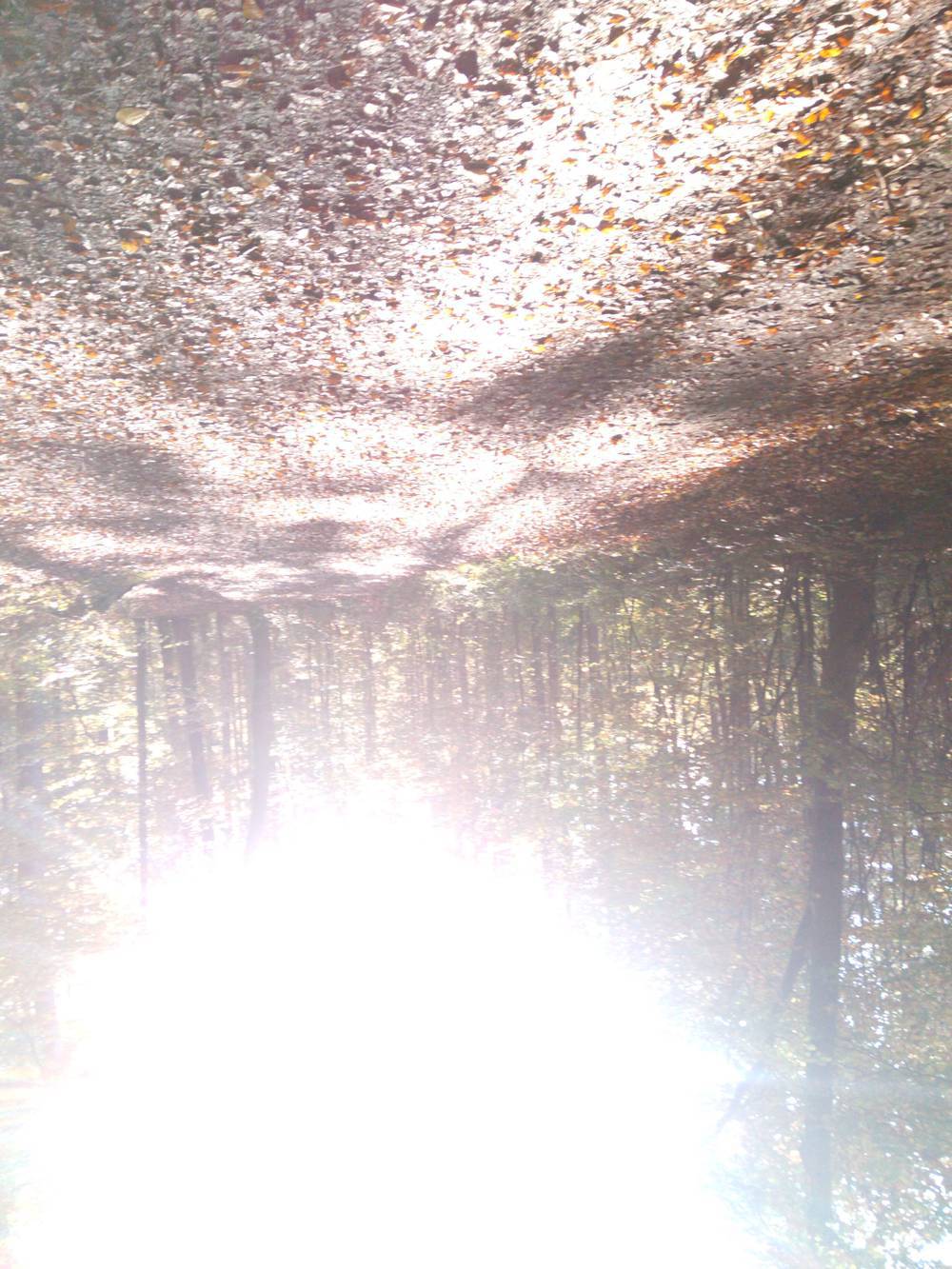} &
		\includegraphics[width=0.23\linewidth]{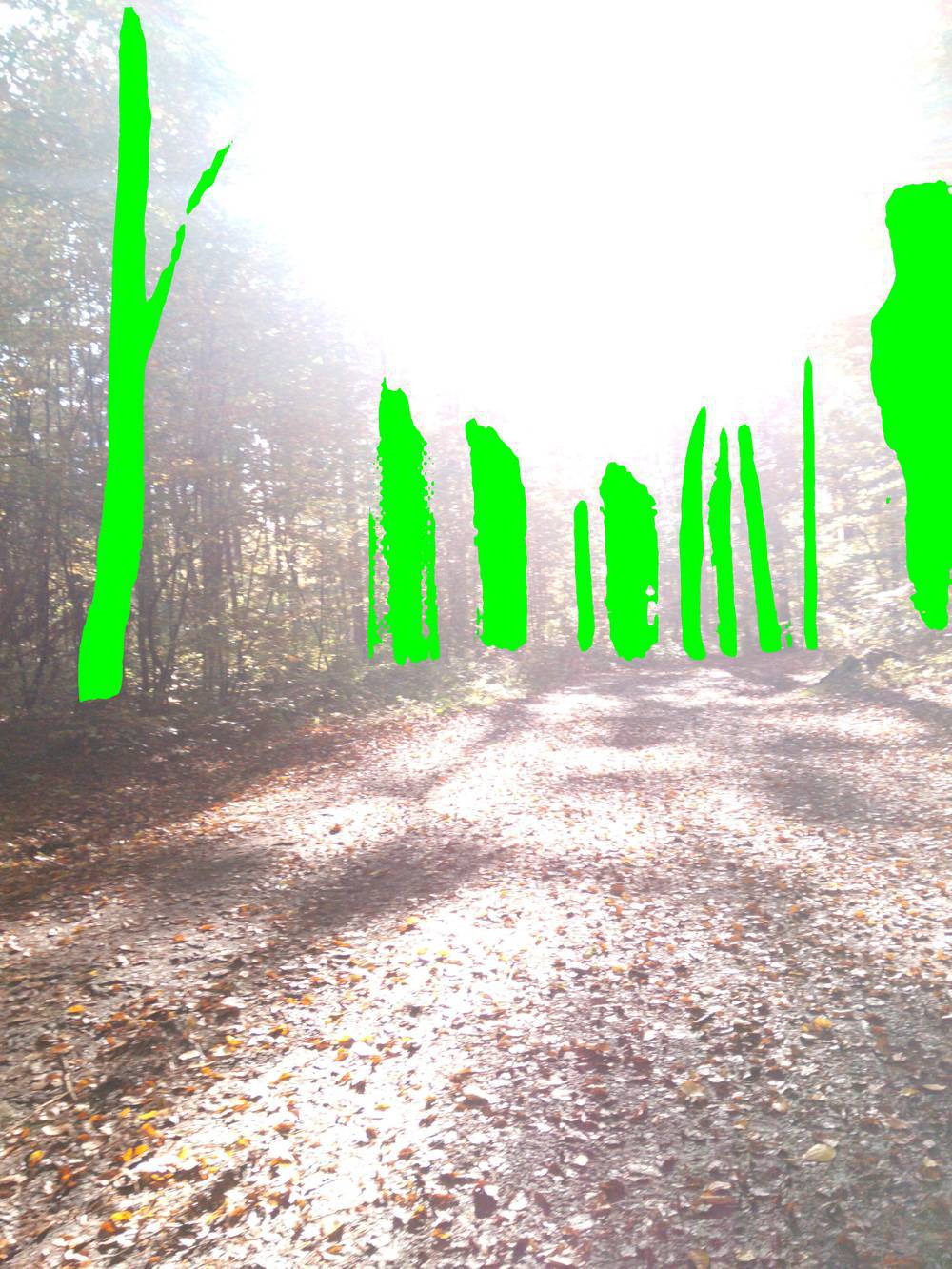} &
		\includegraphics[width=0.23\linewidth]{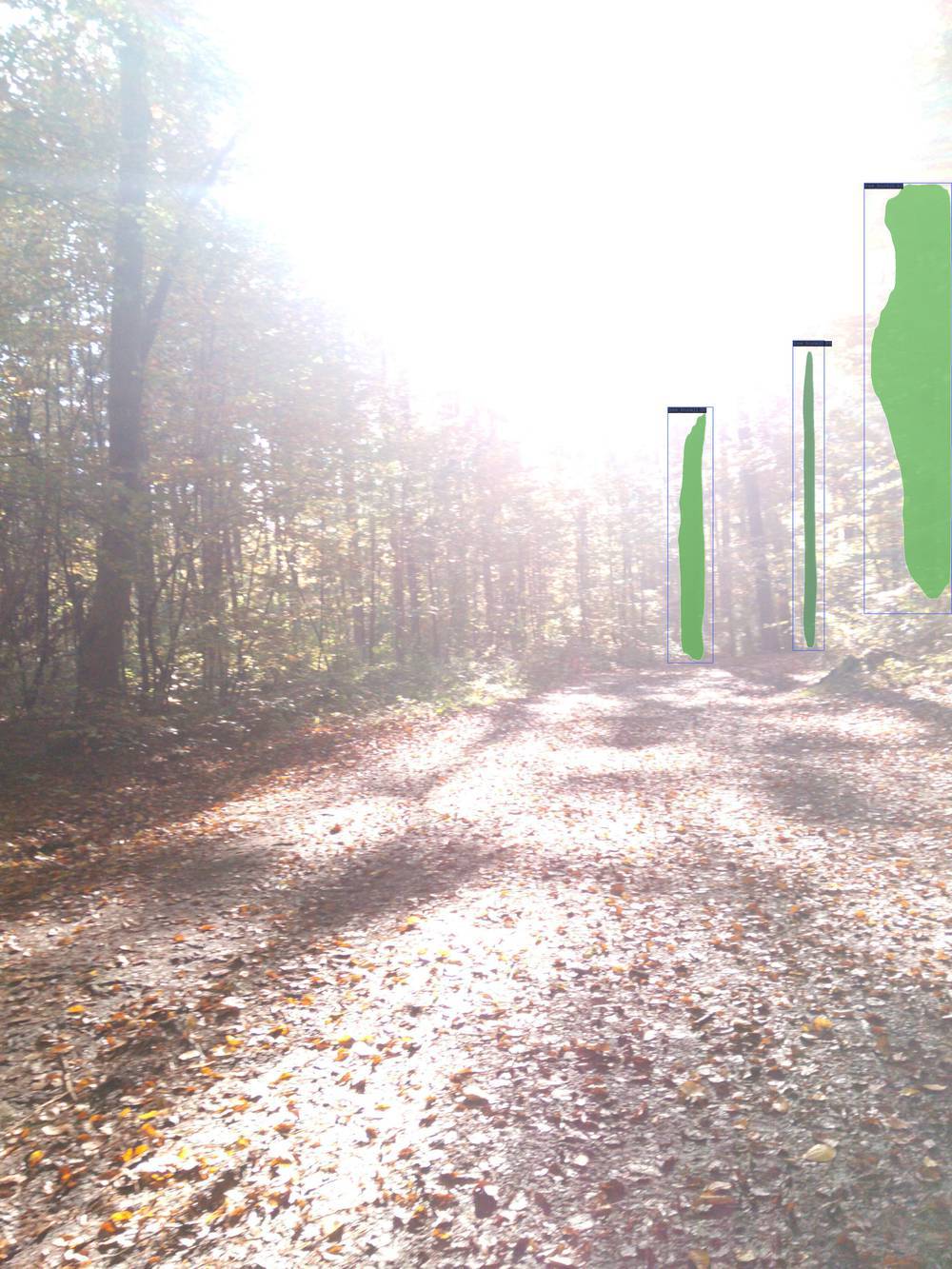} &
		\includegraphics[width=0.23\linewidth]{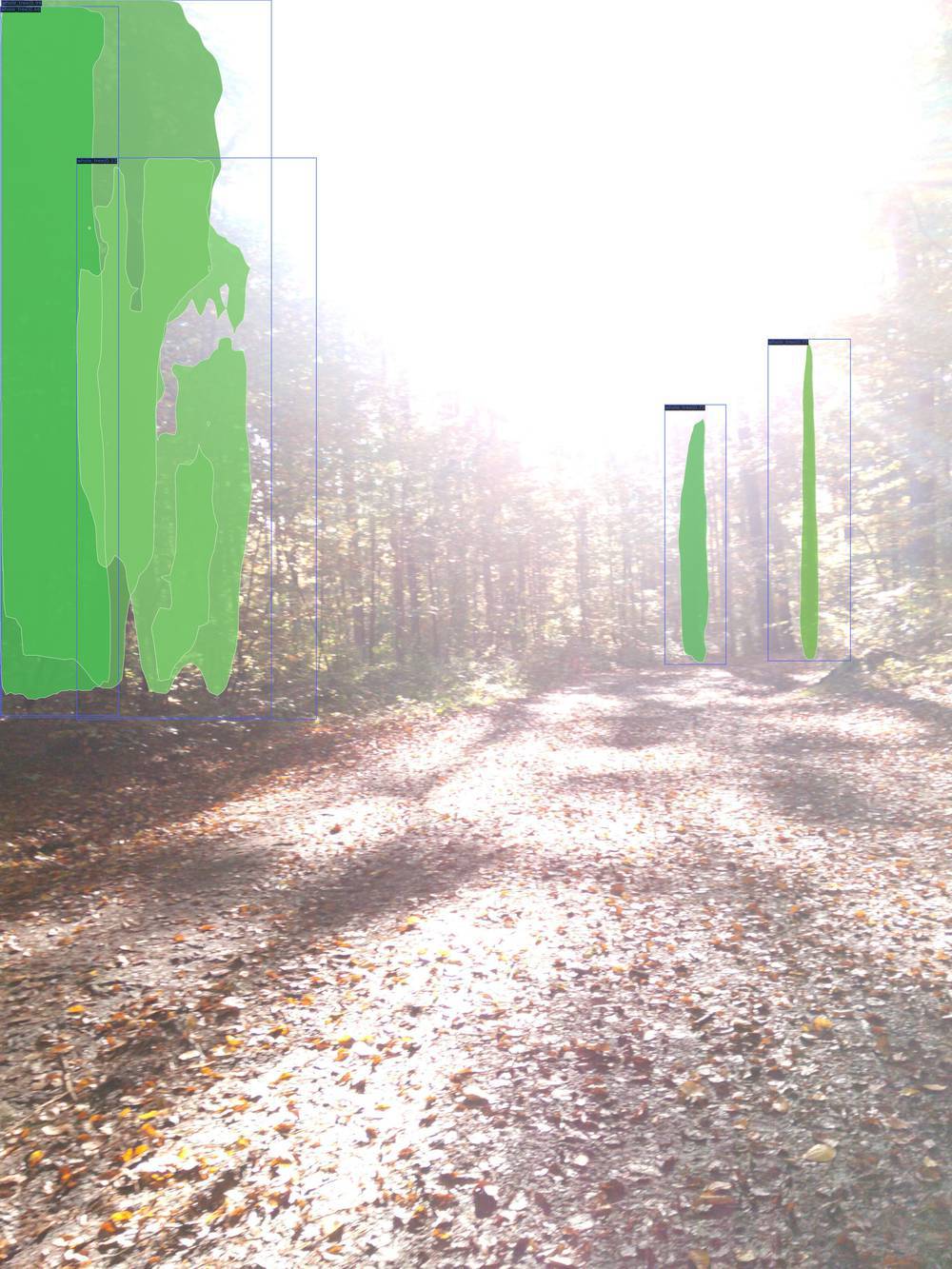} \\
		
	\end{tabular}
	\caption{Phase~2: Domain transfer. Comparison of predictions obtained from 
		\textbf{tree trunk} and \textbf{whole tree} teachers when evaluated on real 
		forest images.}
	\label{fig:phase2_trunk_vs_whole}
\end{figure}

\paragraph{Phase 2: Zero-shot Sim$\rightarrow$Real transfer.}
Table~\ref{tab:phase2} reports the performance of the synthetic teachers when 
evaluated zero-shot on real forest images using the label-agnostic AP metric. 
Despite strong synthetic performance, both trunk and whole-tree teachers suffer 
severe degradation when transferred to real data. For the trunk teacher,
mAP@0.5:0.95 (bbox) drops from \textbf{0.788} to \textbf{0.225}, and 
mAP@0.5:0.95 (segm) from \textbf{0.753} to \textbf{0.147}. The whole-tree 
teacher is even more affected, with bbox mAP@0.5:0.95 decreasing from 
\textbf{0.880} to \textbf{0.066}. These large relative drops (often above 
70-90\%) highlight the strength of the domain gap, even under label-agnostic 
evaluation. Figure~\ref{fig:phase2_trunk_vs_whole} illustrates typical failure 
modes, where illumination changes and background clutter lead to missed trunks 
and fragmented crowns.

\begin{table}[t]
	\centering
		\caption{Phase~2 (Sim$\rightarrow$Real). Synthetic teachers evaluated 
		zero-shot on the real validation set with label-agnostic AP. We report 
		performance on synthetic vs.\ real data along with absolute and relative 
		drops. We use the Swin-T backbone for the below-mentioned tests.}
	\resizebox{0.95\linewidth}{!}{
		\begin{tabular}{|c|c|c|c|c|c|}
			\toprule
			\textbf{Model} & \textbf{Type} & \textbf{Task} & \textbf{Sim Val} & \textbf{Sim$\rightarrow$Real} & \textbf{Abs. Gap (Rel. Drop)} \\
			\midrule
			Mask-RCNN  & Tree trunk & BBox  & 0.788 & 0.225 & 0.563 \, (71.4\%) \\
			Mask-RCNN & Tree trunk & Segm  & 0.753 & 0.147 & 0.606 \, (80.5\%) \\
			\midrule
			Mask-RCNN & Whole tree & BBox  & 0.880 & 0.066 & 0.814 \, (92.5\%) \\
			Mask-RCNN & Whole tree & Segm  & 0.688 & 0.039 & 0.649 \, (94.3\%) \\
			\bottomrule
	\end{tabular}}

	\label{tab:phase2}
\end{table}

\paragraph{Phase 3: Real-only baseline.}
We train Mask R-CNN models with different backbones (ResNet-50, ResNeXt-101, Swin-T) 
directly on real images annotated with the single coarse label \emph{Tree}. 
Table~\ref{tab:phase3_phase4} reports performance on the real validation set. 
Mask R-CNN with Swin-T achieves the best performance, reaching 
\textbf{0.500} mAP$_{50:95}$ (bbox) and \textbf{0.420} mAP$_{50:95}$ (segm), 
outperforming ResNet-50 (by +0.013 (bbox), +0.024 (segm)) and ResNeXt-101 (by +0.039 (bbox), +0.057 (segm)). 

Unlike Phases~1 and 2, Phase~3 does not aim to reduce the Sim$\rightarrow$Real 
gap but establishes an independent baseline trained exclusively on real data. 
As expected, real-only training substantially outperforms zero-shot transfer 
(Phase~2), confirming that exposure to target-domain appearance is essential. 
However, qualitative analysis (refer to Phase~3 column in Fig.~\ref{fig:phase4_qualitative}) 
reveals persistent challenges: fine-grained trunk structures, small trees, 
and heavily occluded instances remain difficult, indicating that real-only 
supervision alone cannot recover the detailed structural priors available 
in synthetic fine-grained annotations.

\begin{table}[t]
	\centering
	\caption{Phase~3 (real-only) results.
		Students are evaluated on the real validation set. ``BBox'' reports COCO-style
		bounding-box AP, and ``Segm'' reports mask AP. mAP@0.5:0.95 averages AP across IoU
		thresholds, while mAP@0.5 is AP at IoU~=~0.5.}
	\label{tab:phase3_phase4}
	\renewcommand{\arraystretch}{1.1}
	\setlength{\tabcolsep}{6pt}
	\resizebox{\linewidth}{!}{
	\begin{tabular}{|l| l| l| c| c| c|}
		\toprule
		\textbf{Model} & \textbf{Type} & \textbf{Backbone} &
		\textbf{mAP@0.5:0.95 (BBox)} &
		\textbf{mAP@0.5 (BBox)} &
		\textbf{mAP@0.5:0.95 (Segm)} \\
		\midrule
		Mask R-CNN & Tree & R50 & 0.487 & 0.817 & 0.396 \\
		Mask R-CNN & Tree & X101   & 0.461 & 0.810 & 0.363 \\
		Mask R-CNN & Tree & Swin-T & 0.500 & 0.792 & 0.420 \\
		
		\bottomrule
	\end{tabular}
}
\end{table}

\begin{figure}[htp!]
	\centering
	\renewcommand{\arraystretch}{0.6}
	\setlength{\tabcolsep}{1pt}
	\begin{tabular}{c c c c}
		RGB Image & GT & Phase 3 & Phase 4\\
		\includegraphics[width=0.24\linewidth, angle=180, origin=c]{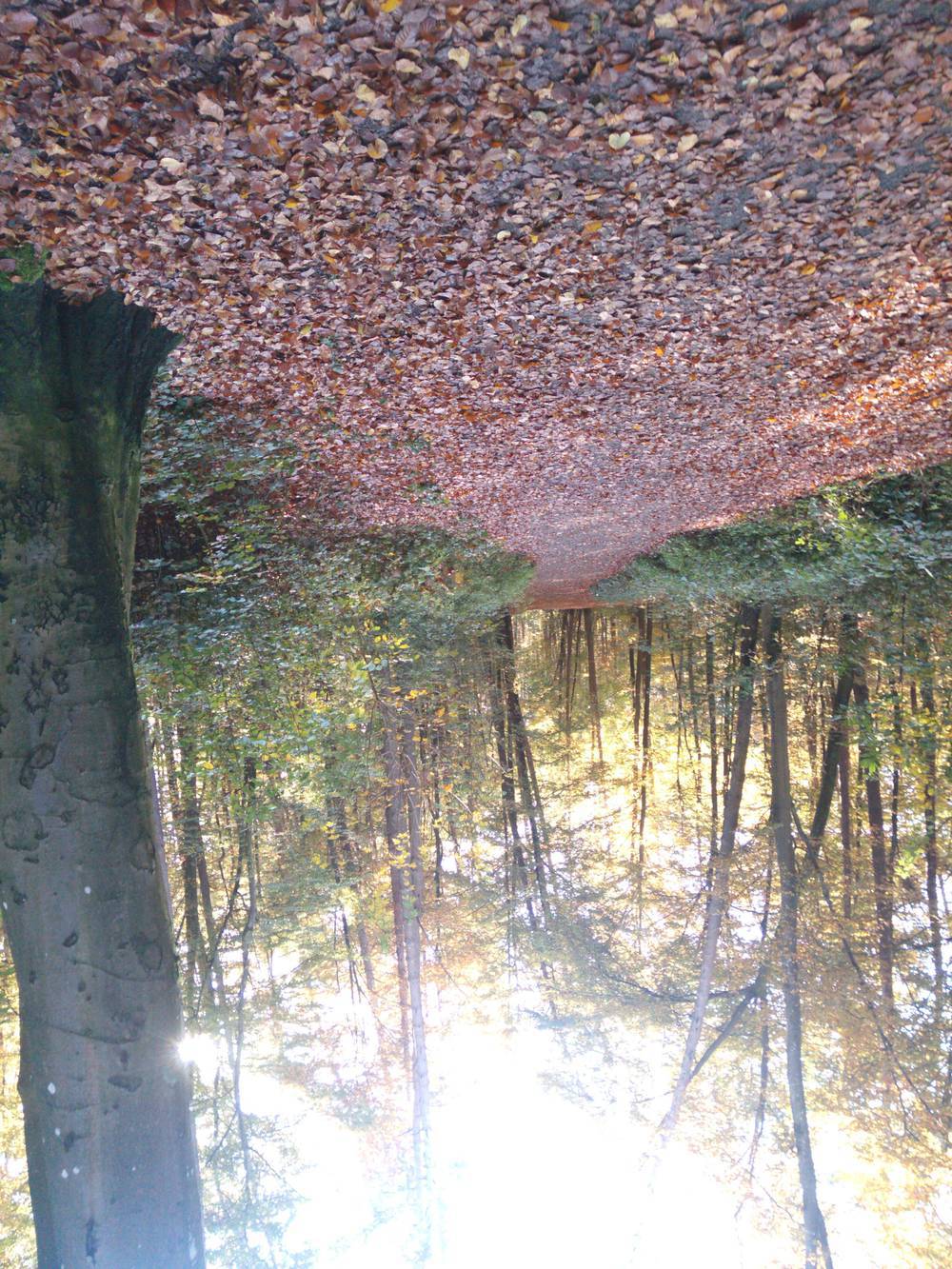} &
		\includegraphics[width=0.24\linewidth]{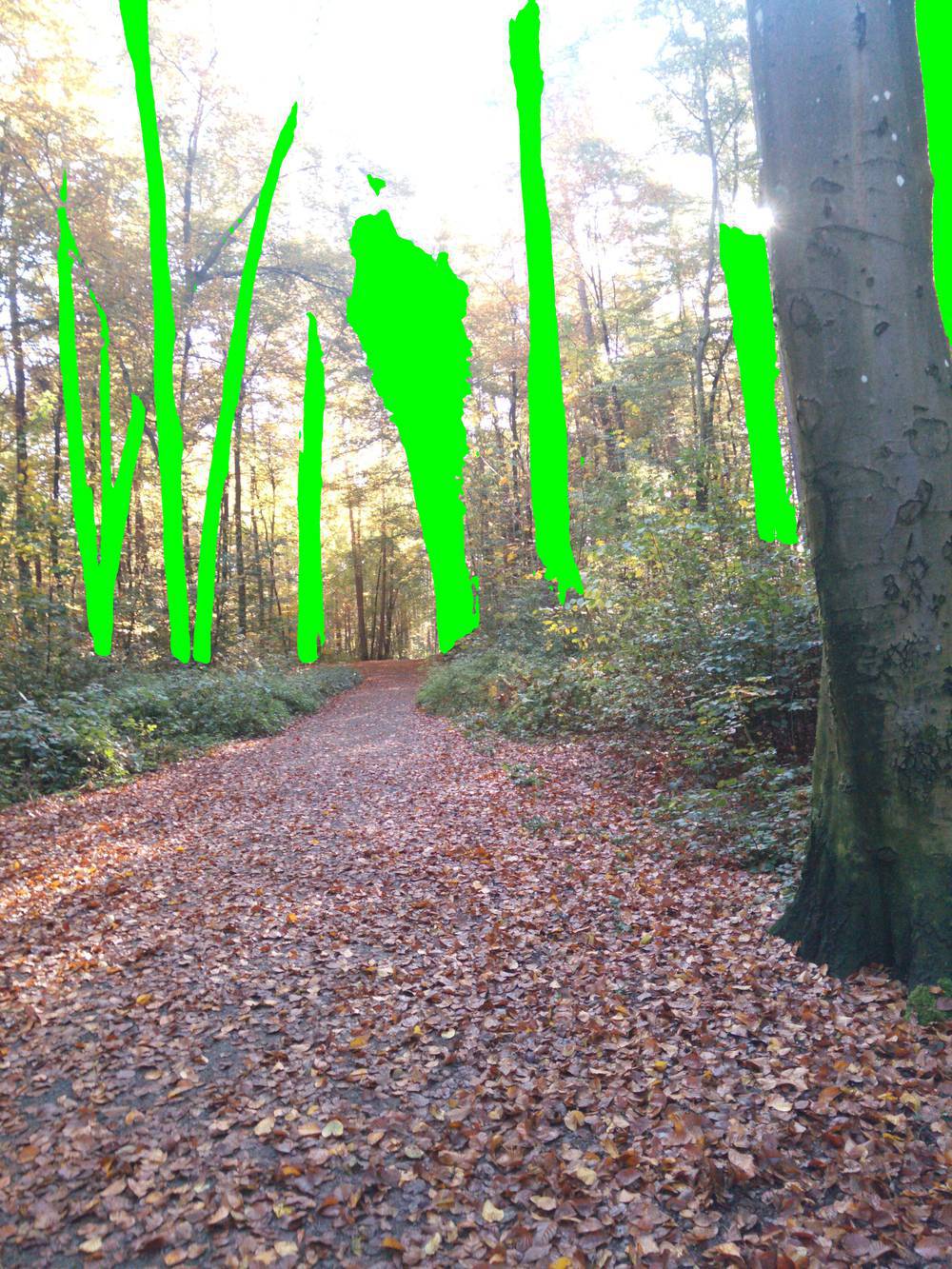} &
		\includegraphics[width=0.24\linewidth]{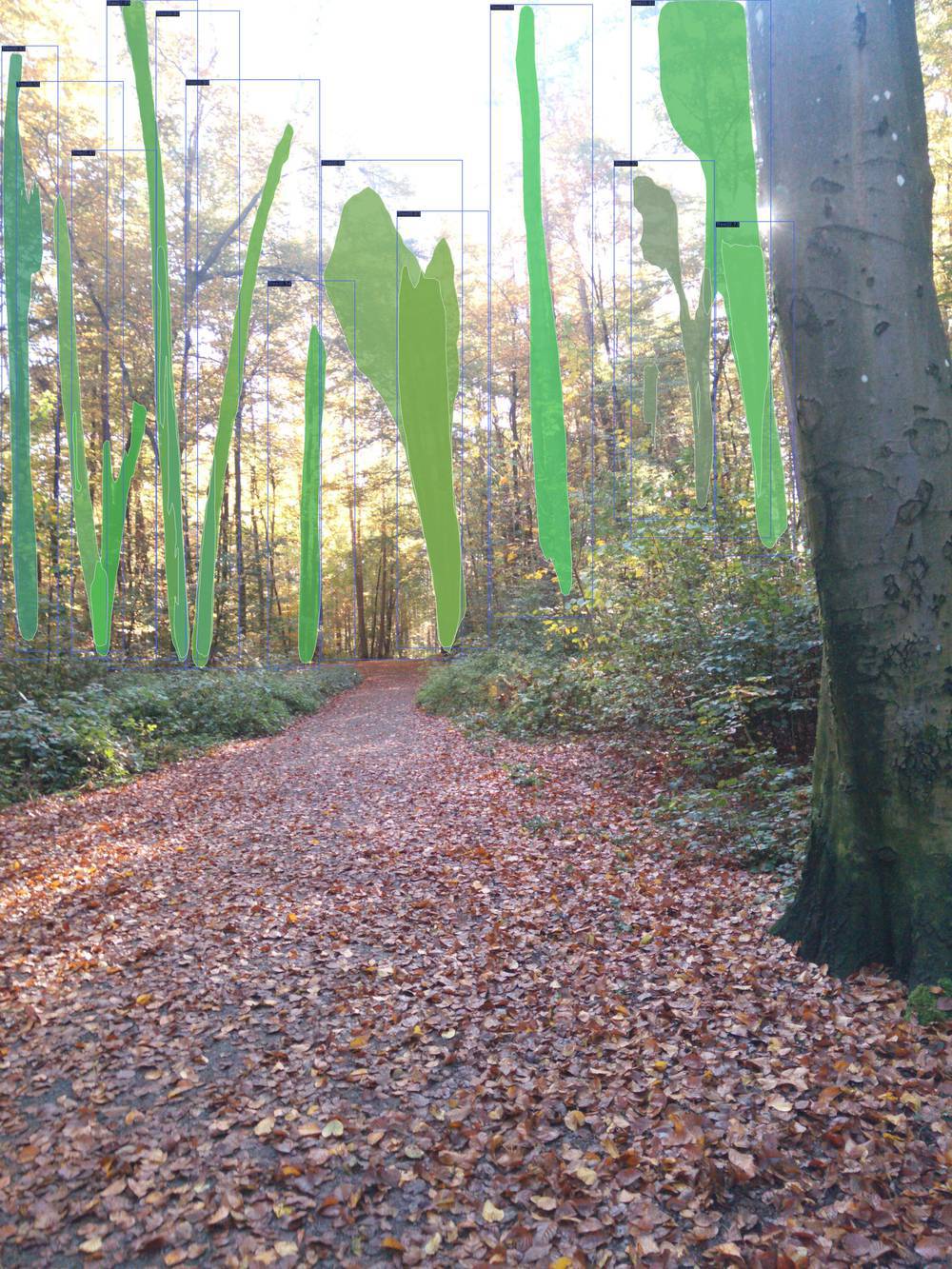} &
		\includegraphics[width=0.24\linewidth]{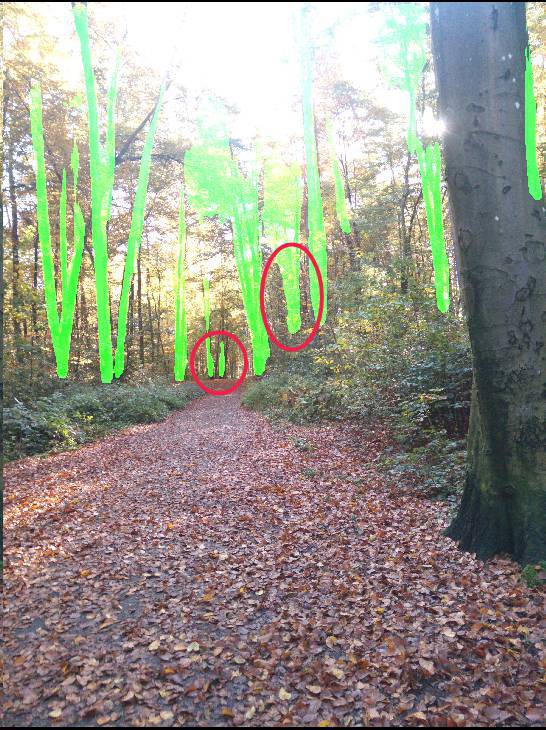}\\
		\includegraphics[width=0.24\linewidth, angle=180, origin=c]{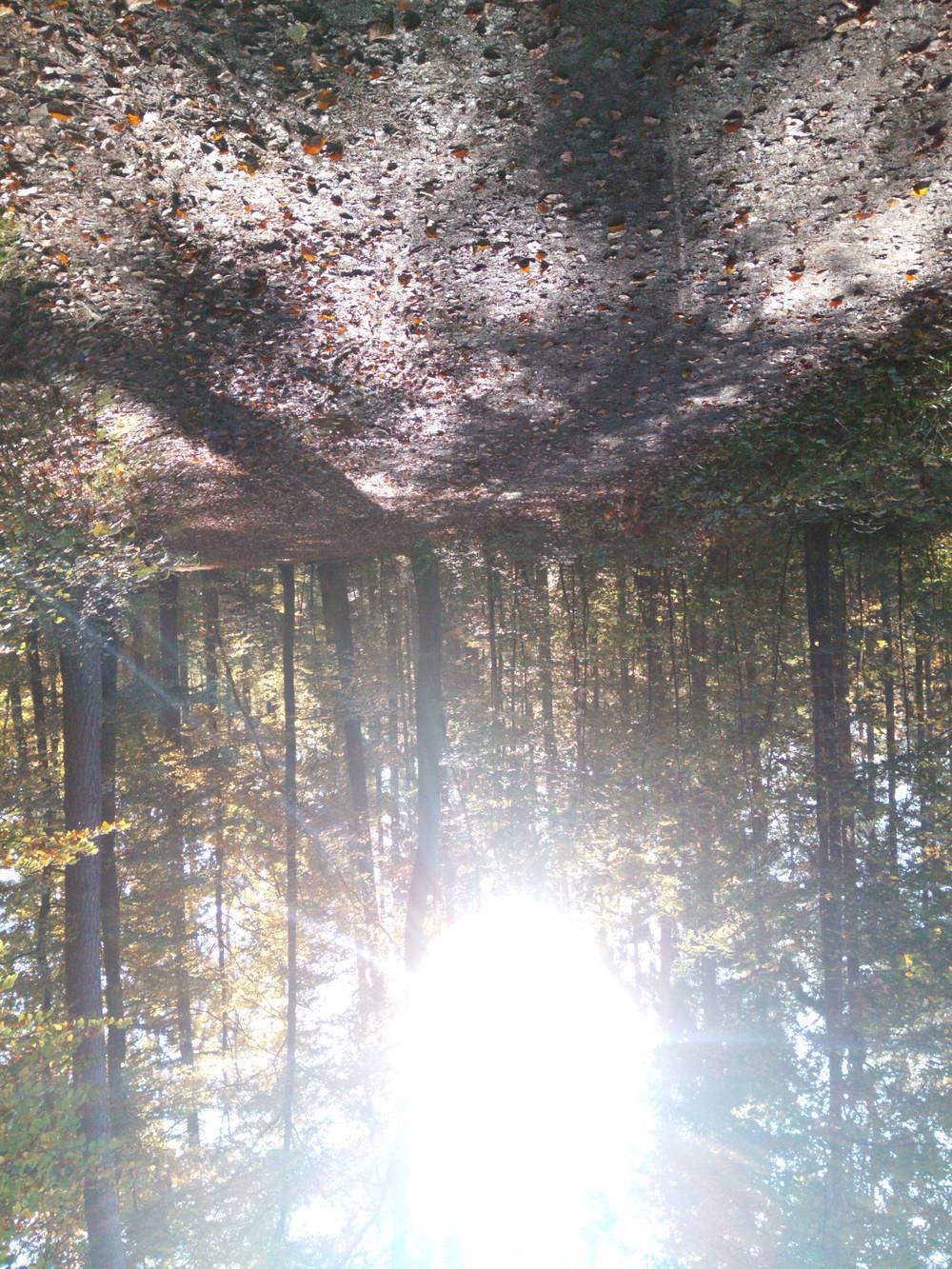} &
		\includegraphics[width=0.24\linewidth]{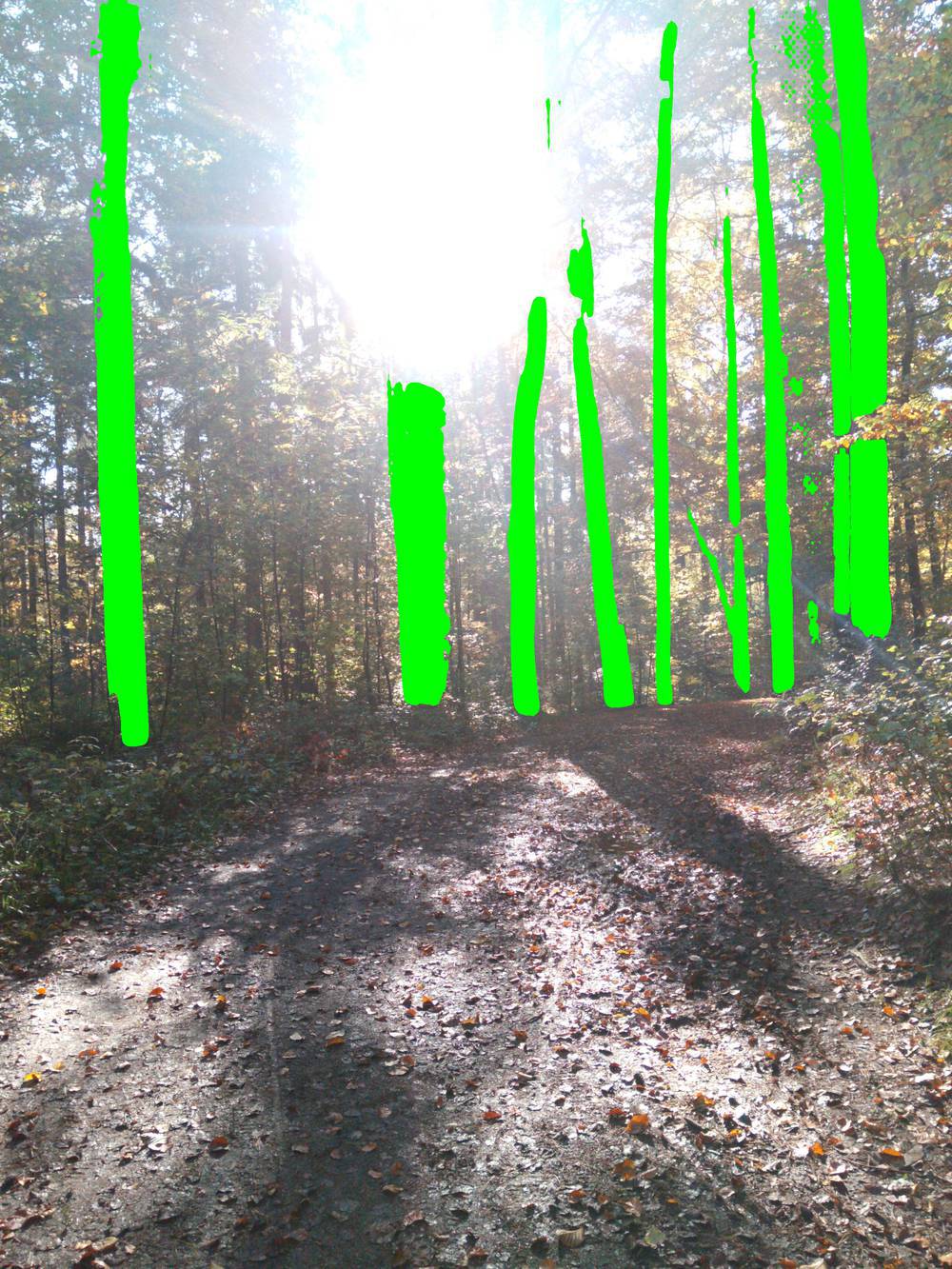} &
		\includegraphics[width=0.24\linewidth]{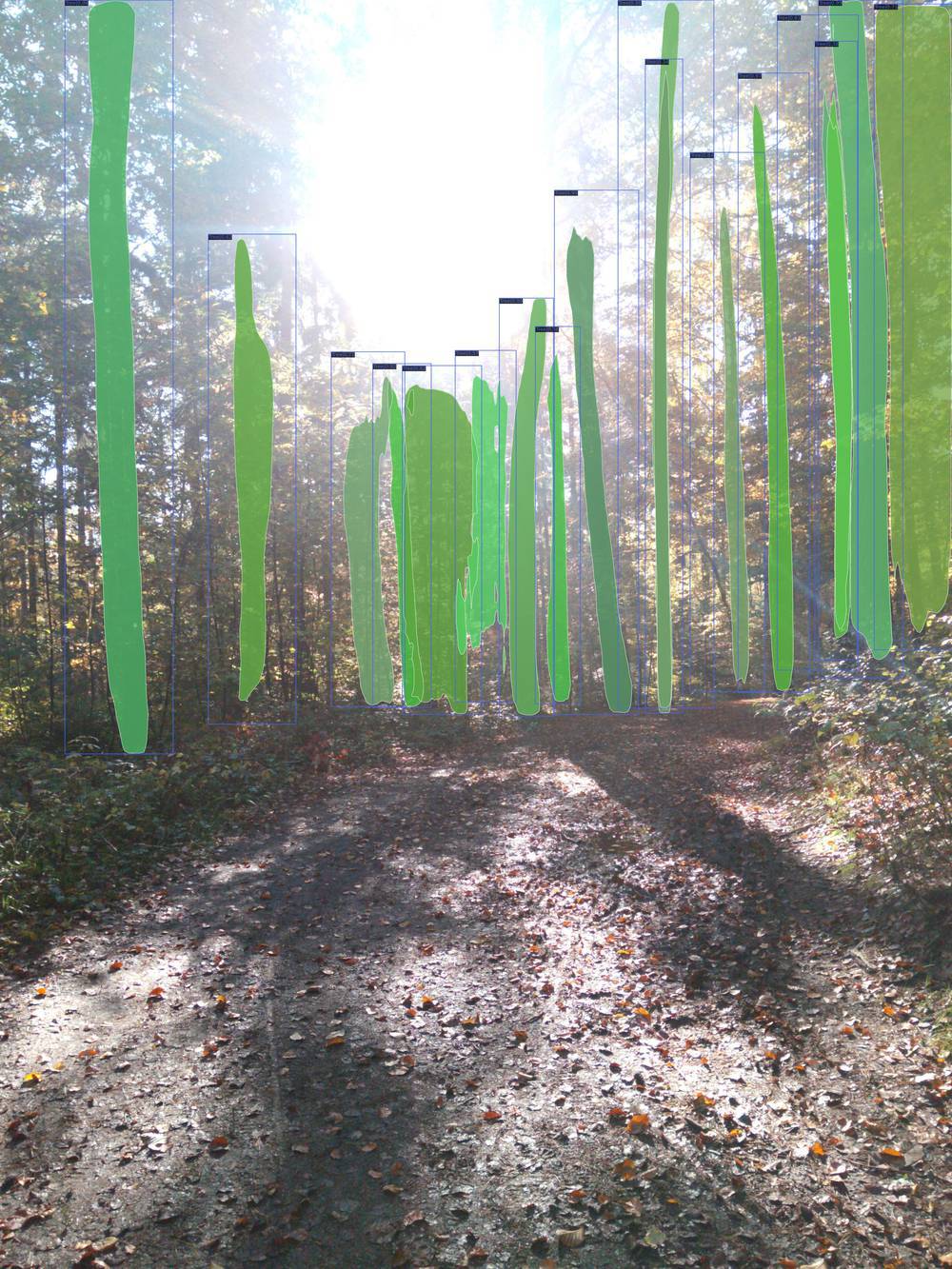} &
		\includegraphics[width=0.24\linewidth]{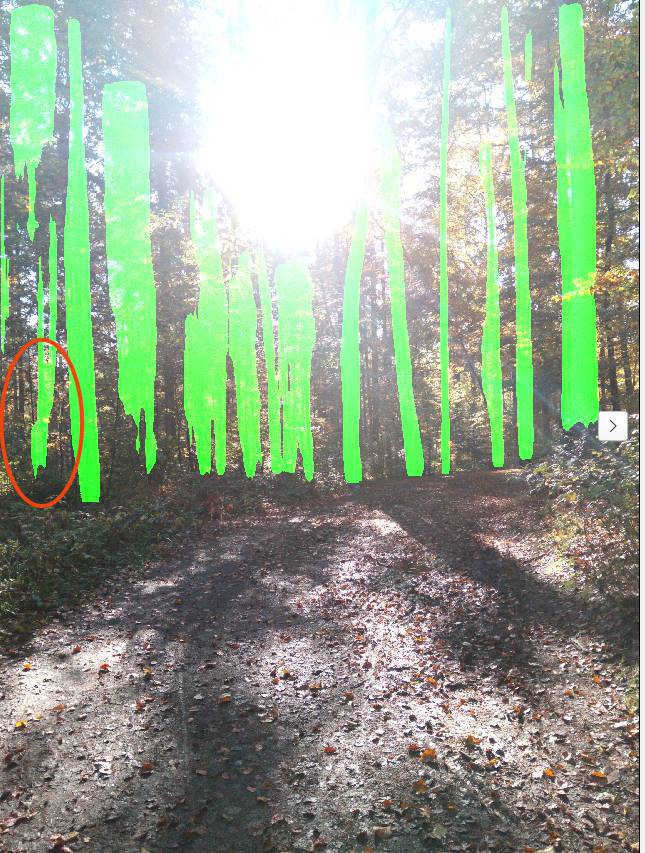}\\
		\includegraphics[width=0.24\linewidth, angle=180, origin=c]{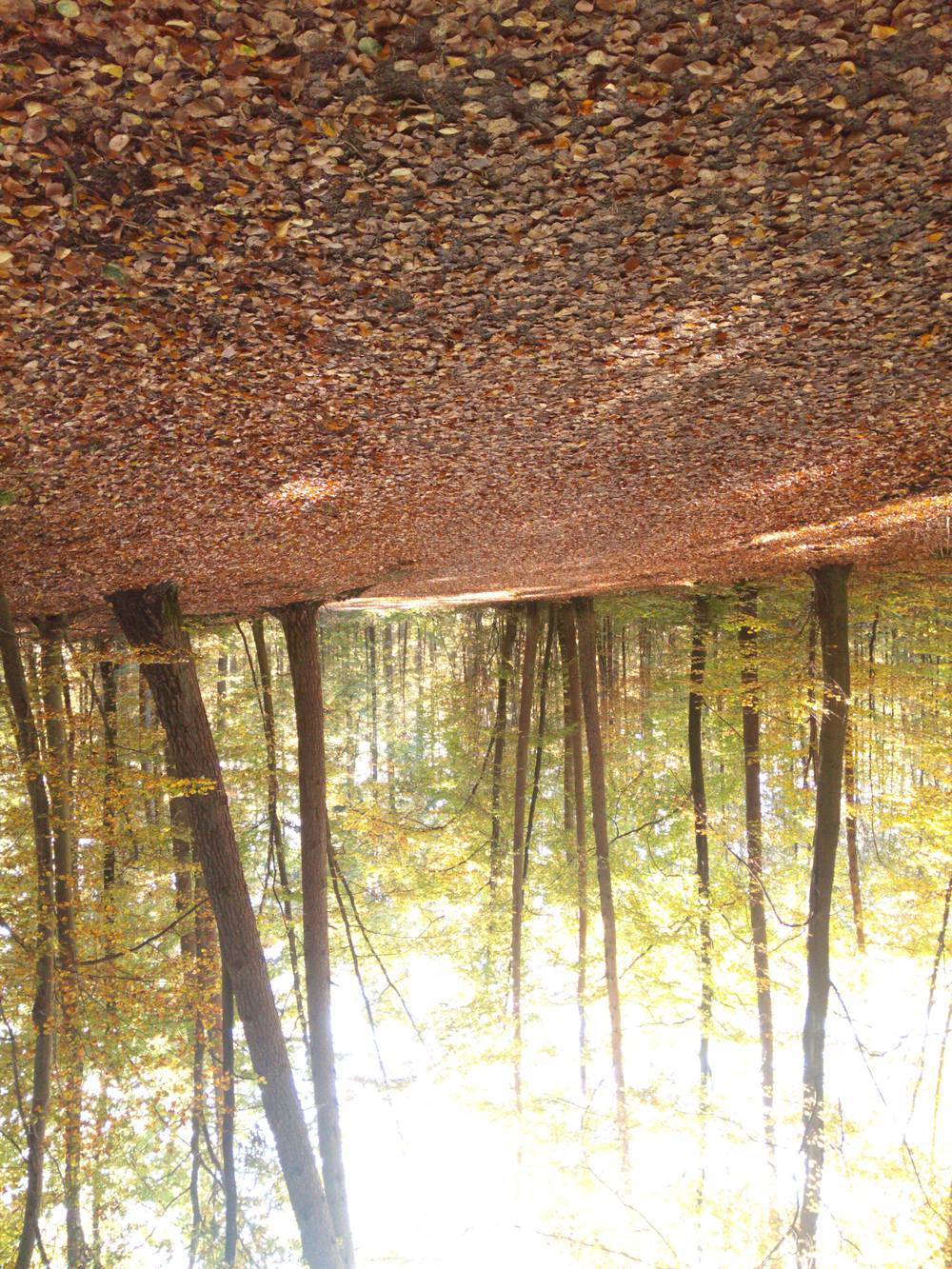} &
		\includegraphics[width=0.24\linewidth]{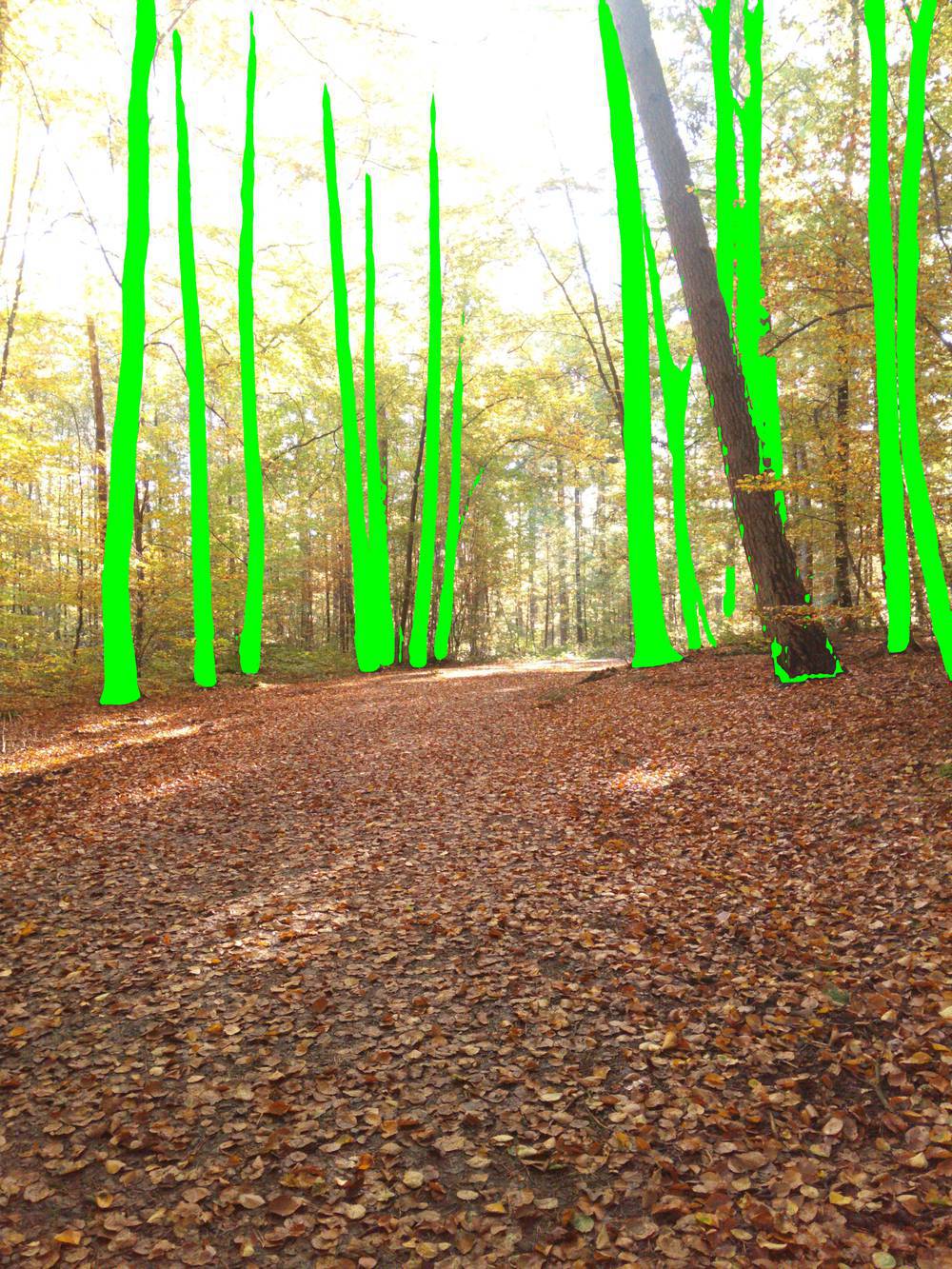} &
		\includegraphics[width=0.24\linewidth]{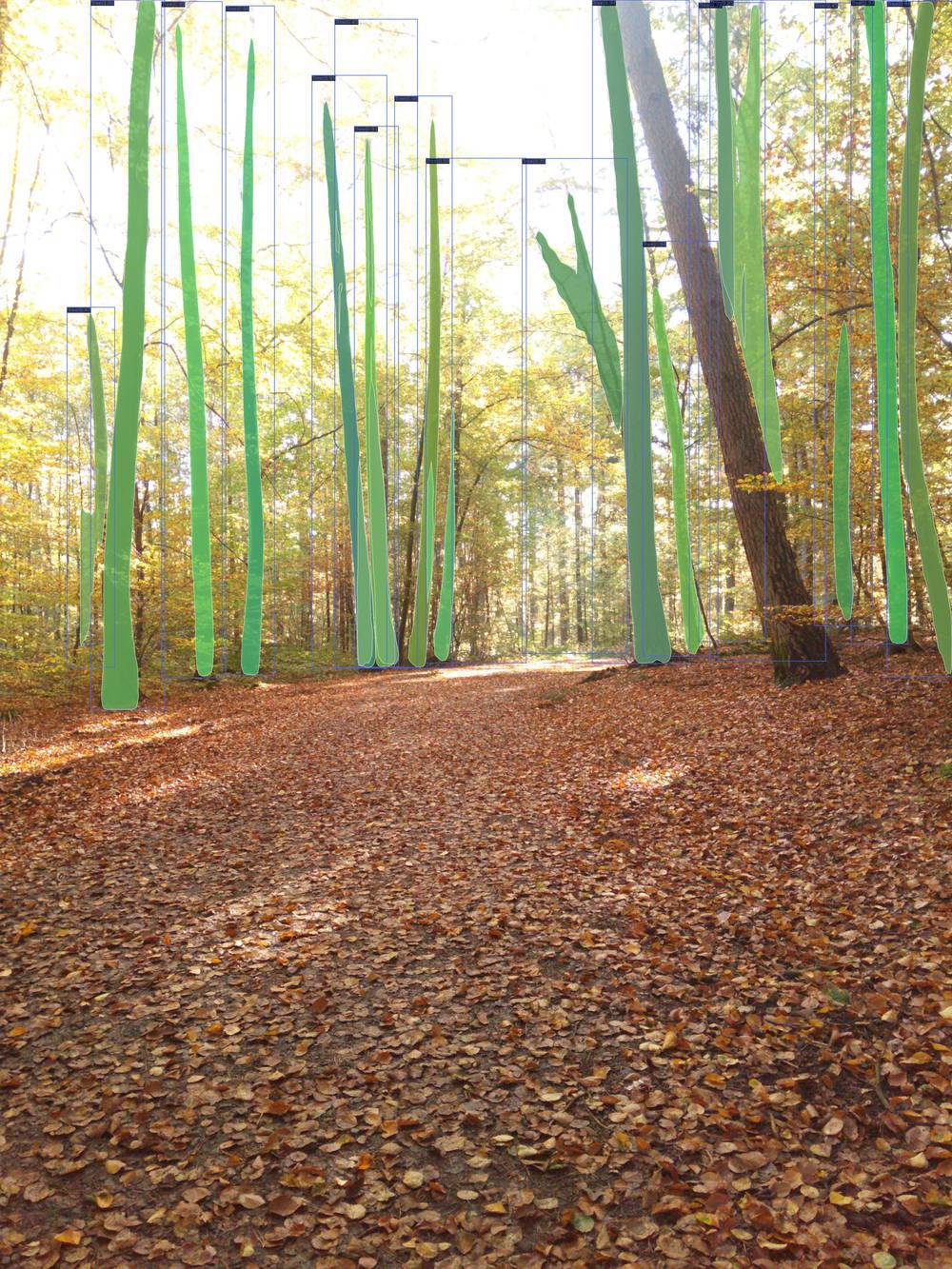} &
		\includegraphics[width=0.24\linewidth]{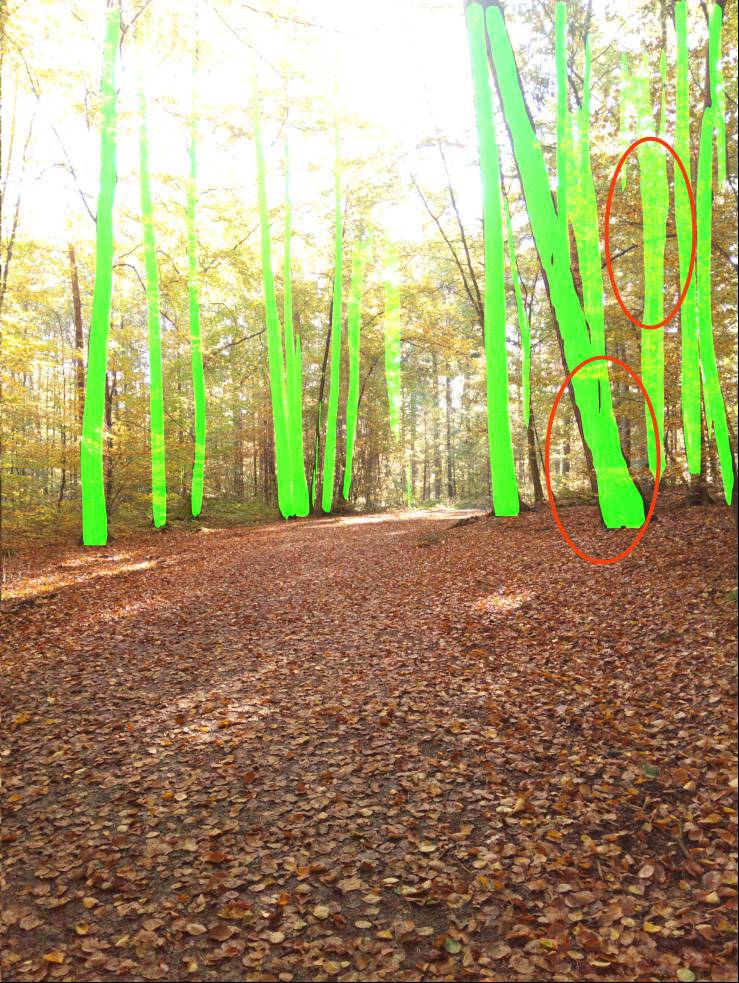}\\
		\includegraphics[width=0.24\linewidth, angle=180, origin=c]{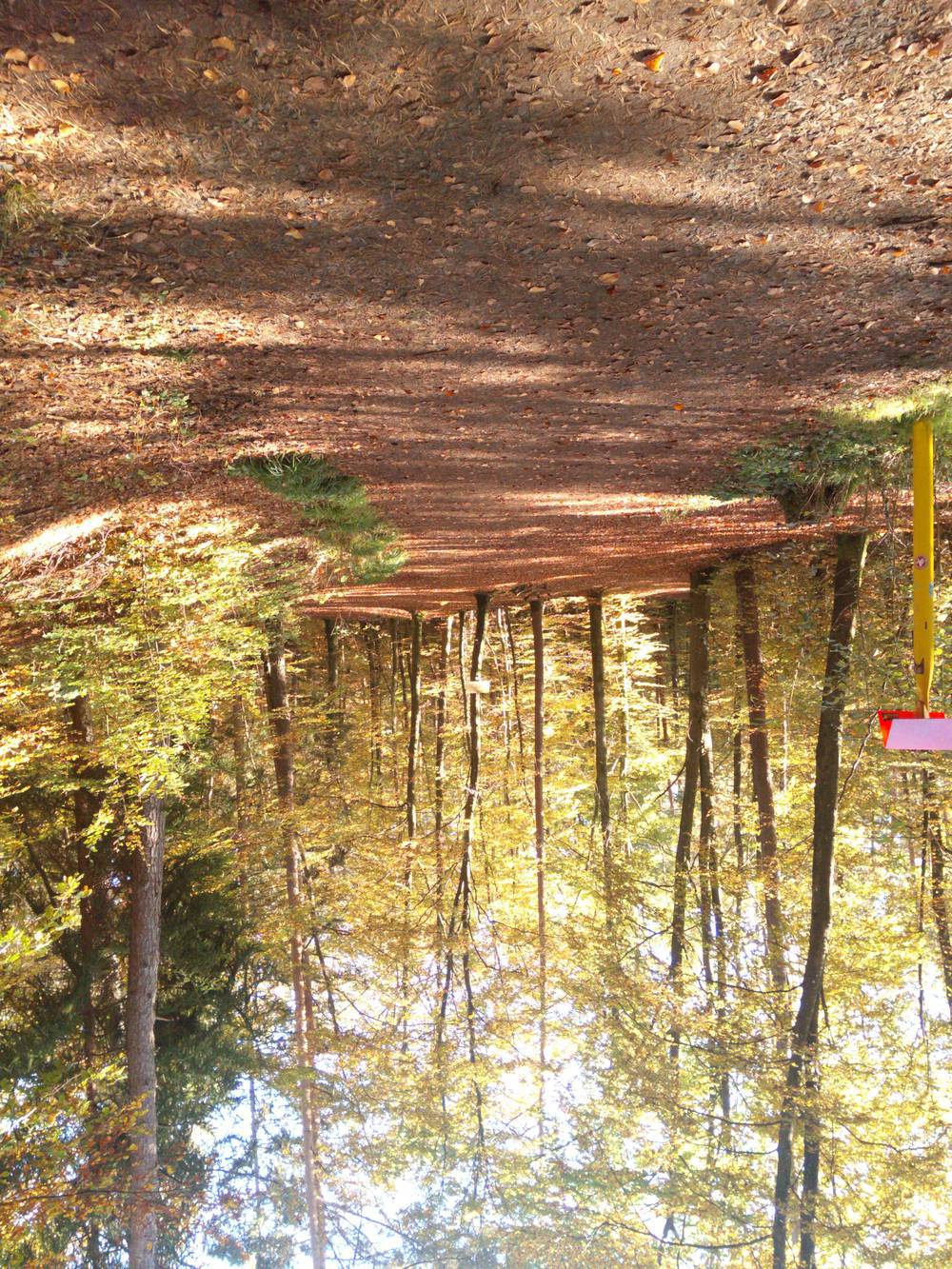} &
		\includegraphics[width=0.24\linewidth]{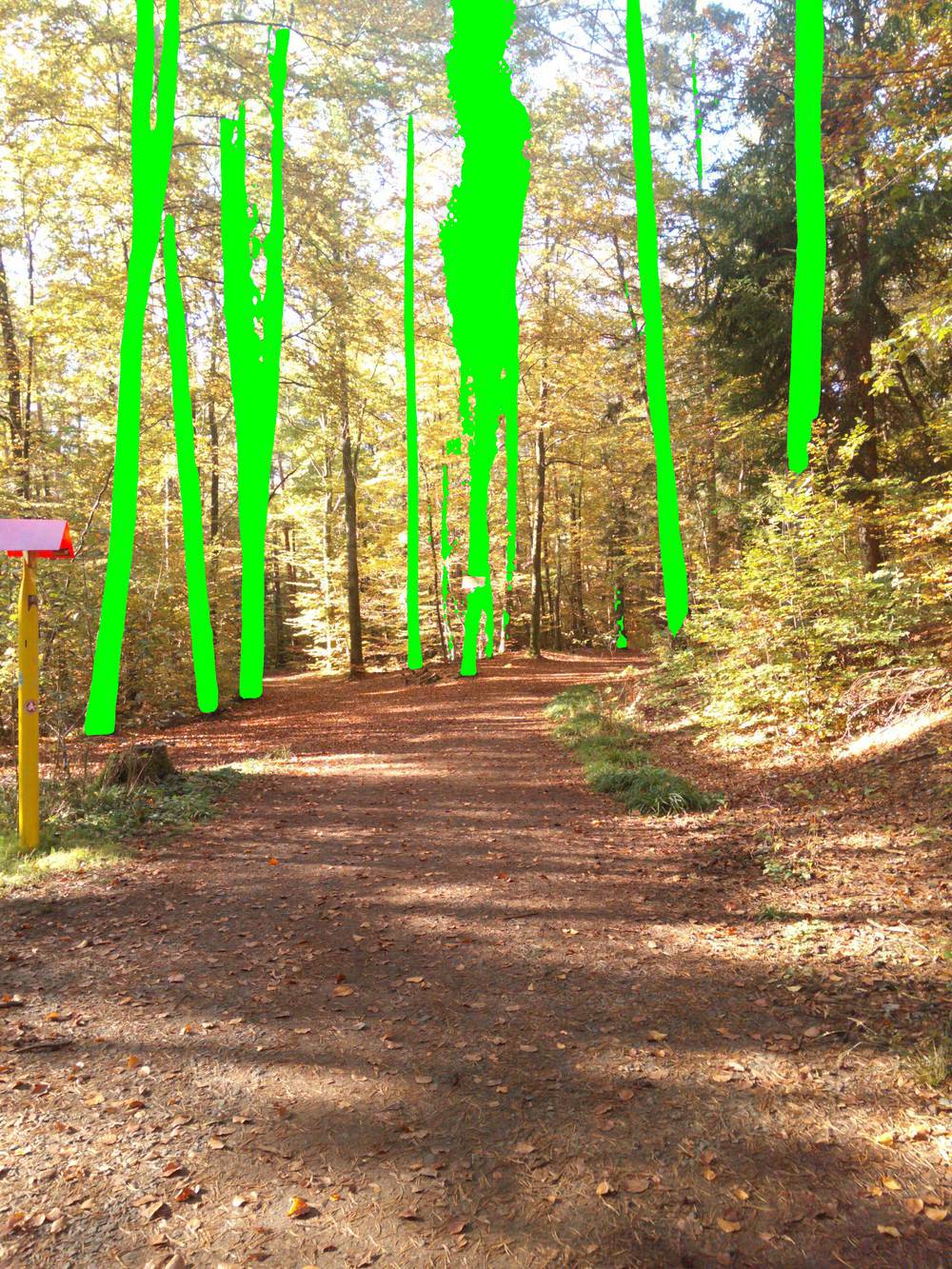} &
		\includegraphics[width=0.24\linewidth]{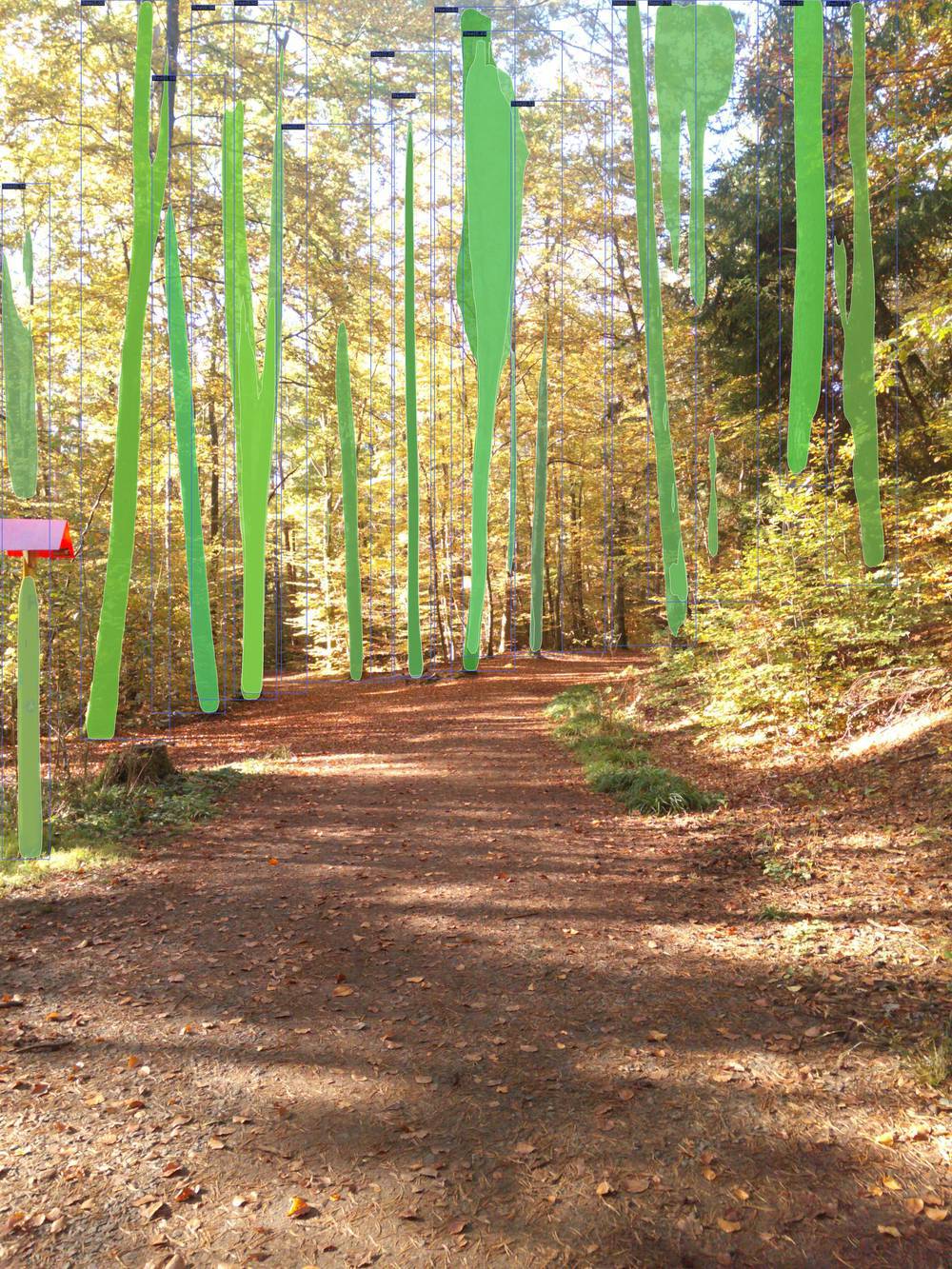} &
		\includegraphics[width=0.24\linewidth]{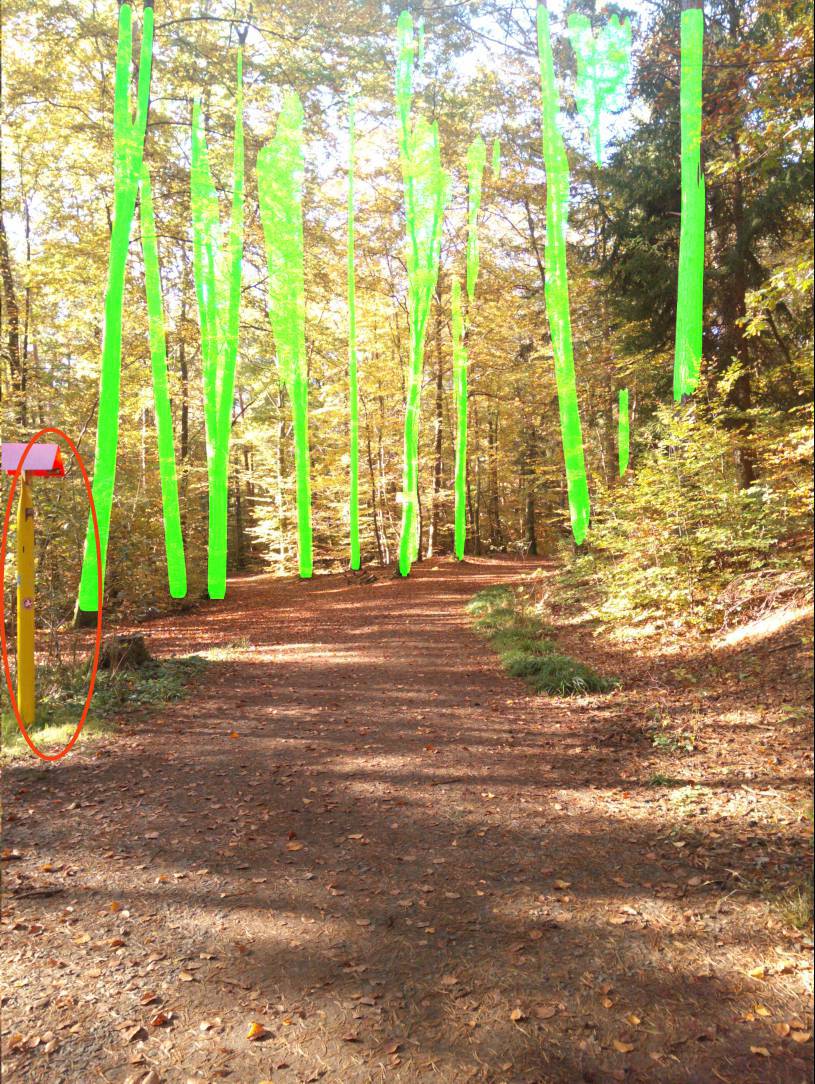}\\
	\end{tabular}
	\caption{Qualitative comparison of real-only training (Phase~3) and
		granularity-aware distillation (Phase~4). Phase~4 recovers additional thin and
		distant tree trunks and exhibits stronger vertical consistency, while Phase~3
		often misses such instances. We also see an improvement in distinguishing capabilities (row 4 for e.g.).}
	\label{fig:phase4_qualitative}
\end{figure}

\paragraph{Phase 4: Granularity-aware distillation.}

Fig.~\ref{fig:phase4_qualitative} additionally presents a qualitative comparison between
the real-only baseline (Phase~3) and the proposed granularity-aware
distillation (Phase~4). Compared to Phase~3, the distilled student exhibits
consistently higher recall, particularly for thin, distant, and partially
occluded tree trunks that are frequently missed by the real-only model. This
indicates that structural priors learned from simulation are successfully
transferred to the real-domain student.

While Phase~4 predictions occasionally exhibit over-extended masks that
partially fill the predicted regions of interest, these errors represent a
shift from \emph{missed detections} toward \emph{boundary imprecision}. Such
behavior is expected during early distillation, as simulated whole-tree
teachers encourage spatial continuity and vertical consistency. Importantly,
the distilled model recovers substantially more valid tree instances across
challenging illumination conditions and cluttered forest scenes. We also observe improved discrimination 
between actual tree structures and tree-like vertical objects (e.g., poles, 
narrow background elements), as seen in the fourth row of table.~\ref{tab:phase4_vs_phase3}. 
This suggests distillation not only increases recall but also enhances 
category-specific structural understanding.

While standard COCO metrics confirm the quantitative advantage of Phase~4, they under-represent two key qualitative improvements. \textbf{First}, the distilled model consistently recovers \emph{thin, distant, or partially occluded trunks} that are systematically missed by real-only training. \textbf{Second}, it exhibits stronger \emph{structural consistency}, producing masks that better align with vertical tree geometry despite occasional boundary imprecision. In forestry perception, where annotation completeness is inherently limited by occlusion and scale variation, these qualitative gains translate to improved operational reliability for downstream robotics tasks.

Overall, granularity-aware distillation improves robustness and recall under
domain shift. These results validate the
core premise of Phase~4: structural knowledge from simulation complements real
appearance statistics and leads to more reliable forest perception.

\begin{table}[t]
	\centering
	\caption{Phase~4 (Granularity-aware distillation) vs. Phase~3 (Real-only baseline). Despite using a lighter ResNet-50 backbone, distillation surpasses all real-only models.}
	\label{tab:phase4_vs_phase3}
	\setlength{\tabcolsep}{6pt}
	\renewcommand{\arraystretch}{1.1}
	\resizebox{\linewidth}{!}{
	\begin{tabular}{|l|c|c|c|}
		\toprule
		\textbf{Model (Backbone)} & \textbf{Training Regime} & \textbf{mAP$_{50:95}$ (Segm)} & \textbf{Improvement} \\
		\midrule
		Mask R-CNN (ResNet-50)   & Phase 3 (Real-only) & 0.396 & -- \\
		Mask R-CNN (ResNeXt-101) & Phase 3 (Real-only) & 0.363 & -- \\
		Mask R-CNN (Swin-T)      & Phase 3 (Real-only) & 0.420 & -- \\
		\midrule
		Mask R-CNN (ResNet-50)   & Phase 4 (Distillation) & \textbf{0.429} & \textbf{+8.3\%} vs. same backbone \\
		\midrule
		Mask R-CNN (ResNet-50)   & Phase 4 (Distillation) & \textbf{0.429} & \textbf{Outperforms} ResNeXt-101 (0.363) \\
		& & & \textbf{Outperforms} Swin-T (0.420) \\
		\bottomrule
	\end{tabular}
}
\end{table}

\subsection{Discussion}
\label{discussion}

Our four-phase analysis reveals critical insights about Sim$\rightarrow$Real transfer under granularity mismatch. While prior work has primarily addressed appearance differences (e.g., texture, illumination), we demonstrate that \emph{structural information loss} due to coarse labeling is an equally significant barrier to real-world performance.

\paragraph{Granularity matters more than model capacity.}
The most striking finding is that a distilled ResNet-50 student (Phase~4) outperforms real-only models with significantly more powerful backbones (ResNeXt-101, Swin-T). This challenges the prevailing assumption that closing the Sim$\rightarrow$Real gap requires increasingly complex architectures. Instead, we show that \emph{how supervision is structured} - specifically, preserving fine-grained geometric priors from simulation - can be more impactful than raw model capacity.

\paragraph{Two teachers are better than one.}
Our ablation of individual teachers (trunk-only vs. whole-tree-only distillation) reveals complementary strengths (more details in supplementary (section 3.4)). The trunk teacher provides strong vertical priors for nearby navigation, while the whole-tree teacher captures overall silhouette for distant perception. Their combination via logit-space merging yields the best performance, confirming that both granularities convey distinct, valuable structural information.

\paragraph{The domain gap is hierarchical.}
Our results suggest the Sim$\rightarrow$Real challenge decomposes into two components: (1) an \emph{appearance gap} (addressed by training on real images), and (2) a \emph{structural gap} (addressed by distillation). Phase~3 closes only the first, while Phase~4 addresses both. This explains why distillation achieves superior recall, particularly for challenging instances where structural cues (e.g., trunk linearity) are critical but poorly captured in coarse real annotations.

\paragraph{Limitations and future directions.}
While effective, our approach has limitations. The over-extended masks observed in early distillation epochs indicate that simulated whole-tree annotations can bias toward spatial continuity at the expense of boundary precision. Future work could explore boundary-aware distillation losses or adaptive weighting between teachers. Additionally, while our logit-space merging handles hierarchical relationships, more sophisticated merging strategies (e.g., attention-based fusion) could further optimize granularity alignment.

\paragraph{Broader implications.}
Our framework demonstrates that synthetic data's greatest value may not be in augmenting training samples, but in providing \emph{structural supervision} unavailable in real annotations. This insight extends beyond forestry to domains like medical imaging (where simulation provides organ substructure labels), agriculture (plant component segmentation), and industrial inspection (part defect hierarchies). By treating granularity mismatch as a first-class problem rather than collapsing labels, we enable more effective knowledge transfer across domains.

\section{Conclusion}
\label{conclusion}

We have presented the first systematic study of Sim$\rightarrow$Real instance segmentation under label granularity mismatch, a common but under-explored challenge in practical perception systems. Our contributions are threefold:

First, we introduced \textbf{MGTD}, a mixed-granularity dataset combining 53k synthetic images with fine-grained trunk/whole-tree annotations and 3.6k real images with only coarse ``Tree'' labels. This benchmark enables controlled analysis of domain shift and granularity effects.

Second, we proposed a \textbf{four-stage evaluation protocol} that isolates key factors: synthetic upper-bound performance, zero-shot domain transfer, real-only baseline training, and granularity-aware distillation. This protocol provides a structured framework for future research in mixed-granularity transfer.

Third, we developed \textbf{granularity-aware distillation}, which transfers structural priors from fine-grained synthetic teachers to a coarse-label student via logit-space merging and mask unification. Experiments demonstrate that this approach not only improves mask AP over real-only baselines but, remarkably, enables a lightweight ResNet-50 student to outperform significantly more powerful architectures trained solely on real data.

Our work reframes Sim$\rightarrow$Real transfer as a \emph{granularity alignment problem}, highlighting that structural knowledge from simulation can compensate for missing real-world annotations. This insight offers a practical path toward deploying robust perception systems in domains where fine-grained labeling is infeasible at scale.


\title{Supplementary Material for \\ Granularity-Aware Transfer for Tree Instance Segmentation in Synthetic and Real Forests.}
%
%
\institute{Robotics Research Lab, University of Kaiserslautern-Landau, Kaiserslautern, Germany \\
	\url{https://rrlab.cs.rptu.de/en} \and
	University of Kaiserslautern-Landau, Kaiserslautern, Germany}
\maketitle

\section{Dataset insights}
\label{dataset_insights}

\subsection{Simulated data}
\paragraph{Tree trunks.}
Across the four dataset parts (we denote dawn, dusk, noon and snow as 1,2,3,4 respectively), the tree trunk category provides dense supervision with an average of 10-13 instances per image. Part 1 contains the largest split with over 158k annotated trunks, while Parts 3 and 4 show slightly higher per-image densities ($\sim$11-13) than Parts 1 and 2 ($\sim$10-11). The train/val/test splits remain balanced, preserving consistent object statistics across evaluation stages. In total, the tree trunk subsets comprise more than 48k images and over 554k annotated trunk instances, making this category particularly well suited for studying small, repetitive object detection in cluttered forest scenes.

\paragraph{Whole trees.}
In contrast, the whole tree category captures coarser structural information with only 3-8 instances per image on average. Part 1 provides the most sparse distribution ($\sim$3 objects per image), while Parts 3 and 4 reflect denser forest compositions with up to $\sim$7-8 trees per frame. Despite lower densities, the whole tree splits still encompass more than 46k images and 252k annotated instances in total, offering complementary supervision at a broader structural scale. These annotations highlight larger context and global tree structure, making them suitable for evaluating detectors under scale variation and long-range occlusions.
The precise constitution of the dataset can be further seen with the table:\ref{tab:dataset_details}

\begin{table}[htp!]
	\centering
	\small
	\caption{Dataset statistics for the simulated part. Each entry shows 
		\#images / \#objects / average objects per image (I/O/A). 
		Subsets 1--4 correspond to dawn, dusk, noon, and snow respectively.}
	\setlength{\tabcolsep}{3pt}
	\renewcommand{\arraystretch}{1.15}
	
	\resizebox{\textwidth}{!}{
		\begin{tabular}{|l|l|l|l|l|}
			\toprule
			\textbf{Subset} & \textbf{Train (I/O/A)} & \textbf{Val (I/O/A)} 
			& \textbf{Test (I/O/A)} & \textbf{Total (I/O)} \\
			\midrule
			1 / Tree Trunk   & 10036 / 111068 / 11.07 & 2867 / 31832 / 11.10 & 1434 / 15632 / 10.90 & 14337 / 158532 \\
			1 / Whole Tree   & 8921 / 29282 / 3.28    & 2549 / 8338 / 3.27   & 1274 / 4184 / 3.28    & 12744 / 41704 \\
			2 / Tree Trunk   & 9712 / 101677 / 10.47  & 2775 / 29301 / 10.56 & 1388 / 14474 / 10.43  & 13875 / 145452 \\
			2 / Whole Tree   & 9400 / 47411 / 5.04    & 2686 / 13714 / 5.11  & 1343 / 6811 / 5.07    & 13429 / 67936 \\
			3 / Tree Trunk   & 7447 / 84651 / 11.37   & 2128 / 23852 / 11.21 & 1064 / 12314 / 11.57  & 10639 / 120817 \\
			3 / Whole Tree   & 7337 / 45580 / 6.21    & 2096 / 12606 / 6.01  & 1049 / 6486 / 6.18    & 10482 / 64672 \\
			4 / Tree Trunk   & 7080 / 90864 / 12.83   & 2023 / 25731 / 12.72 & 1011 / 12823 / 12.68  & 10114 / 129418 \\
			4 / Whole Tree   & 7024 / 55033 / 7.83    & 2007 / 15633 / 7.79  & 1003 / 7755 / 7.73    & 10034 / 78421 \\
			\bottomrule
		\end{tabular}
	}
	
	\label{tab:dataset_details}
\end{table}

\begin{table*}[!htbp]
	\centering
	\small
	\caption{Dataset statistics for the real part of the dataset, divided into two subsets. 
		Each entry shows \#images / \#objects / average objects per image (I/O/A).}
	\renewcommand{\arraystretch}{1.15}
	\setlength{\tabcolsep}{6pt}
	
	\begin{tabular}{|l|c|c|c|c|}
		\toprule
		\textbf{Subset} & \textbf{Train (I/O/A)} & \textbf{Val (I/O/A)} 
		& \textbf{Test (I/O/A)} & \textbf{Total (I/O)} \\
		\midrule
		Part~1 & 1400 / 14065 / 10.05 & 400 / 4064 / 10.16 & 200 / 1974 / 9.87 & 2000 / 20103 \\
		Part~2 & 1120 / 9156 / 8.18   & 320 / 2663 / 8.32  & 160 / 1285 / 8.03 & 1600 / 13104 \\
		\bottomrule
	\end{tabular}

	\label{tab:real_dataset_details}
\end{table*}

\subsection{Real data}
\paragraph{Trees.}
As shown in Table \ref{tab:real_dataset_details}, the real part of the dataset is divided into two parts with a consistent 70/20/10 train/val/test split. The objects are annotated with ``Trees" label rather than ``Tree trunk" or ``Whole tree". Part 1 contains a total of 2000 images with 20,103 annotated objects, averaging about 10 objects per image across all splits. On the other hand, part 2 consists of 1600 images with 13,104 objects, corresponding to an average of about 8 objects per image. We can observe that the object density is highly consistent across train, validation, and test sets within each part, which indicates that the splits are well balanced and representative. Notably, Part 1 exhibits denser forest scenes compared to Part 2, resulting in more annotated objects per image and posing a comparatively greater challenge for detection and segmentation models.

\begin{figure}[htp!]
    \centering
    \setlength{\tabcolsep}{1pt} 
    \renewcommand{\arraystretch}{0.9} 
    \begin{tabular}{ccc}
         \includegraphics[width=0.28\linewidth, angle=180, origin=c]{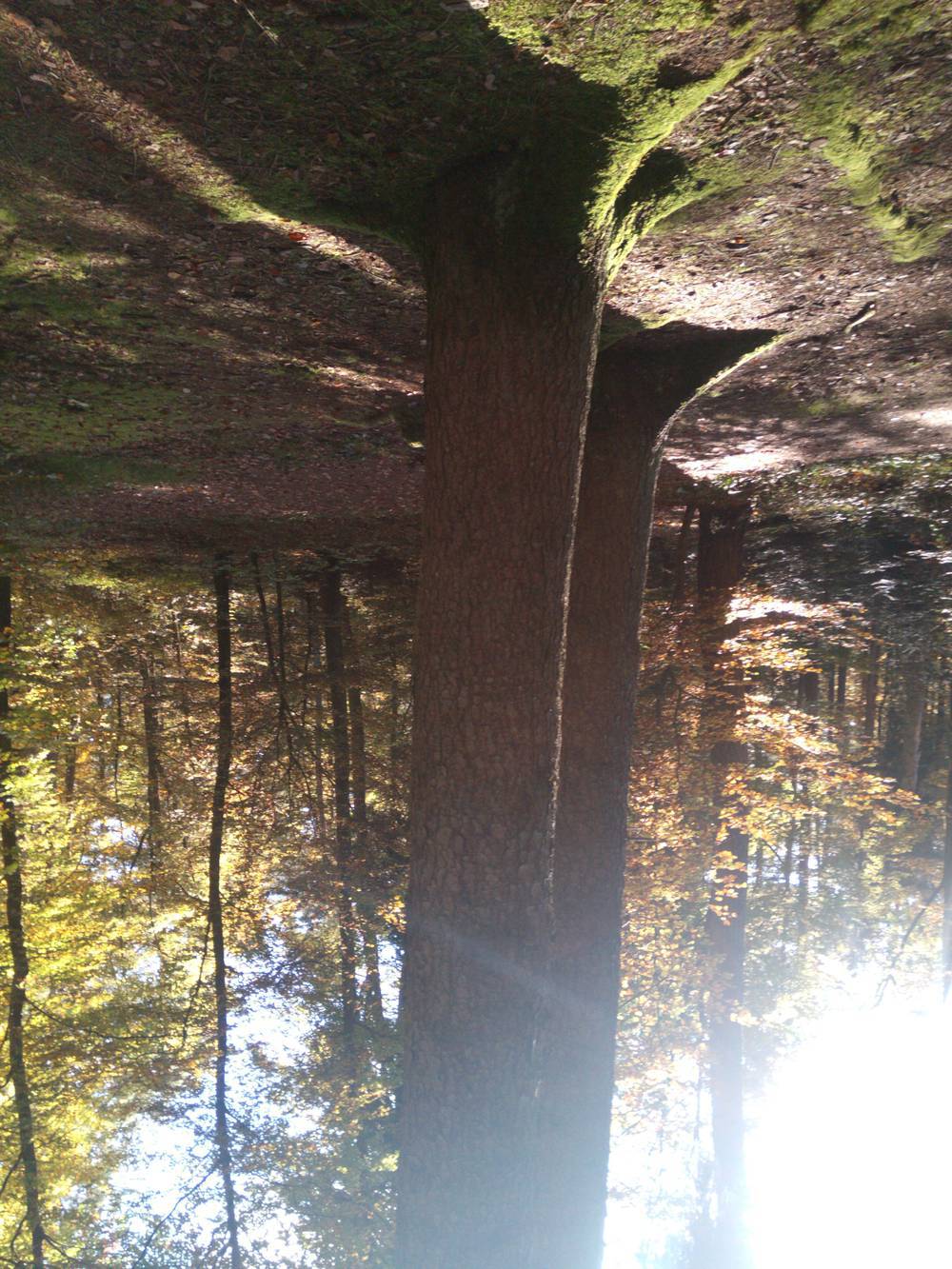}&
         \includegraphics[width=0.28\linewidth, angle=180, origin=c]{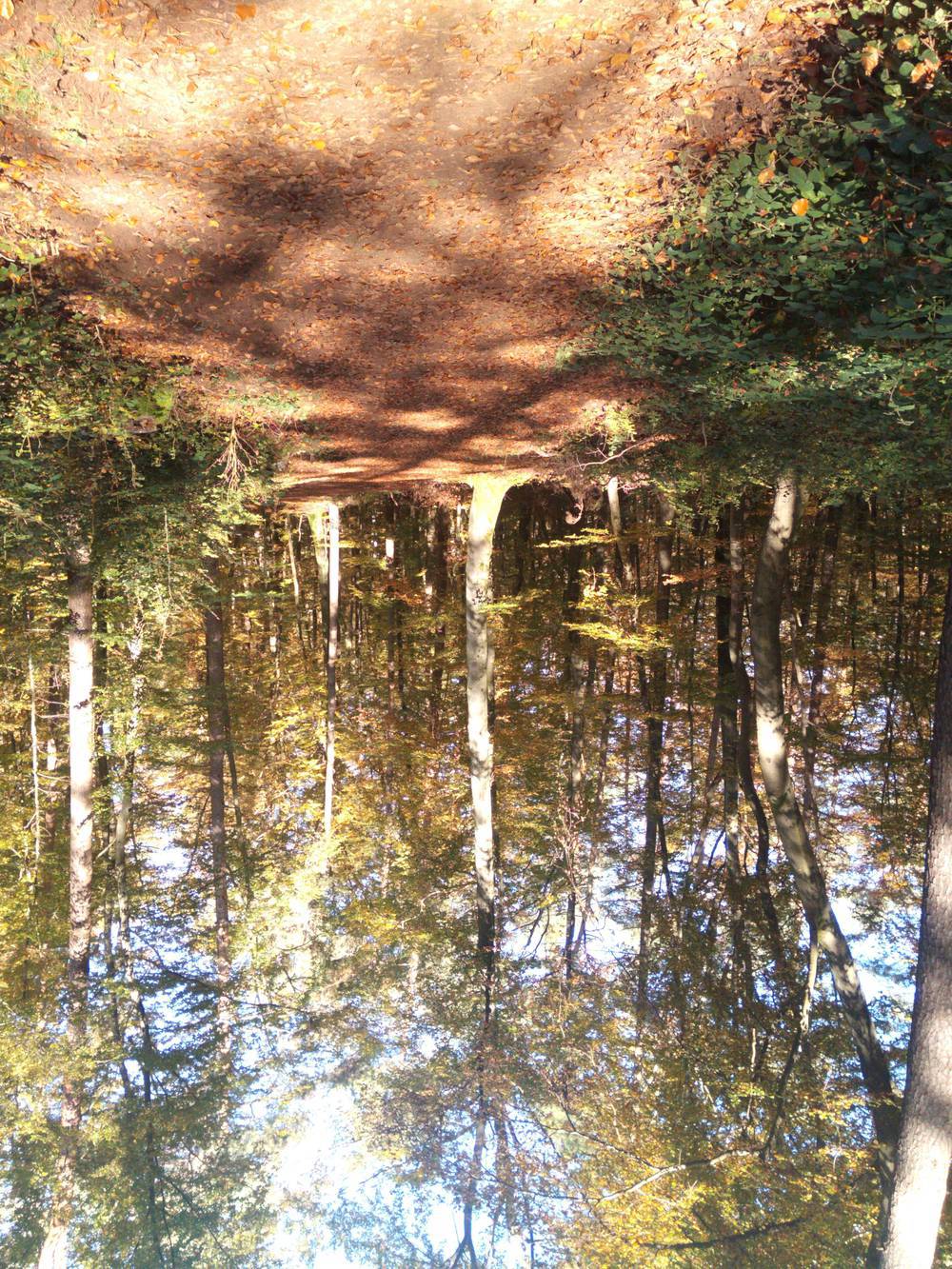}&
         \includegraphics[width=0.28\linewidth, angle=180, origin=c]{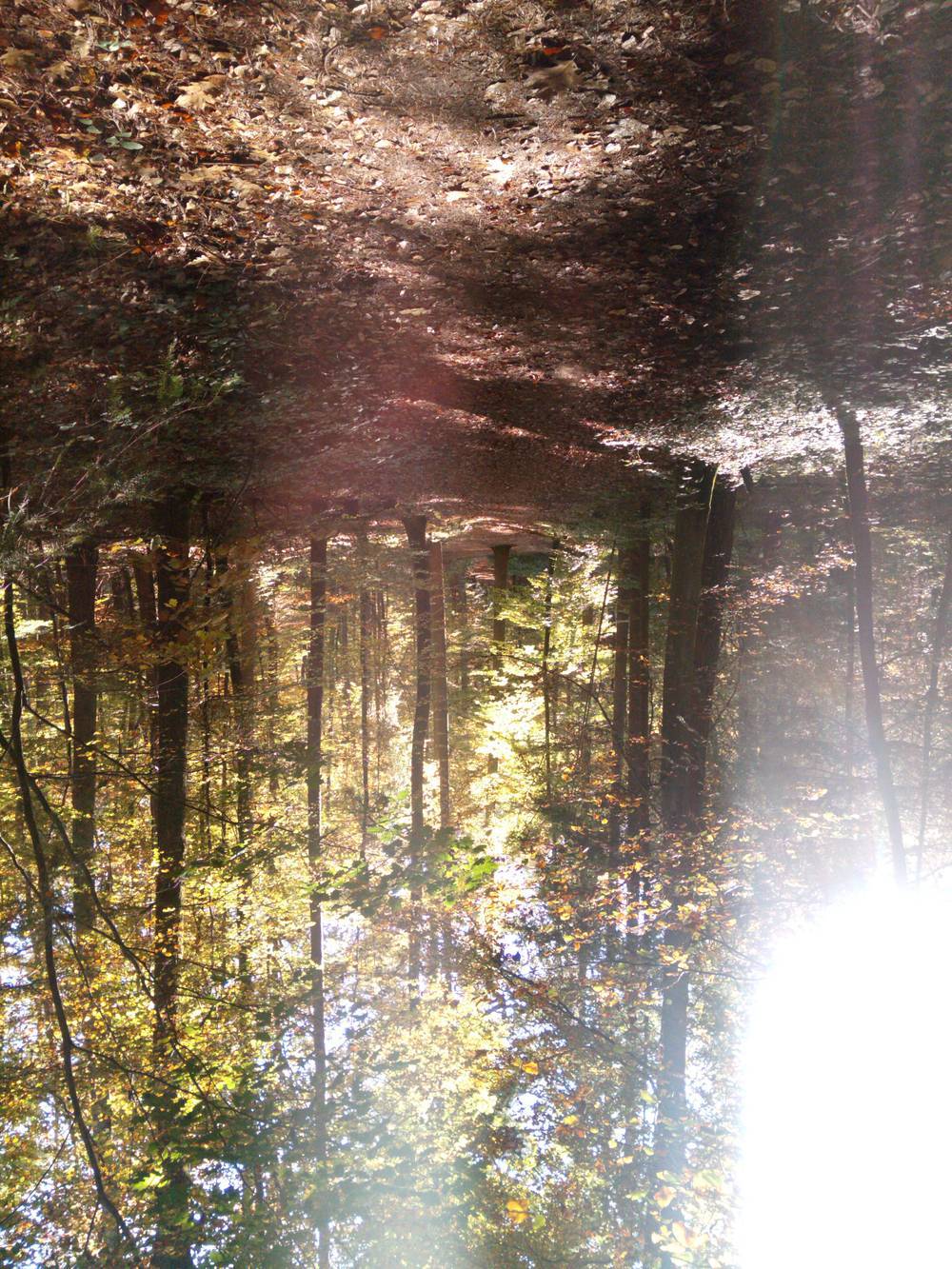}\\
         \includegraphics[width=0.28\linewidth, angle=180, origin=c]{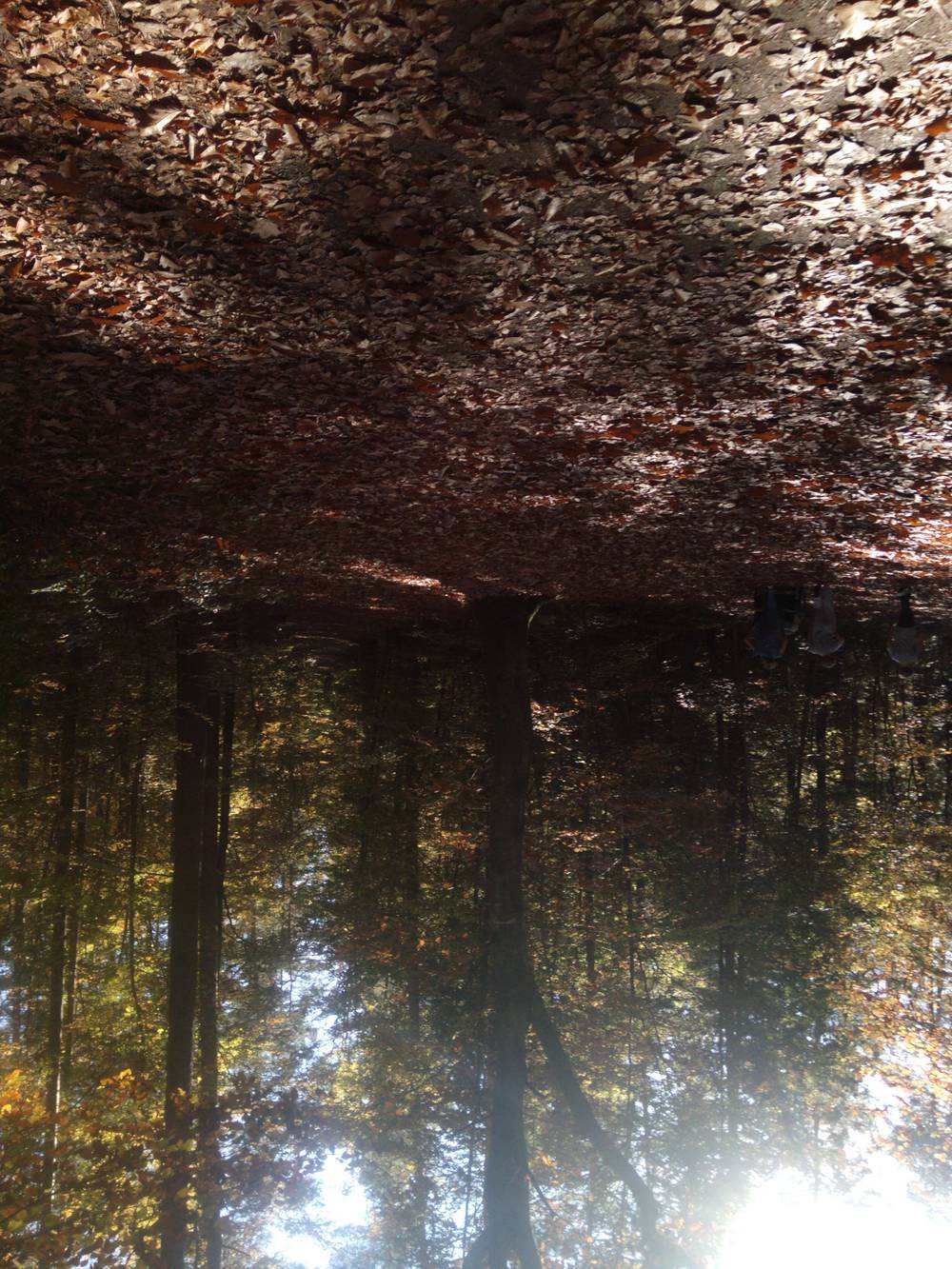}&
         \includegraphics[width=0.28\linewidth, angle=180, origin=c]{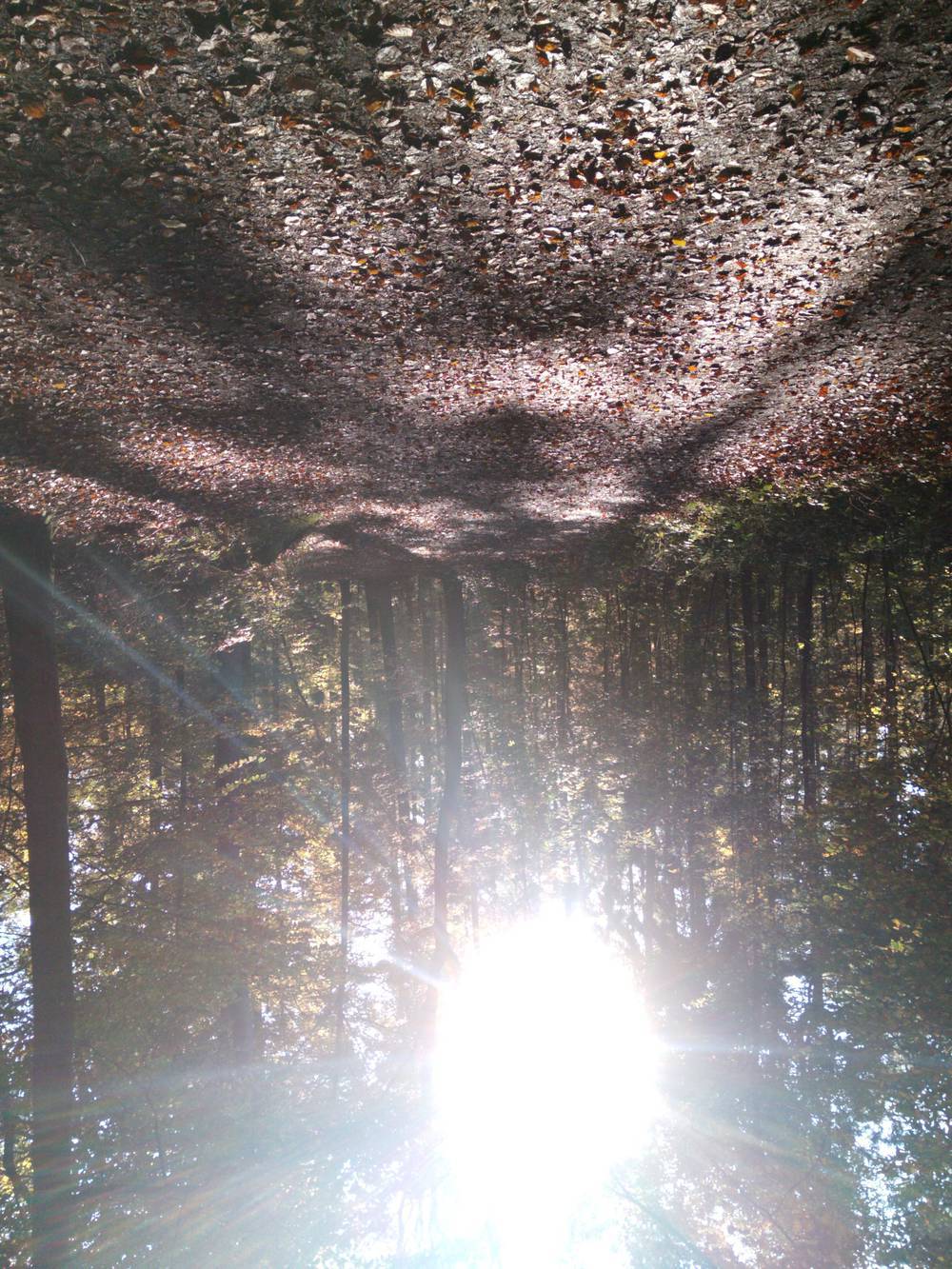}&
         \includegraphics[width=0.28\linewidth, angle=180, origin=c]{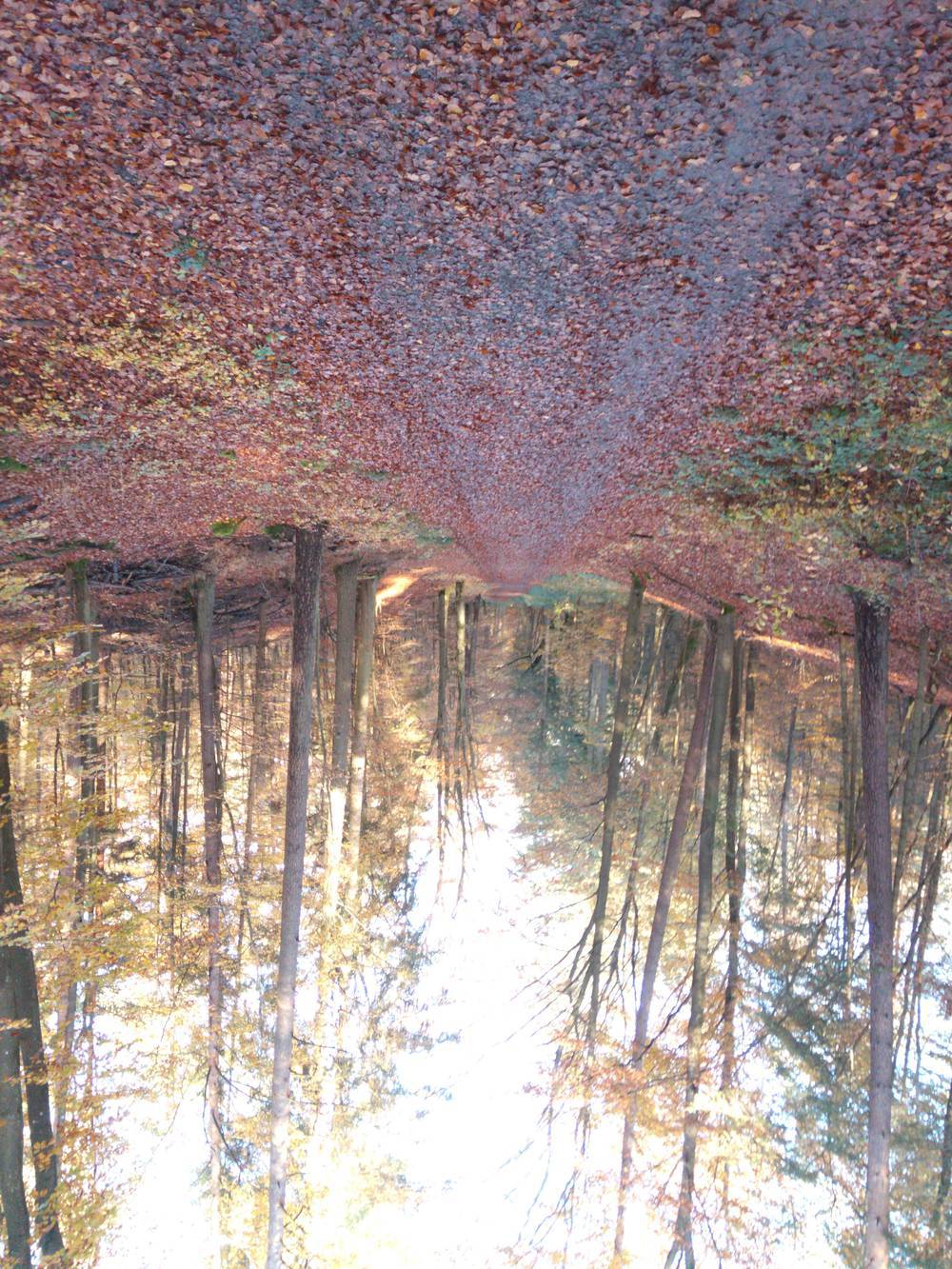}\\
         \includegraphics[width=0.28\linewidth]{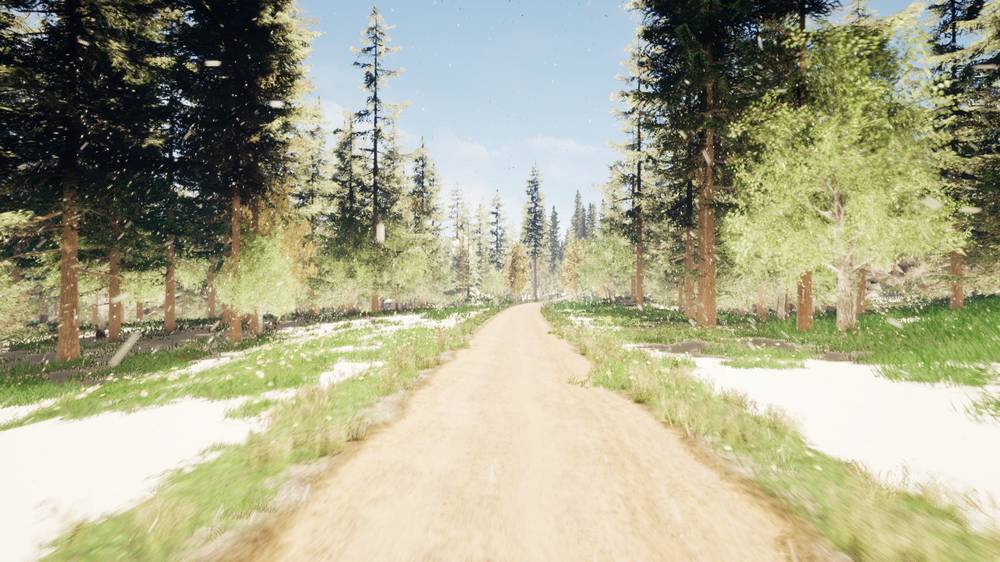}&
         \includegraphics[width=0.28\linewidth]{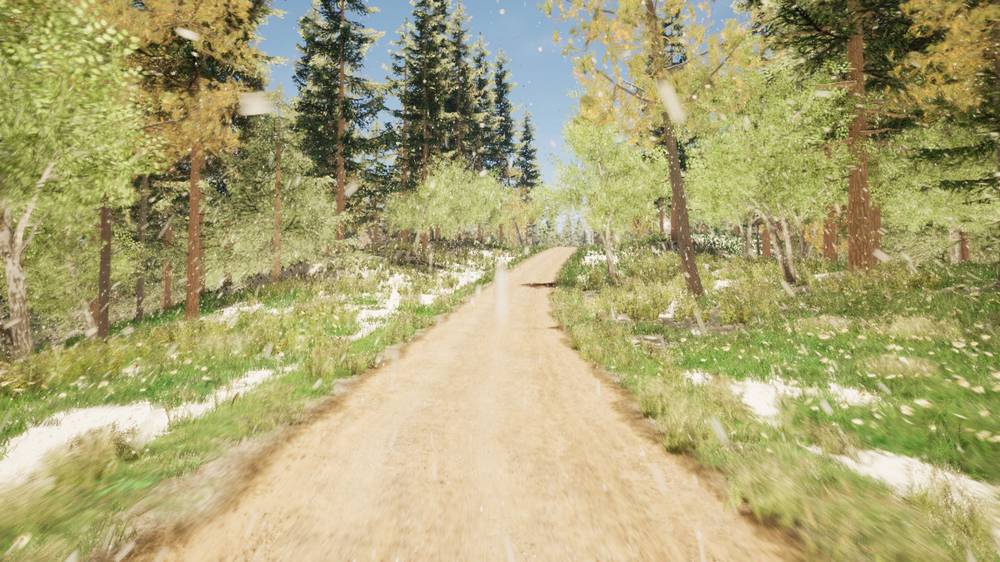}&
         \includegraphics[width=0.28\linewidth]{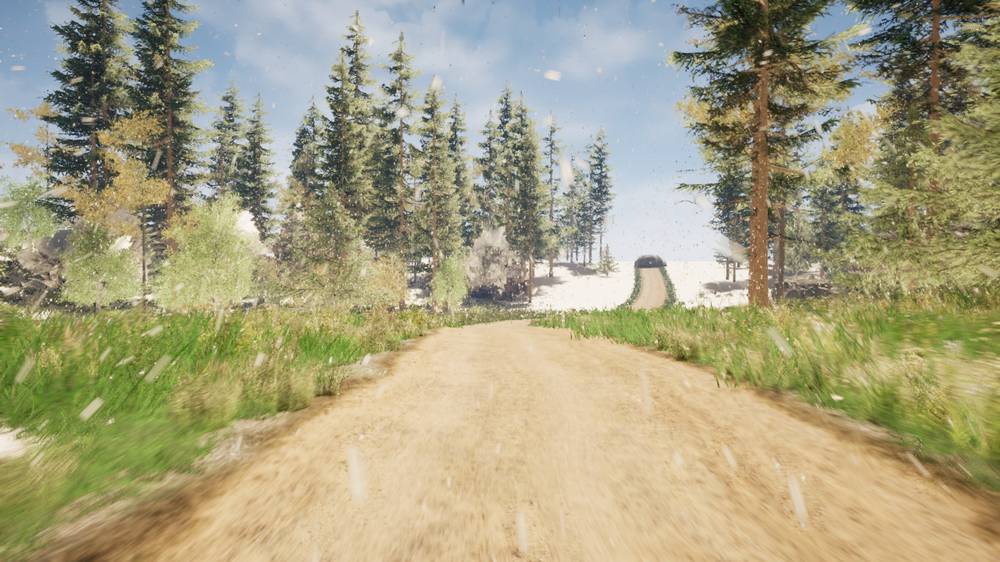}\\
         \includegraphics[width=0.28\linewidth]{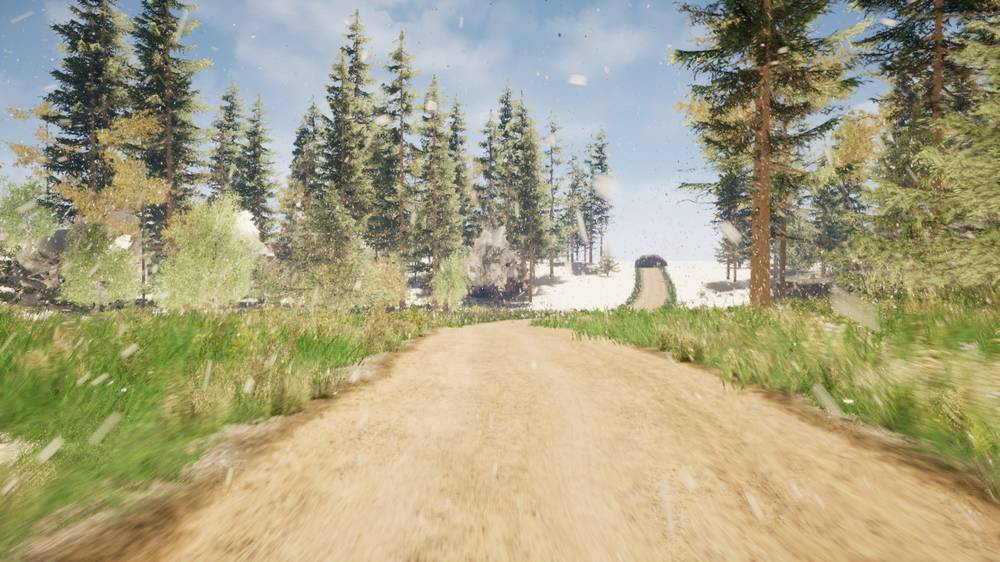}&
         \includegraphics[width=0.28\linewidth]{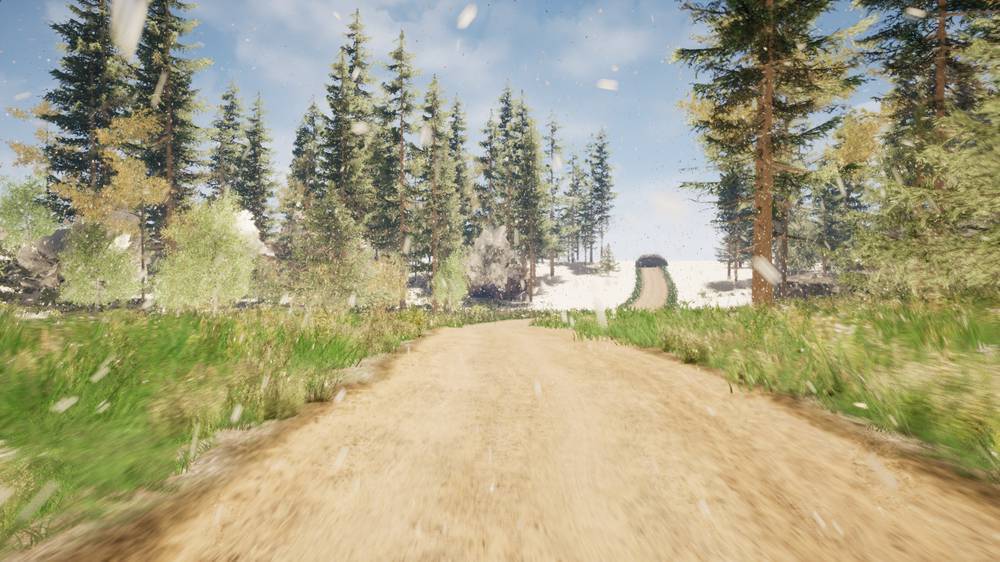}&
         \includegraphics[width=0.28\linewidth]{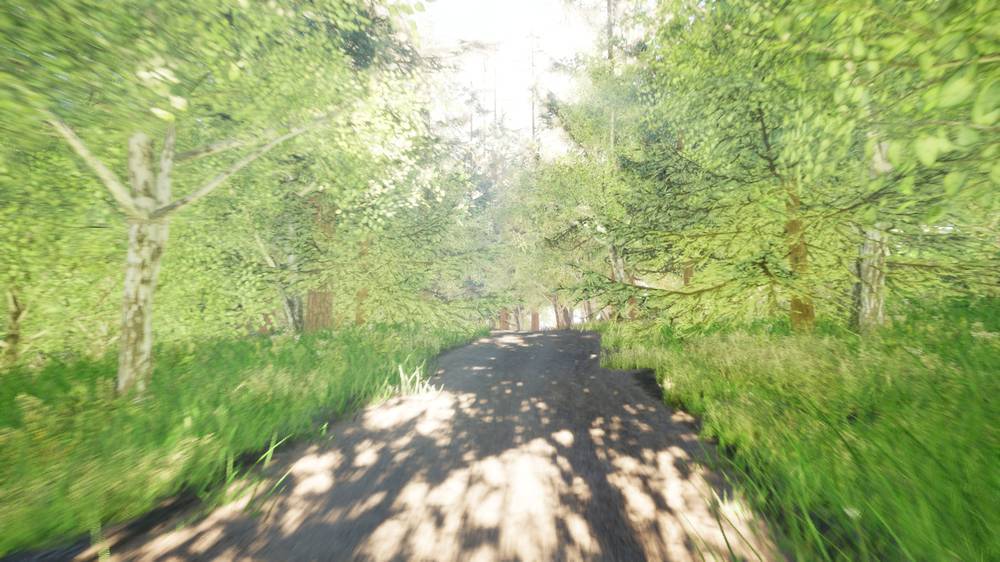}\\
         \includegraphics[width=0.28\linewidth]{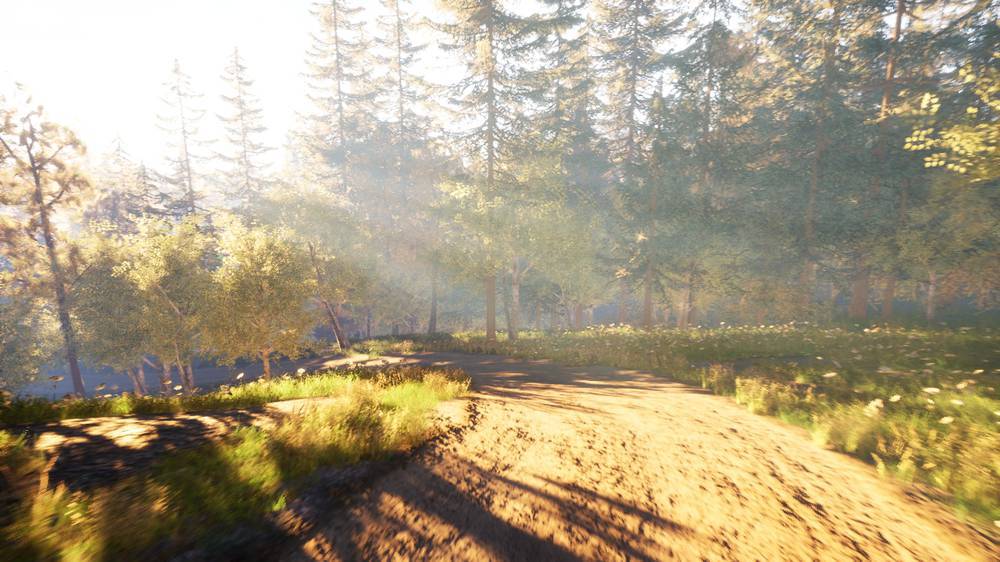}&
         \includegraphics[width=0.28\linewidth]{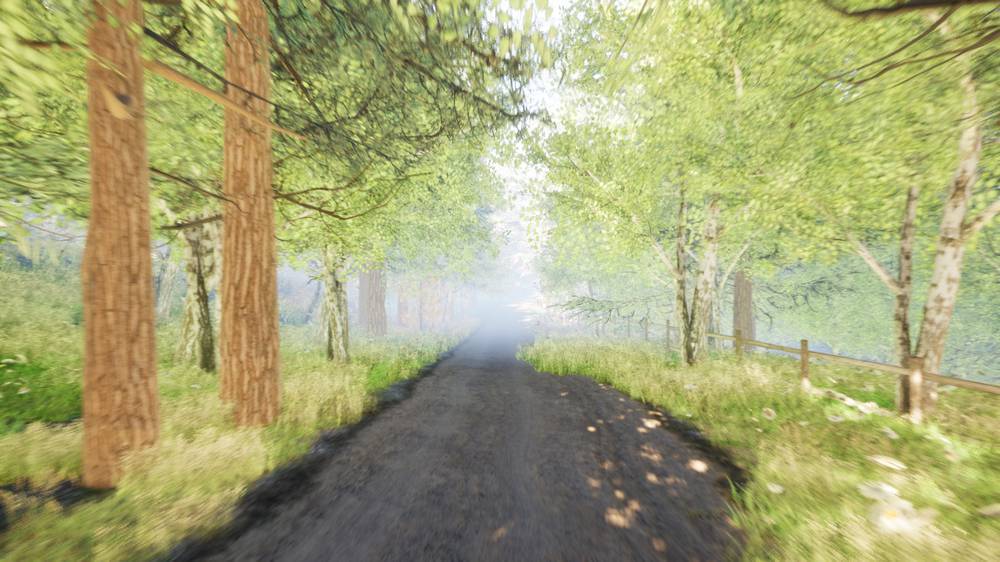}&
         \includegraphics[width=0.28\linewidth]{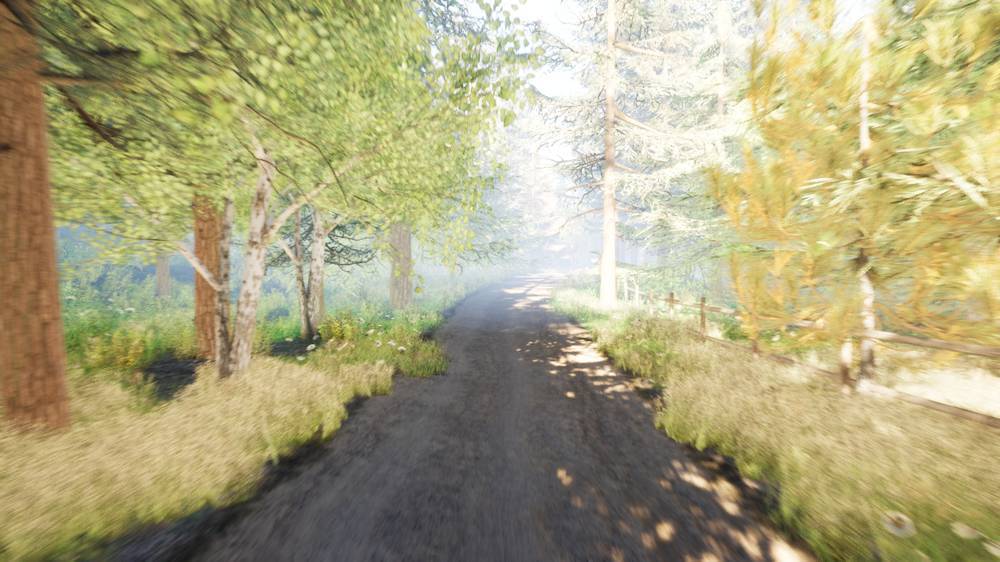}\\
         \includegraphics[width=0.28\linewidth]{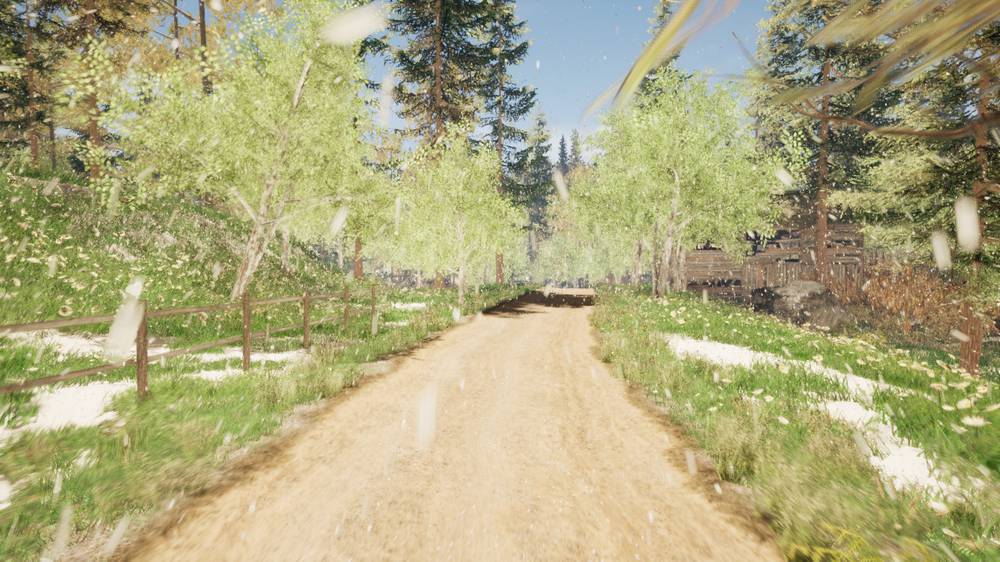}&
         \includegraphics[width=0.28\linewidth]{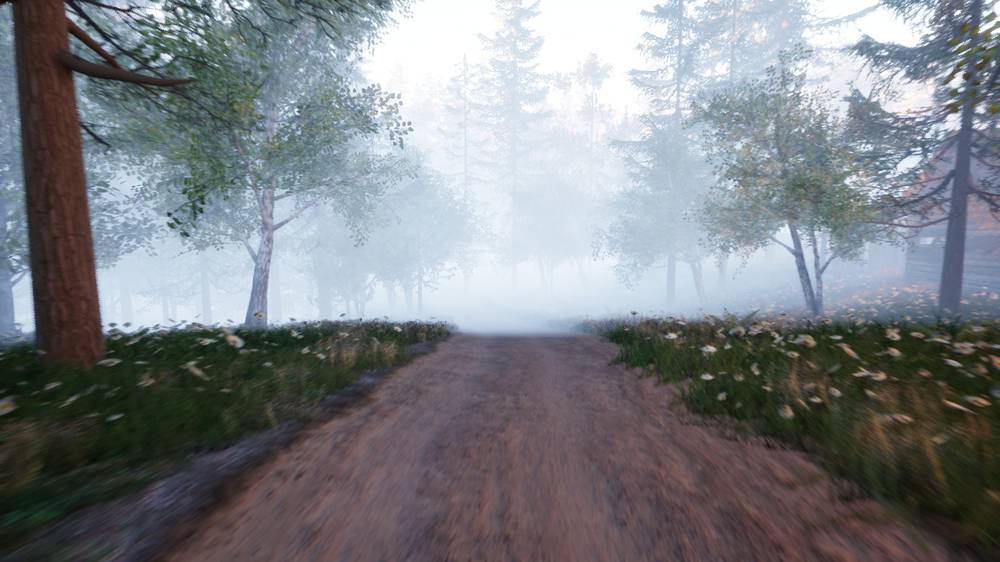}&
         \includegraphics[width=0.28\linewidth]{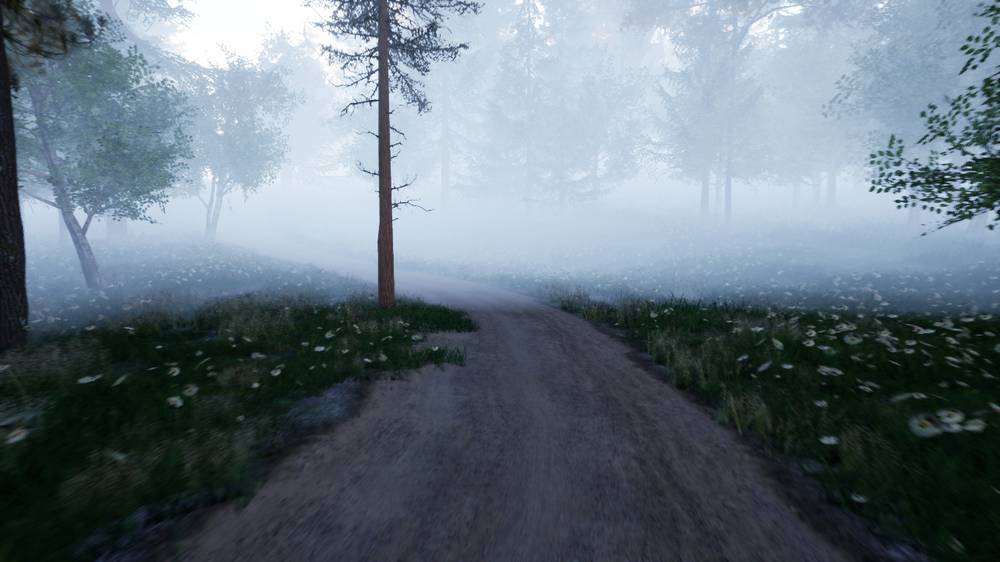}\\
    \end{tabular}
    \caption{Sample images from the MGTD dataset. Includes both the real and simulated part.}
    \label{fig:dataset_overview}
\end{figure}  
\clearpage

\section{Training configurations}
\label{training_configs}

\textbf{Training Configurations Overview}
This section details the complete hyperparameter setups used for all experiments. Tables \ref{tab:yolo_parameters}, \ref{tab:yolo_hyper} (same for vertical/ oriented bbox object detection) show the setup for object detection with  YOLOv8/YOLOv11 (n, s, m, l, x variants) in simulated and real images respectively. Table \ref{tab:rtdetr_hparams} shows the hyperparameters for object detection with RT-DETR (with Regnet backbone) and \ref{tab:instance_segmentation_hyper} lists the full settings for instance segmentation using Mask-RCNN and Cascade Mask-RCNN.

\begin{table*}[ht]
	\centering
	\small
	\caption{Hyperparameters for YOLOv8 / YOLOv11 (n, s, m, l, x) on simulated trees for both axis-aligned and oriented bounding boxes.\protect\footnotemark}
	\renewcommand{\arraystretch}{1.15}
	
	\begin{tabularx}{\textwidth}{p{3.0cm} X}
		\toprule
		\textbf{Parameter Group} & \textbf{Setting} \\
		\midrule
		
		Image size &
		1024$\times$1024 (fixed), multi\_scale = false \\
		
		Batching &
		Batch size = 1 (auto-batch), single GPU; no gradient accumulation \\
		
		Optimizer &
		SGD (momentum = 0.937, weight decay = 5e--4) \\
		
		Learning rate schedule &
		Linear LR; base LR = 0.01; final LR fraction = 0.01 $\rightarrow$ final LR = $1\times10^{-4}$; 
		Warmup: 3 epochs (momentum = 0.8, bias\_lr = 0.0) \\
		
		Augmentations &
		Mosaic: on (prob = 1.0), disabled in last 10 epochs; 
		MixUp: off; 
		HSV jitter: h = 0.015, s = 0.7, v = 0.4; 
		RandomAffine: translate = 0.1, scale = 0.5; 
		Flips: fliplr = 0.5, flipud = 0.0; 
		Random Erasing: prob = 0.4; 
		CutMix / Copy-Paste: off; 
		AutoAugment: RandAugment \\
		
		Losses &
		Box: 7.5 (CIoU style); 
		Class: 0.5 (BCE); 
		DFL: 1.5 (distributional regression); 
		Label smoothing: none \\
		
		Epochs &
		100 total; last 10 epochs with Mosaic disabled \\
		
		NMS &
		IoU threshold = 0.7; 
		conf = default; 
		agnostic\_nms = false; 
		nms = false in export (no WBF) \\
		
		\bottomrule
	\end{tabularx}
	\label{tab:yolo_parameters}
\end{table*}

\begin{table*}[ht]
	\centering
	\small
		\caption{Hyperparameters for YOLOv8 / YOLOv11 (n, s, m, l, x) – Real Trees\protect\footnotemark[\value{footnote}]}
	\renewcommand{\arraystretch}{1.15}
	
	\begin{tabularx}{\textwidth}{p{3.0cm} X}
		\toprule
		\textbf{Parameter Group} & \textbf{Setting} \\
		\midrule
		
		Image size &
		640$\times$640 (fixed, multi\_scale: false) \\
		
		Batching &
		Batch size = 16, single/multi-GPU depending on device, no gradient accumulation \\
		
		Optimizer &
		SGD (momentum: 0.937, weight decay: 0.0005) \\
		
		Learning rate schedule &
		Linear LR (cos\_lr: false), base LR = 0.01, final LR fraction = 0.01 $\rightarrow$ ends at $1\times10^{-4}$; 
		warmup (epochs = 3.0, momentum = 0.8, bias\_lr = 0.1) \\
		
		Augmentations &
		Same as simulated trees: Mosaic on (prob = 1.0), off in last 10 epochs (close\_mosaic = 10); 
		MixUp off; HSV jitter (h = 0.015, s = 0.7, v = 0.4); 
		RandomAffine (degrees = 0.0, translate = 0.1, scale = 0.5, shear = 0.0, perspective = 0.0); 
		Flips (flipud = 0.0, fliplr = 0.5); 
		CutMix off; Random Erasing on (prob = 0.4); 
		Copy-Paste off; AutoAugment: RandAugment \\
		
		Losses &
		Box loss: 7.5 (CIoU-style); 
		Class loss: 0.5 (BCE, no focal); 
		DFL: 1.5 (distributional regression on); 
		Label smoothing: 0.0 \\
		
		Epochs &
		50 total, last 10 epochs no-Mosaic fine-tune \\
		
		NMS &
		IoU threshold = 0.7, conf = framework default; 
		agnostic\_nms = false; 
		nms = false in export (no WBF) \\
		
		\bottomrule
	\end{tabularx}
	\label{tab:yolo_hyper}
\end{table*}

\begin{table*}[ht]
	\centering
	\small
	\caption{Hyperparameter setup for RT-DETR used in our experiments.}
	\renewcommand{\arraystretch}{1.15}
	
	\begin{tabularx}{\textwidth}{p{3.2cm} X}
		\toprule
		\textbf{Parameter Group} & \textbf{Setting} \\
		\midrule
		
		Backbone \& Init &
		RegNet (return\_idx = [1,2,3]); initialized with ImageNet-1K weights (torchvision) \\
		
		Epochs &
		130 total \\
		
		Optimizer &
		AdamW; LR = 1e-4 (all modules); weight decay = 1e-4 (bias/norm excluded); gradient clipping = 0.1 \\
		
		LR Schedule &
		MultiStepLR (milestones = [1000], $\gamma=0.1$ → no effective decay); no warmup \\
		
		Queries \& Matcher &
		300 object queries; Hungarian costs — cls: 2, L1: 5, GIoU: 2;  
		focal matcher $(\alpha=0.25, \gamma=2.0)$ \\
		
		Losses &
		Varifocal classification loss $(\alpha=0.75, \gamma=2.0)$;  
		Box: L1 (w=5) + GIoU (w=2);  
		Denoising with 100 queries \\
		
		Augmentations &
		Multi-scale short-side = \{480–800, steps 32\};  
		RandomPhotometricDistort (p=0.5), RandomZoomOut, RandomIoUCrop (p=0.8), RandomHorizontalFlip;  
		final Resize 640$\times$640 \\
		
		Multi-scale Features &
		3 levels from HybridEncoder (feat\_strides = [8,16,32], hidden\_dim = 256);  
		two-stage: no \\
		
		Inference &
		Single-scale 640$\times$640; top-300 queries; no NMS (DETR-style) \\
		
		\bottomrule
	\end{tabularx}

	\label{tab:rtdetr_hparams}
\end{table*}

\begin{table}[htp!]
\small
\captionof{table}{Hyperparameter setup for instance segmentation on real and simulated trees (MMDetection). Models are grouped by family, repeated values are merged for compactness.}
\centering
\begin{minipage}{\textwidth}
\centering
\resizebox{\textwidth}{!}{
\begin{tabular}{l l c c c l}
\toprule
\textbf{Model} & \textbf{Backbone} & \textbf{Optimizer} & \textbf{LR} & \textbf{Weight Decay} & \textbf{Schedule / Augmentations} \\
\midrule
\multirow{2}{*}{Mask R-CNN} 
  & ResNeXt-101-64x4d + FPN & SGD ($m=0.9$)  & 0.02    & 0.0001 & Step LR + linear warmup ($r=0.001$), Resize + Flip + Norm + Pad \\
  & Swin-T + FPN            & AdamW          & 1e-4    & 0.05   & Step LR + linear warmup ($r=0.001$), Resize + Flip + Norm + Pad \\
\midrule
\multirow{2}{*}{Cascade Mask R-CNN} 
  & ResNeXt-101-64x4d + FPN & SGD ($m=0.9$)  & 0.02    & 0.0001 & Step LR + linear warmup ($r=0.001$), Resize + Flip + Norm + Pad \\
  & Swin-T + FPN            & AdamW          & 1e-4    & 0.05   & Step LR + linear warmup ($r=0.001$), Resize + Flip + Norm + Pad \\
\midrule
\multirow{2}{*}{Mask2Former} 
  & R50\_lsj\_50e           & AdamW          & 1e-4    & 0.05   & Step LR + linear warmup ($r=0.001$), Resize + Flip + Norm + Pad \\
  & R101\_lsj\_50e / tuned  & AdamW          & 1e-4    & 0.05   & Step LR + linear warmup ($r=0.001$), Resize + Flip + Norm + Pad \\
\bottomrule
\end{tabular}
}
\vspace{0.3em}
\label{tab:instance_segmentation_hyper}
\end{minipage}
\end{table}

All experiments were conducted on two NVIDIA RTX 6000 Ada GPUs with mixed-precision training (AMP) enabled. Distributed data-parallel (DDP) training was used for multi-GPU runs. The real tree dataset (4608$\times$3456) and the simulated tree dataset (1280$\times$720) were trained with different epochs to balance convergence and overfitting risk (see footnote). Unless otherwise specified, default optimizer settings and learning rate schedules were used as defined by the respective frameworks (Ultralytics for YOLO, MMDetection for instance segmentation).

\begin{table*}[htp!]
	\centering
	\small
	\renewcommand{\arraystretch}{1.1}
	\setlength{\tabcolsep}{5pt}
	\caption{Training hyperparameters for Phase~4 (granularity-aware distillation).}
	\label{tab:phase4_hyperparams}
	\resizebox{\textwidth}{!}{
		\begin{tabular}{p{3.3cm} p{12.7cm}}
			\toprule
			\textbf{Hyperparameter} & \textbf{Value} \\
			\midrule
			Teacher architecture & Mask R-CNN (Swin-T, MMDetection). \\
			Student architecture & Mask R-CNN (ResNet-50 + FPN). \\
			Training epochs & 100 \\
			Batch size (total) & 48 \\
			Labeled batch size & 16 \\
			Data loader workers & 8 \\
			Warm-up epochs & 10 \\
			Optimizer & AdamW \\
			Learning rate & $1\times10^{-4}$ \\
			Minimum learning rate & $1\times10^{-6}$ \\
			Weight decay & 0.01 \\
			Gradient clipping & 1.0 \\
			LR schedule & Cosine decay \\
			KD weight $\lambda_{\text{KD}}$ & 1.0 \\
			KD ramp-up epochs & 10 \\
			KD temperature $\tau$ & 2.0 \\
			Confidence threshold (early $\rightarrow$ late) & $0.6 \rightarrow 0.4$ \\
			Input resolution & $512 \times 512$ \\
			Weak augmentations & Color jitter: brightness=0.1, contrast=0.1, saturation=0.1, hue=0.05 \\
			Strong augmentations & Color jitter: brightness=0.6, contrast=0.6, saturation=0.6, hue=0.2; Gaussian blur: radius $\in[0.1, 2.0]$ \\
			Copy-paste & Probability=0.5; scale $\in[0.8, 1.2]$ \\
			\bottomrule
		\end{tabular}
	}
\end{table*}

\footnotetext{Simulated dataset trained for 100 epochs, real dataset trained for 50 epochs due to higher overfitting risk and greater per-iteration computational cost.}

\clearpage
\section{Instance Segmentation experiments}
\subsection{Phase 1}

\paragraph{Training on simulated tree trunks and whole trees}
In the first phase, we trained two synthetic teachers independently, one specialized for the tree trunk class and one for the whole tree class. Validation on the synthetic splits establishes an upper bound for what can be achieved with perfect domain alignment. For Mask-RCNN, Metrics show (from table\ref{tab:sim_results_combined}) that the Swin-T backbone consistently outperforms ResNeXt-101, particularly in AP at higher IoU thresholds (AP) This confirms that fine-grained structural priors can be captured reliably in simulation.
To complement the quantitative results, we provide representative qualitative examples from the synthetic test set. 

\begin{table*}[htp!]
\centering
\caption{\textbf{Phase 1:} Bounding-box (top) and segmentation (bottom) results on simulated datasets (tree trunk and whole tree). Metrics include AP/AR at multiple IoU thresholds and object scales. Bold numbers highlight best values in each block.}
\resizebox{\textwidth}{!}{
\begin{tabular}{l l l c c c c c c c c c c c c c}
\toprule
Type & Model & Backbone & Epochs & mAP@[.5:.95] (Box) & mAP@50 (Box) & mAP@75 (Box) & APs (Box) & APm (Box) & APl (Box) & mAR@100 & mAR@300 & mAR@1000 & ARs & ARm & ARl \\
\midrule
Tree Trunk & Mask-RCNN         & X101   & 24 & 0.732 & 0.904 & 0.858 & 0.694 & 0.769 & 0.837 & 0.812 & 0.812 & 0.812 & 0.778 & 0.840 & 0.916 \\
           &                   & Swin-T & 24 & \textbf{0.788} & \textbf{0.936} & \textbf{0.908} & \textbf{0.772} & \textbf{0.801} & \textbf{0.868} & \textbf{0.847} & \textbf{0.847} & \textbf{0.847} & \textbf{0.822} & \textbf{0.865} & \textbf{0.939} \\
           & Cascade Mask-RCNN & X101   & 24 & 0.762 & 0.905 & 0.872 & 0.728 & 0.801 & 0.877 & 0.829 & 0.829 & 0.829 & 0.797 & 0.858 & 0.937 \\
           &                   & Swin-T & 24 & 0.764 & 0.912 & 0.877 & 0.729 & 0.803 & 0.874 & 0.837 & 0.837 & 0.837 & 0.798 & 0.844 & \textbf{0.942} \\
\midrule
Whole Tree & Mask-RCNN         & X101   & 24 & 0.859 & 0.947 & 0.930 & 0.717 & 0.880 & 0.856 & 0.905 & 0.905 & 0.905 & 0.764 & 0.895 & 0.909 \\
           &                   & Swin-T & 24 & 0.880 & 0.948 & 0.936 & 0.798 & 0.903 & 0.874 & 0.923 & 0.923 & 0.923 & 0.823 & 0.916 & 0.926 \\
           & Cascade Mask-RCNN & X101   & 24 & \textbf{0.887} & \textbf{0.994} & \textbf{0.934} & 0.768 & 0.908 & \textbf{0.882} & 0.923 & 0.923 & 0.923 & 0.715 & \textbf{0.919} & 0.924 \\
           &                   & Swin-T & 24 & 0.883 & 0.950 & 0.937 & \textbf{0.815} & 0.901 & 0.879 & 0.922 & 0.922 & 0.922 & \textbf{0.859} & 0.917 & \textbf{0.924} \\
\midrule\midrule
Type & Model & Backbone & Epochs & mAP@[.5:.95] (Segm) & mAP@50 (Segm) & mAP@75 (Segm) & APs (Segm) & APm (Segm) & APl (Segm) & mAR@100 & mAR@300 & mAR@1000 & ARs & ARm & ARl \\
\midrule
Tree Trunk & Mask-RCNN         & X101   & 24 & 0.686 & 0.902 & 0.820 & 0.620 & 0.742 & 0.812 & 0.767 & 0.767 & 0.767 & 0.735 & 0.797 & 0.864 \\
           &                   & Swin-T & 24 & \textbf{0.753} & \textbf{0.935} & \textbf{0.880} & \textbf{0.727} & \textbf{0.780} & \textbf{0.834} & \textbf{0.814} & \textbf{0.814} & \textbf{0.814} & \textbf{0.795} & \textbf{0.830} & \textbf{0.884} \\
           & Cascade Mask-RCNN & X101   & 24 & 0.663 & 0.902 & 0.810 & 0.614 & 0.719 & 0.821 & 0.732 & 0.732 & 0.732 & 0.692 & 0.748 & 0.889 \\
           &                   & Swin-T & 24 & 0.673 & 0.909 & 0.818 & 0.621 & 0.729 & 0.815 & 0.743 & 0.743 & 0.743 & 0.704 & 0.782 & \textbf{0.890} \\
\midrule
Whole Tree & Mask-RCNN         & X101   & 24 & 0.670 & 0.911 & 0.775 & 0.265 & 0.557 & 0.694 & 0.751 & 0.751 & 0.751 & 0.465 & 0.667 & 0.768 \\
           &                   & Swin-T & 24 & \textbf{0.688} & \textbf{0.920} & \textbf{0.797} & \textbf{0.377} & \textbf{0.585} & 0.708 & \textbf{0.766} & \textbf{0.766} & \textbf{0.766} & \textbf{0.568} & \textbf{0.689} & 0.781 \\
           & Cascade Mask-RCNN & X101   & 24 & 0.686 & 0.914 & 0.793 & 0.315 & 0.585 & 0.709 & 0.762 & 0.762 & 0.762 & 0.541 & 0.663 & 0.776 \\
           &                   & Swin-T & 24 & 0.681 & 0.916 & 0.794 & 0.317 & 0.583 & 0.704 & 0.756 & 0.756 & 0.756 & 0.544 & 0.683 & \textbf{0.794} \\
\bottomrule
\end{tabular}
}

\label{tab:sim_results_combined}
\end{table*}

\begin{figure}[h!]
    \centering
    \renewcommand{\arraystretch}{1.2} 
    \setlength{\tabcolsep}{1pt} 

    \begin{tabular}{c c c}
        \textbf{RGB Image} & \textbf{Prediction} & \textbf{Ground Truth} \\
        \includegraphics[width=0.3\linewidth]{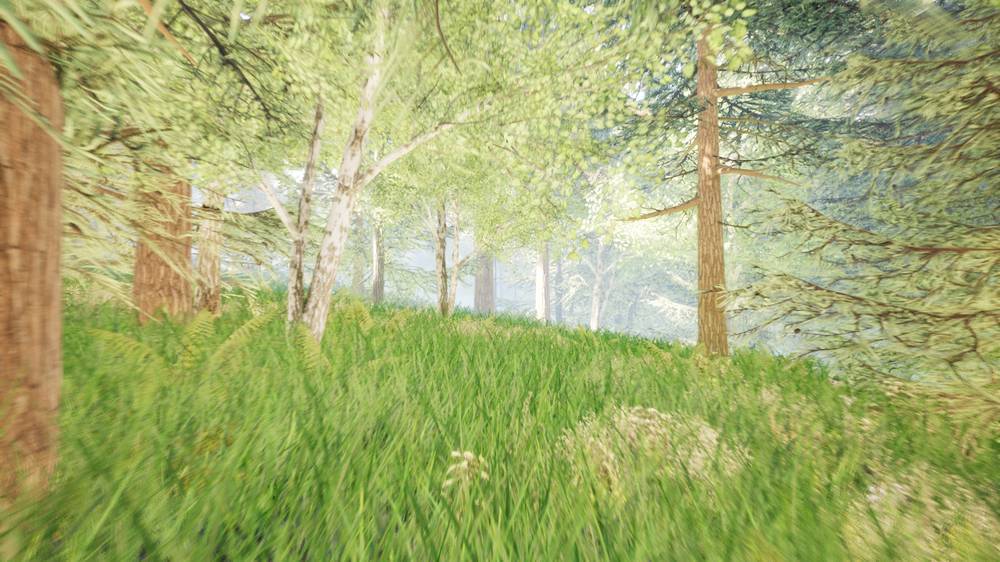} &
        \includegraphics[width=0.3\linewidth]{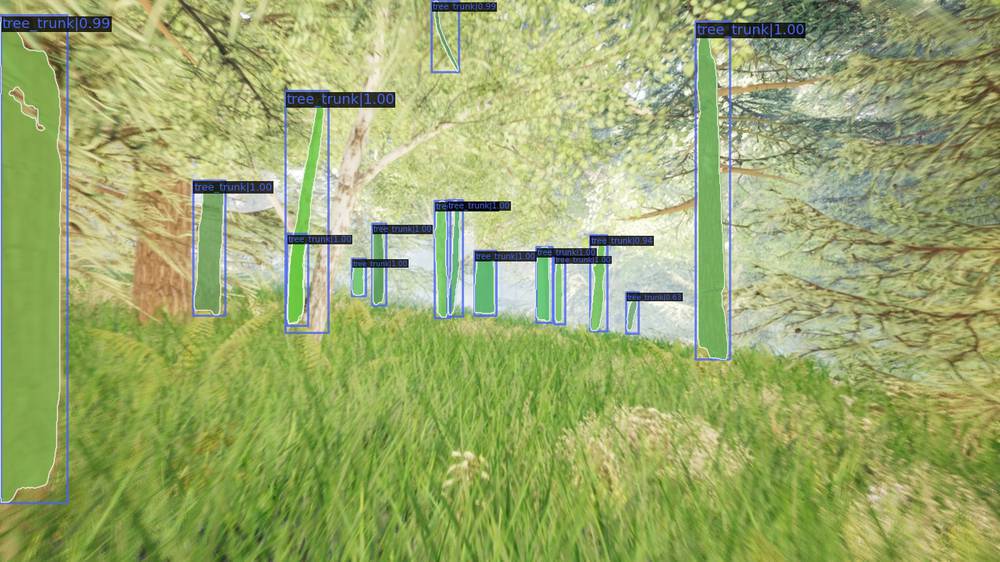} &
        \includegraphics[width=0.3\linewidth]{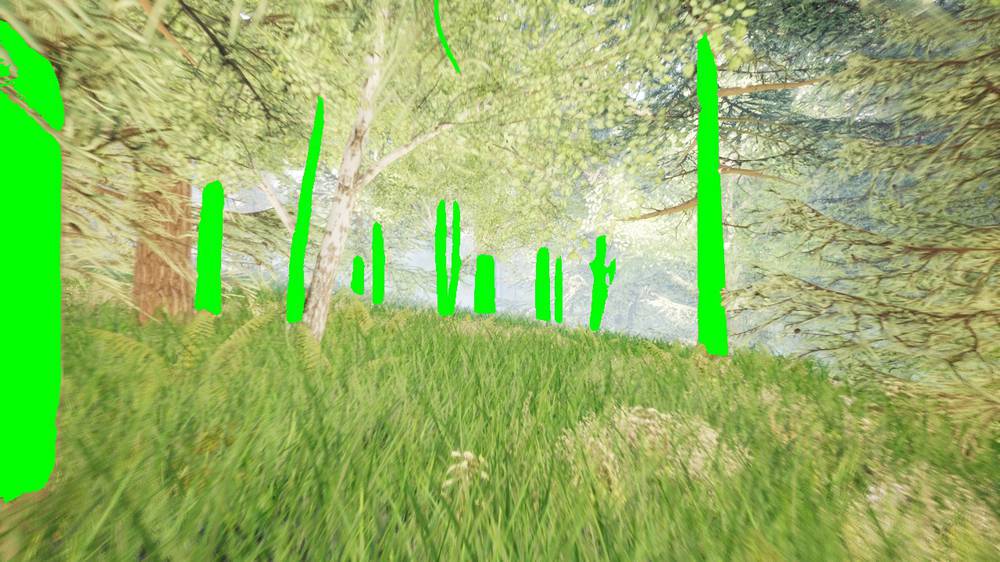} \\
        \includegraphics[width=0.3\linewidth]{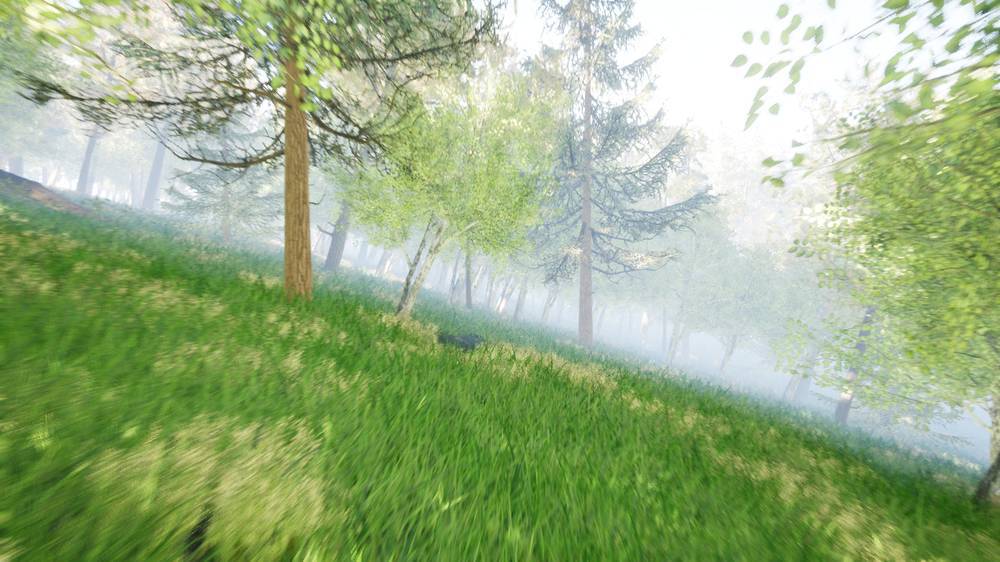} &
        \includegraphics[width=0.3\linewidth]{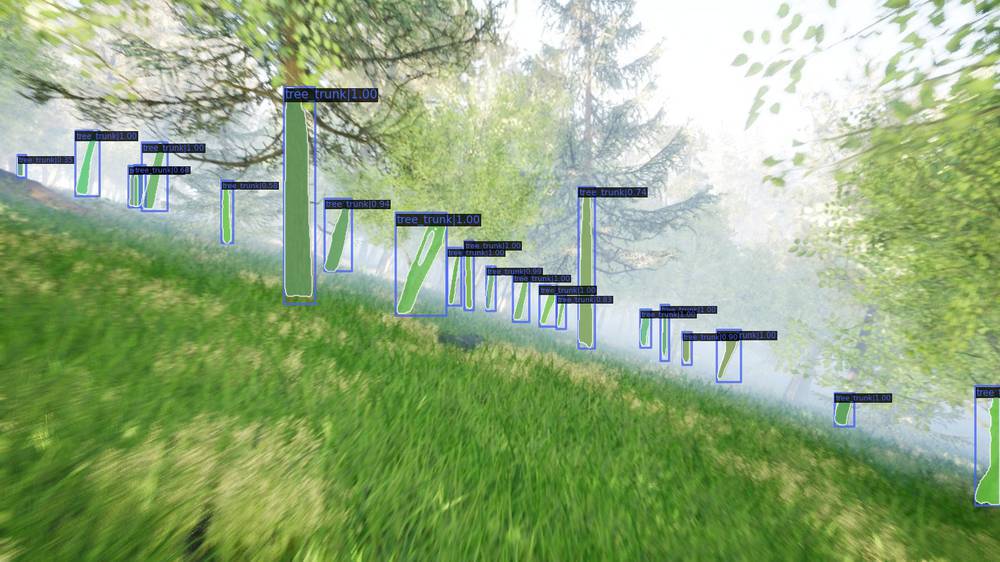} &
        \includegraphics[width=0.3\linewidth]{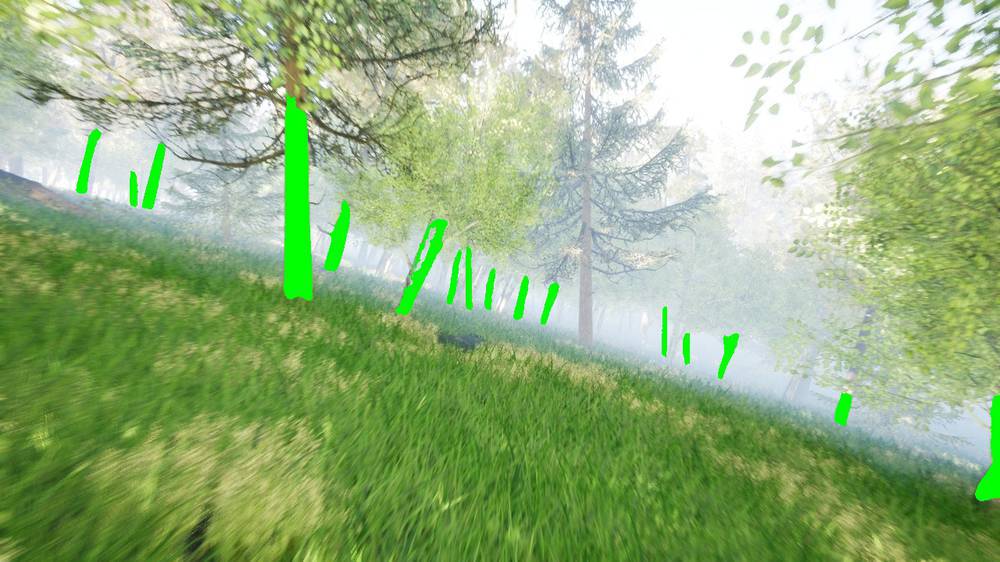} \\
        \includegraphics[width=0.3\linewidth]{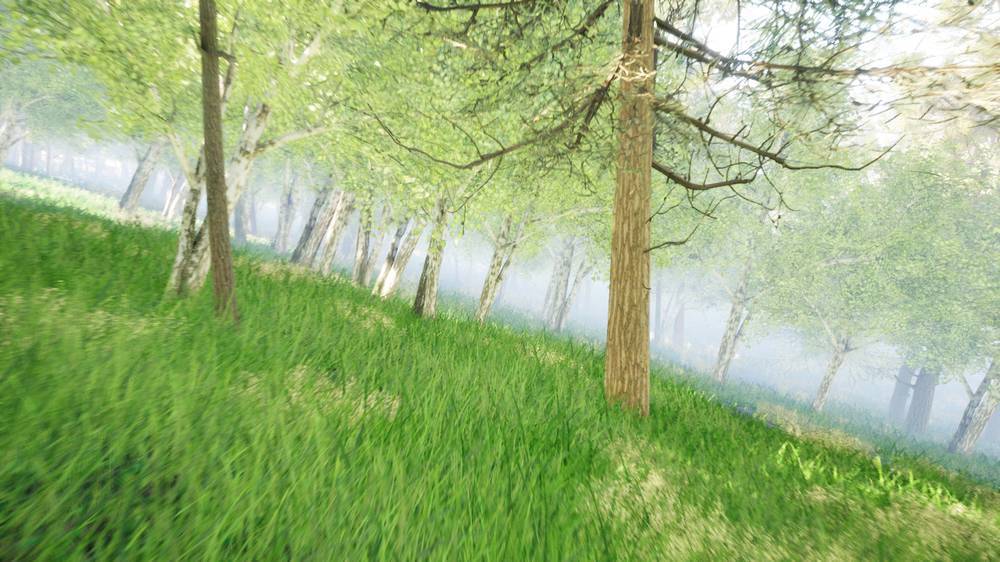} &
        \includegraphics[width=0.3\linewidth]{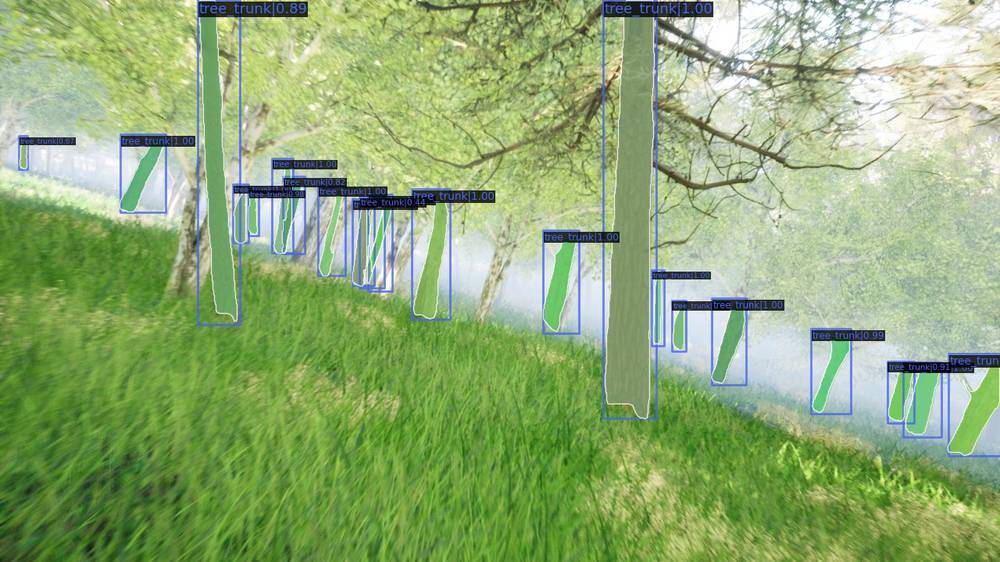} &
        \includegraphics[width=0.3\linewidth]{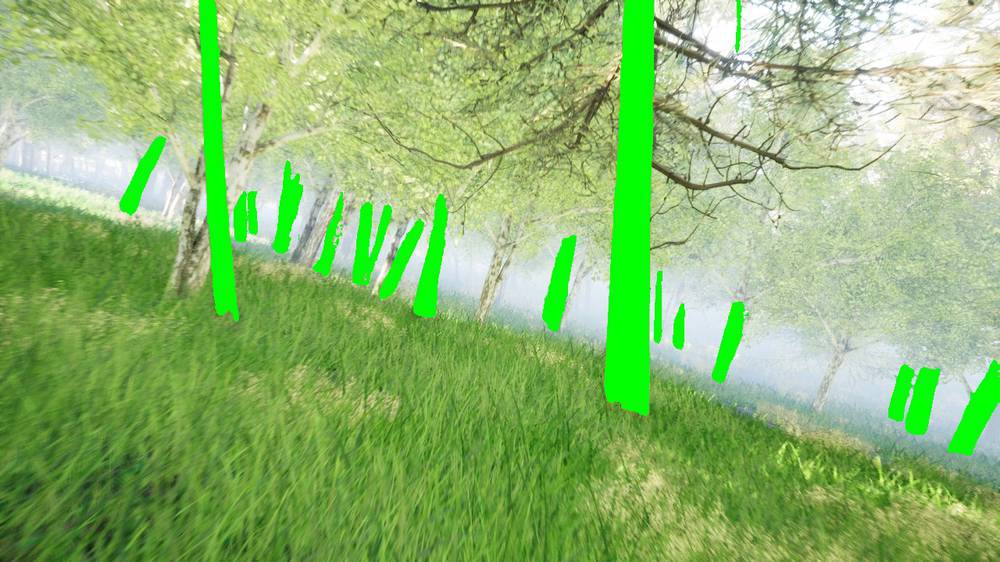} \\
        \includegraphics[width=0.3\linewidth]{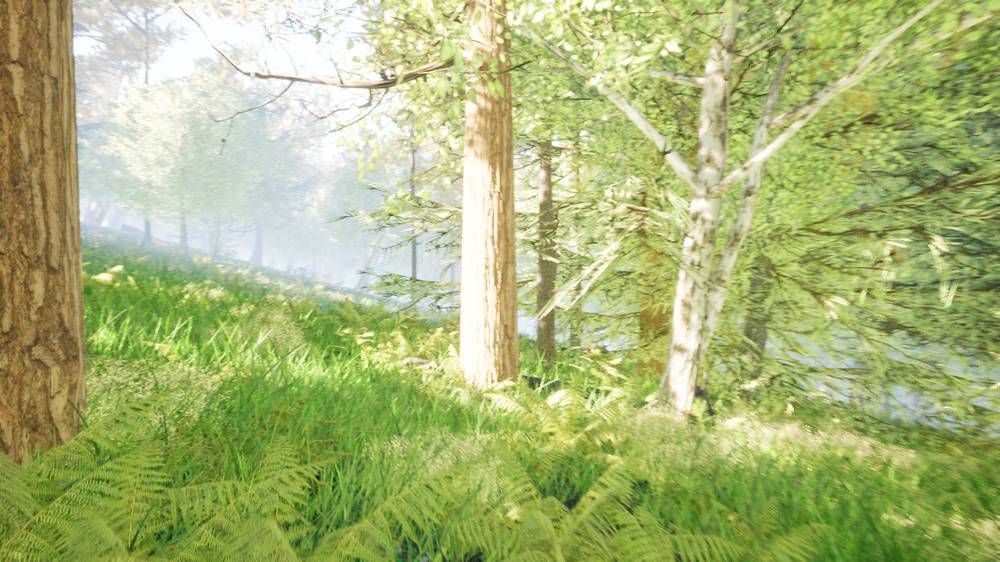} &
        \includegraphics[width=0.3\linewidth]{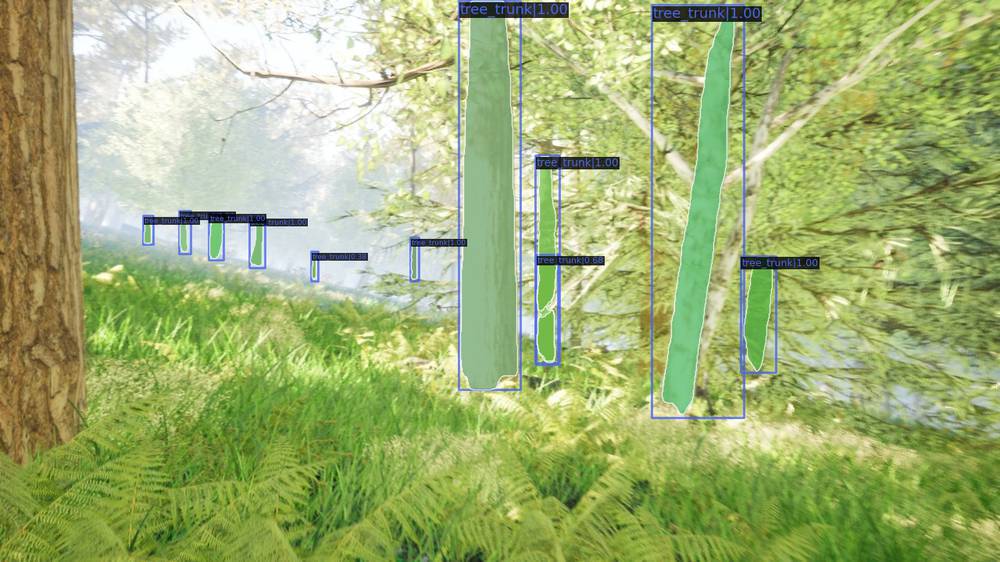} &
        \includegraphics[width=0.3\linewidth]{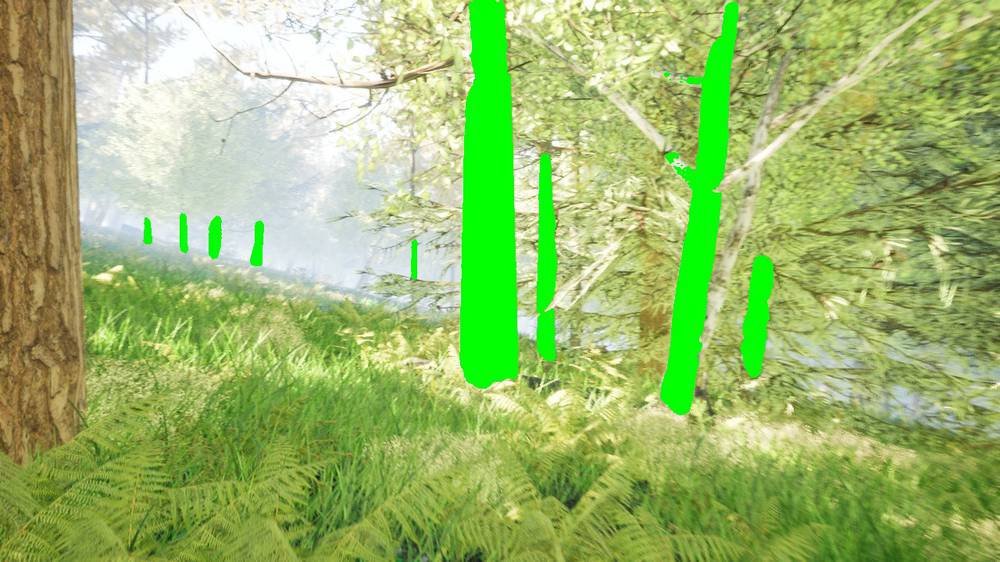} \\
        \includegraphics[width=0.3\linewidth]{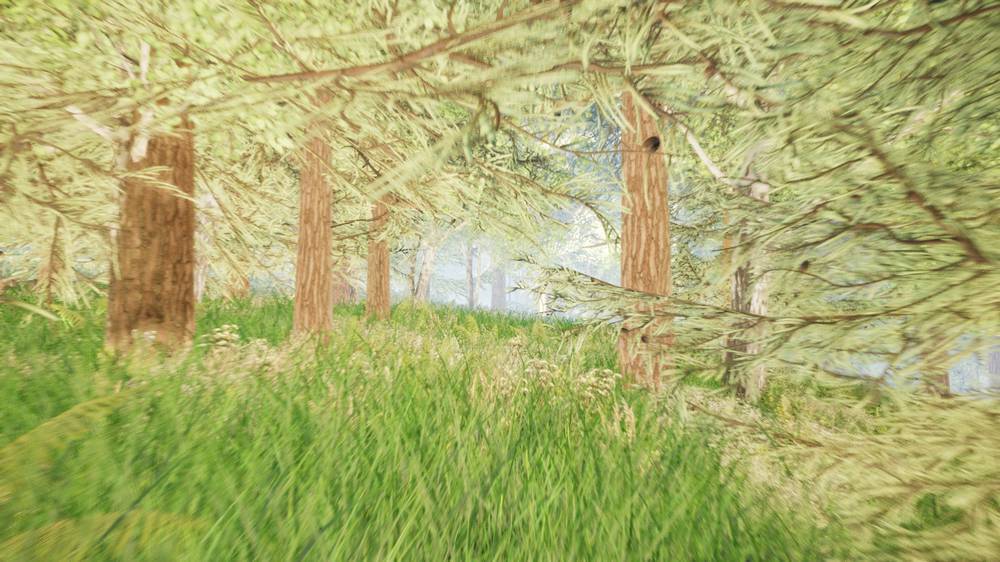} &
        \includegraphics[width=0.3\linewidth]{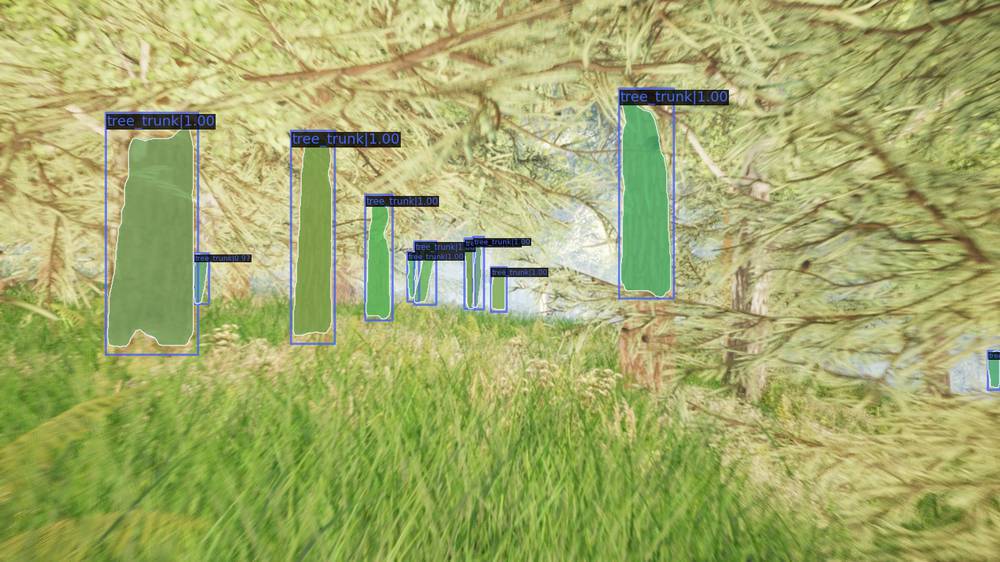} &
        \includegraphics[width=0.3\linewidth]{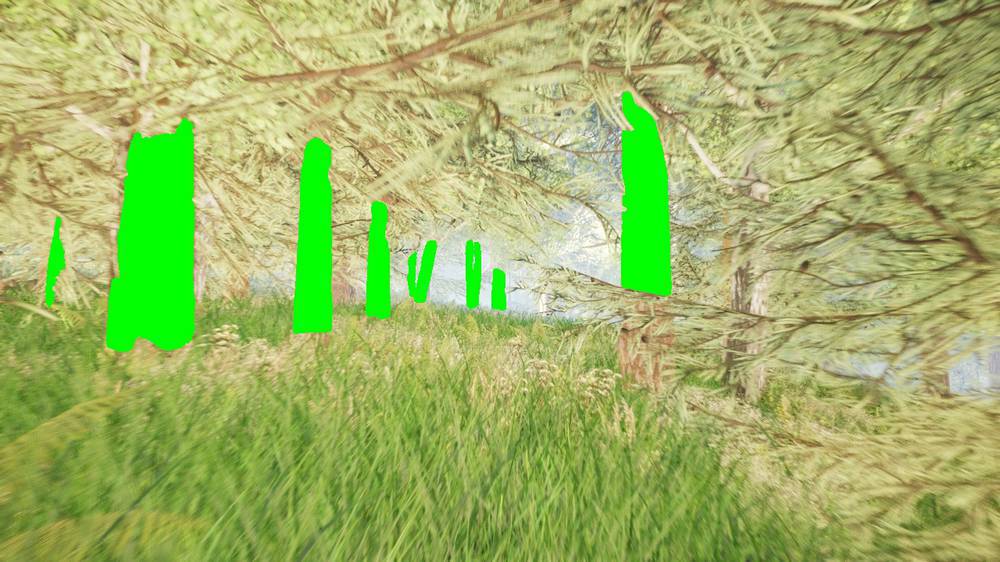} \\
        \includegraphics[width=0.3\linewidth]{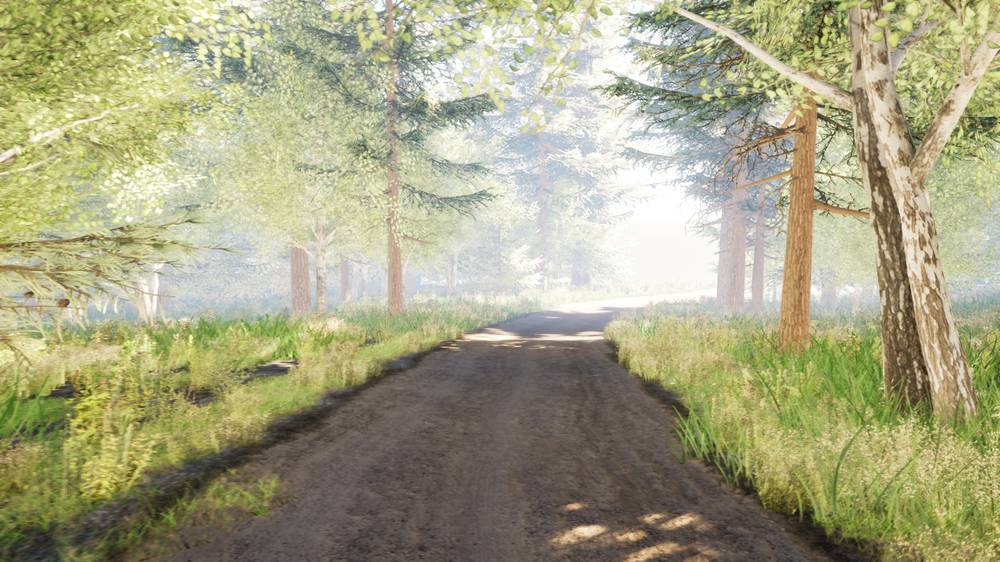} &
        \includegraphics[width=0.3\linewidth]{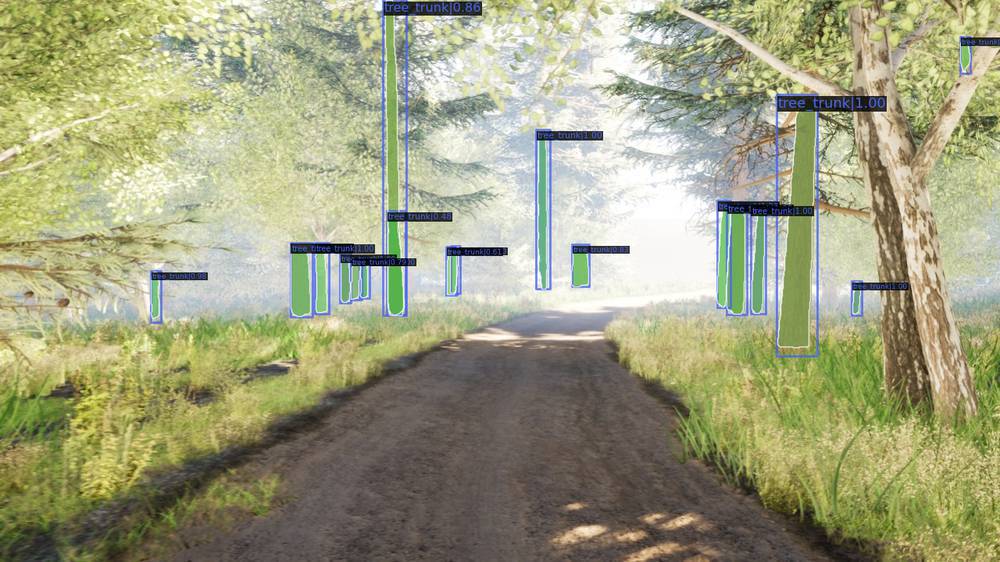} &
        \includegraphics[width=0.3\linewidth]{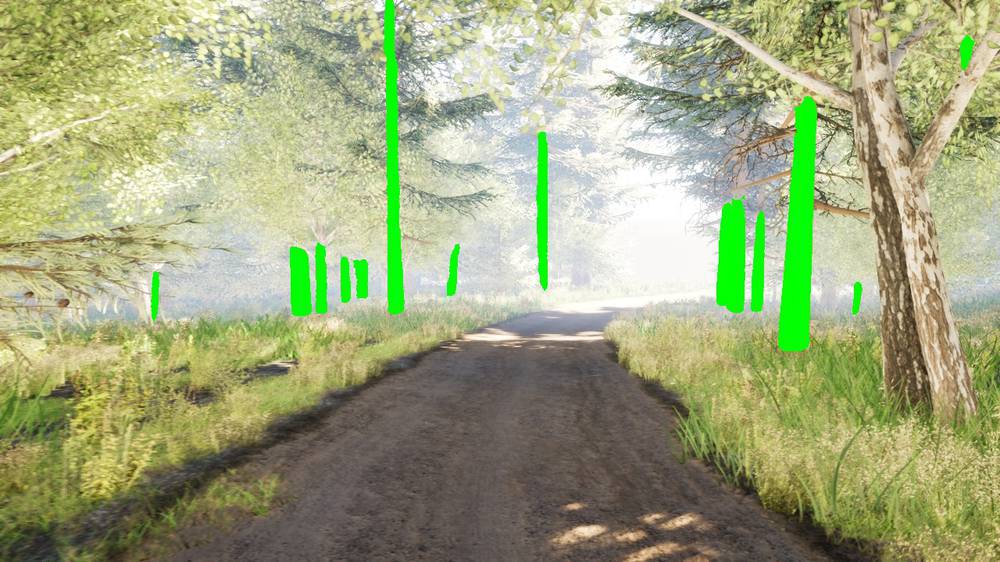} \\
        \includegraphics[width=0.3\linewidth]{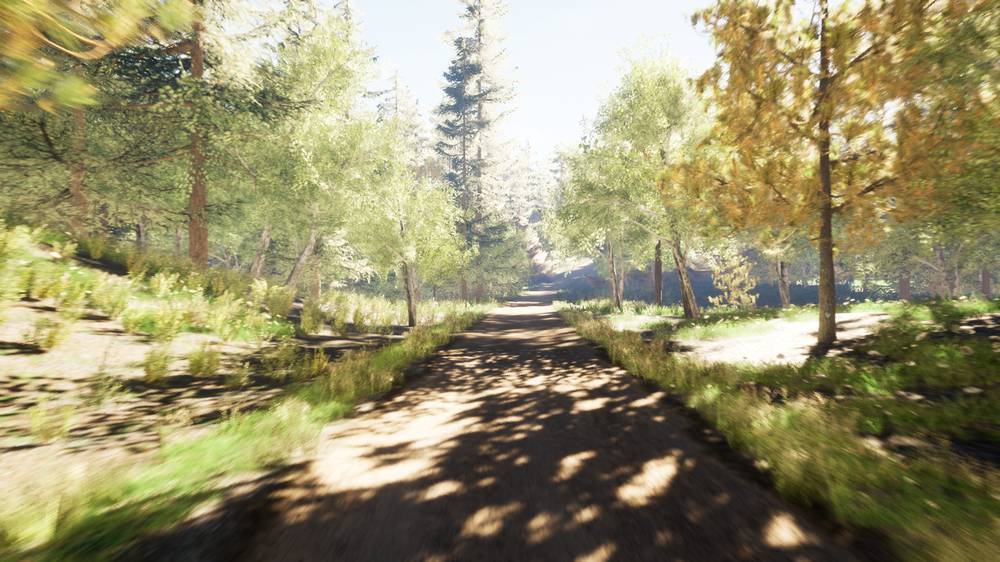} &
        \includegraphics[width=0.3\linewidth]{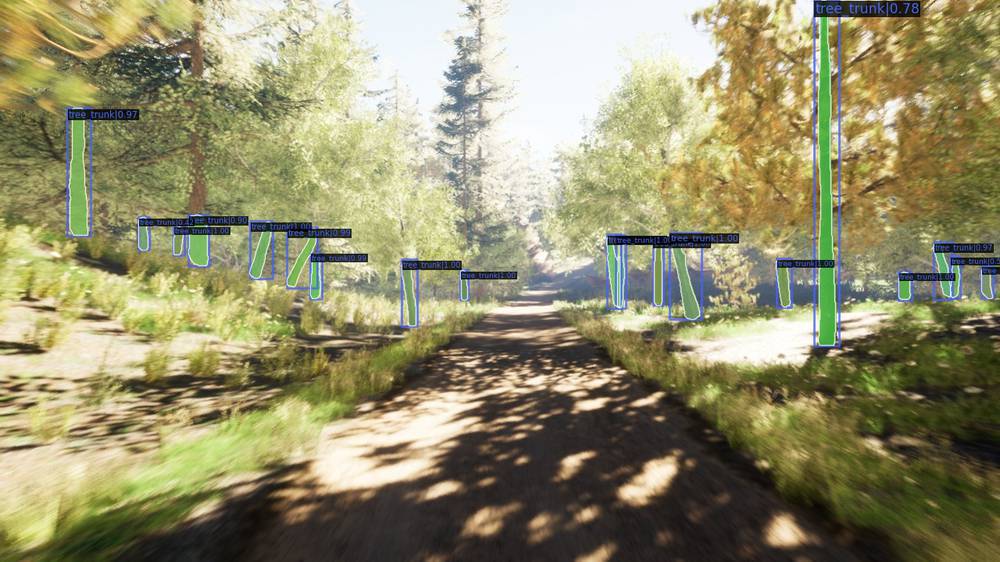} &
        \includegraphics[width=0.3\linewidth]{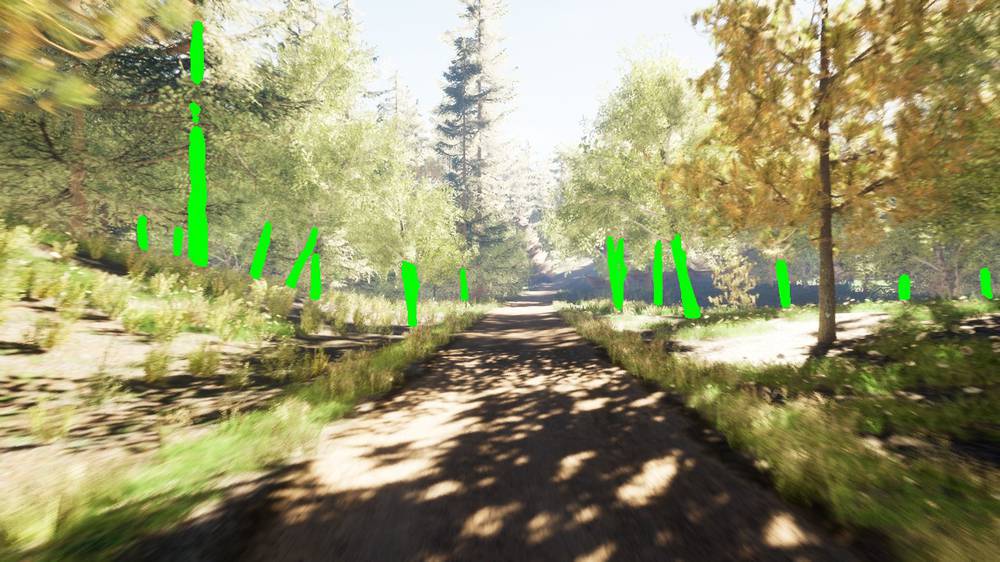} \\
    \end{tabular}
    \caption{\textbf{Phase 1:} Instance segmentation results on \textbf{Tree Trunk} examples. Each row shows the RGB image, model prediction, and ground truth mask.}
    \label{fig:phase1_trunk}
\end{figure}

Figure \ref{fig:phase1_trunk} illustrates qualitative results of Mask~R-CNN (Swin-T) trained on the simulated tree trunk dataset. The model consistently (to some extent) detects and segments prominent trunks in diverse synthetic forest scenes, demonstrating that it has learned reliable vertical structure cues. Large and medium trunks are localized with high confidence, and multiple instances are detected per frame, confirming the advantage of Swin-T’s strong multi-scale feature extraction. At the same time, \textbf{several limitations} become evident. \textbf{First}, the predicted trunk masks tend to be coarser and thicker than the ground-truth annotations, with blurred boundaries that fail to capture the precise silhouette of thin tree stems. \textbf{Second}, thin or distant trunks are frequently missed, as seen in the background of some rows, which highlights the well-known difficulty of Mask-RCNN in handling small objects. Taken together, these examples indicate that while the synthetic-trained trunk teacher is effective at providing structural priors (e.g., trunk verticality, spatial arrangement, relative scale), it is not perfect, its masks are (sometimes) noisy and biased, and its coverage of small/occluded objects is incomplete. These limitations underline the necessity of Phase 4 distillation, where the teacher’s coarse but informative priors can be selectively transferred to a real-trained student rather than being used directly. 
\begin{figure}[htp!]
    \centering
    \renewcommand{\arraystretch}{1.2} 
    \setlength{\tabcolsep}{2pt} 

    \begin{tabular}{c c c}
        \textbf{RGB Image} & \textbf{Prediction} & \textbf{Ground Truth} \\
        \includegraphics[width=0.3\linewidth]{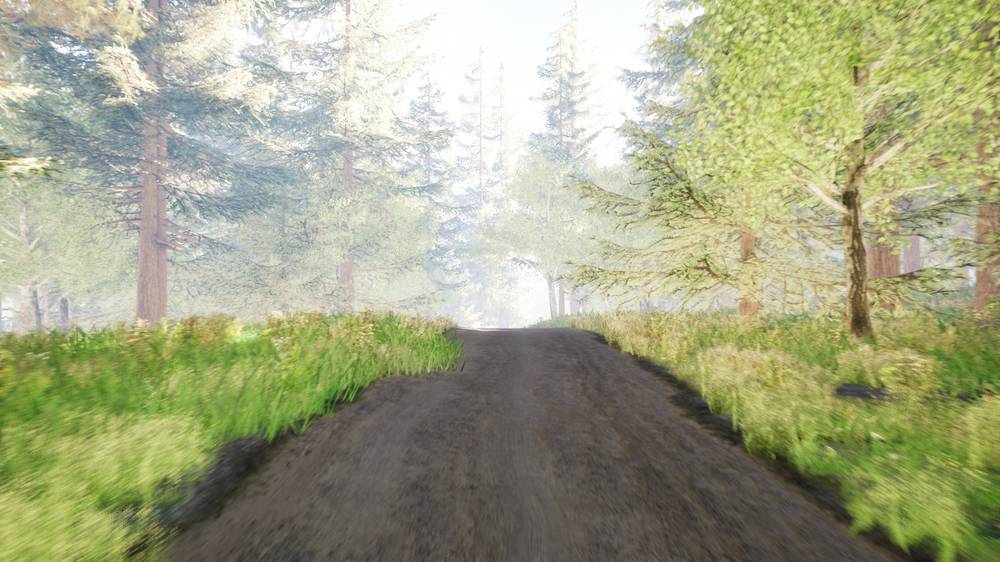} &
        \includegraphics[width=0.3\linewidth]{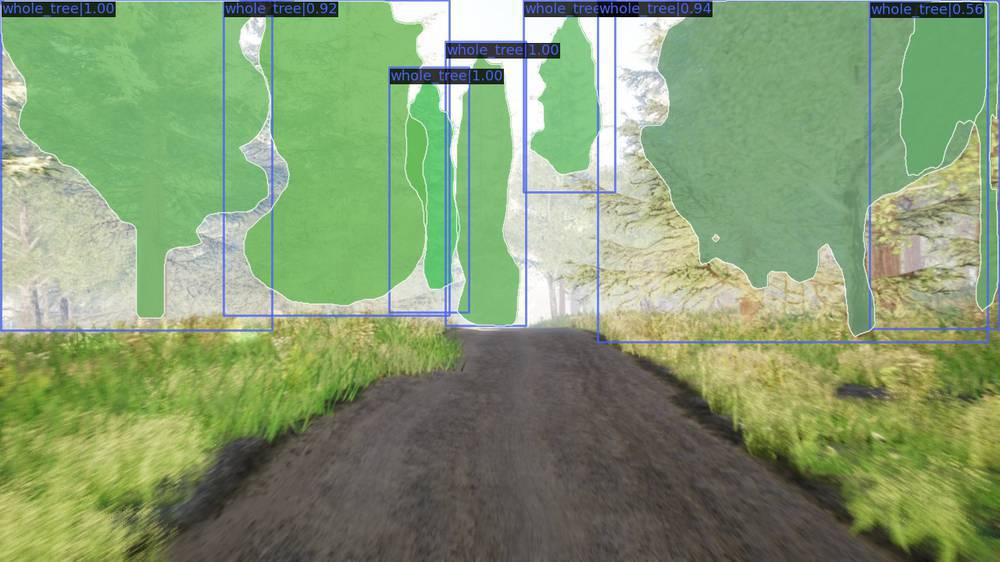} &
        \includegraphics[width=0.3\linewidth]{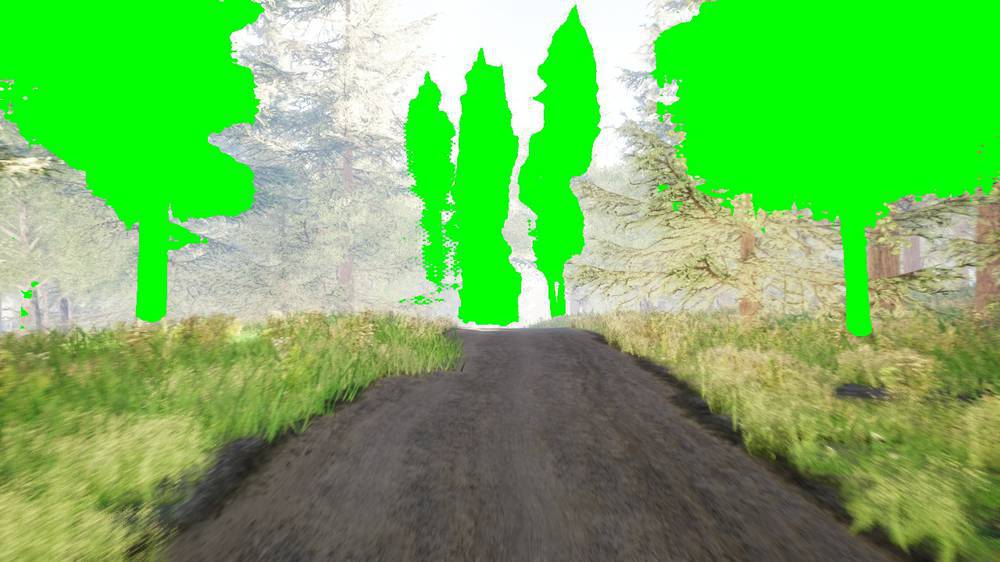} \\
        \includegraphics[width=0.3\linewidth]{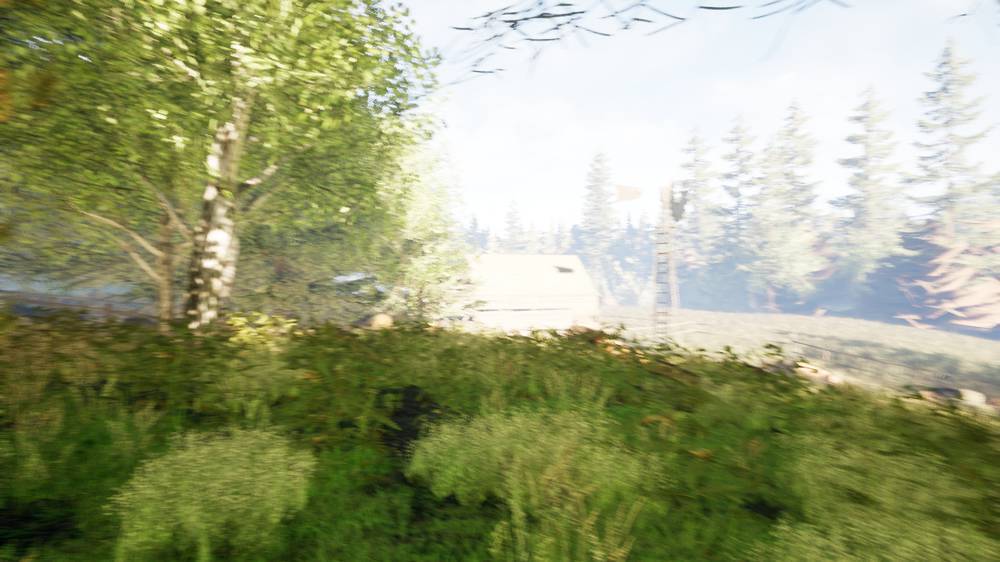} &
        \includegraphics[width=0.3\linewidth]{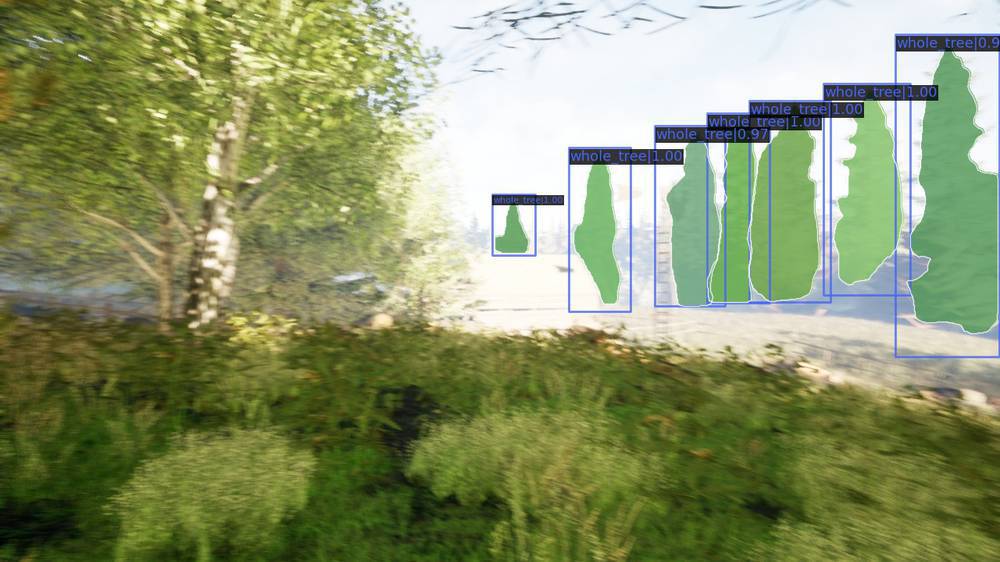} &
        \includegraphics[width=0.3\linewidth]{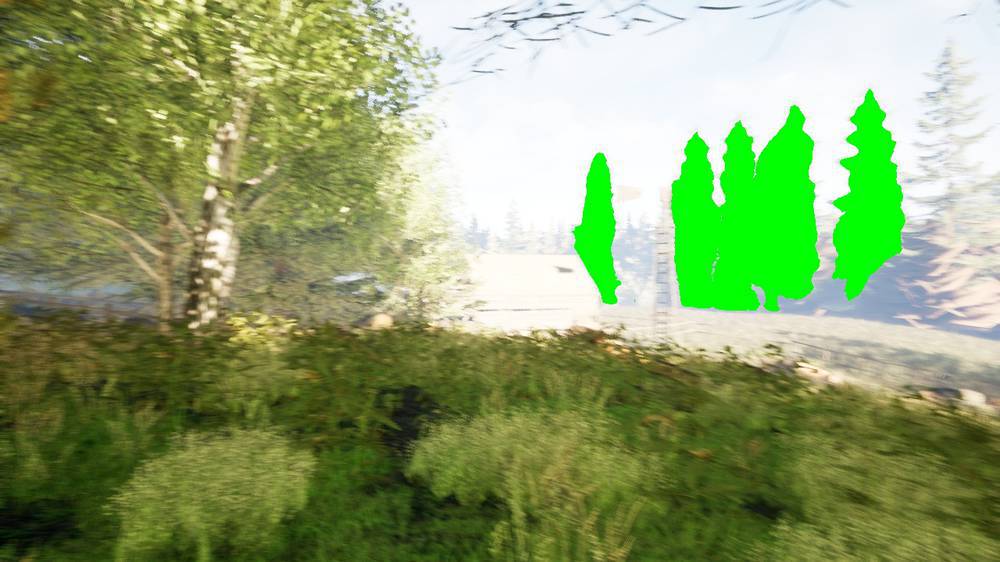} \\
        \includegraphics[width=0.3\linewidth]{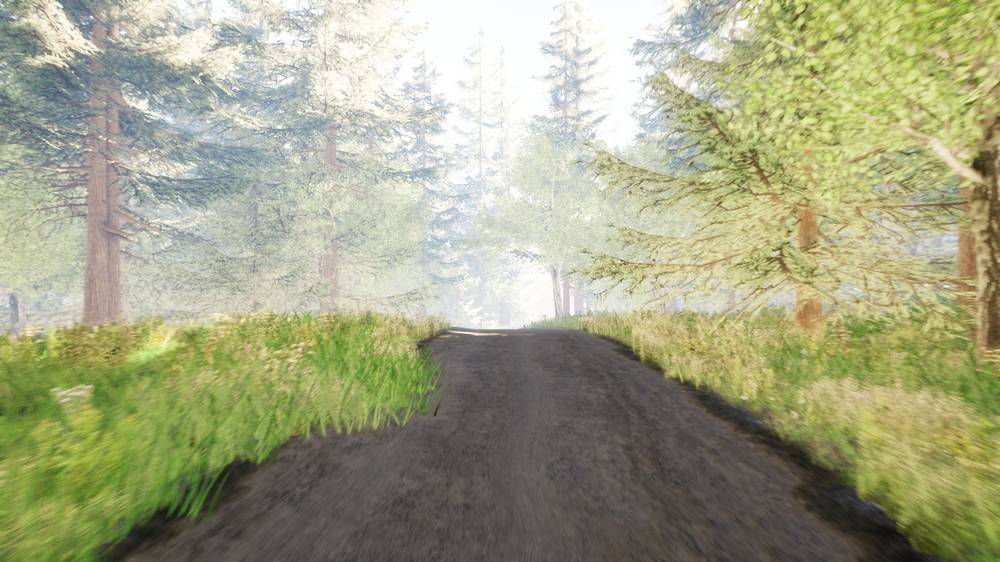} &
        \includegraphics[width=0.3\linewidth]{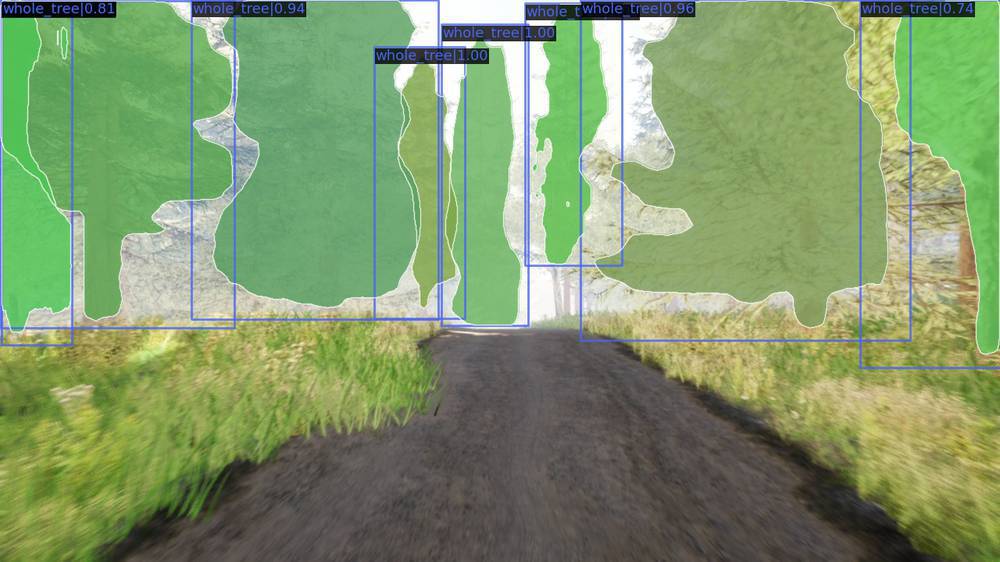} &
        \includegraphics[width=0.3\linewidth]{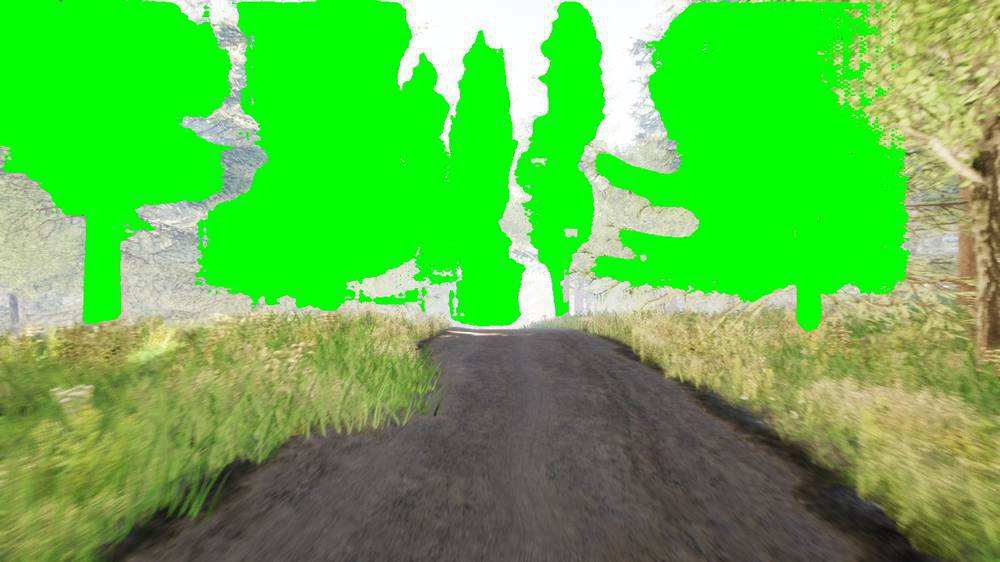} \\
        \includegraphics[width=0.3\linewidth]{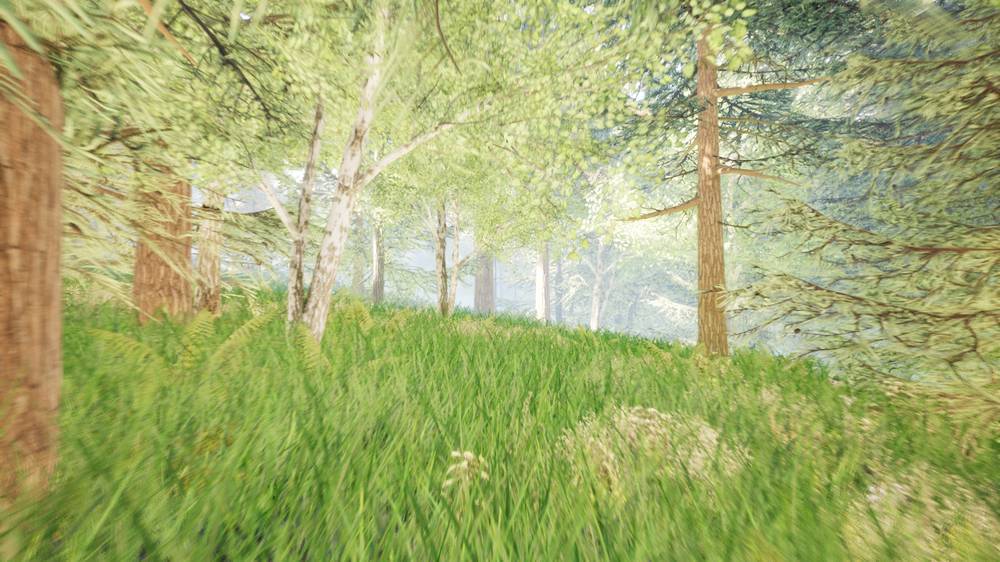} &
        \includegraphics[width=0.3\linewidth]{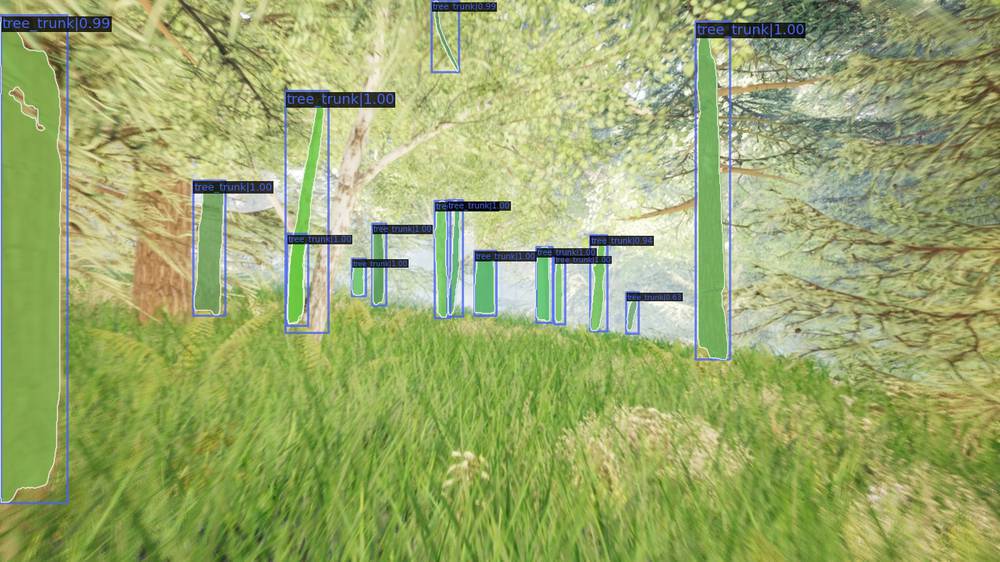} &
        \includegraphics[width=0.3\linewidth]{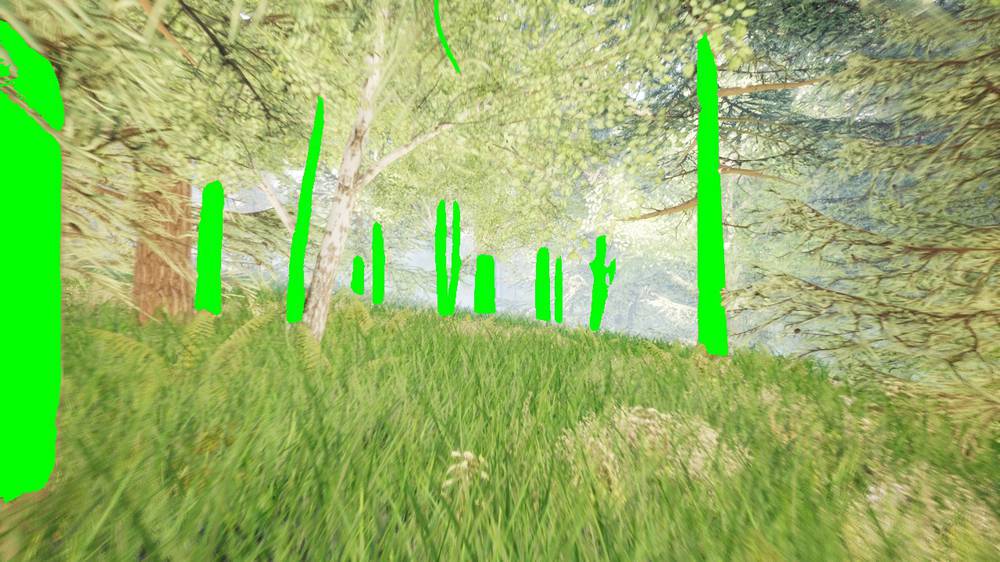} \\
        \includegraphics[width=0.3\linewidth]{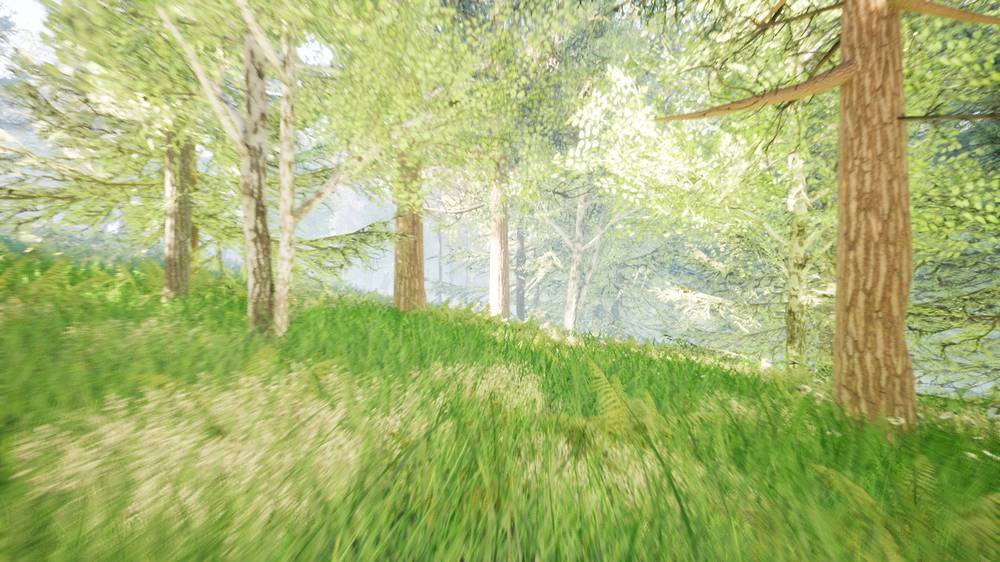} &
        \includegraphics[width=0.3\linewidth]{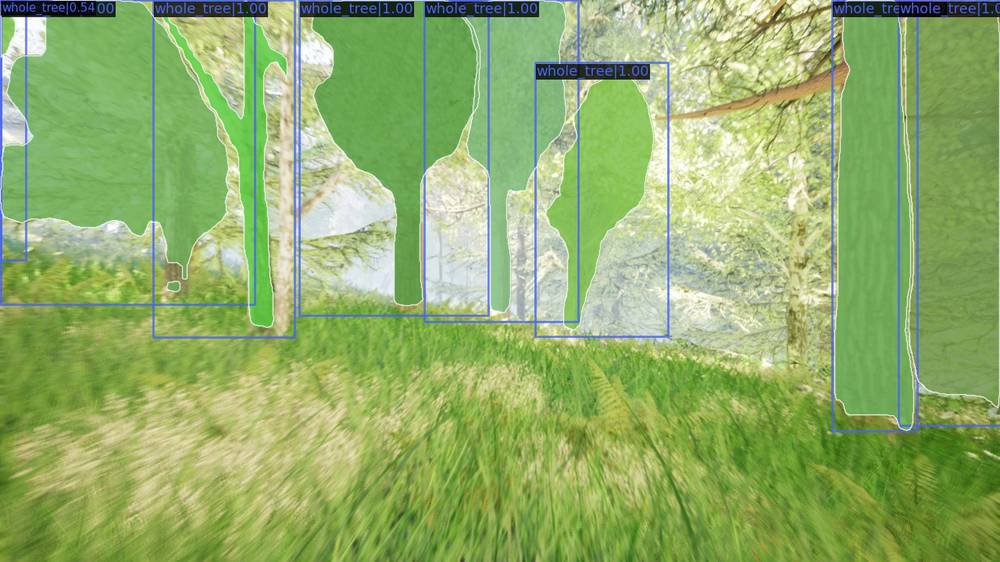} &
        \includegraphics[width=0.3\linewidth]{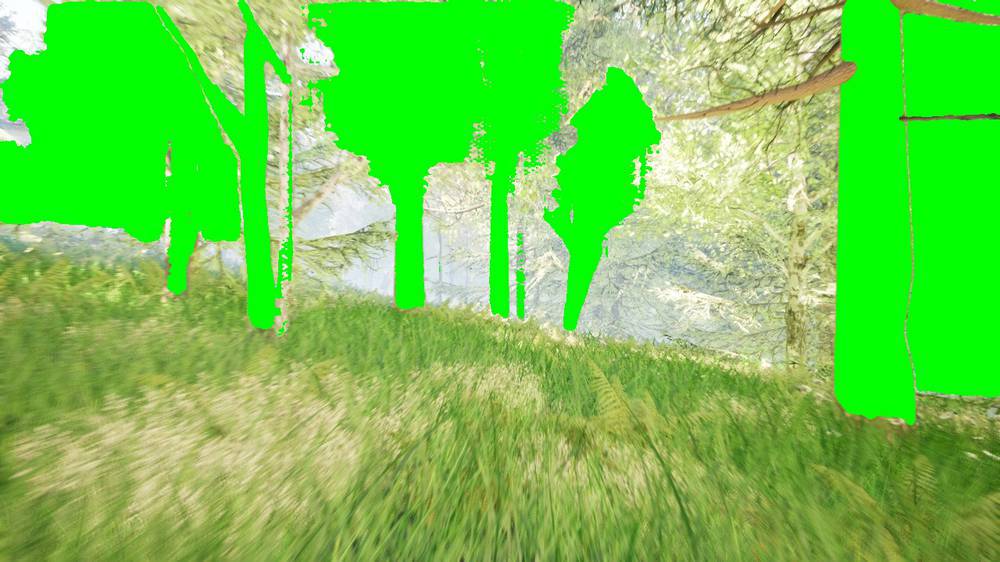} \\
        \includegraphics[width=0.3\linewidth]{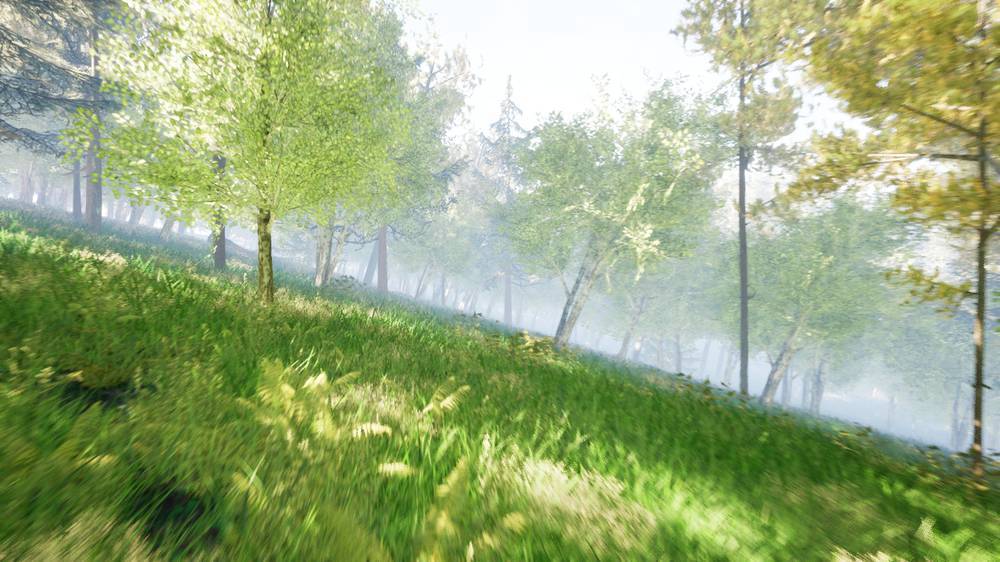} &
        \includegraphics[width=0.3\linewidth]{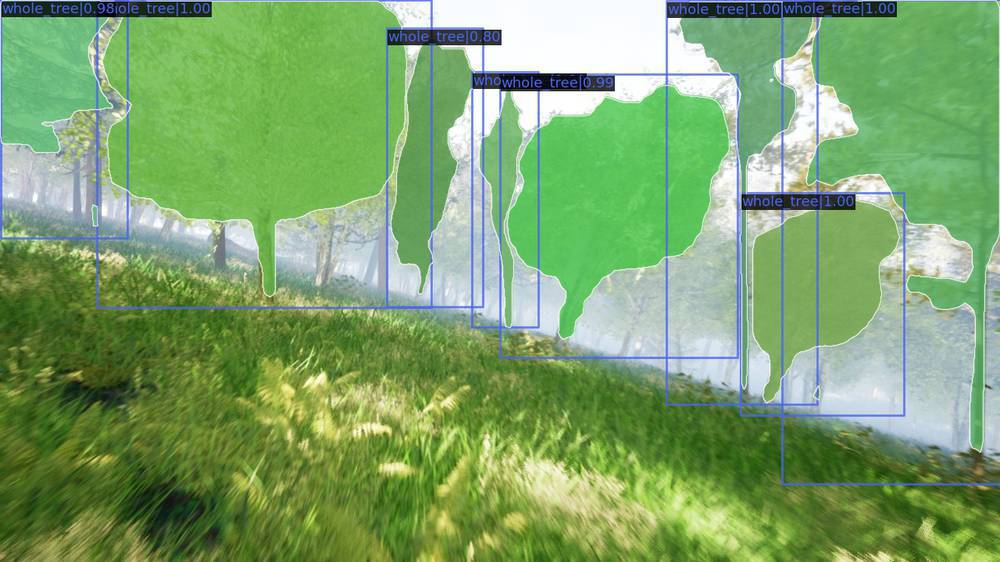} &
        \includegraphics[width=0.3\linewidth]{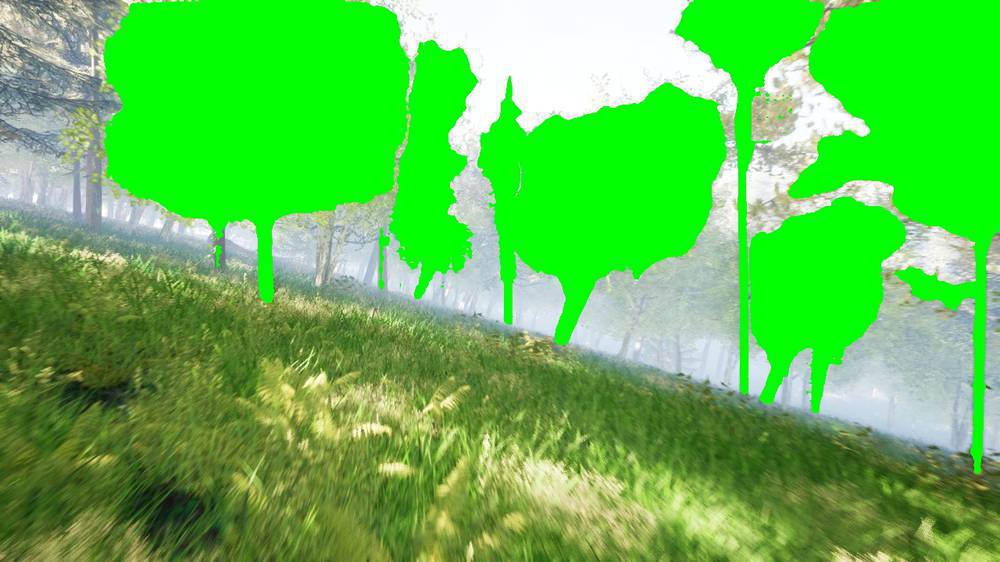} \\
        \includegraphics[width=0.3\linewidth]{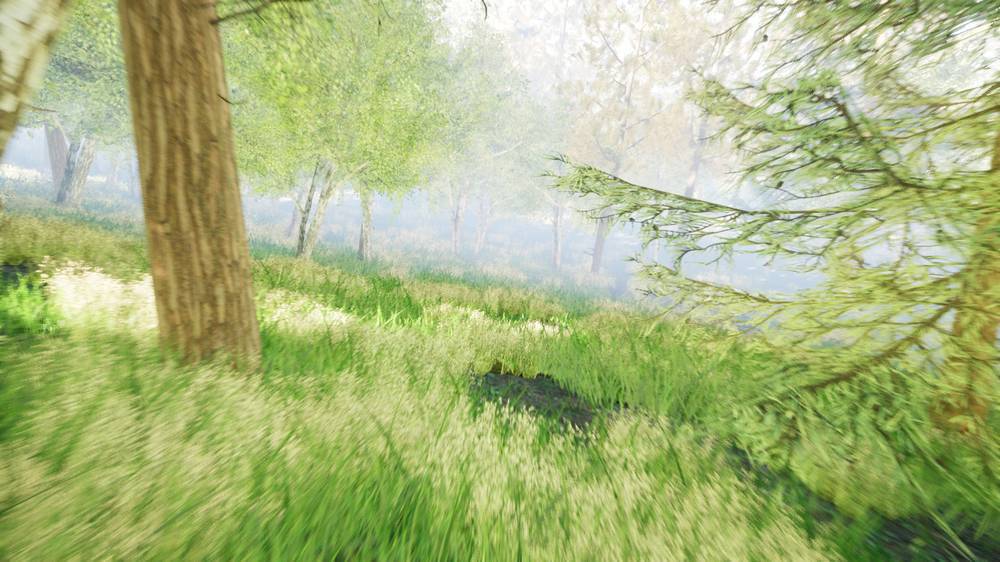} &
        \includegraphics[width=0.3\linewidth]{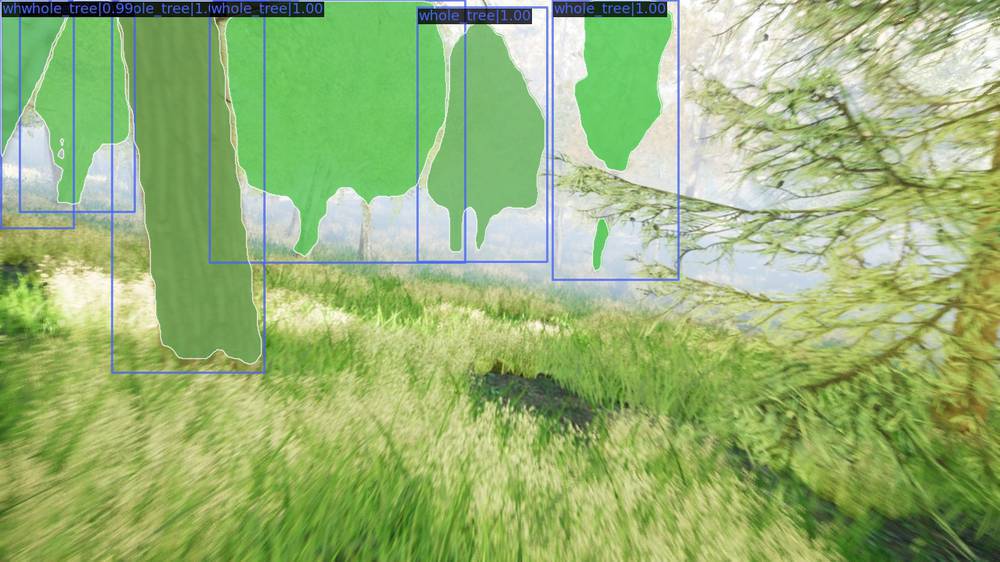} &
        \includegraphics[width=0.3\linewidth]{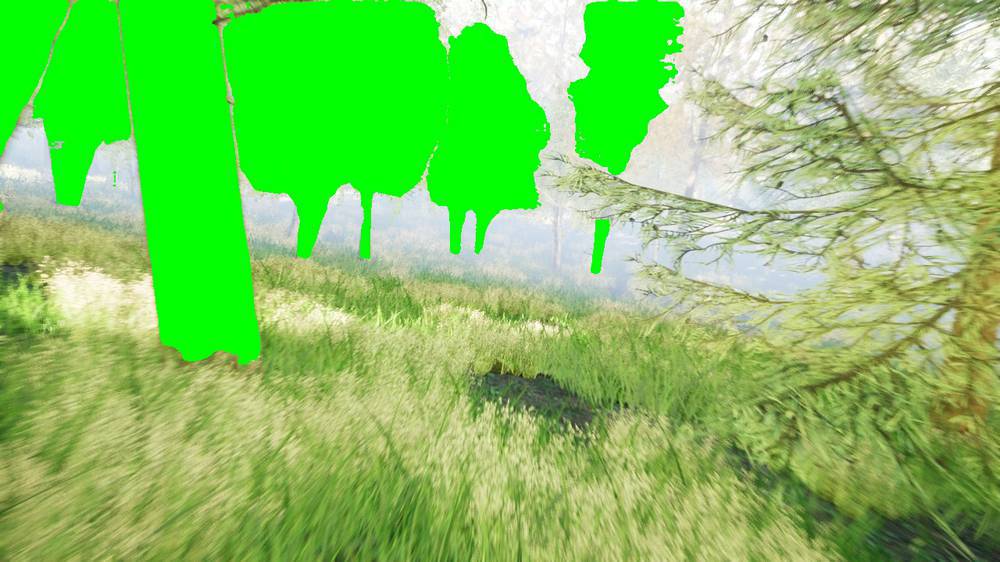} \\
    \end{tabular}
    \caption{\textbf{Phase 1:} Instance segmentation results on \textbf{Whole tree} examples. Each row shows the RGB image, model prediction, and ground truth mask.}
    \label{fig:phase1_whole_tree}
\end{figure}

Figure \ref{fig:phase1_whole_tree} shows qualitative predictions of the Mask-RCNN (Swin-T) trained on the simulated whole tree annotations. Unlike the trunk-trained teacher, this model is tasked with capturing the full crown–trunk structure, which results in substantially larger and more complex masks. Overall, the model successfully detects and segments most foreground trees, often producing complete masks that resemble the overall tree silhouettes. This confirms that the network can exploit synthetic supervision to capture both vertical stems and broader canopy regions. However, the qualitative examples also highlight several limitations. First, the predicted masks are frequently over-extended compared to the ground truth, with crowns spreading into background vegetation or sky regions. This indicates that while the model learns to cover the general tree outline, its boundaries are imprecise, particularly in regions of fine leaf structure. Second, the model tends to miss thin or distant trees in cluttered scenes, where the GT annotates multiple small instances. Third, there is evidence of mask fragmentation: in some cases, large trees are detected but split into multiple overlapping predictions, reflecting uncertainty in defining object extents. Finally, we observe that the model occasionally produces coarse vertical shapes for occluded trees, suggesting that the crown prior is weak when visibility is low. Taken together, these examples suggest that the whole-tree teacher is effective at providing global tree structure priors but less reliable at capturing fine detail, particularly for small or highly occluded instances. In the context of distillation, this means that the whole-tree teacher offers complementary information to the trunk teacher while the trunk model excels at precise stem localization, the whole-tree model provides richer, but noisier, canopy context.

\clearpage
\subsection{Phase 2}

\begin{figure}[ht]
	\centering
	\renewcommand{\arraystretch}{1.2}
	\setlength{\tabcolsep}{1pt}
	
	\begin{tabular}{c c c}
		\textbf{RGB Image} & \textbf{Prediction} & \textbf{Ground Truth} \\
		\includegraphics[width=0.28\linewidth, angle=180, origin=c]{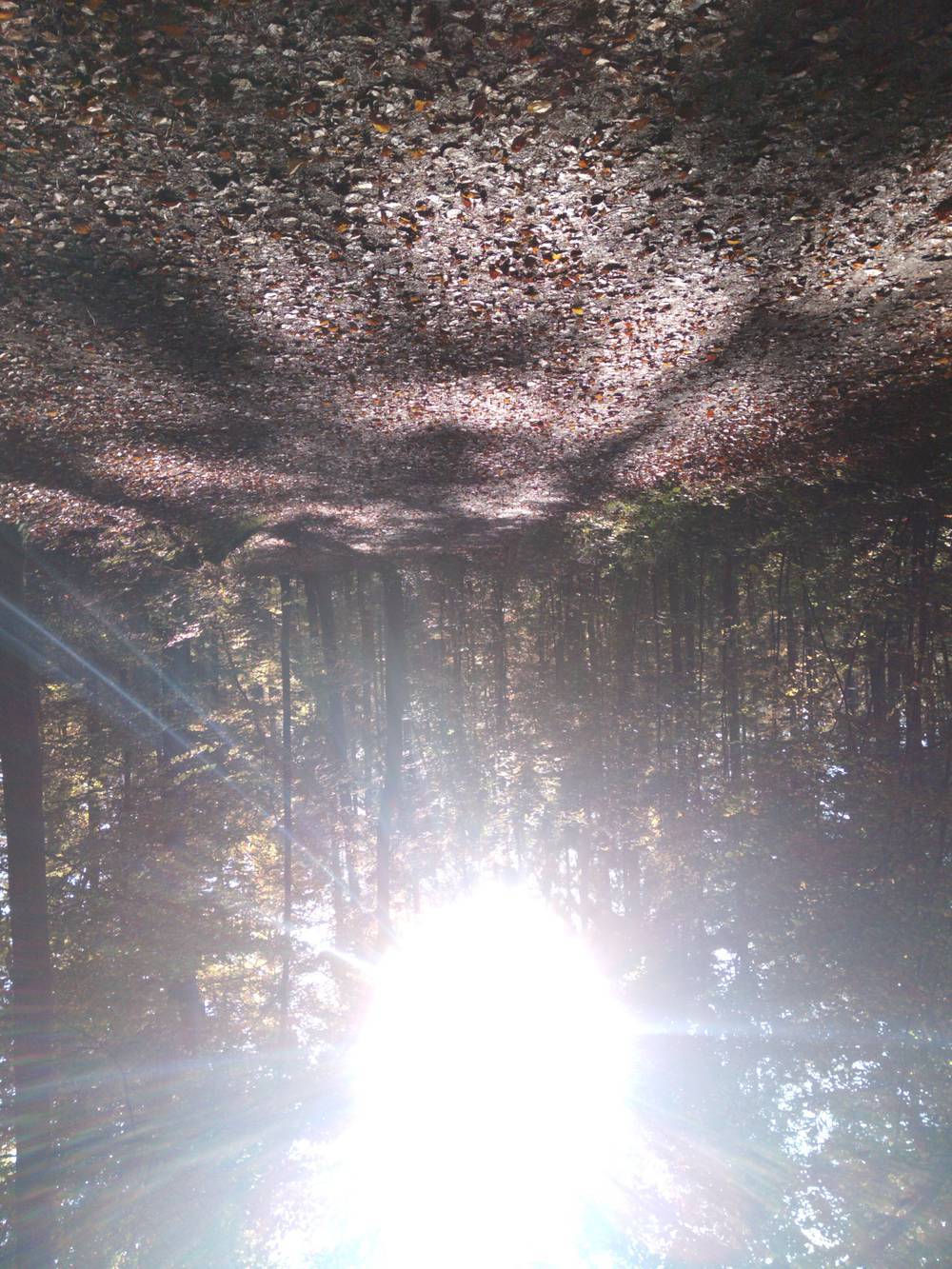} &
		\includegraphics[width=0.28\linewidth]{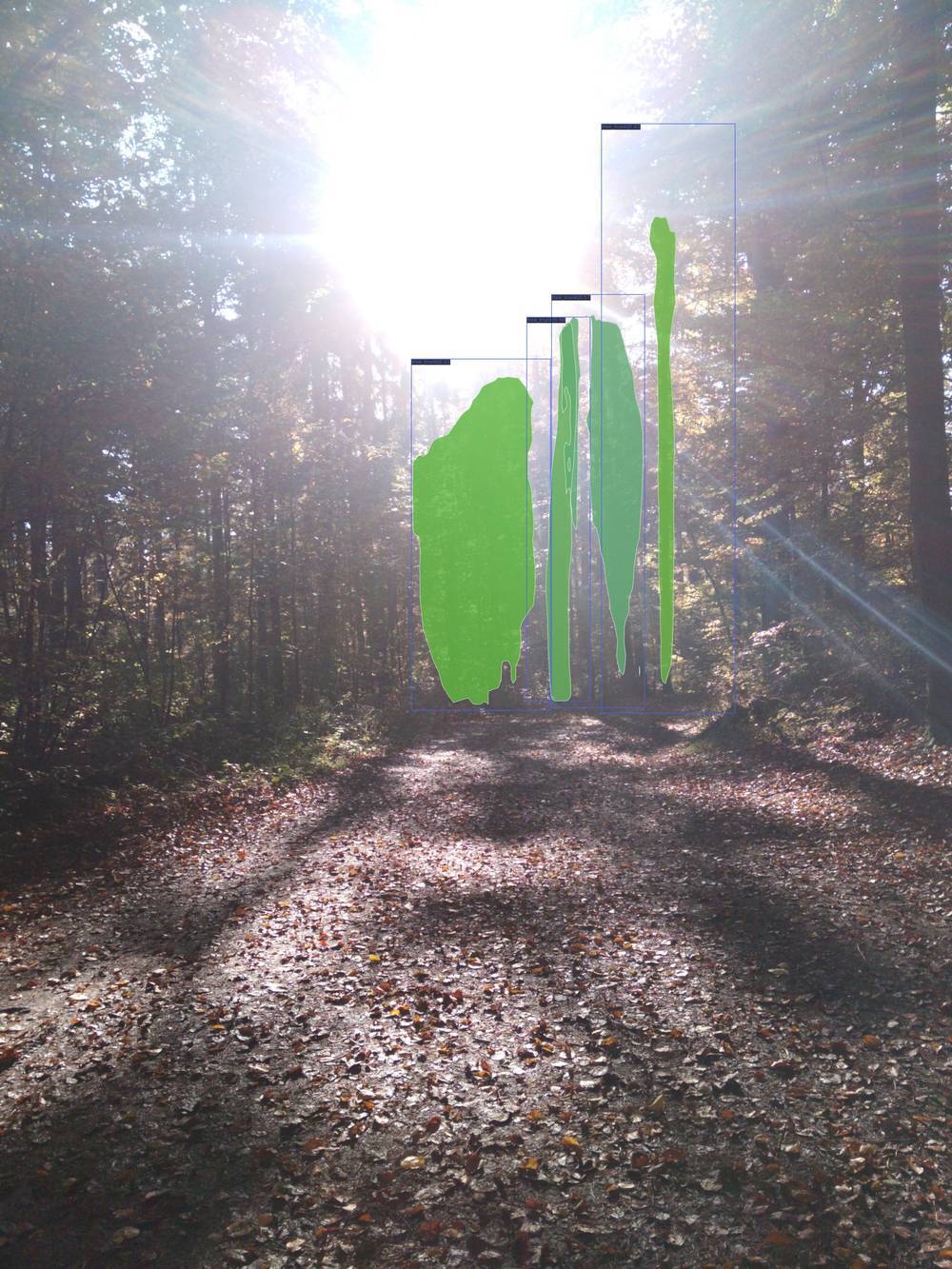} &
		\includegraphics[width=0.28\linewidth]{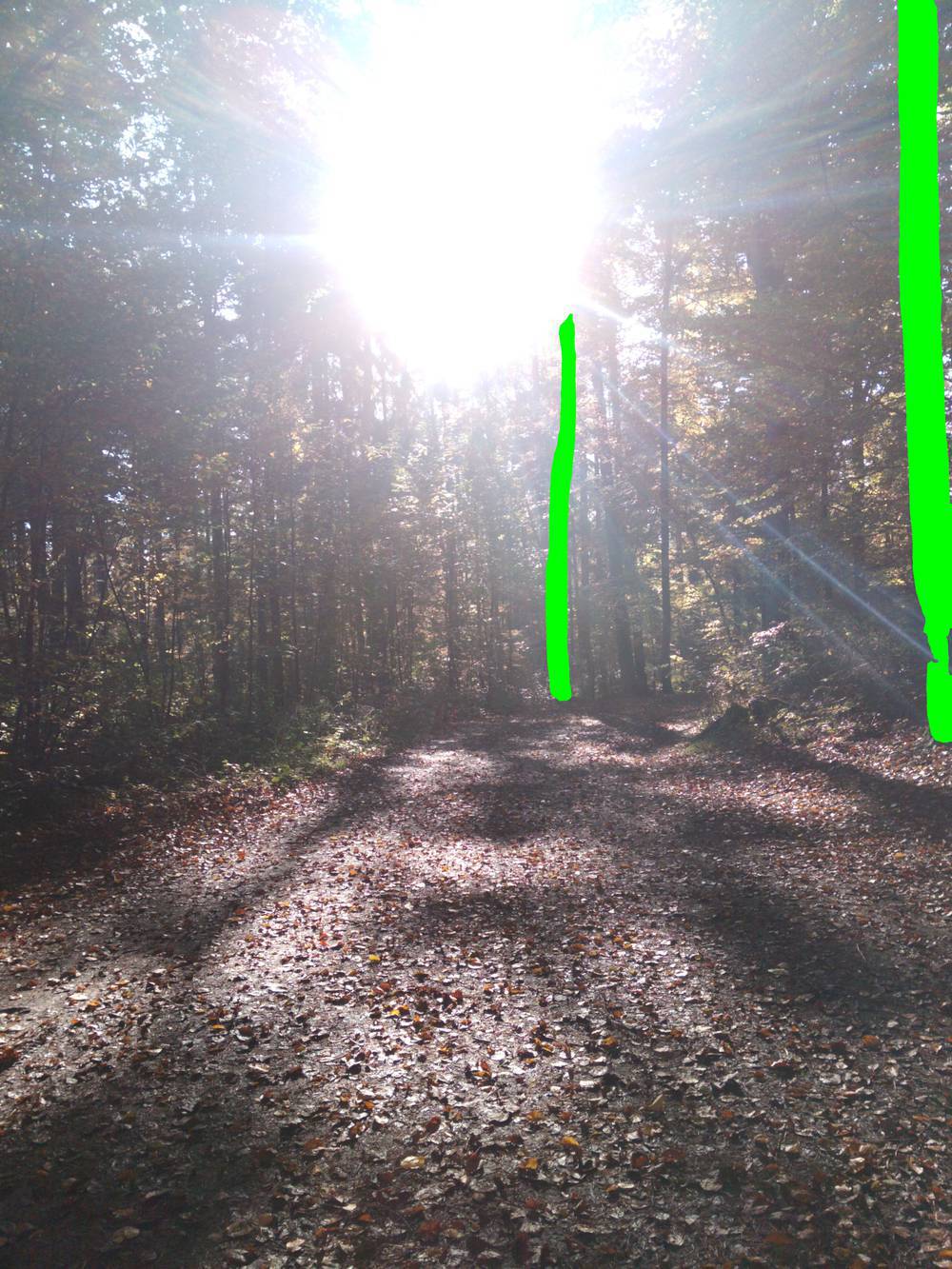} \\
		\includegraphics[width=0.28\linewidth, angle=180, origin=c]{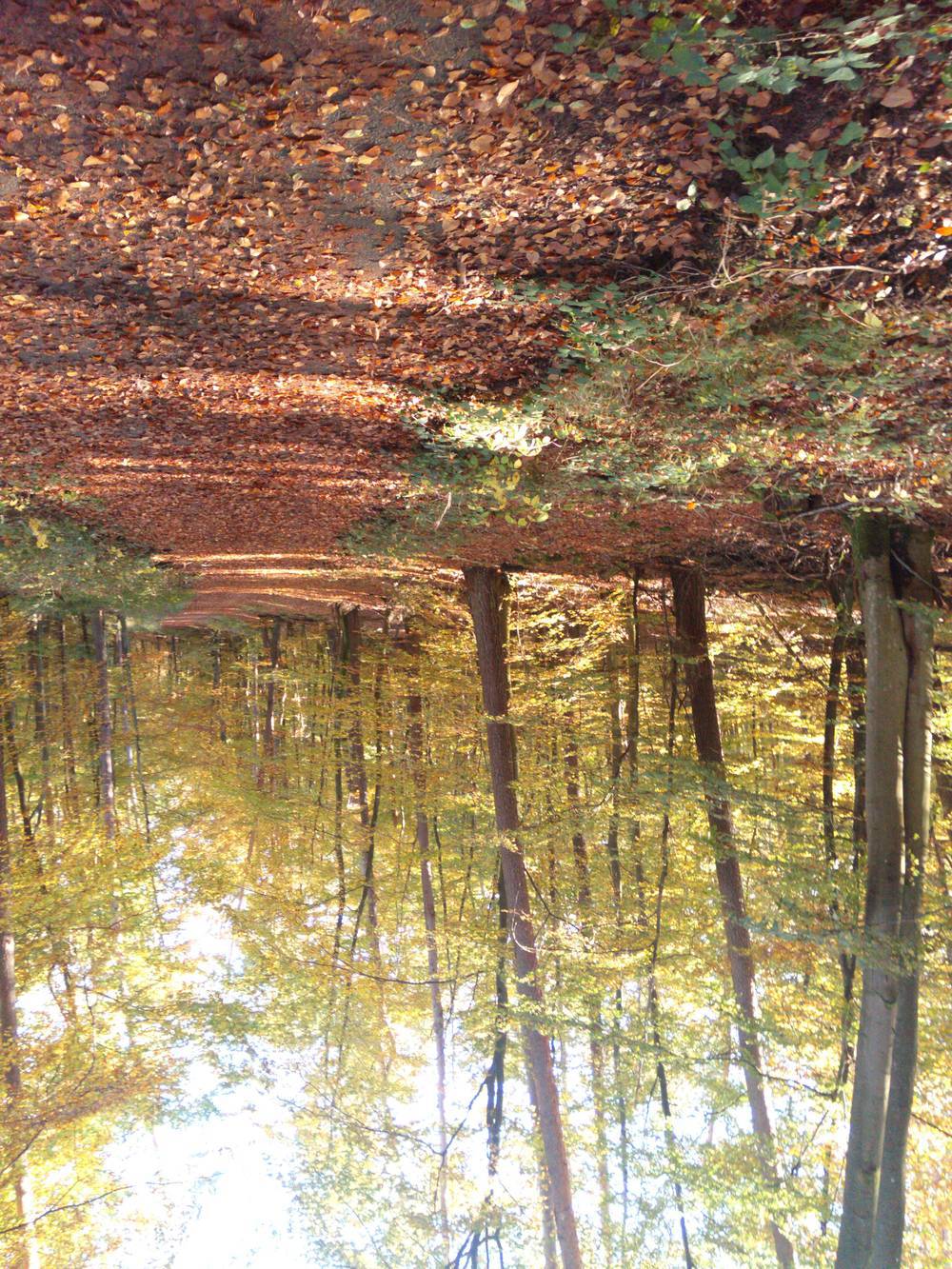} &
		\includegraphics[width=0.28\linewidth]{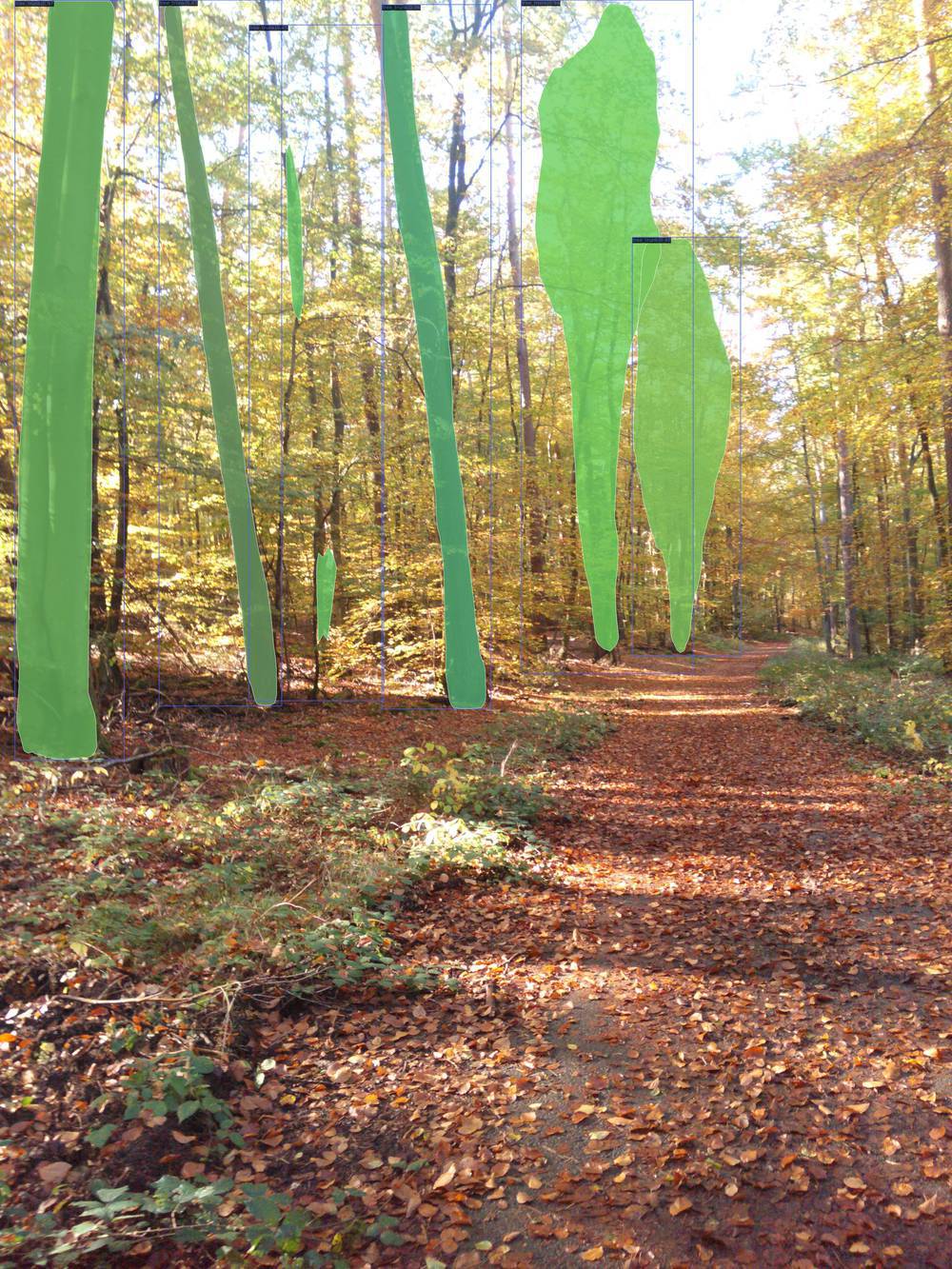} &
		\includegraphics[width=0.28\linewidth]{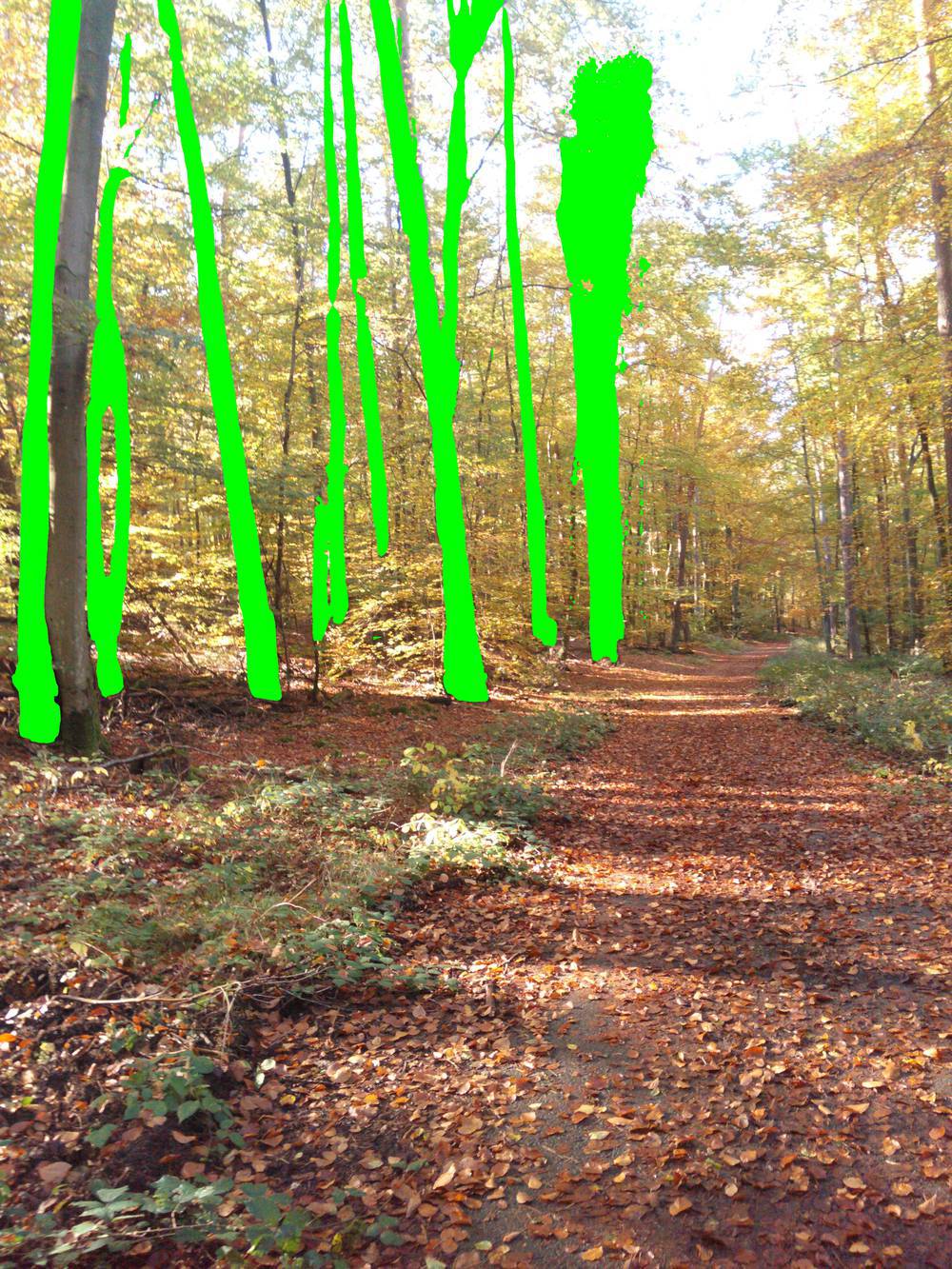} \\
		\includegraphics[width=0.28\linewidth, angle=180, origin=c]{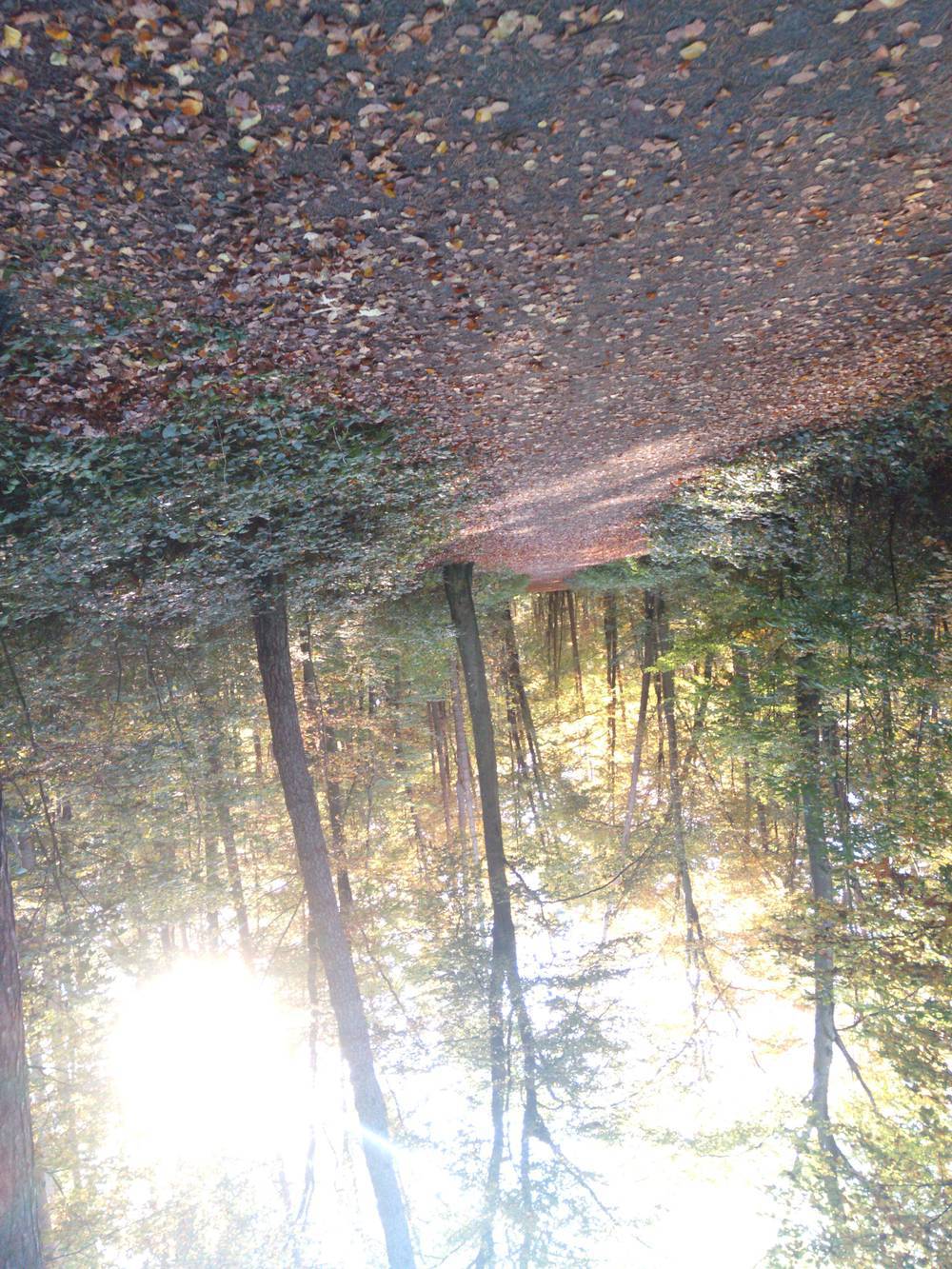} &
		\includegraphics[width=0.28\linewidth]{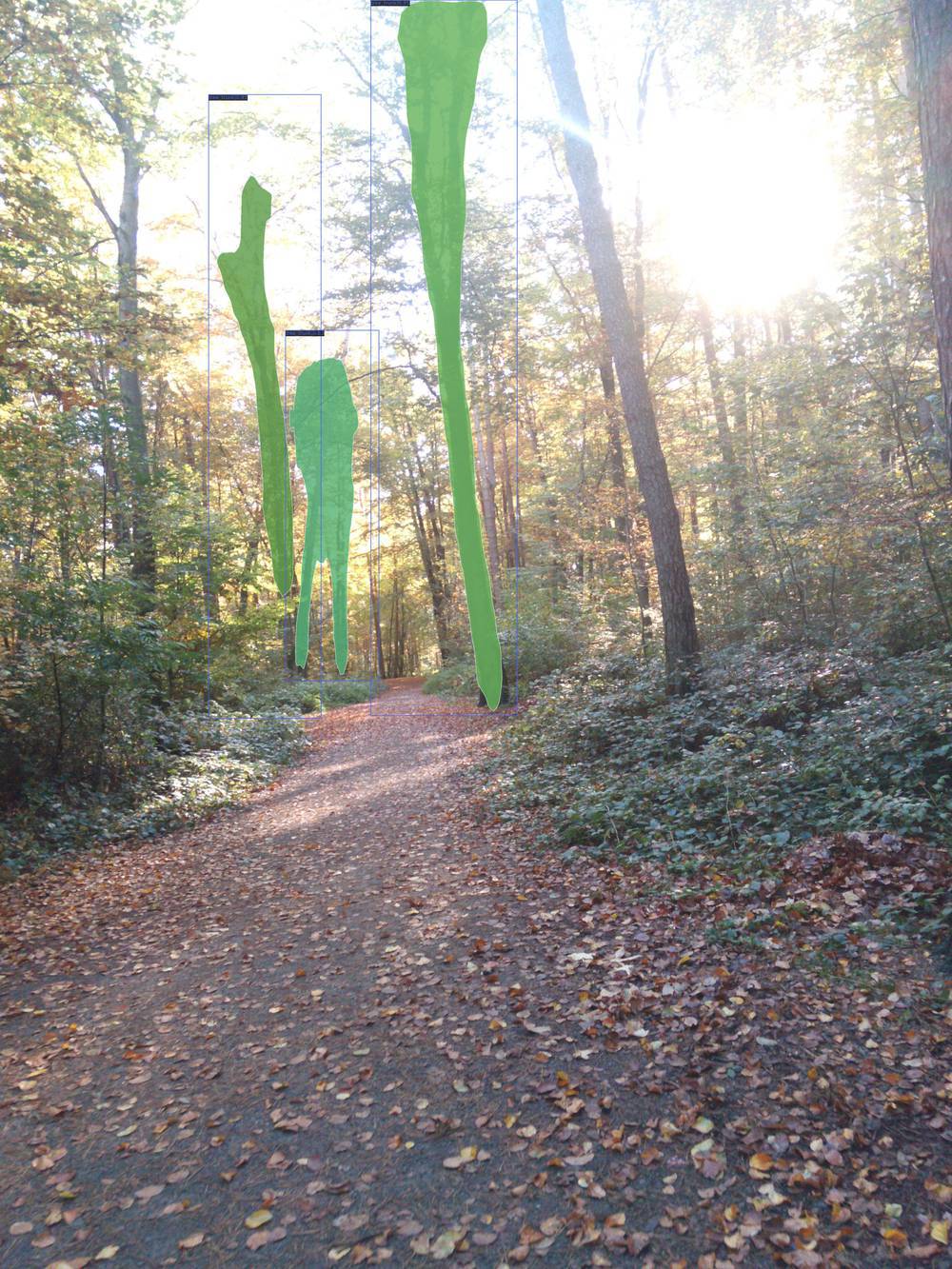} &
		\includegraphics[width=0.28\linewidth]{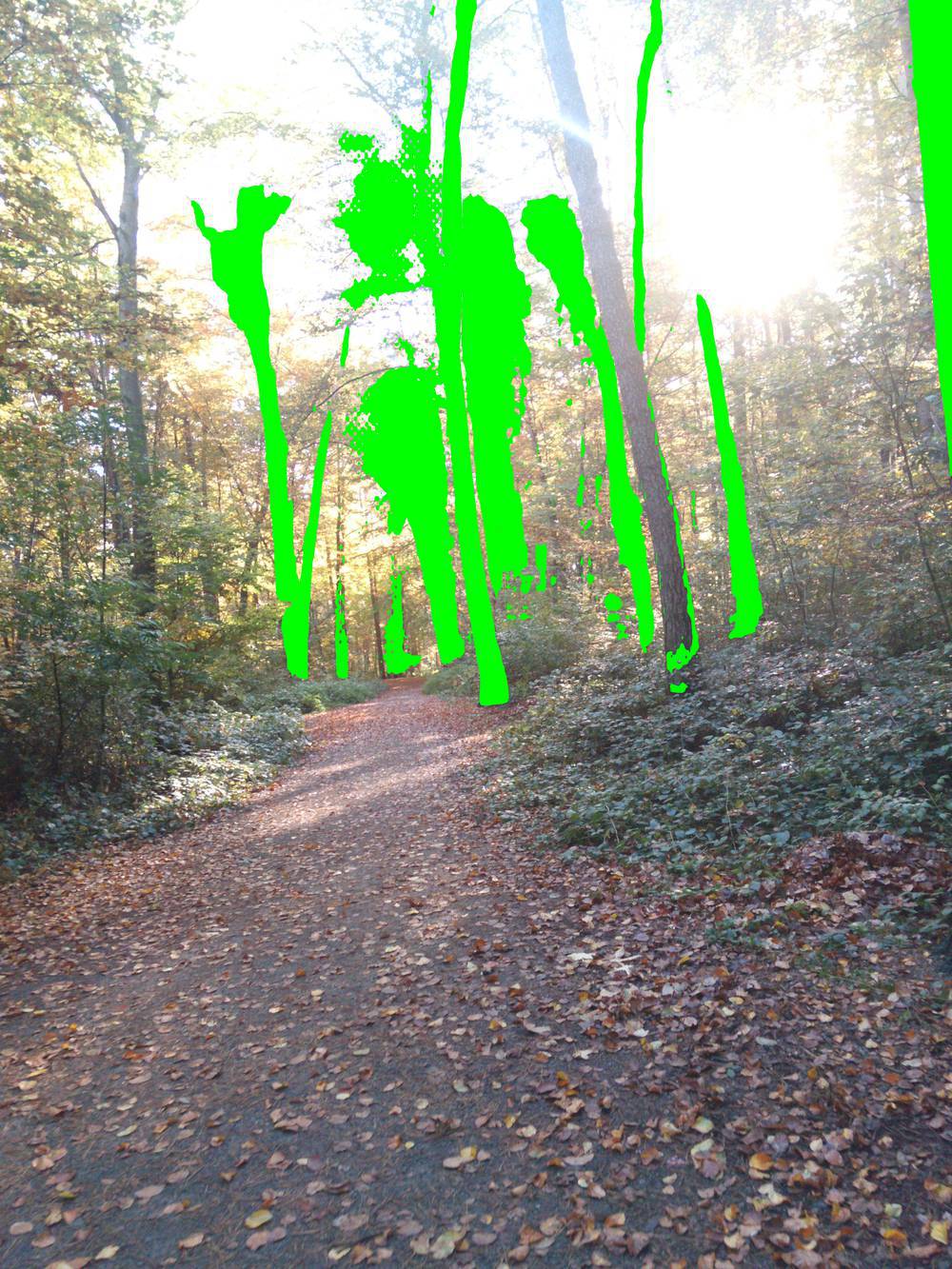} \\
		\includegraphics[width=0.28\linewidth, angle=180, origin=c]{images/inference_results/tt_real/IMG_160101_000659_0113_RGB.JPG.jpg} &
		\includegraphics[width=0.28\linewidth]{images/inference_results/tt_real/IMG_160101_000659_0113_RGB.JPG_pred.jpg} &
		\includegraphics[width=0.28\linewidth]{images/inference_results/tt_real/IMG_160101_000659_0113_RGB.JPG_gt.jpg} \\
	\end{tabular}
	\caption{\textbf{Phase 2:} Domain transfer (Tree Trunk $\rightarrow$ Real Trees).}
	\label{fig:phase2_trunk2real}
\end{figure}

Figure \ref{fig:phase2_trunk2real} presents qualitative examples of the synthetic trunk-trained teacher evaluated on real forest images (and table \ref{tab:sim2real_combined} the quantitative results). The domain transfer clearly exposes the limitations of cross-domain generalization. While the model occasionally detects a subset of prominent vertical structures, the predictions are often coarse, incomplete, or misaligned compared to the annotated ground truth. Overexposure, causes the model to collapse multiple thin trunks into a few thick, blob-like predictions, demonstrating sensitivity to extreme lighting conditions absent in the synthetic training data. In other examples (row 4), only 1-2 large trunks are segmented while the majority of thinner stems remain undetected, reflecting a consistent failure on small and cluttered instances. Where detections do occur, the masks are overly thick and lack fine boundary detail, in contrast to the thin GT annotations. This shows that the model has learned a bias towards detecting \textbf{“generic vertical masses”} rather than precise tree stem structures when transferred to real complex forestry images.

\begin{table*}[t]
	\centering
		\caption{\textbf{Phase 2: } Validation metrics for direct Sim$\rightarrow$Real transfer on the validation set. Bounding box and segmentation metrics are separated by a dashed line.}
	\resizebox{\textwidth}{!}{
		\begin{tabular}{l l c c c c c c c c c c c}
			\toprule
			\textbf{Model} & \textbf{Type} & \textbf{Epochs} 
			& \textbf{mAP@0.5:0.95} & \textbf{mAP@0.5} & \textbf{mAP@0.75} 
			& \textbf{APs} & \textbf{APm} & \textbf{API} 
			& \textbf{mAR@100} & \textbf{AR@300} & \textbf{AR@1000} & \textbf{ARI} \\
			\midrule
			\multicolumn{13}{c}{\textbf{Bounding Box Metrics}} \\
			\midrule
			Mask-RCNN (Swin-T) & Tree trunk & 24 & \textbf{0.225} & 0.445 & 0.205 & -1.000 & -1.000 & 0.225 & 0.364 & 0.364 & 0.364 & 0.364 \\
			Mask-RCNN (Swin-T) & Whole tree & 24 & 0.066 & 0.191 & 0.038 & -1.000 & -1.000 & 0.066 & 0.178 & 0.178 & 0.178 & 0.178 \\
			\hdashline
			\multicolumn{13}{c}{\textbf{Segmentation Metrics}} \\
			\midrule
			Mask-RCNN (Swin-T) & Tree trunk & 24 & \textbf{0.147} & 0.300 & 0.127 & -1.000 & -1.000 & 0.147 & 0.249 & 0.249 & 0.249 & 0.249 \\
			Mask-RCNN (Swin-T) & Whole tree & 24 & 0.039 & 0.104 & 0.024 & -1.000 & -1.000 & 0.039 & 0.131 & 0.131 & 0.131 & 0.131 \\
			\bottomrule
	\end{tabular}}
	\label{tab:sim2real_combined}
\end{table*}

These results illustrate \textbf{two important insights}: (i) the trunk teacher provides some localization capability even in the target domain, suggesting transferable structural priors, but (ii) the predictions are noisy and incomplete due to exposure gaps (lighting, texture, background clutter) between simulation and reality. This validates our Phase 2 analysis i.e. domain gap is a major bottleneck, and direct teacher transfer cannot serve as a deployable solution without adaptation. Instead, its role is most useful in distillation (as further explored in Phase 4), where coarse but informative priors can be combined with real supervision to yield a more reliable student.

\begin{figure}[htp!]
    \centering
    \renewcommand{\arraystretch}{1.2}
    \setlength{\tabcolsep}{1pt}
    \begin{tabular}{c c c}
        \textbf{RGB Image} & \textbf{Prediction} & \textbf{Ground Truth} \\
        \includegraphics[width=0.28\linewidth, angle=180, origin=c]{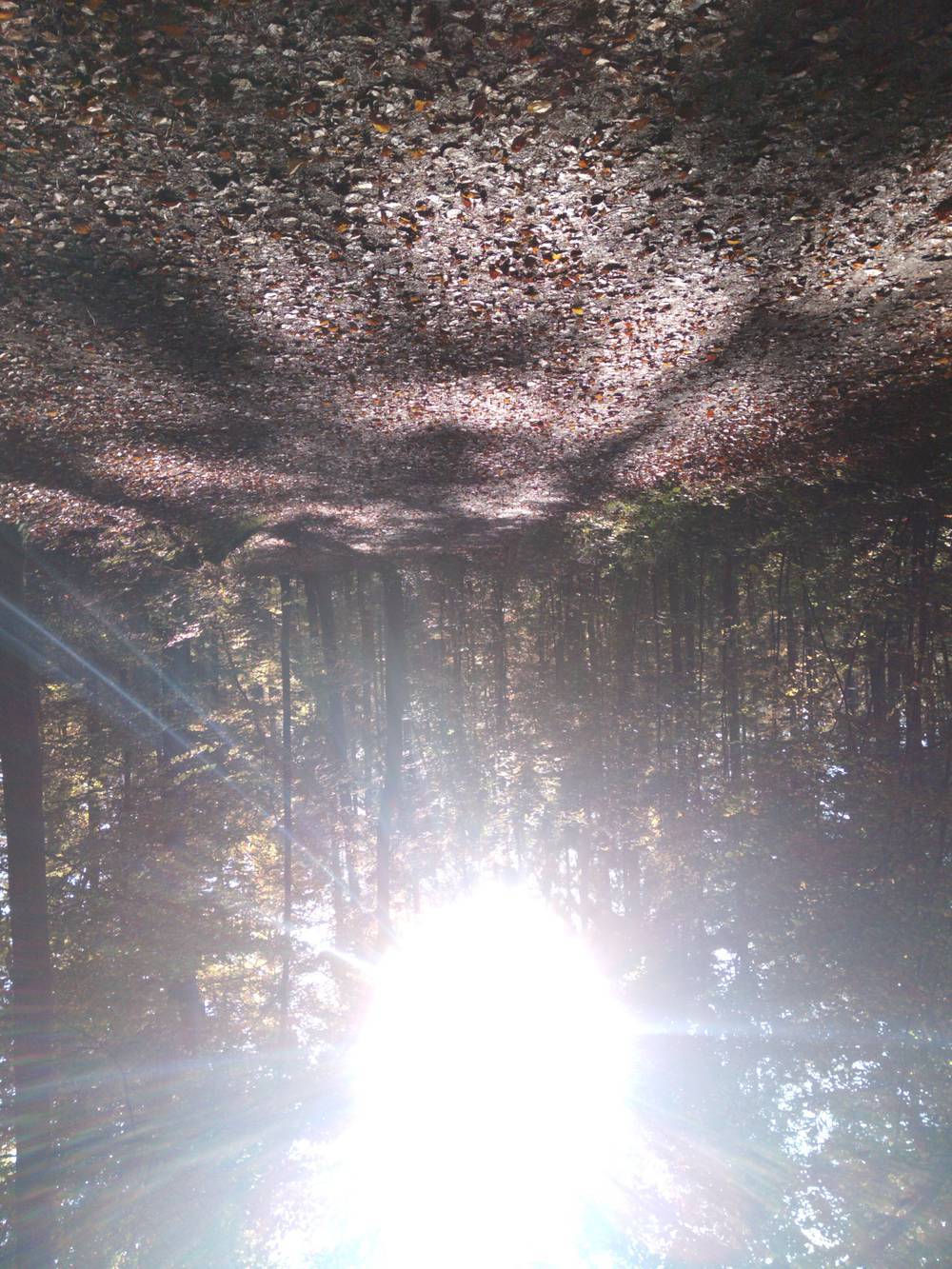} &
        \includegraphics[width=0.28\linewidth]{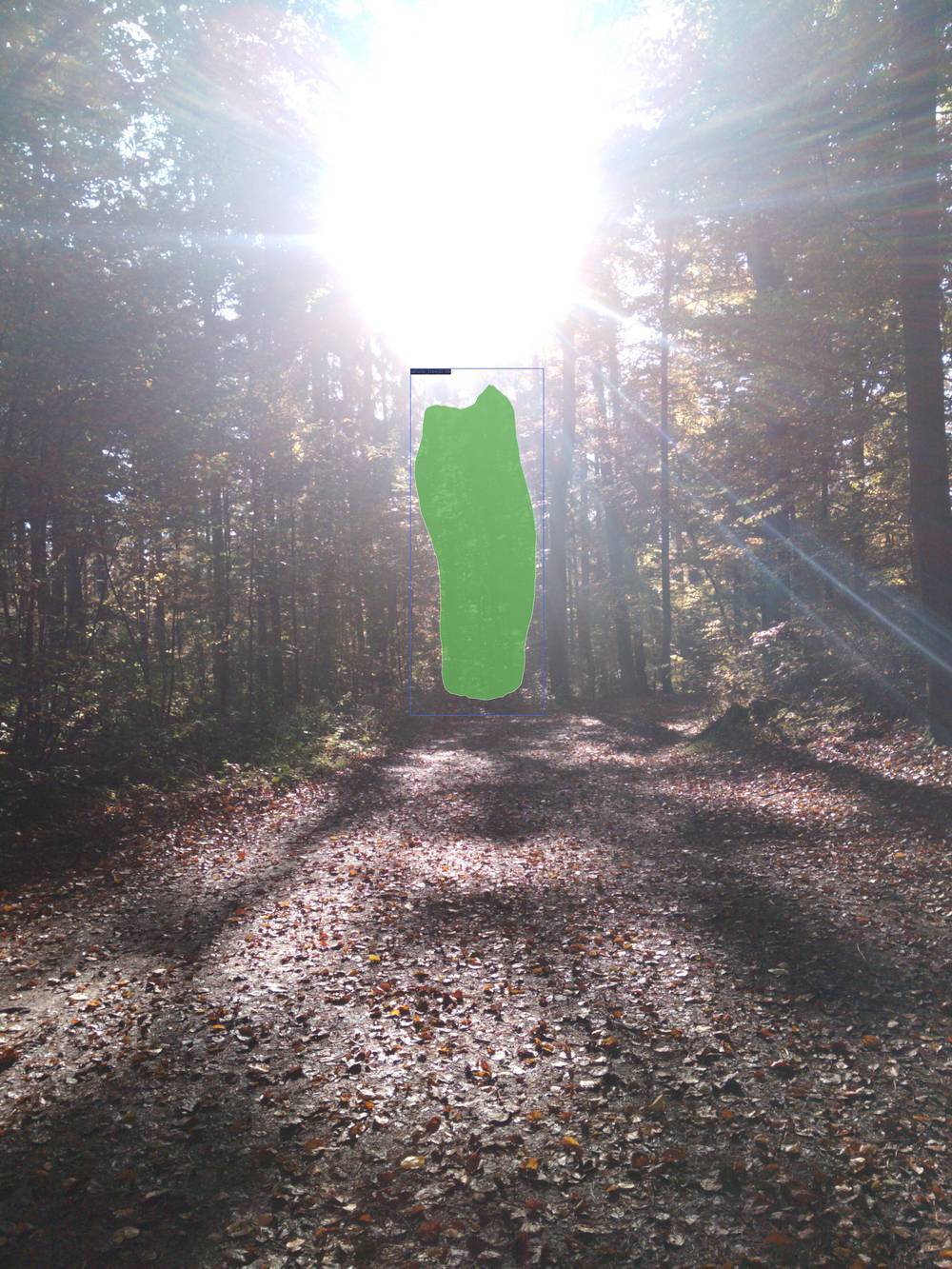} &
        \includegraphics[width=0.28\linewidth]{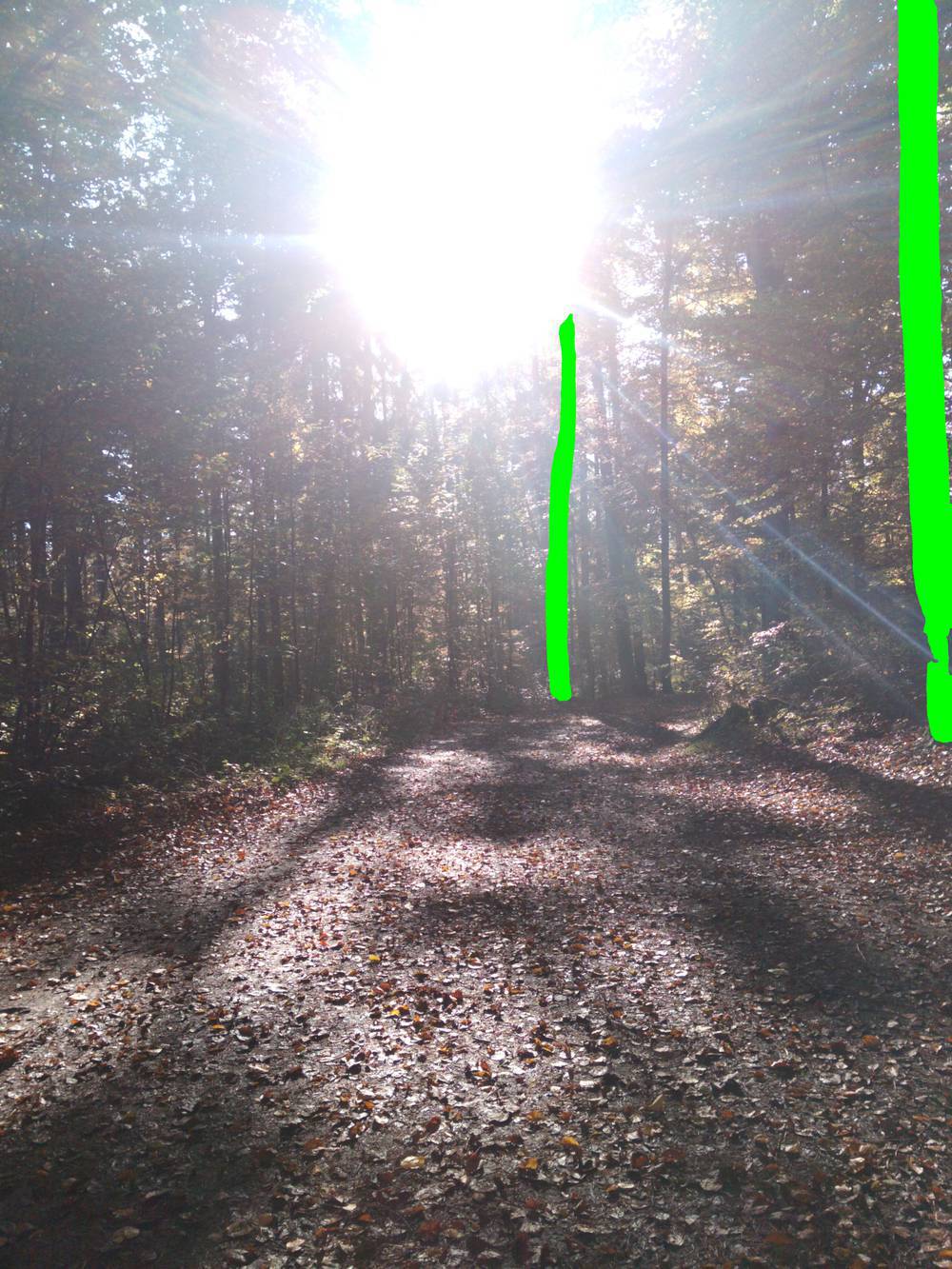} \\
        \includegraphics[width=0.28\linewidth, angle=180, origin=c]{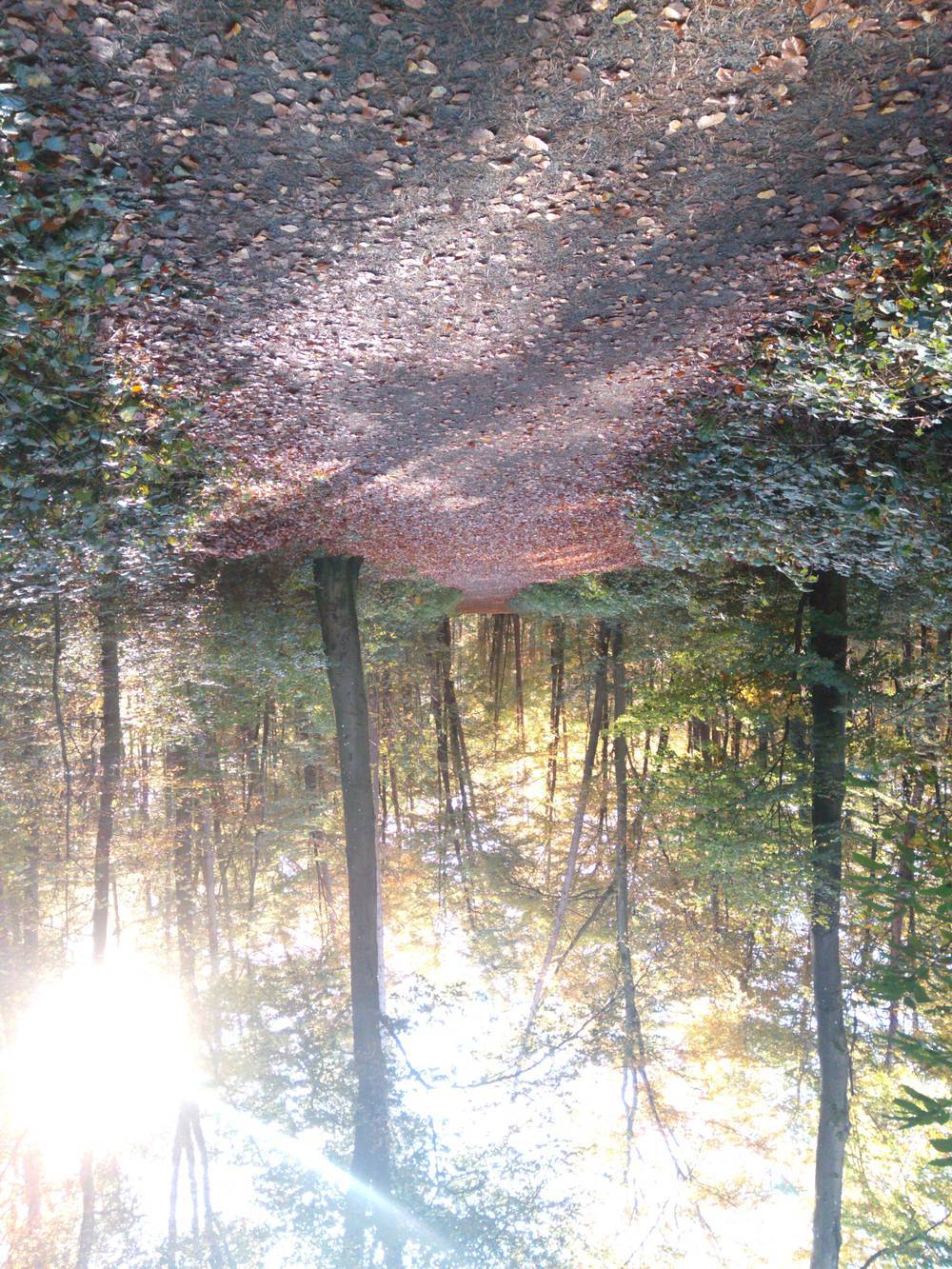} &
        \includegraphics[width=0.28\linewidth]{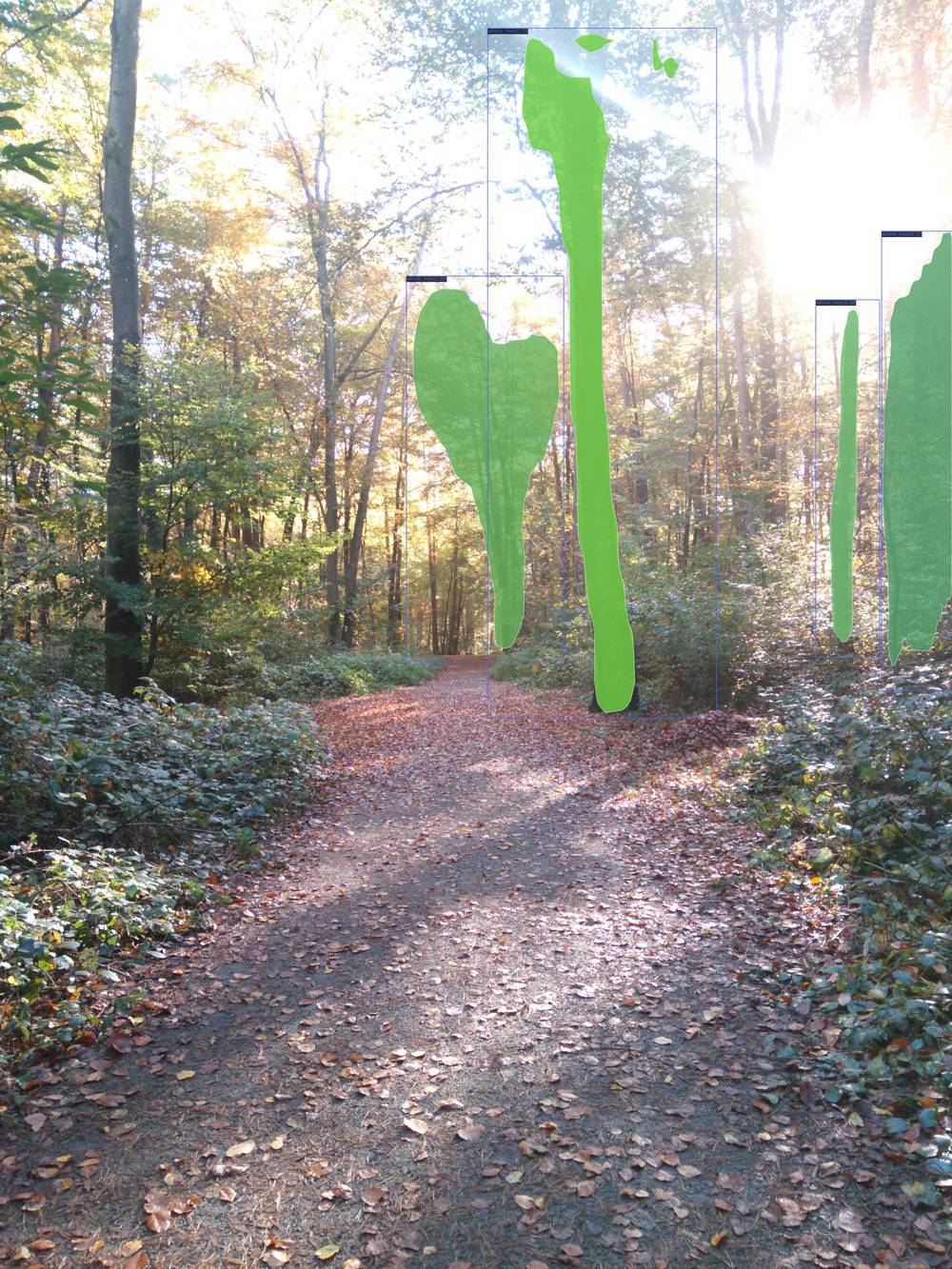} &
        \includegraphics[width=0.28\linewidth]{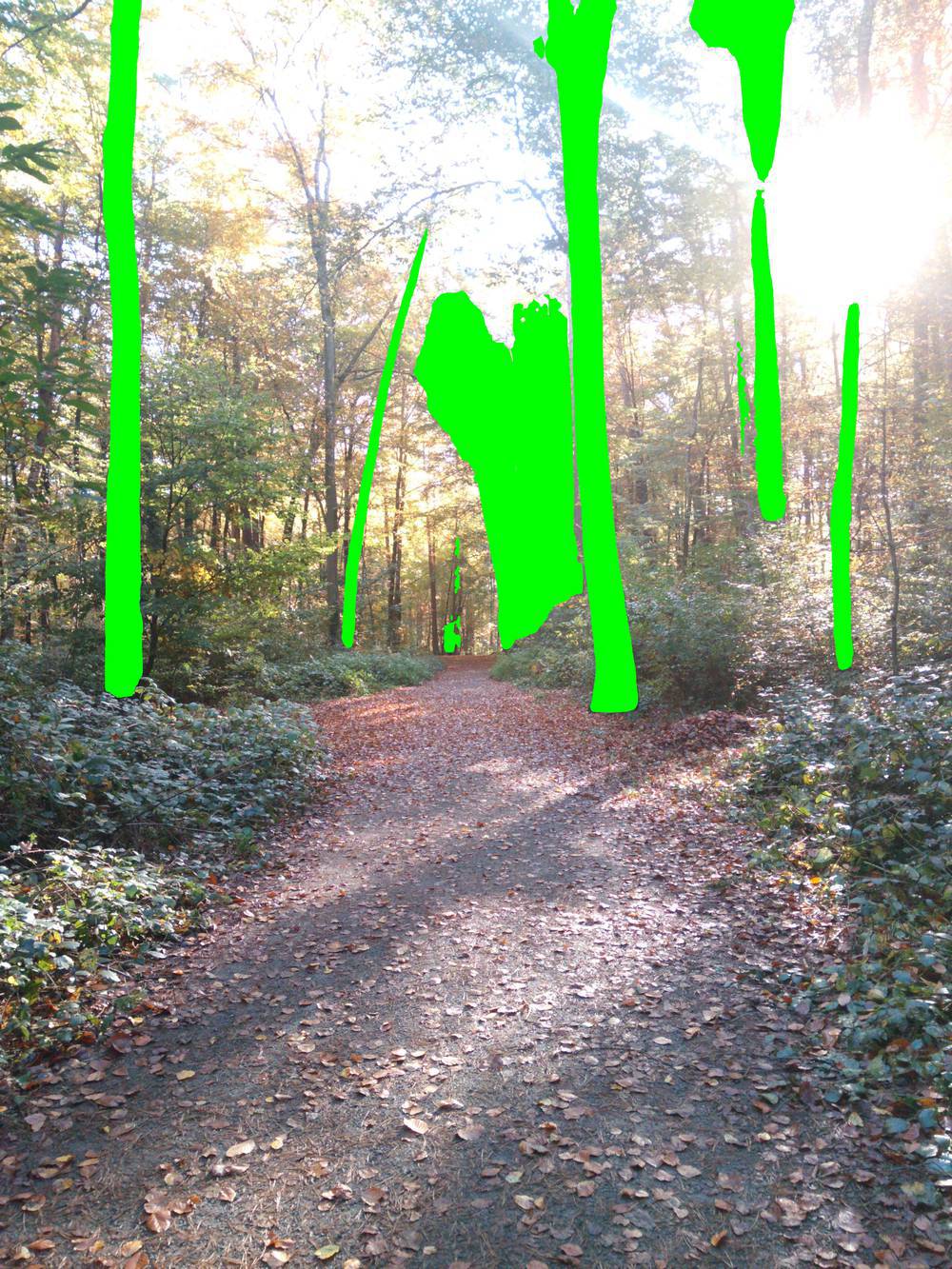} \\
        \includegraphics[width=0.28\linewidth, angle=180, origin=c]{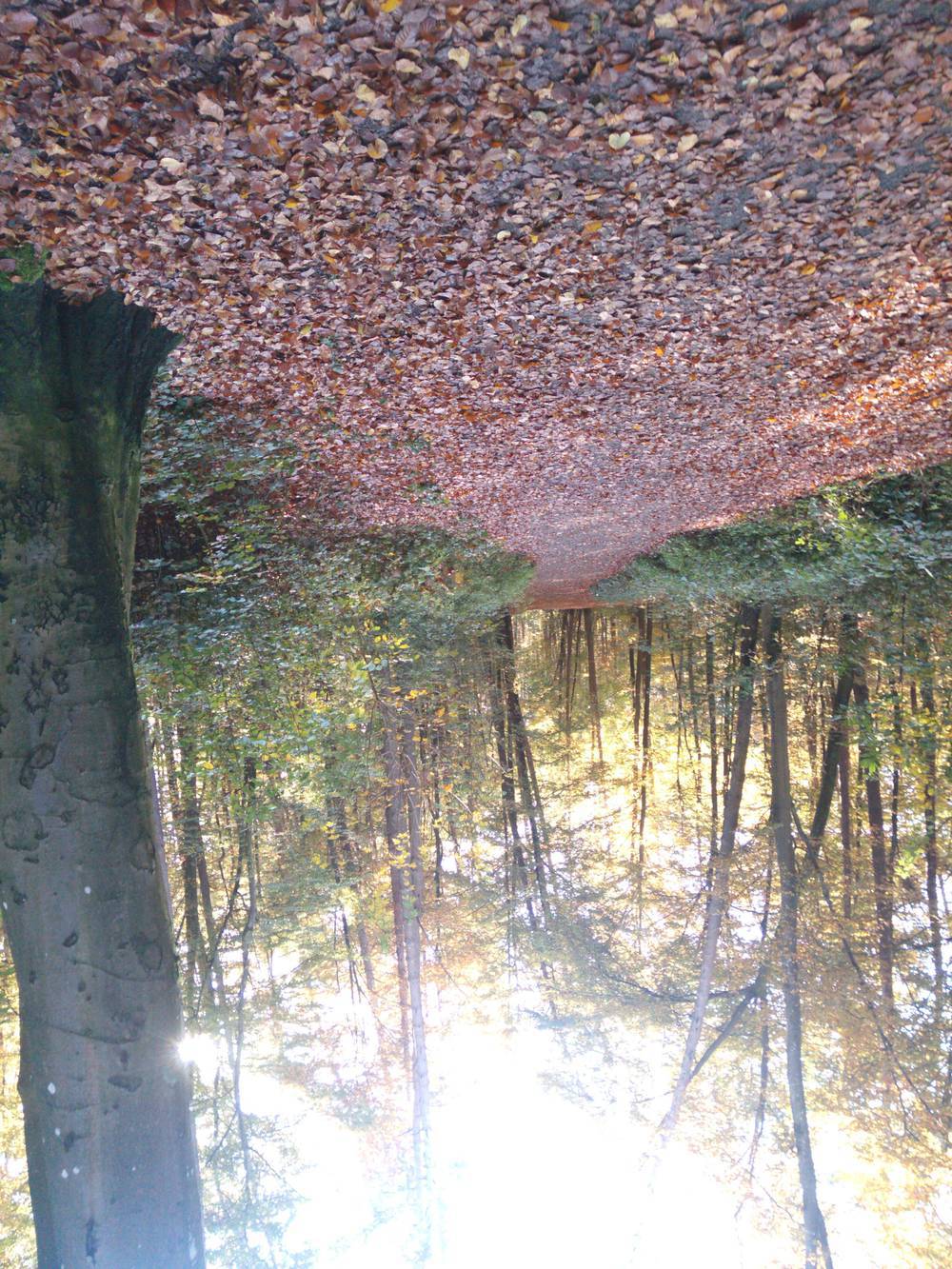} &
        \includegraphics[width=0.28\linewidth]{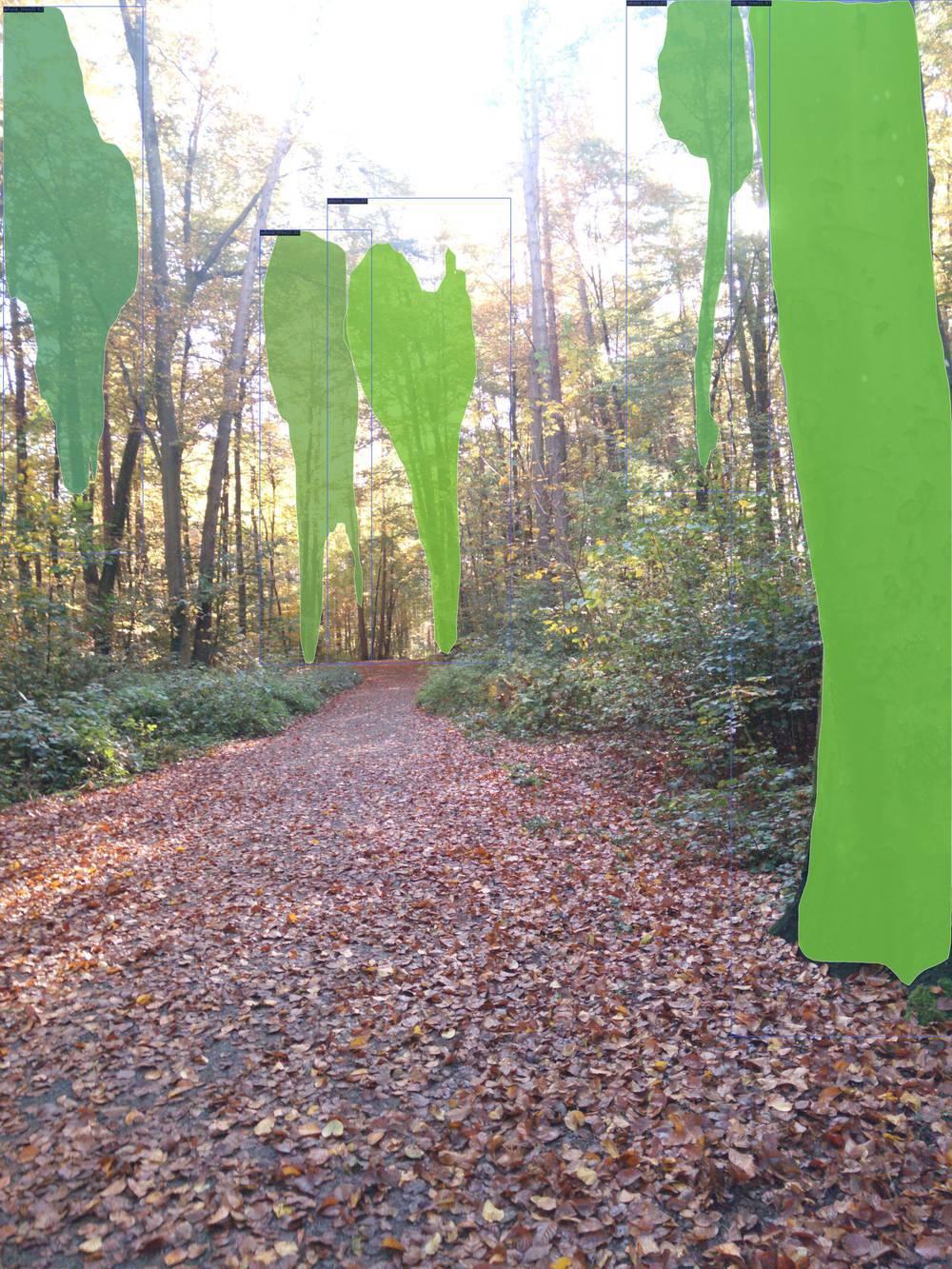} &
        \includegraphics[width=0.28\linewidth]{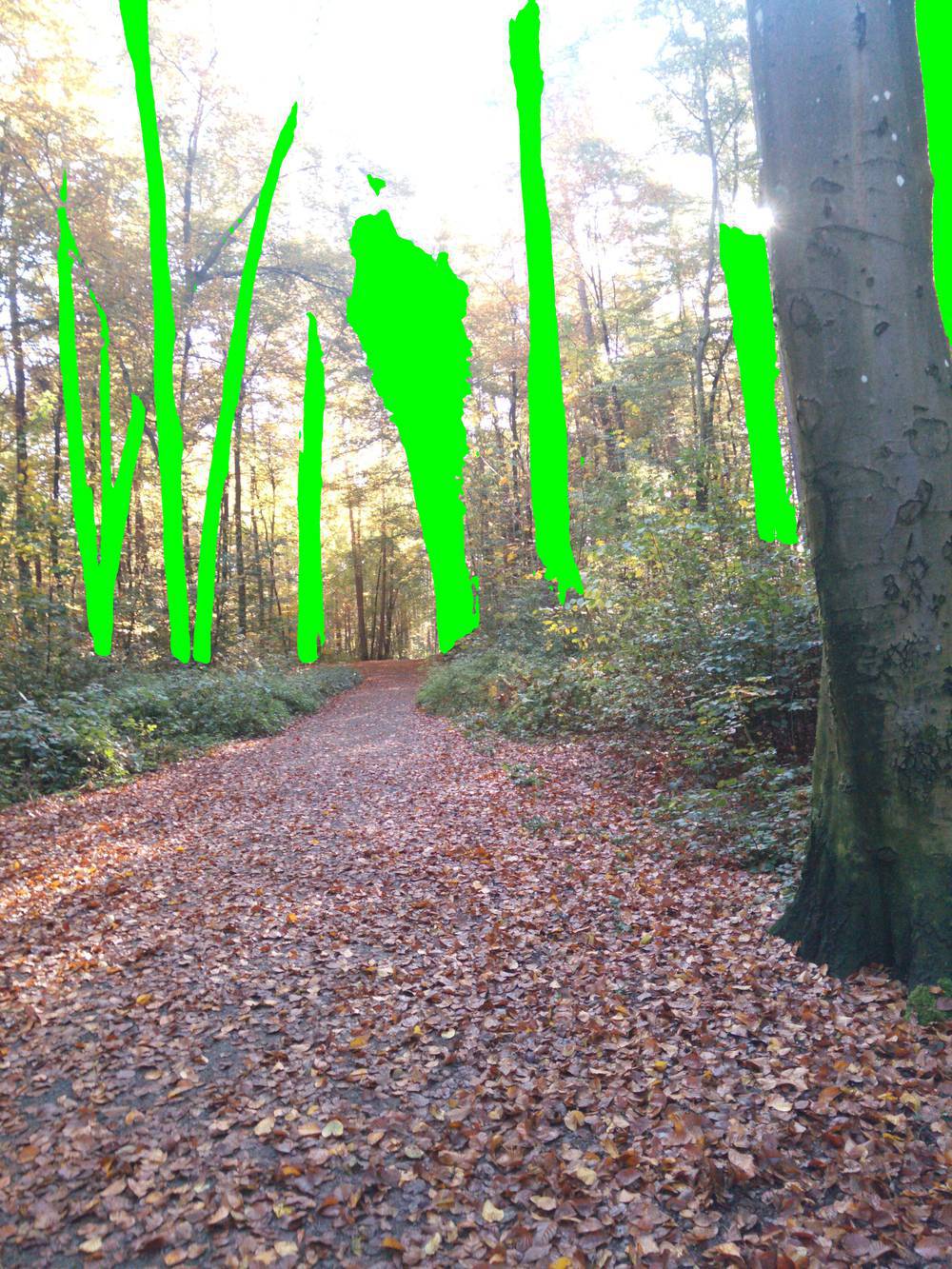} \\
        \includegraphics[width=0.28\linewidth, angle=180, origin=c]{images/inference_results/wt_real/IMG_160101_000703_0117_RGB.JPG} &
        \includegraphics[width=0.28\linewidth]{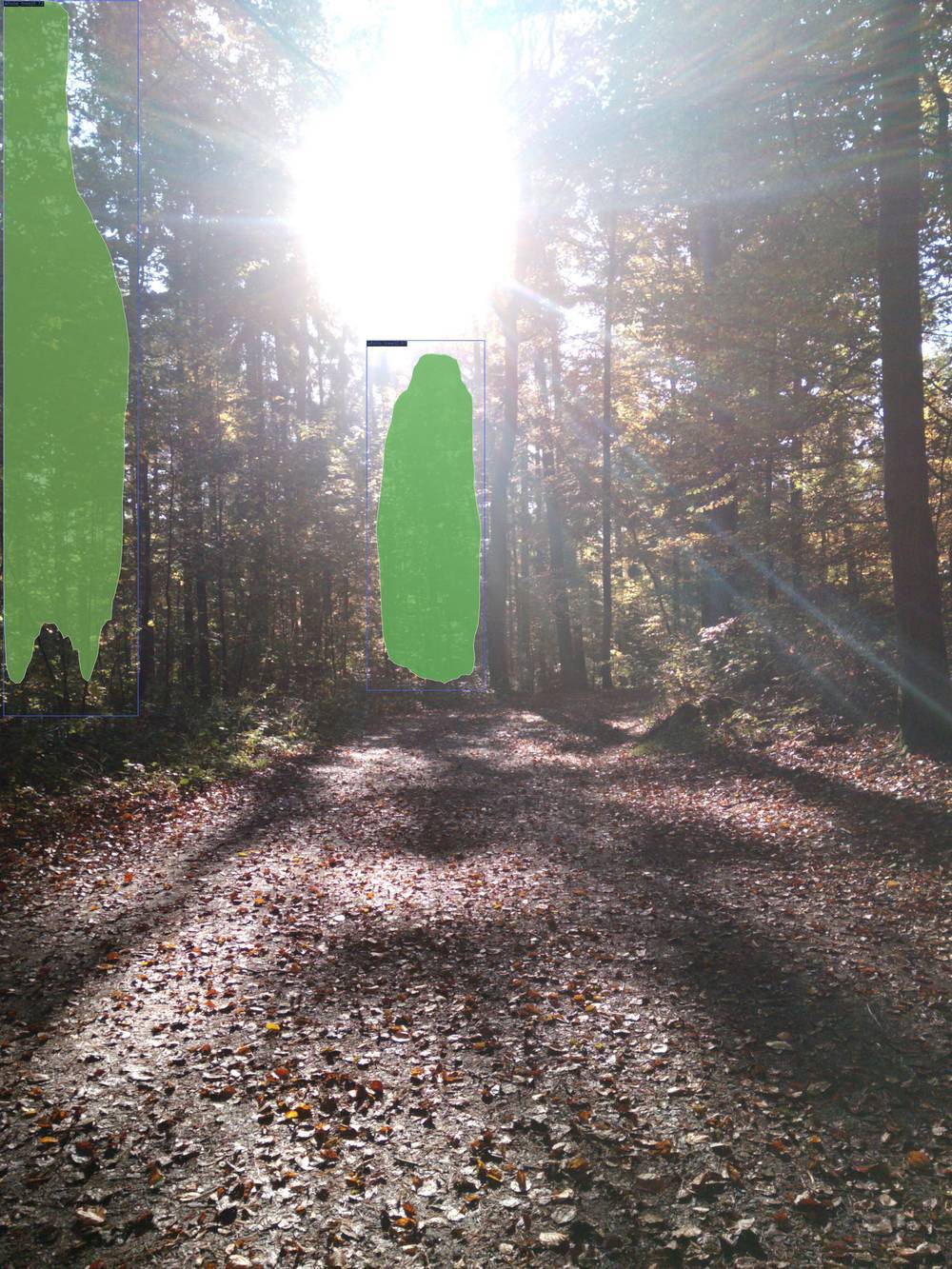} &
        \includegraphics[width=0.28\linewidth]{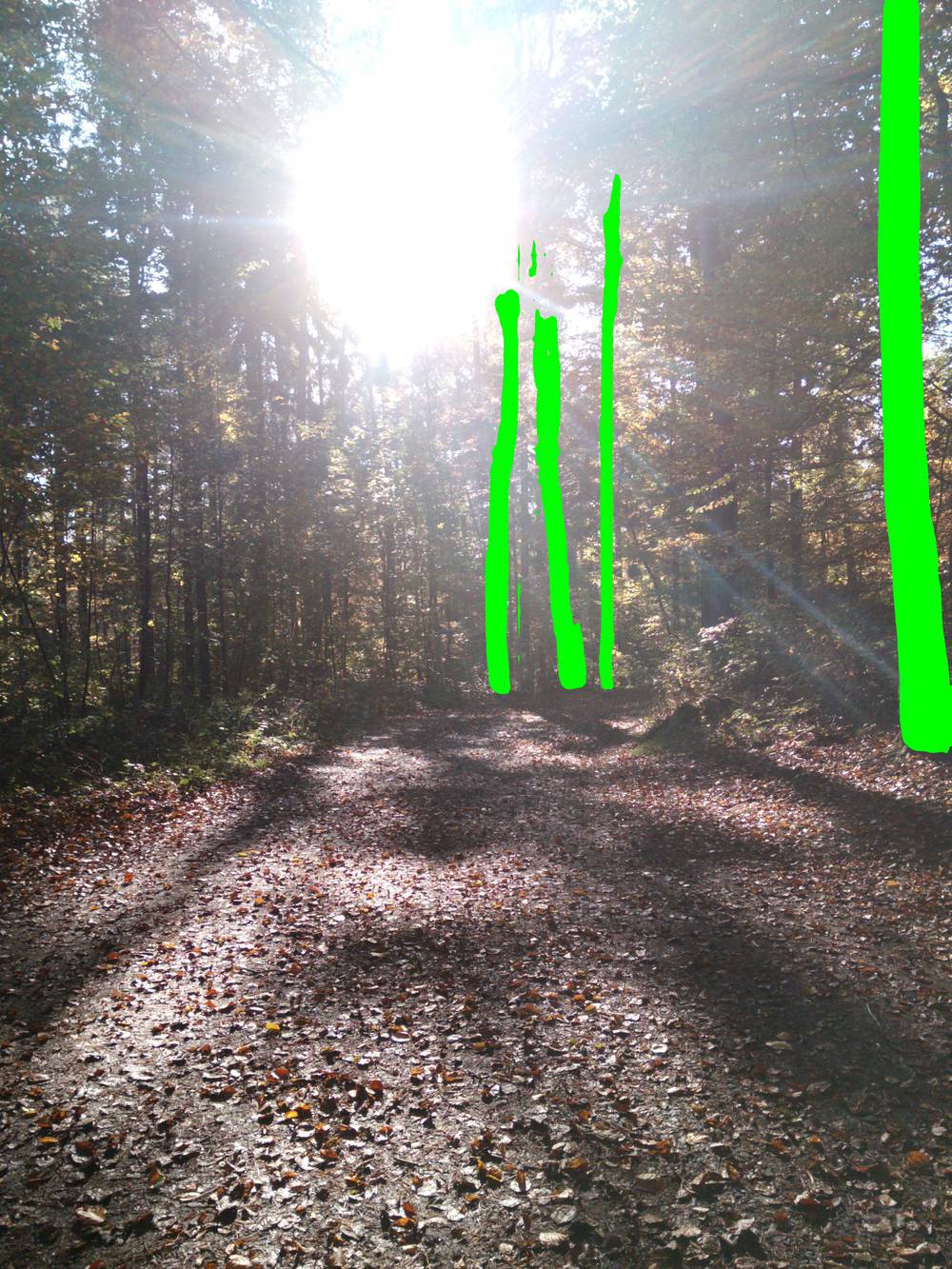} \\
    \end{tabular}
    \caption{\textbf{Phase 2:} Domain transfer (Whole Tree $\rightarrow$ Real Trees).}
    \label{fig:phase2_wholetree2real}
\end{figure}
Figure \ref{fig:phase2_wholetree2real} shows qualitative examples of the whole-tree teacher (trained on simulation) applied directly to real forest imagery (and table \ref{tab:sim2real_combined}). Compared to the trunk teacher (as shown previously), the whole-tree model produces much larger, canopy-shaped masks that attempt to capture full tree silhouettes. However, in the real domain these predictions are often highly imprecise. Large blobs are placed around dominant vertical structures, but the contours rarely align with the thin crowns or fine-grained tree boundaries observed in the ground truth. In several cases, the model collapses multiple adjacent trees into a single wide mask, overestimating spatial extent and ignoring tree separability. At the same time, numerous smaller and thinner trees clearly annotated in the GT remain undetected, indicating weak recall in cluttered conditions. Illumination changes further degrade these issues, overexposed regions lead to entire areas being ignored or segmented as random shapes, highlighting the sensitivity of the whole-tree teacher to domain-specific artifacts.

Overall, the whole-tree teacher transfers global structure priors (recognizing that trees occupy large vertical and canopy-like regions) \textbf{but at the cost of boundary precision and instance differentiation}. This complements the trunk teacher, where the trunk model focuses on precise vertical stems but struggles with coverage, the whole-tree model covers broad structures but loses detail and granularity. Together, these observations motivate Phase 4 distillation, where the complementary strengths of both teachers can be integrated with real annotations to train a student that balances coverage with precision.

\clearpage
\subsection{Phase 3}
\begin{figure}[ht]
	\centering
	\renewcommand{\arraystretch}{1.2}
	\setlength{\tabcolsep}{1pt}
	
	\begin{tabular}{c c c}
		\textbf{RGB Image} & \textbf{Prediction} & \textbf{Ground Truth} \\
		\includegraphics[width=0.28\linewidth, angle=180, origin=c]{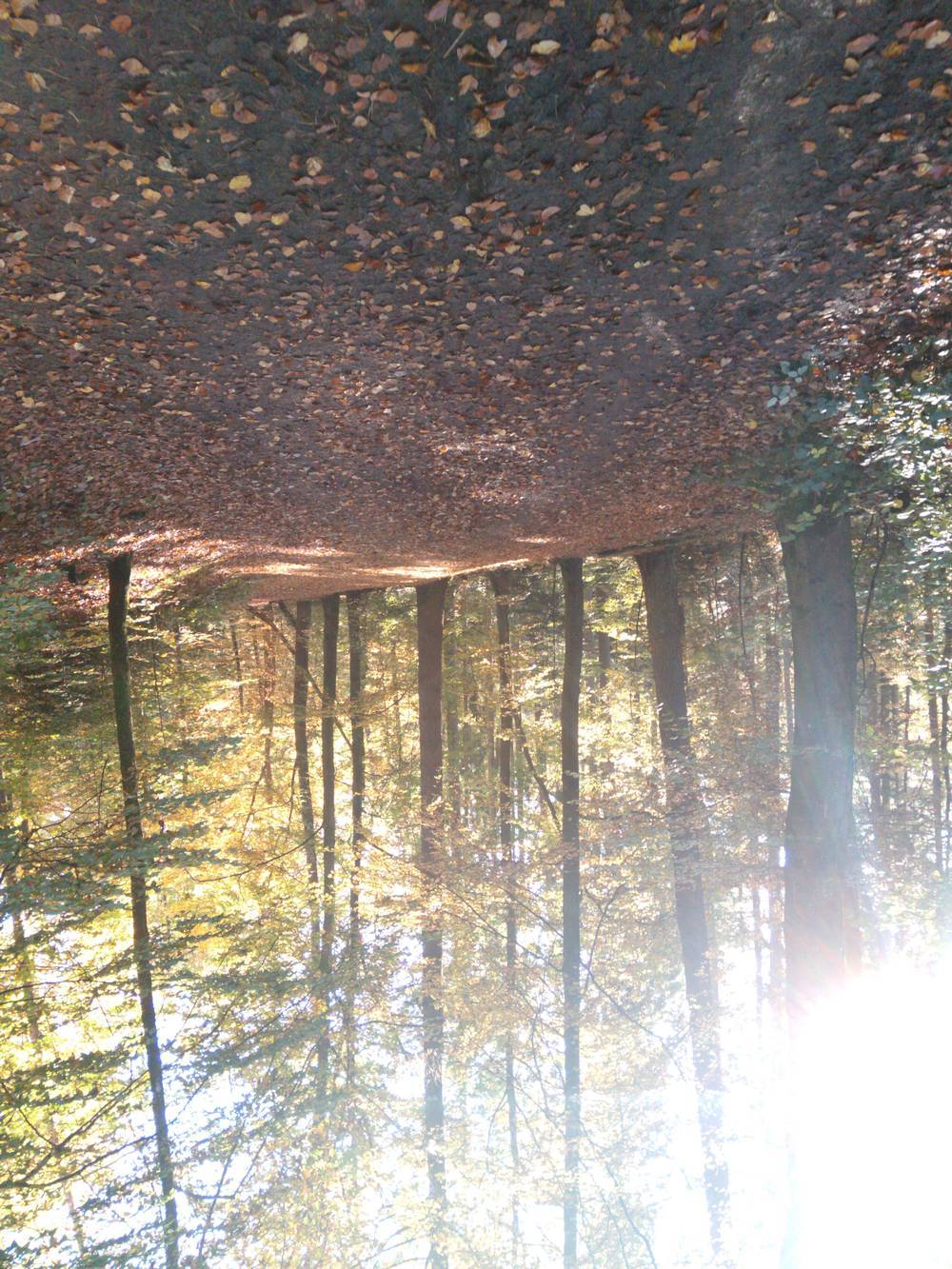} &
		\includegraphics[width=0.28\linewidth]{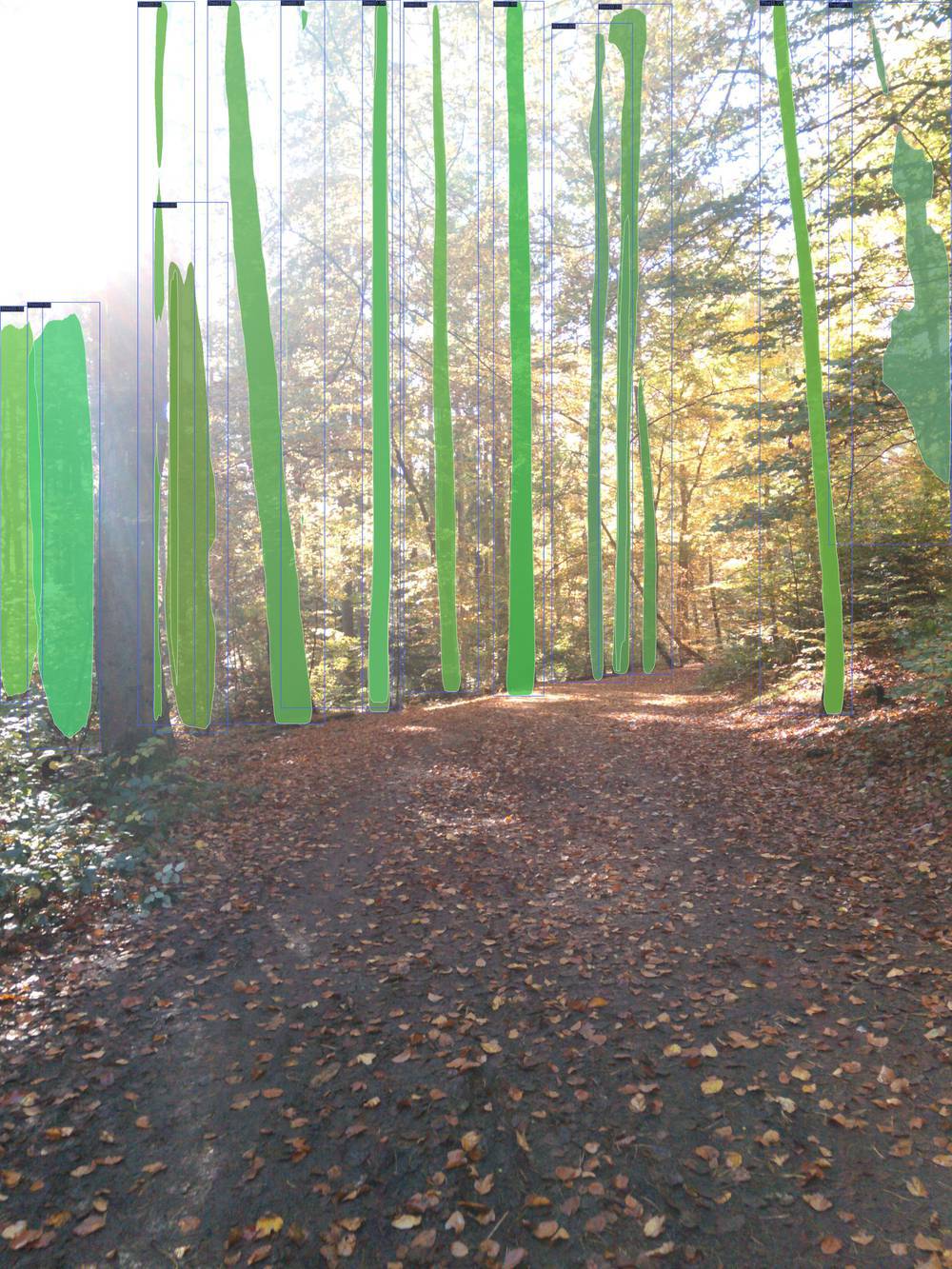} &
		\includegraphics[width=0.28\linewidth]{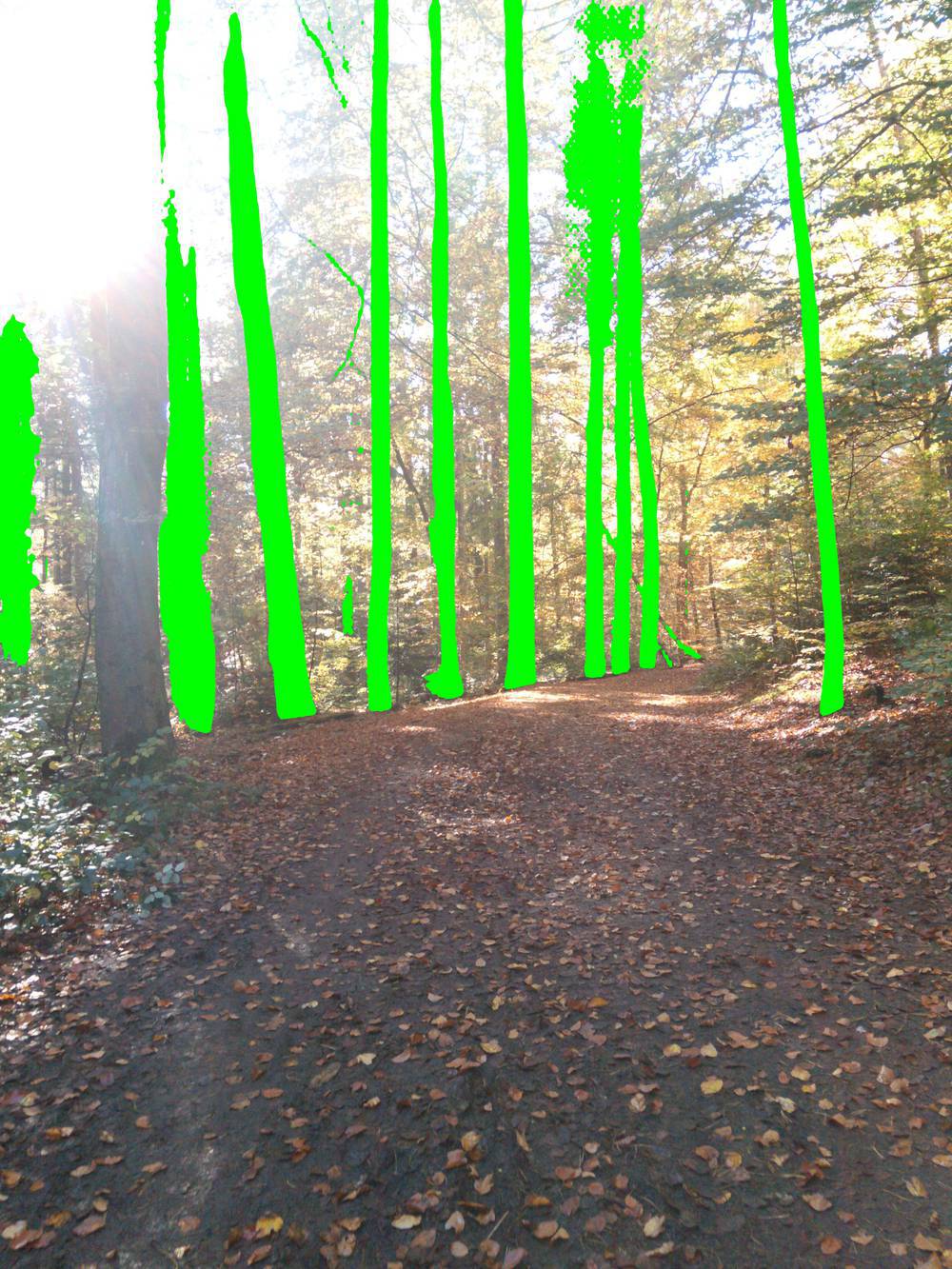} \\
		\includegraphics[width=0.28\linewidth, angle=180, origin=c]{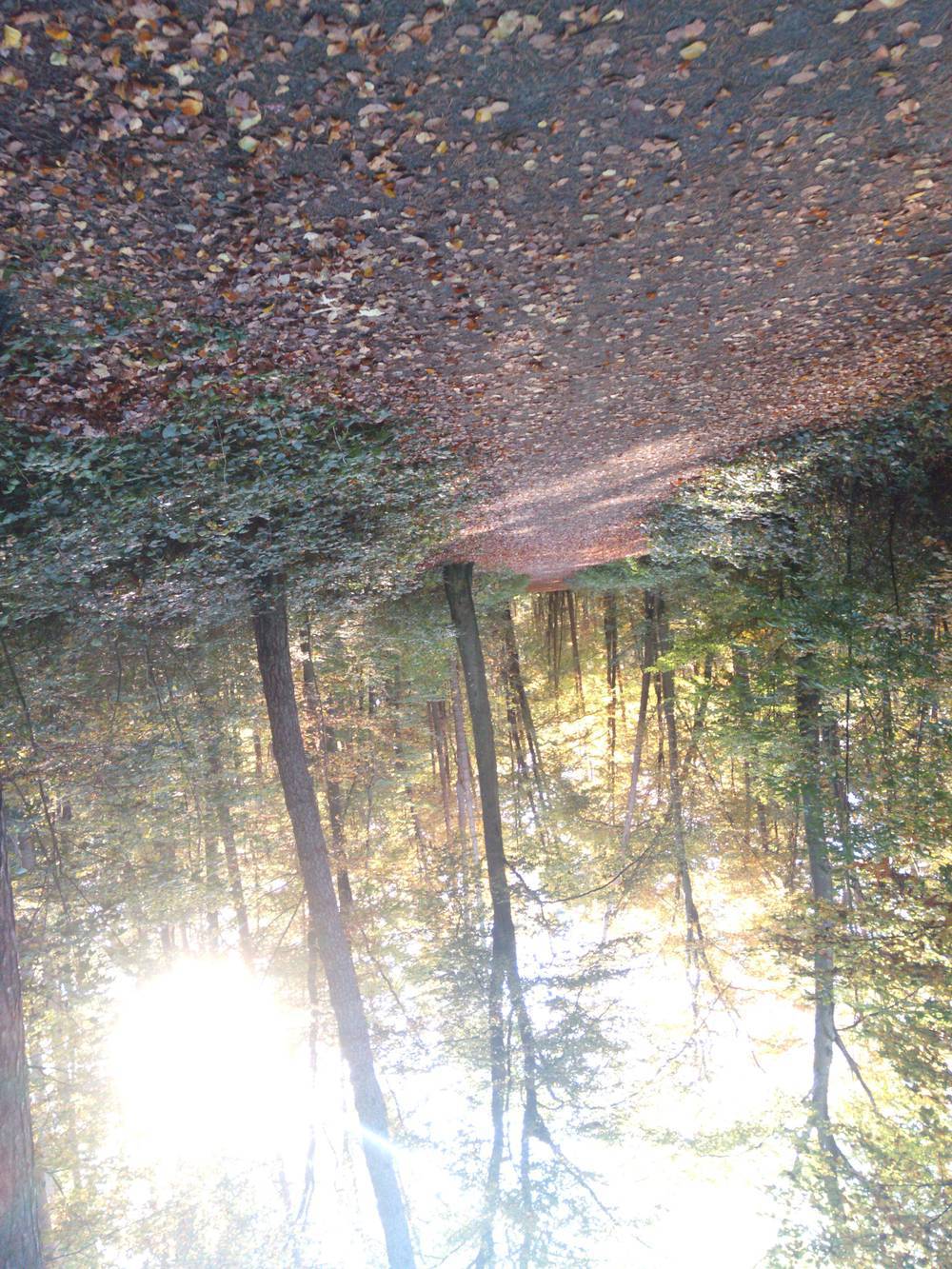} &
		\includegraphics[width=0.28\linewidth]{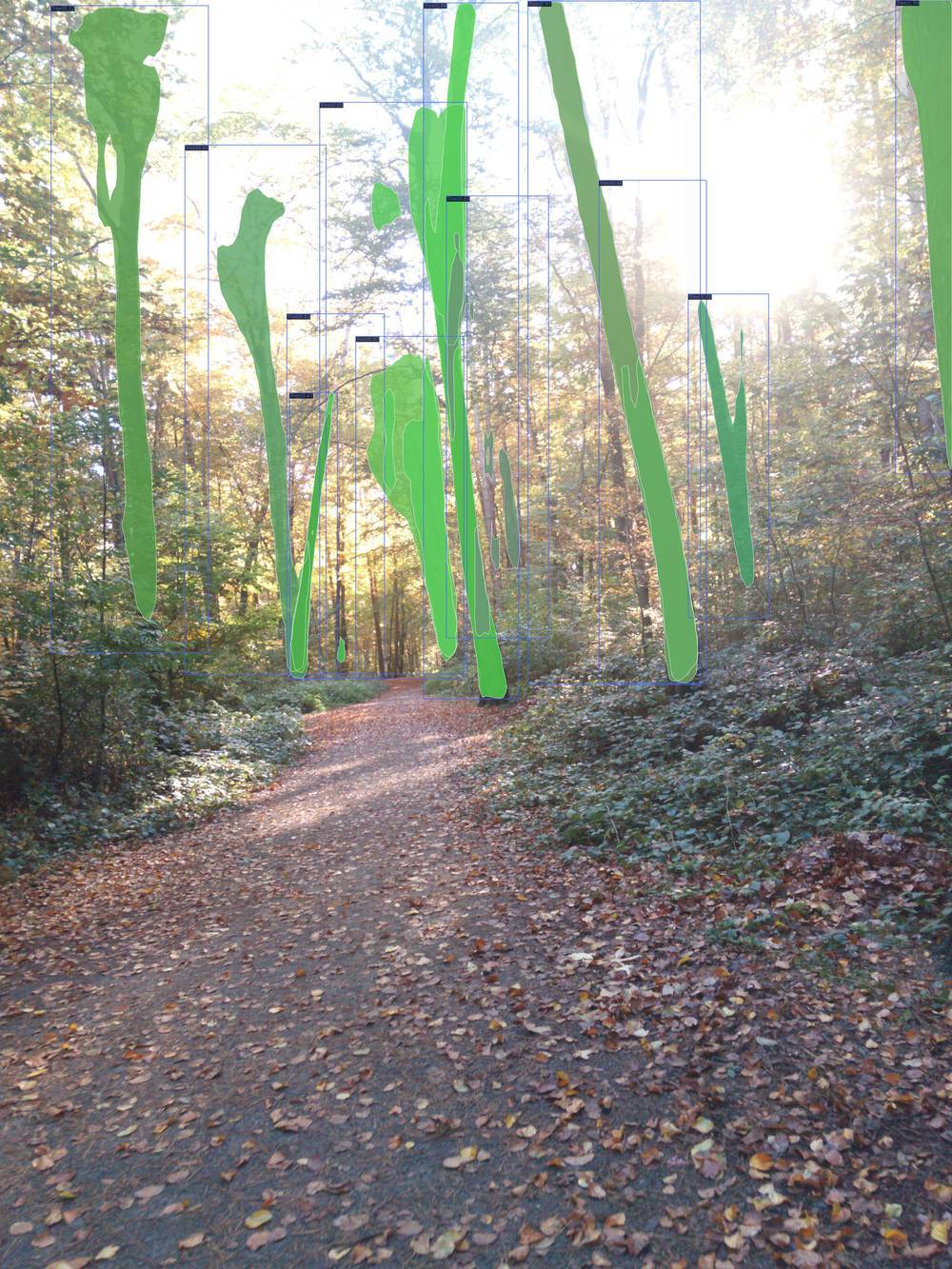} &
		\includegraphics[width=0.28\linewidth]{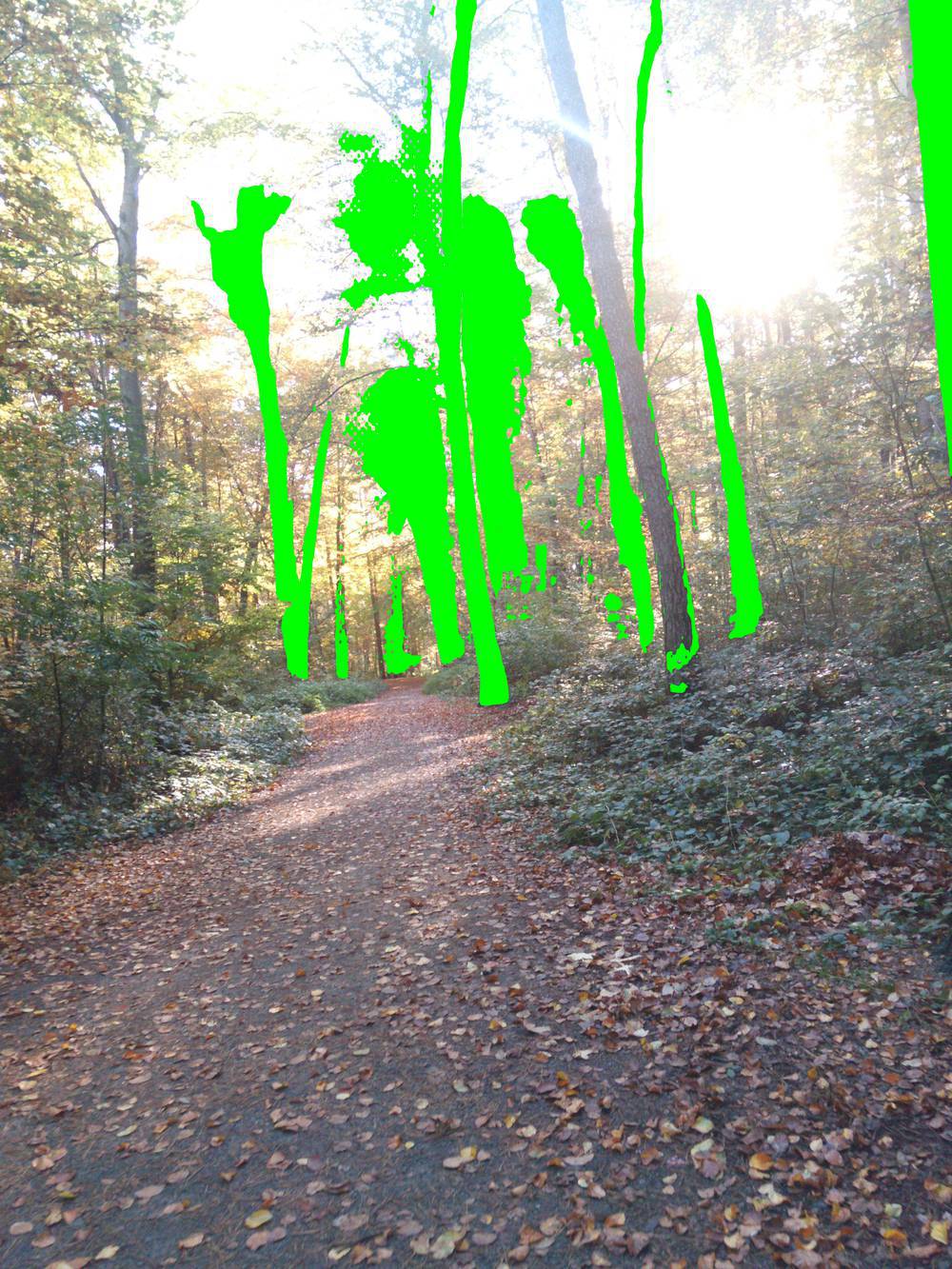} \\
		\includegraphics[width=0.28\linewidth, angle=180, origin=c]{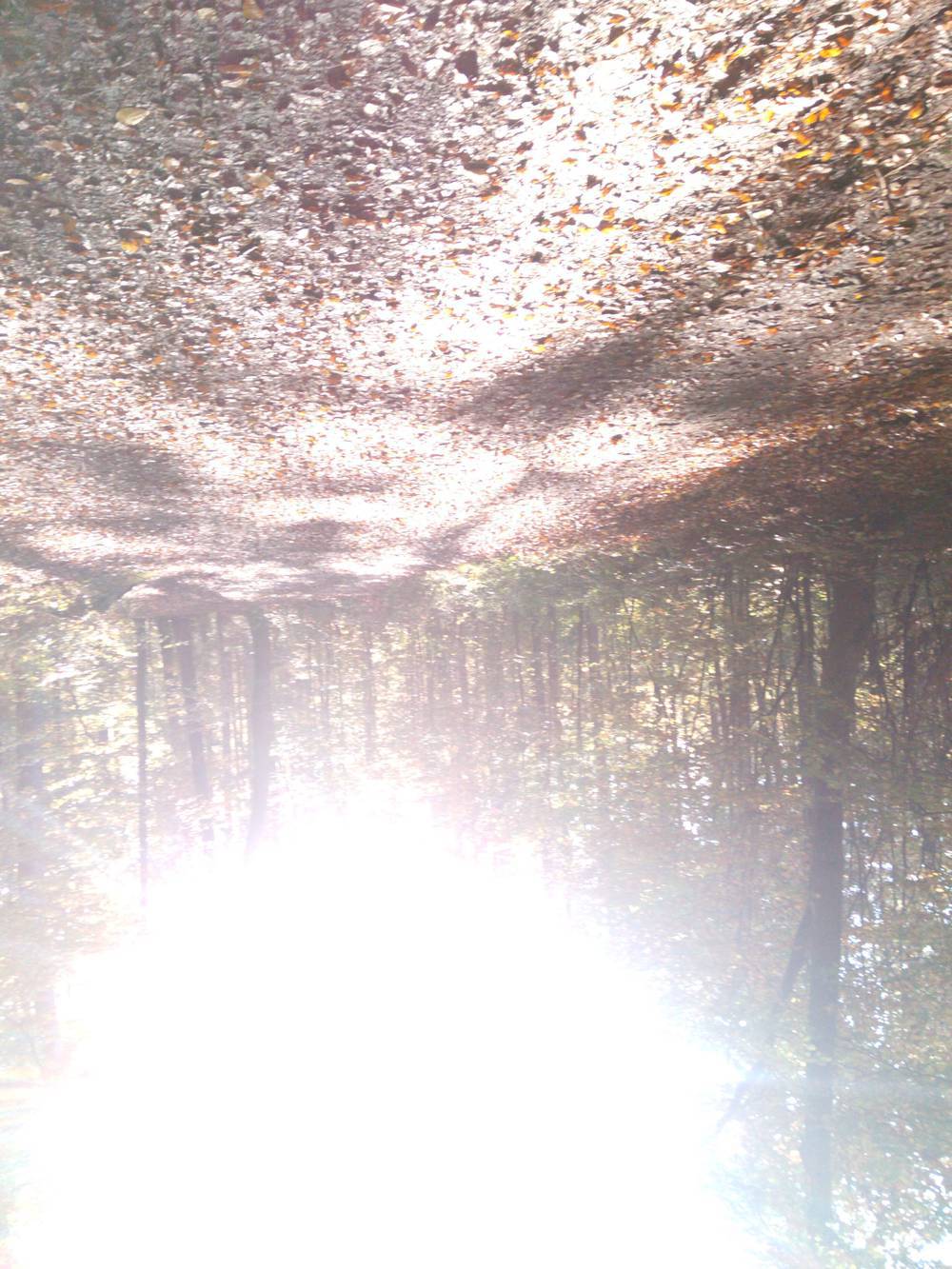} &
		\includegraphics[width=0.28\linewidth]{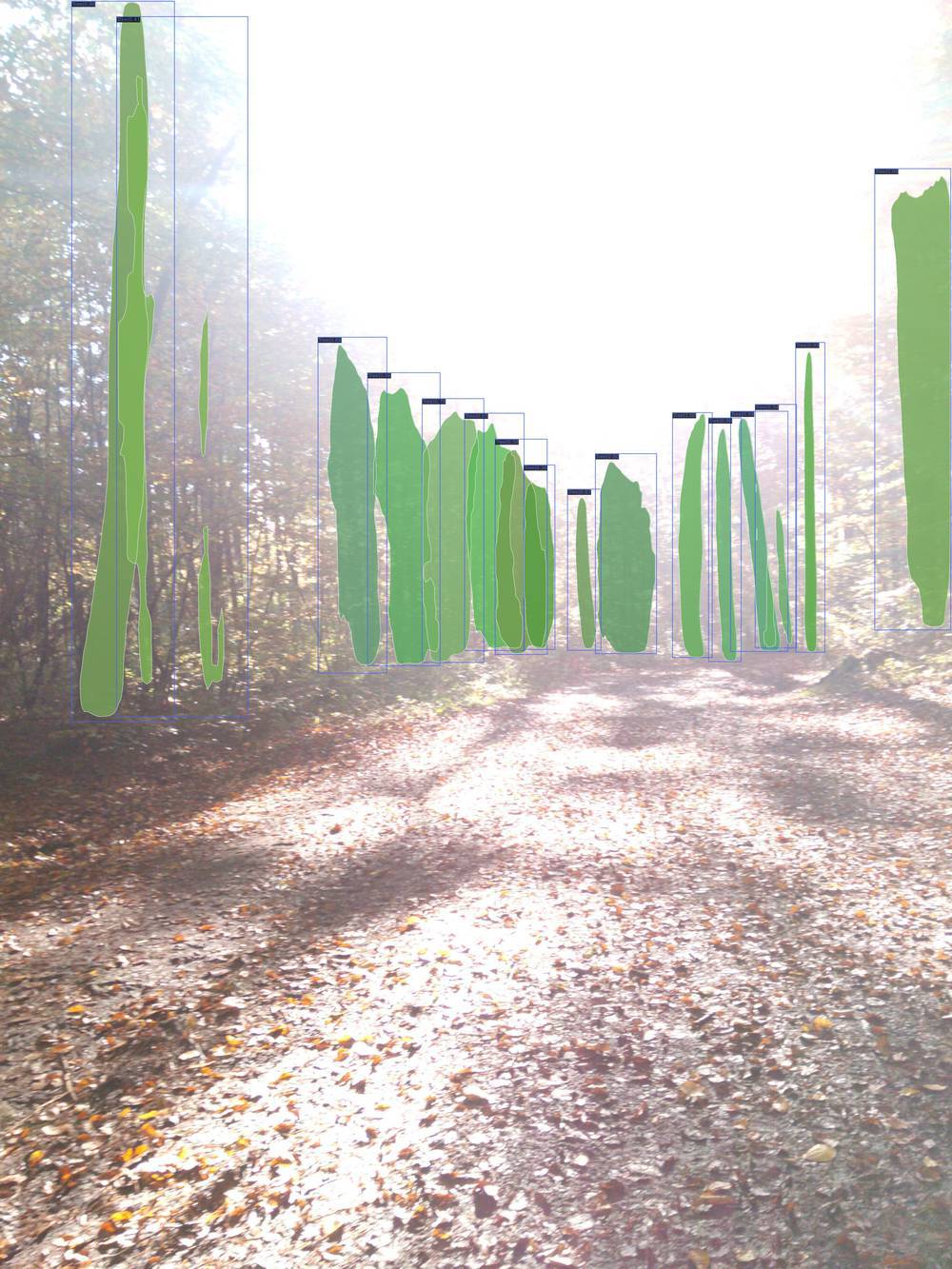} &
		\includegraphics[width=0.28\linewidth]{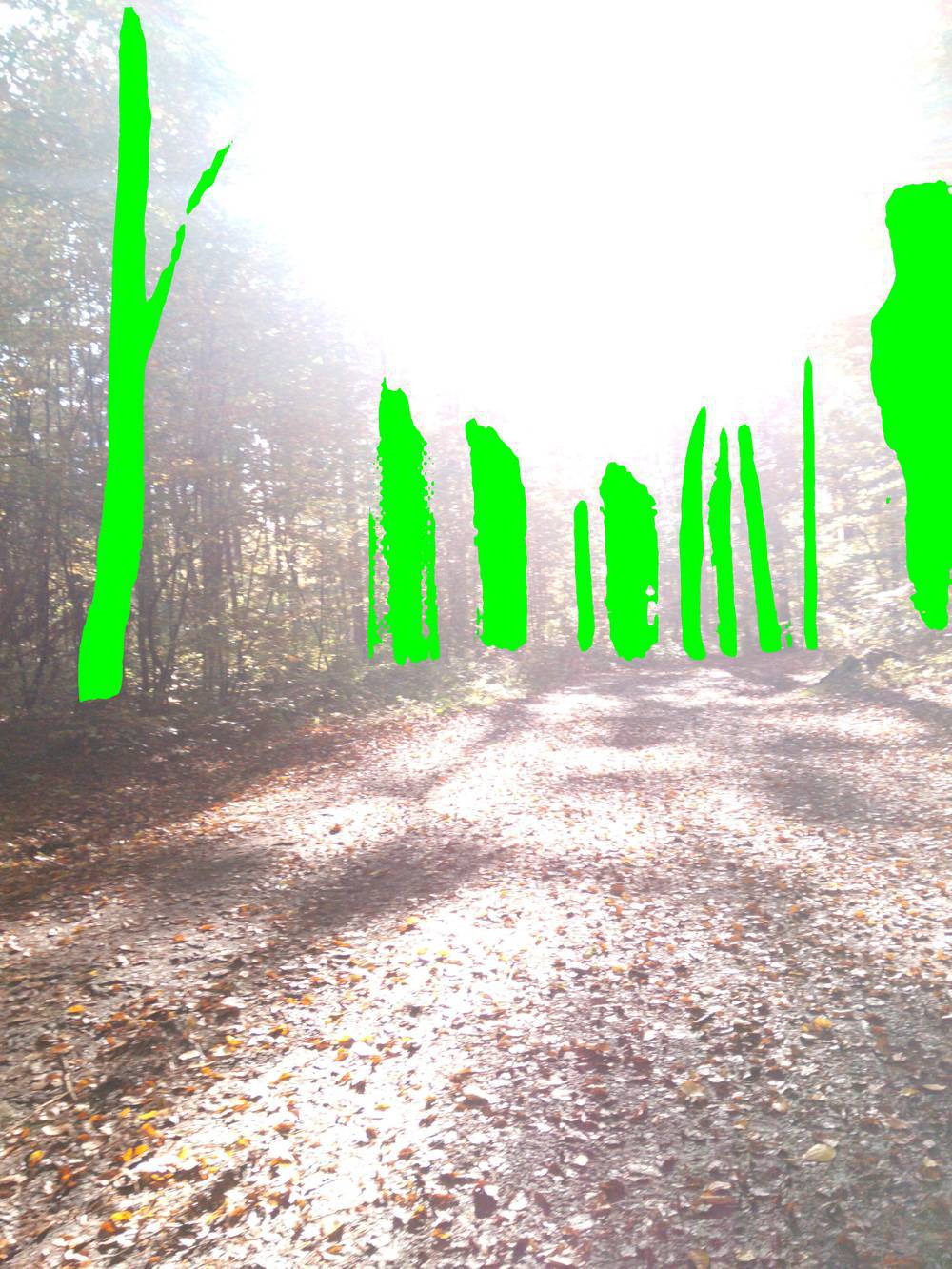} \\
		\includegraphics[width=0.28\linewidth, angle=180, origin=c]{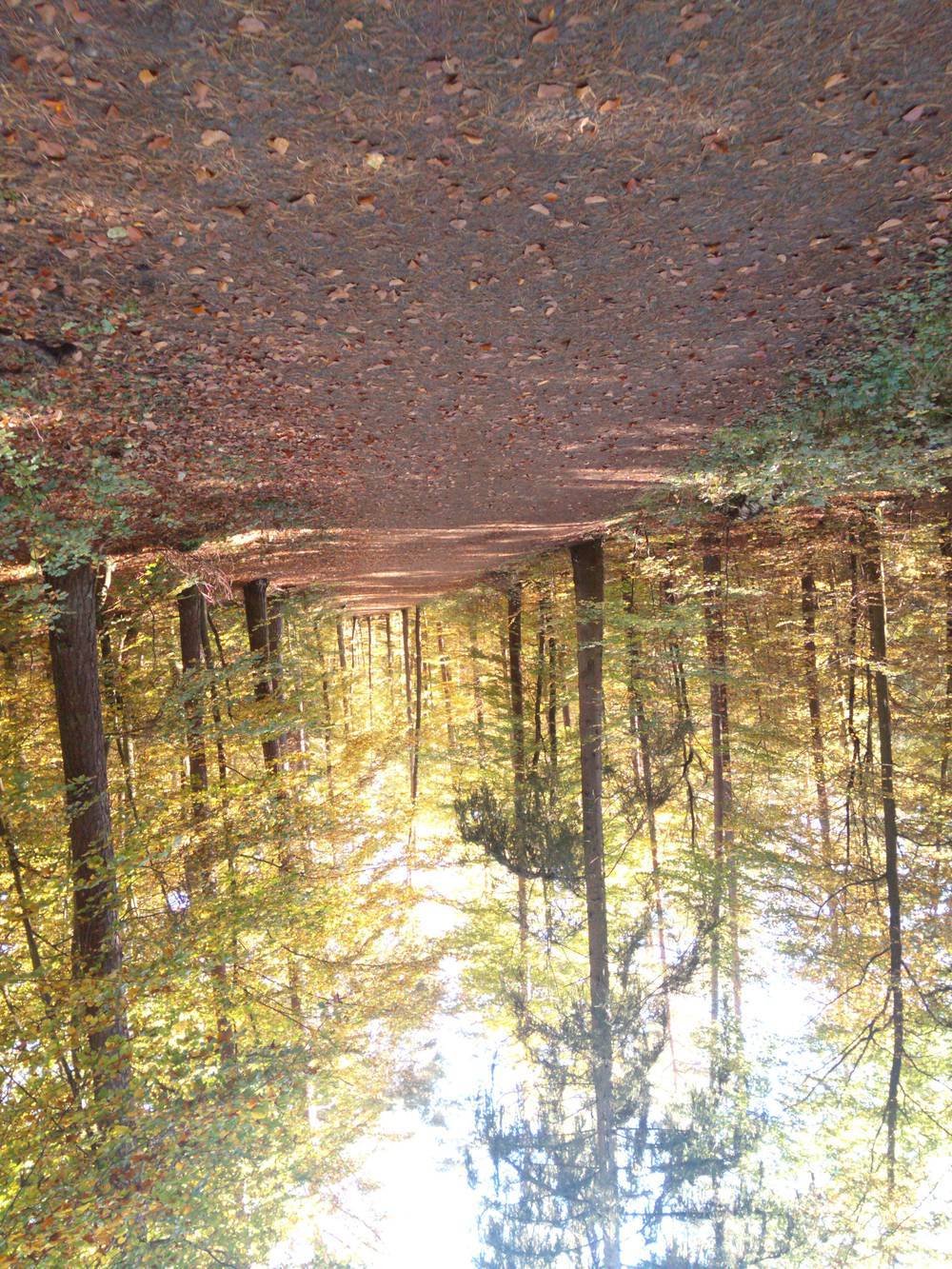} &
		\includegraphics[width=0.28\linewidth]{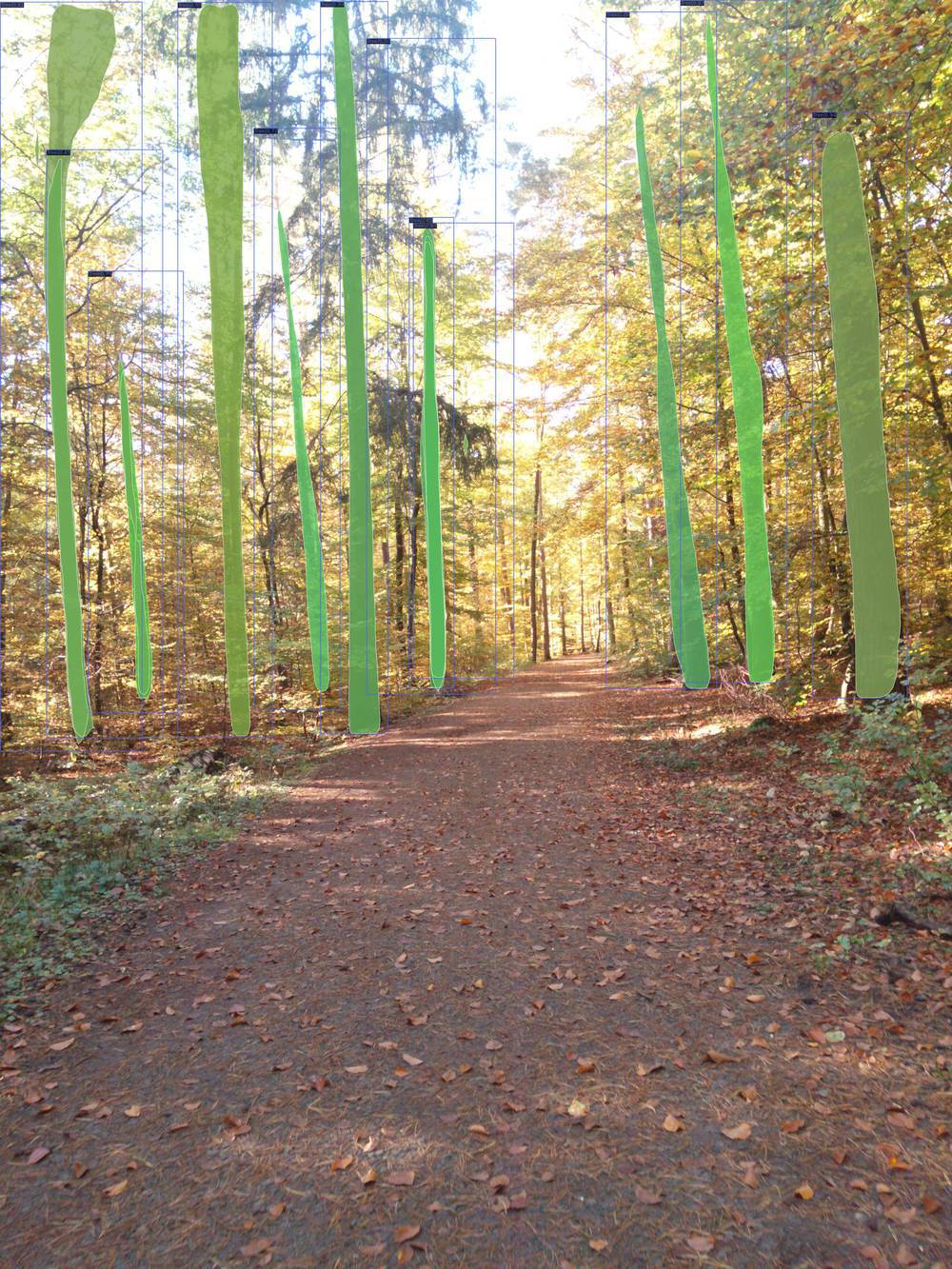} &
		\includegraphics[width=0.28\linewidth]{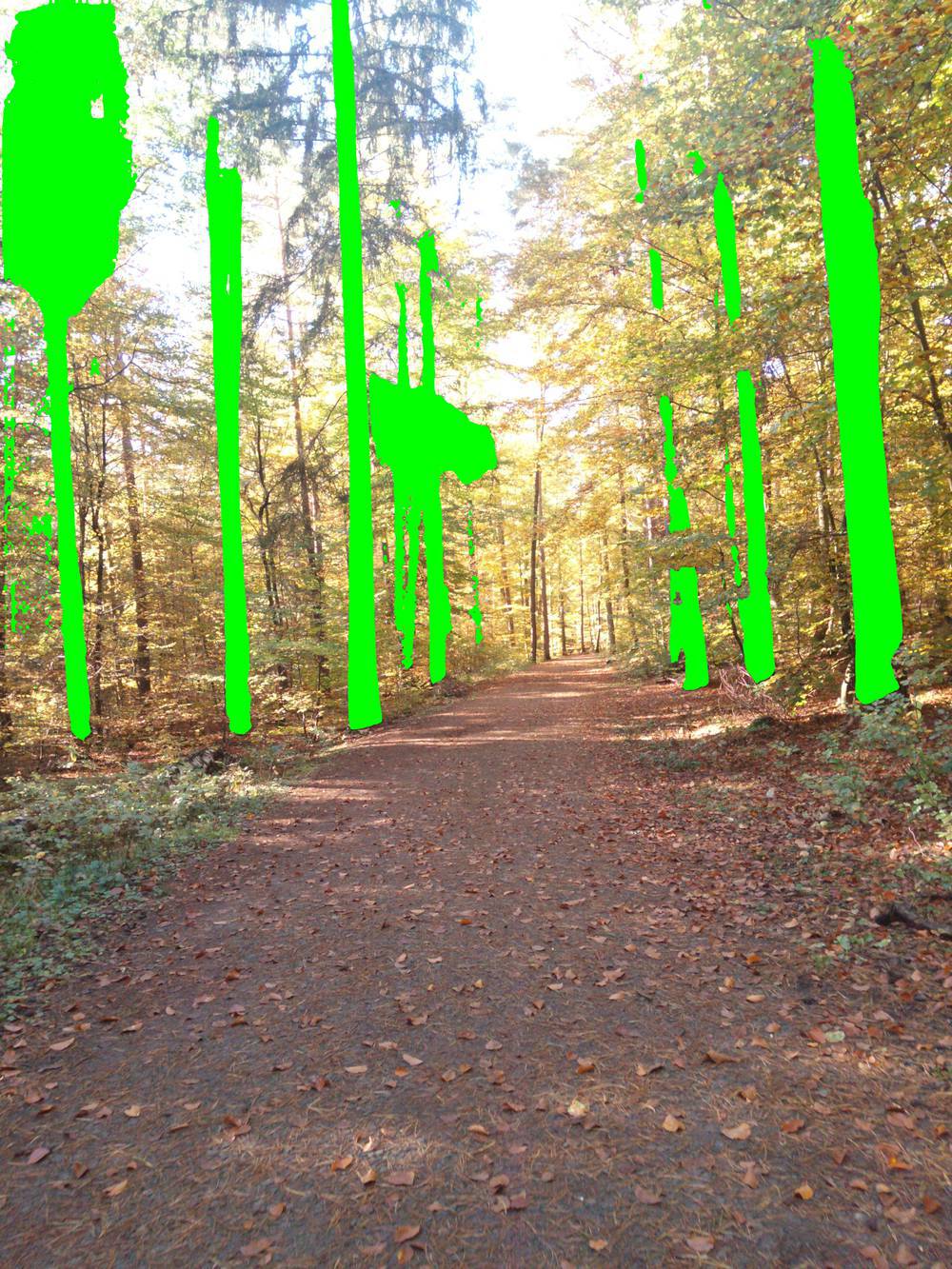} \\
	\end{tabular}
	\caption{\textbf{Phase 3:} Instance segmentation (Mask-RCNN with Swin-T backbone) results on real trees (trained directly on real data). Each row shows RGB image, prediction, and ground truth.}
	\label{fig:phase3_real}
\end{figure}

Figure \ref{fig:phase3_real} shows qualitative results (whereas table \ref{tab:phase_3_real}, the quantitative results) of the Mask~R-CNN student trained exclusively on the real dataset. Compared to the synthetic-trained teachers, the real-only student produces cleaner, better aligned segmentations of tree stems, with masks that more closely match the slender geometry of annotated trunks. Large and medium-sized trees are detected consistently across diverse lighting conditions, demonstrating that the model has adapted well to the target domain. However, several limitations remain. Under strong backlighting or overexposure, the model still misses many small or distant stems, particularly in cluttered backgrounds where contrast is low. In some cases, predictions merge multiple thin trunks into a single thicker mask, echoing the teacher’s coarse segmentation bias. The student also tends to under-segment the scene—focusing primarily on the most salient trunks while ignoring fine-scale structures visible in the ground truth.

Overall, these results highlight the benefit of training directly on real annotations: the student achieves higher precision and produces masks that are semantically closer to ground truth than the synthetic teachers. At the same time, recall for small or occluded trees remains limited, pointing to the restricted scale of the real dataset ($\sim$3000 images) and the absence of fine-grained supervisory cues such as trunk/crown separation.

\begin{table*}[htp!]
	\centering
	\caption{\textbf{Phase 3:} Instance segmentation results on real tree dataset. 
	For Mask2Former, validation results were not fully available; 
	test results are reported for completeness. The test results showed similar pattern as validation.}
	\resizebox{\textwidth}{!}{
		\begin{tabular}{l l l c c c c c c c c c}
			\toprule
			\textbf{Task} & \textbf{Model} & \textbf{Backbone} & \textbf{Epochs} & \textbf{mAP@[.5:.95]} & \textbf{mAP@50} & \textbf{AP50} & \textbf{AP75} & \textbf{APs} & \textbf{APm} & \textbf{APl} & \textbf{AR@[.5:.95]} \\
			\midrule
			\textbf{Bbox} & Mask-RCNN & Swin-T      & 24 & 0.500 & 0.792 & 0.792 & 0.544 & - & - & 0.500 & 0.617 \\
			&           & X-101       & 24 & 0.461 & 0.810 & 0.810 & 0.467 & - & - & 0.461 & 0.592 \\
			& Cascade RCNN & X-101    & 24 & 0.571 & 0.834 & 0.834 & 0.629 & - & - & 0.571 & 0.684 \\
			&           & Swin-T      & 24 & 0.571 & 0.842 & 0.842 & 0.629 & - & - & 0.571 & 0.685 \\
			& Mask2Former & R50       & 24 & 0.265 & 0.623 & 0.623 & 0.186 & - & - & 0.265 & 0.510 \\
			&           & R101        & 24 & 0.280 & 0.662 & 0.662 & 0.194 & - & - & 0.280 & 0.523 \\
			&           & R101\_tuned & 24 & 0.278 & 0.656 & 0.656 & 0.189 & - & - & 0.270 & 0.530 \\
			\midrule
			\textbf{Segm} & Mask-RCNN & Swin-T      & 24 & 0.420 & 0.692 & 0.692 & 0.446 & - & - & 0.420 & 0.529 \\
			&           & X-101       & 24 & 0.363 & 0.674 & 0.674 & 0.352 & - & - & 0.363 & 0.467 \\
			& Cascade RCNN & X-101    & 24 & 0.455 & 0.741 & 0.741 & 0.481 & - & - & 0.455 & 0.540 \\
			&           & Swin-T      & 24 & 0.469 & 0.751 & 0.751 & 0.501 & - & - & 0.469 & 0.552 \\
			& Mask2Former & R50       & 24 & 0.430 & 0.694 & 0.694 & 0.452 & - & - & 0.430 & 0.580 \\
			&           & R101        & 24 & 0.447 & 0.721 & 0.721 & 0.481 & - & - & 0.447 & 0.594 \\
			&           & R101\_tuned & 24 & 0.446 & 0.719 & 0.719 & 0.475 & - & - & 0.440 & 0.604 \\
			\bottomrule
		\end{tabular}
	\label{tab:phase_3_real}
	}

\end{table*}

\clearpage

\subsection{Phase 4: Granularity-Aware Distillation}
\label{sec:phase4}

Phase~4 aims to transfer fine-grained structural priors learned in simulation
into a real-data student trained only with coarse \textit{Tree} annotations.
Two synthetic teachers, trained separately on \textit{tree trunk} and
\textit{whole tree} labels, provide auxiliary supervision to a Mask R-CNN
student operating exclusively on real forest images.

Training follows a staged strategy (with the hyper-parameters mentioned as in \ref{tab:phase4_hyperparams}. The student is first optimized using only
real annotations to adapt to real-world appearance statistics and avoid early
collapse caused by noisy pseudo-labels. Distillation is then gradually
introduced by supervising the student with filtered teacher predictions that
are collapsed into the student's single-class label space. This enables the
student to inherit fine-grained geometric and vertical structure from
simulation while remaining consistent with real-domain supervision.

The final training objective combines supervised learning on real data with a
consistency-based distillation loss:
\begin{equation}
	\label{eq:phase4_loss}
	\mathcal{L}
	= \mathcal{L}_{\text{sup}}
	+ \lambda(t)\,\mathcal{L}_{\text{KD}},
\end{equation}
where $\mathcal{L}_{\text{sup}}$ denotes the standard detection and segmentation
losses on real annotations, $\mathcal{L}_{\text{KD}}$ transfers structural
priors from the synthetic teachers, and $\lambda(t)$ is a time-dependent weight that increases over training.

\subsubsection{Analysis of Individual Teacher Distillation}
\label{app:teacher_analysis}

We further provide a detailed analysis of the two separate distillation experiments conducted in Phase~4: one where the student learns solely from the \textbf{tree trunk teacher} and another from the \textbf{whole tree teacher}. The metrics reveal complementary strengths, justifying our final combined granularity-aware distillation approach.

\subsubsection{Quantitative Comparison of Training Dynamics}

The final and peak performance metrics for the two experimental setups are summarized in Table~\ref{tab:teacher_metrics}.

\begin{table}[htbp]
	\centering
	\caption{Final and peak performance metrics for student models trained with individual teachers. The tree trunk teacher facilitates higher peak precision (mAP), while the whole tree teacher contributes to final overlap accuracy (IoU).}
	\resizebox{\textwidth}{!}{
	\label{tab:teacher_metrics}
	\begin{tabular}{lccc}
		\toprule
		\textbf{Metric} & \textbf{Tree Trunk + Student} & \textbf{Whole Tree + Student} & \textbf{Key Insight} \\
		\midrule
		\textbf{Final IoU} & 0.512 & \textbf{0.630} & Whole tree supervision leads to better final \emph{overlap}. \\
		\textbf{Peak IoU} & \textbf{0.689} (Epoch 97) & 0.667 (Epoch 90) & Trunk supervision enables higher \emph{best-ever} overlap. \\
		\textbf{Final mAP@[.5:.95]} & 0.138 & \textbf{0.326} & Whole tree supervision yields better final \emph{detection quality}. \\
		\textbf{Peak mAP@[.5:.95]} & \textbf{0.427} (Epoch 88) & 0.381 (Epoch 90) & Trunk supervision achieves superior \emph{peak detection precision}. \\
		\textbf{Stability (Post-Epoch 50)} & High & Low & Trunk teacher provides more \emph{consistent learning signals}. \\
		\bottomrule
	\end{tabular}
}
\end{table}

\subsubsection{Interpretation of Learning Patterns}

The pattern reveal distinct dynamics that underscore the unique value of each teacher's granularity:

\begin{itemize}
	\item \textbf{Tree Trunk Teacher Promotes Precision:} The student model in this setup shows a stable, rising trend in mask average precision (mAP) in later epochs. This indicates that the trunk teacher provides a clear, unambiguous signal for \emph{localizing the fundamental vertical structure} of a tree, which is crucial for reliable detection. The higher \textbf{peak mAP} achieved supports this.
	
	\item \textbf{Whole Tree Teacher Provides Context:} While leading to higher final IoU, this setup exhibits significant variance in mAP throughout training. This suggests that learning from the full silhouette is a noisier task, as the model must interpret complex crown shapes and separate them from cluttered backgrounds. However, its contribution to final IoU confirms it successfully teaches the model about the \emph{complete spatial extent} of a tree.
	
	\item \textbf{Synergy in Combined Distillation:} These patterns explain the qualitative results in the main paper (Fig.~3). The trunk teacher's influence is evident in the improved recall of \emph{thin, distant trunks}. The whole tree teacher's contribution manifests as a better grasp of overall tree presence and scale, though it can occasionally lead to \emph{over-extended masks} as the model learns to fill plausible "tree-like" regions - a shift from missed detections to boundary imprecision.
\end{itemize}

This ablation study validates the design of our granularity-aware distillation framework. It demonstrates empirically that:
\begin{enumerate}
	\item Structural knowledge from \emph{both} fine-grained annotations is transferable and beneficial.
	\item The \emph{trunk} teacher is the primary driver for detection precision and stable convergence.
	\item The \emph{whole tree} teacher provides important contextual and spatial cues.
\end{enumerate}
Therefore, merging their guidance via logit-space merging and mask union, as done in our final method, creates a student that benefits from \textbf{precise localization} and \textbf{holistic understanding}, leading to the robust performance gains reported in the main paper.

\clearpage
\section{Object detection experiments}
Apart from Instance segmentation, we also provide trained models for the task of Object detection. Specifically, YOLOV8 (all variants) and YOLOV11 (all variants) were trained separately on tree trunks and whole trees with the specifications (as mentioned on table \ref{tab:yolo_hyper}). We show a systematic analysis of the domain gaps from Sim$\rightarrow$Real in the subsequent phases.

\subsection{Phase 1 results}
\begin{figure}[htp]
	\centering
	\setlength{\tabcolsep}{1pt}     
	\renewcommand{\arraystretch}{0.9}
	
	\caption{\textbf{Phase 1:} Qualitative analysis of yolov11m (trained on simulated tree trunks) on the simulated tree trunks val set.}
	
	\begin{tabular}{ccc}
		\toprule
		\textbf{RGB Image} & \textbf{Prediction} & \textbf{Ground Truth} \\
		\midrule
		
		\includegraphics[width=0.3\linewidth]{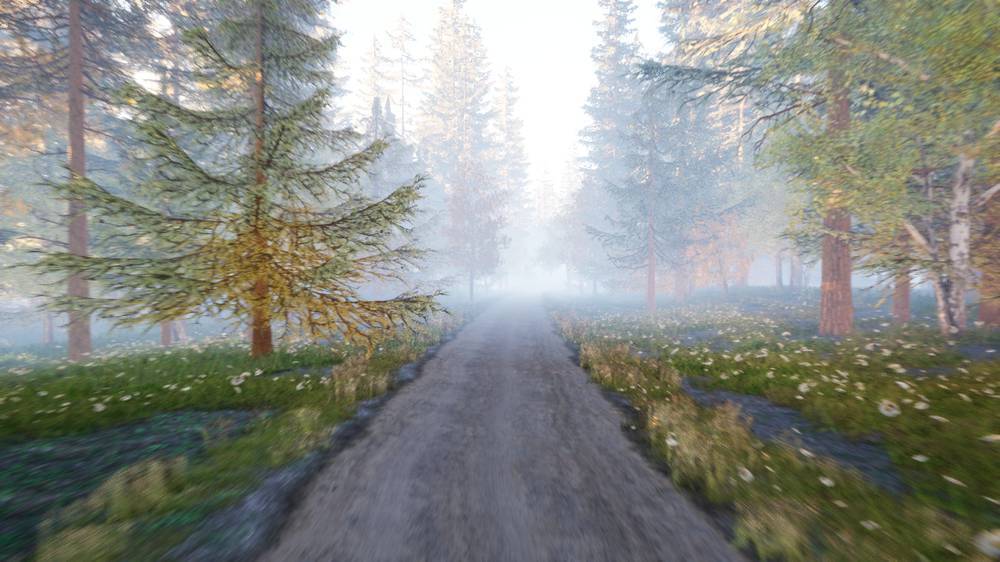} &
		\includegraphics[width=0.3\linewidth]{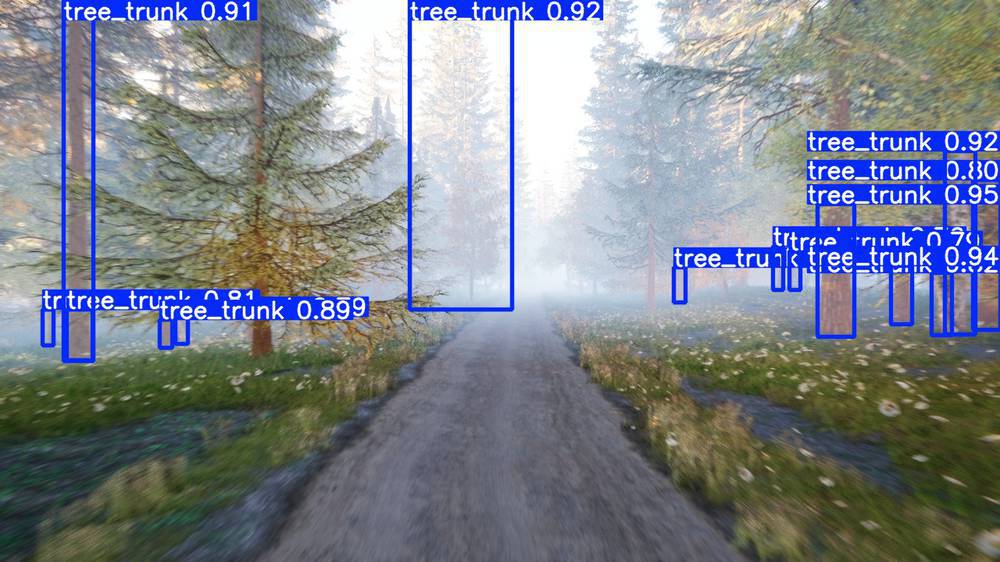} &
		\includegraphics[width=0.3\linewidth]{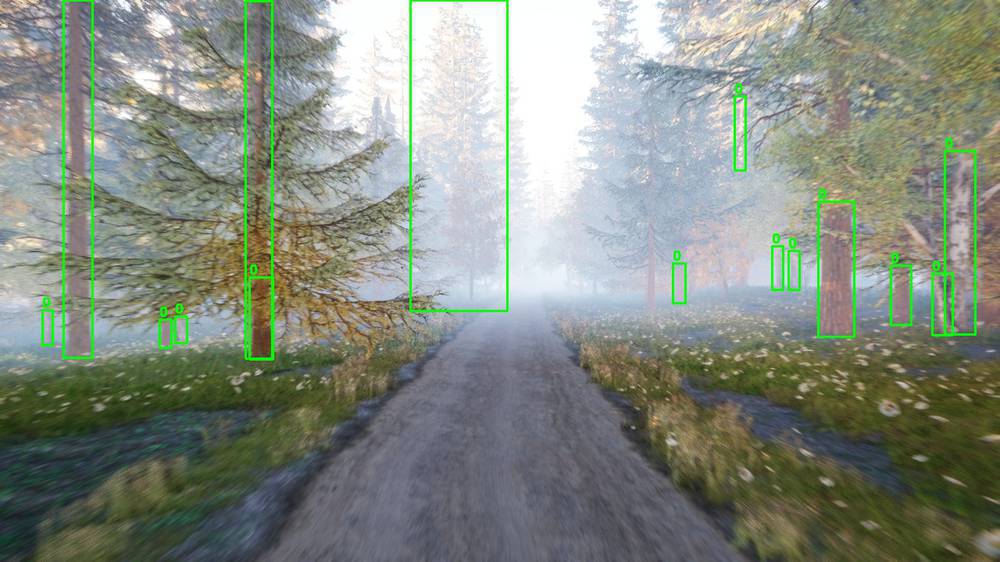} \\
		
		\includegraphics[width=0.3\linewidth]{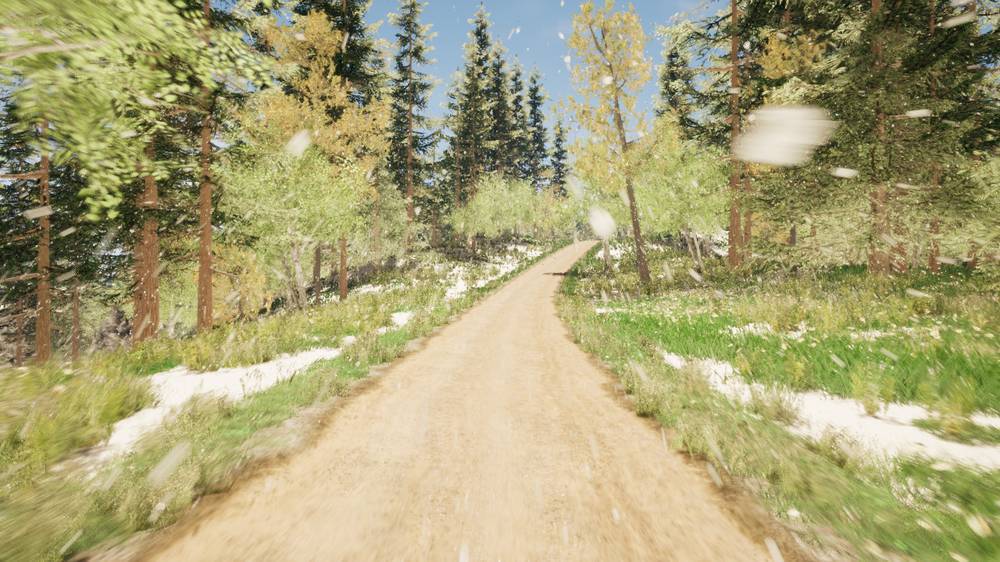} &
		\includegraphics[width=0.3\linewidth]{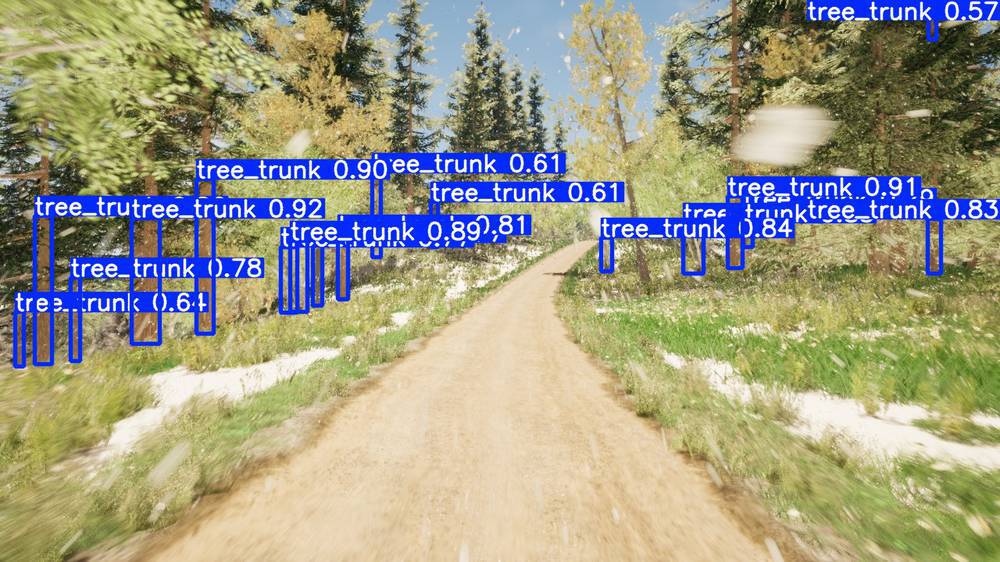} &
		\includegraphics[width=0.3\linewidth]{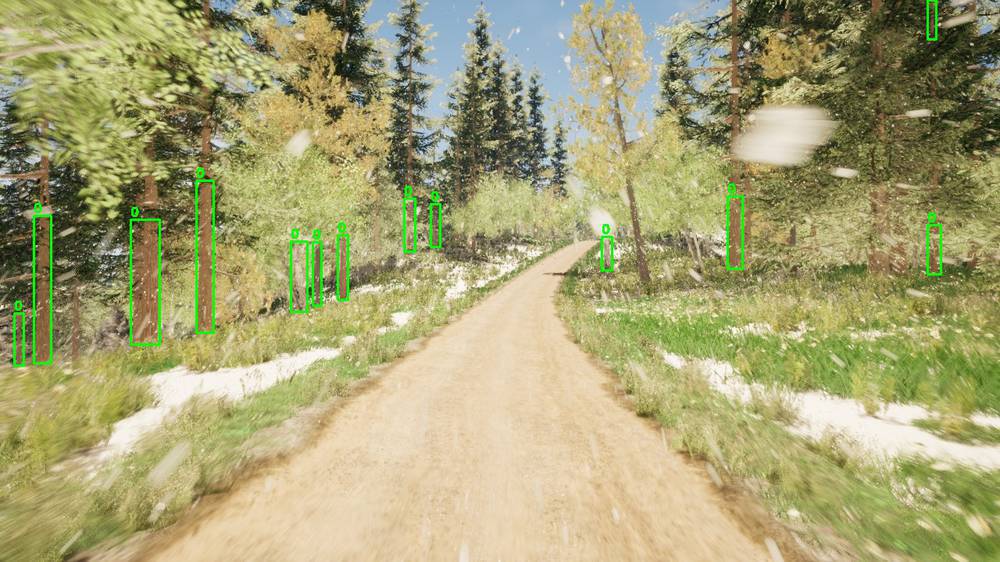} \\
		
		\includegraphics[width=0.3\linewidth]{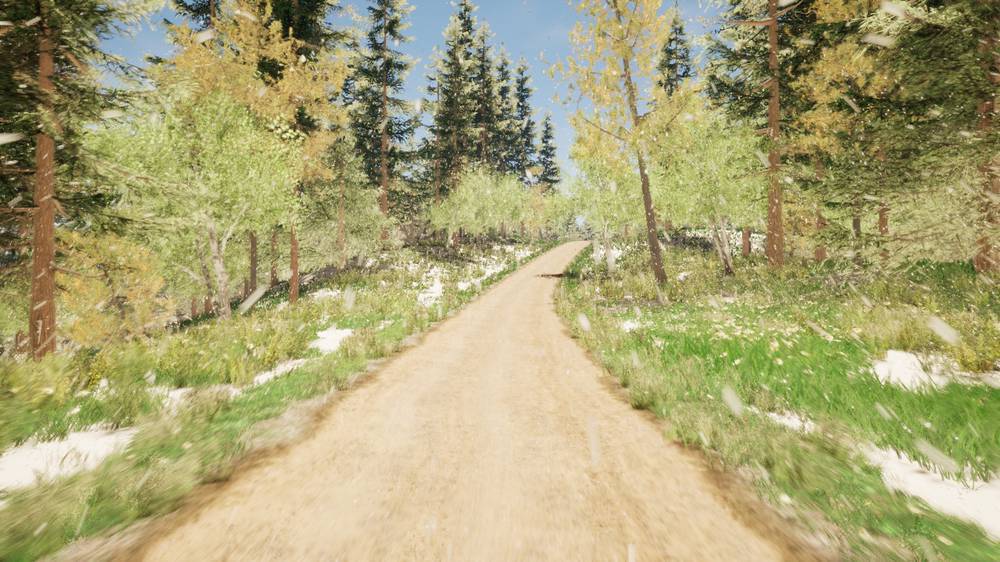} &
		\includegraphics[width=0.3\linewidth]{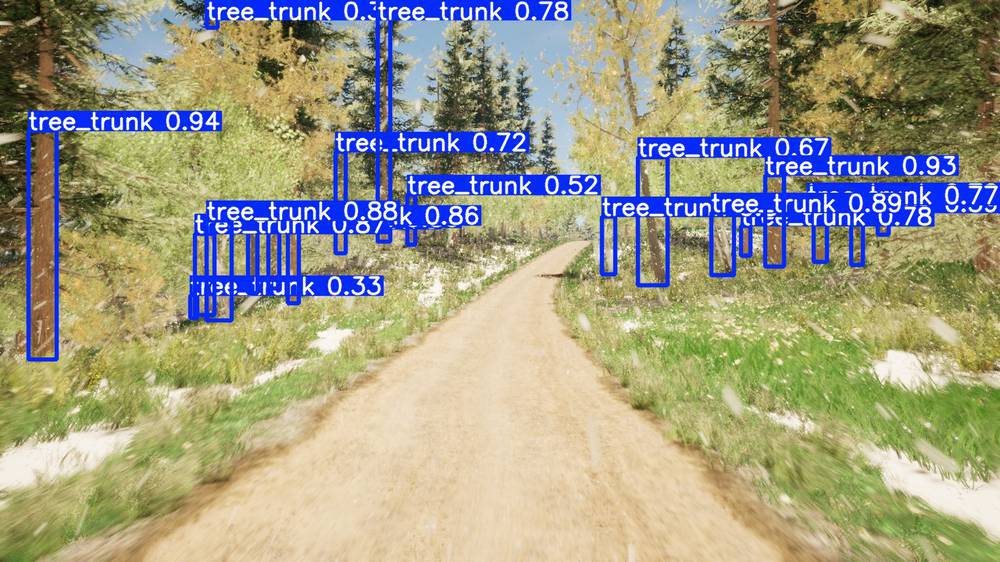} &
		\includegraphics[width=0.3\linewidth]{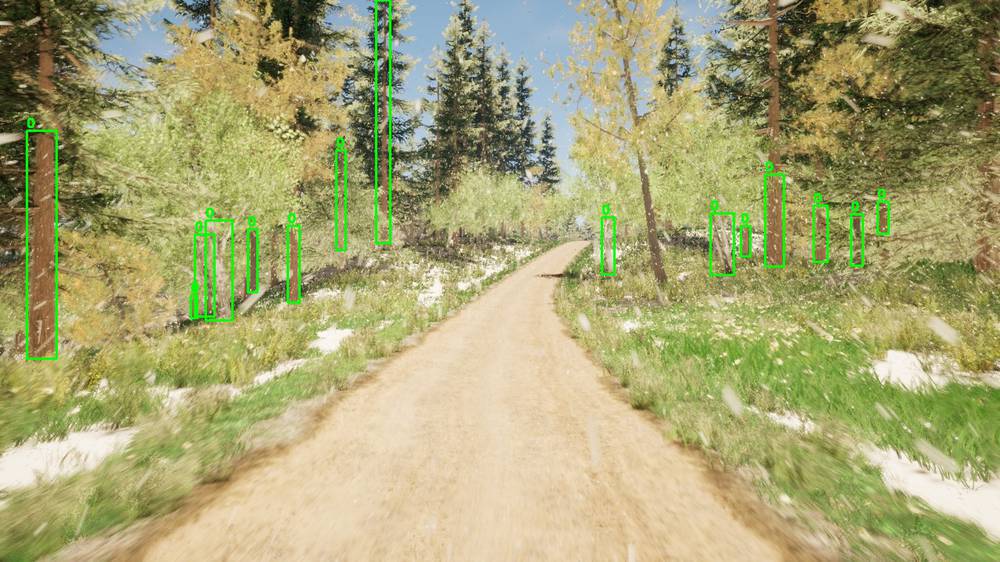} \\
		
		\bottomrule
	\end{tabular}
	
	\label{fig:yolo_phase_1_tree_trunk}
\end{figure}

\begin{figure}[ht]
	\centering
	\setlength{\tabcolsep}{1pt}      
	\renewcommand{\arraystretch}{0.9} 
	
	\caption{\textbf{Phase 1:} Qualitative analysis of yolov8m (trained on simulated whole trees) on the simulated whole trees val set.}
	
	\begin{tabular}{ccc}
		\toprule
		\textbf{RGB Image} & \textbf{Prediction} & \textbf{Ground Truth} \\
		\midrule
		
		\includegraphics[width=0.3\linewidth]{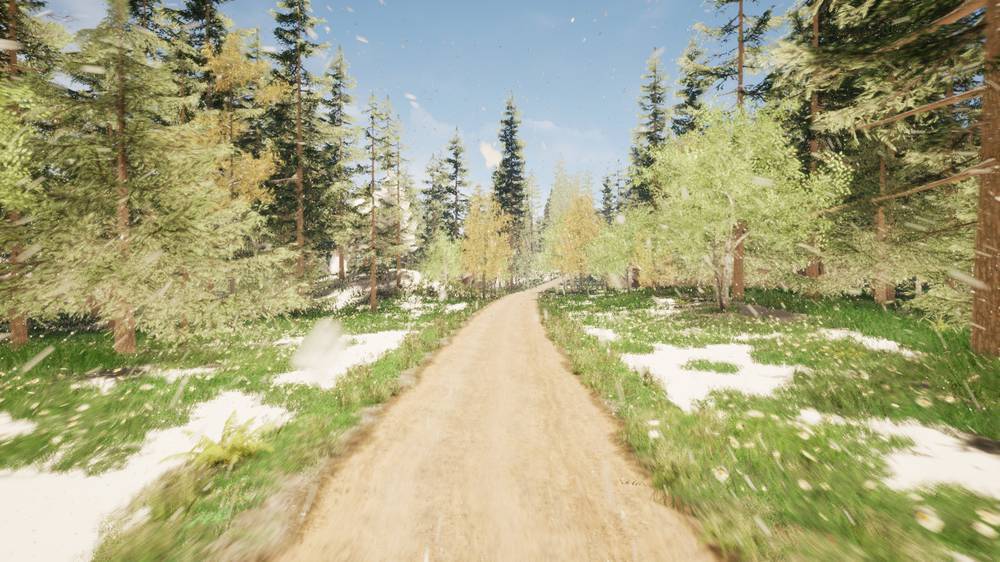} &
		\includegraphics[width=0.3\linewidth]{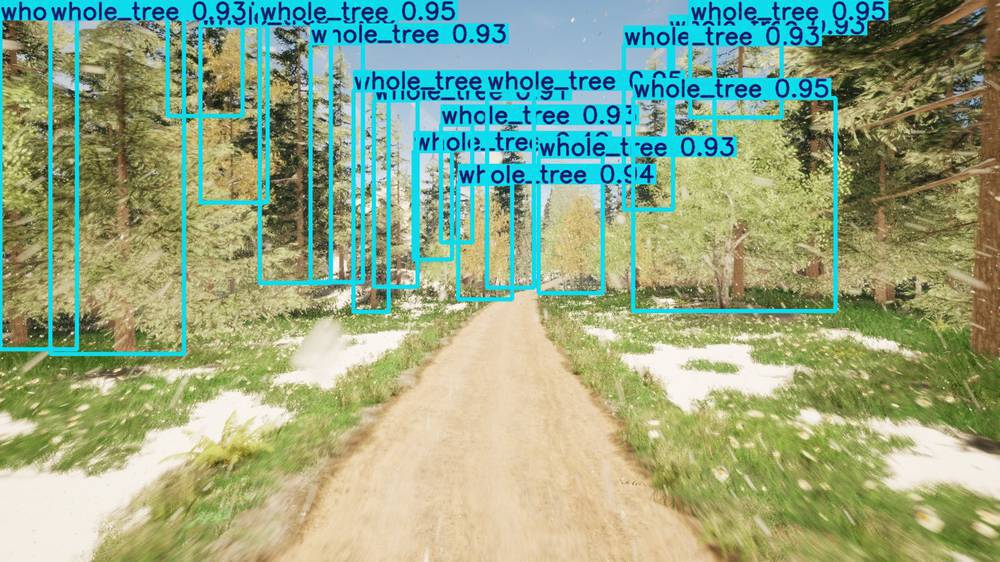} &
		\includegraphics[width=0.3\linewidth]{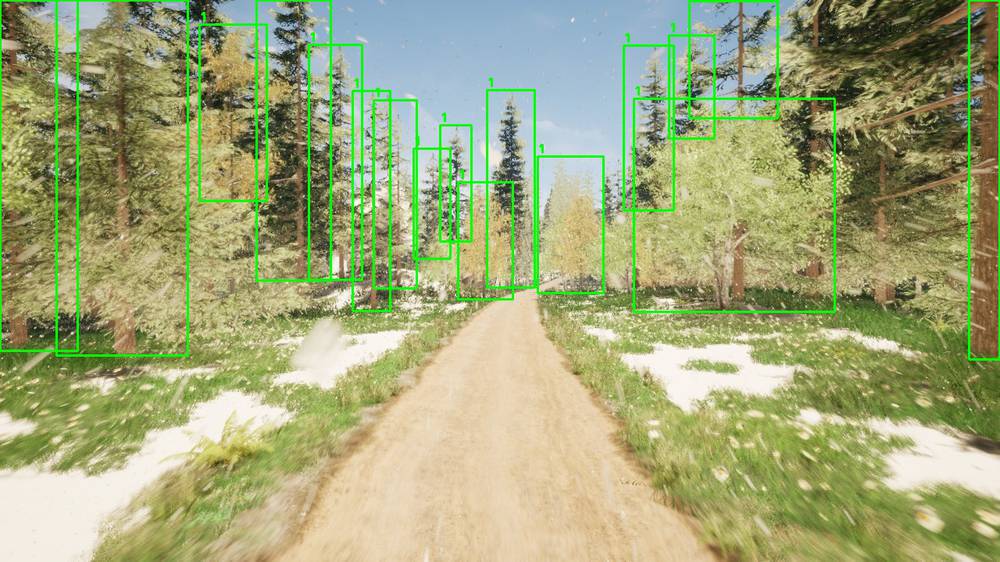} \\
		
		\includegraphics[width=0.3\linewidth]{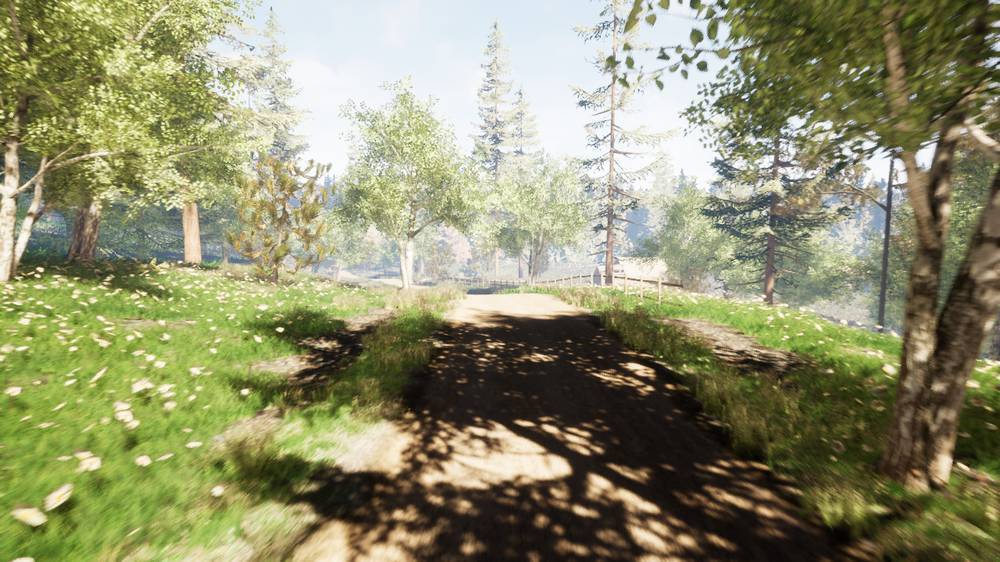} &
		\includegraphics[width=0.3\linewidth]{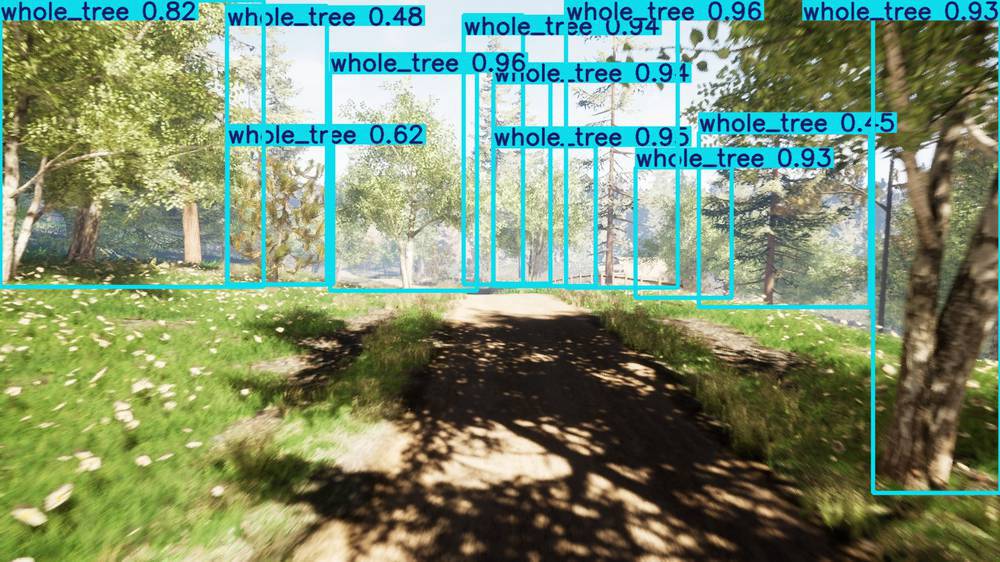} &
		\includegraphics[width=0.3\linewidth]{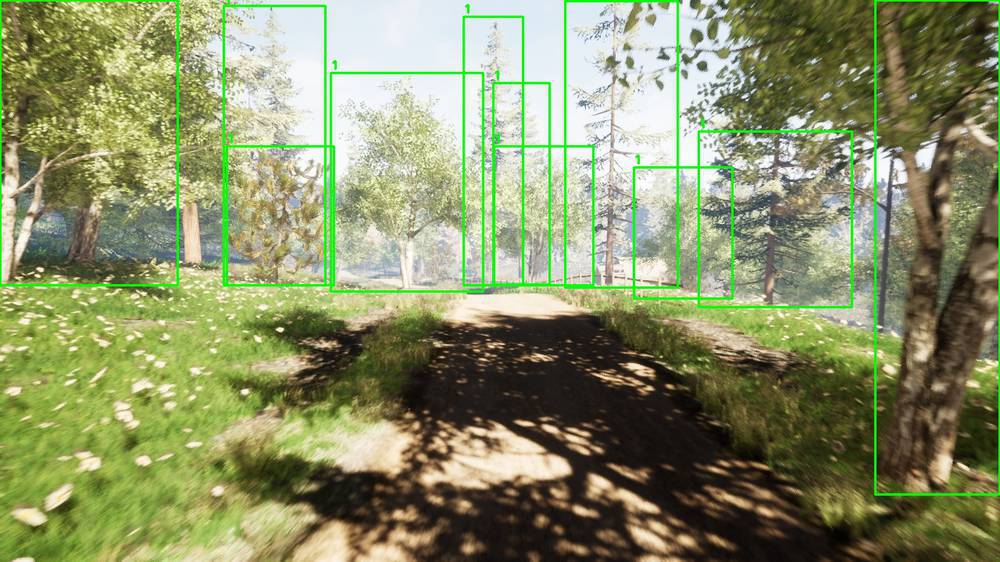} \\
		
		\includegraphics[width=0.3\linewidth]{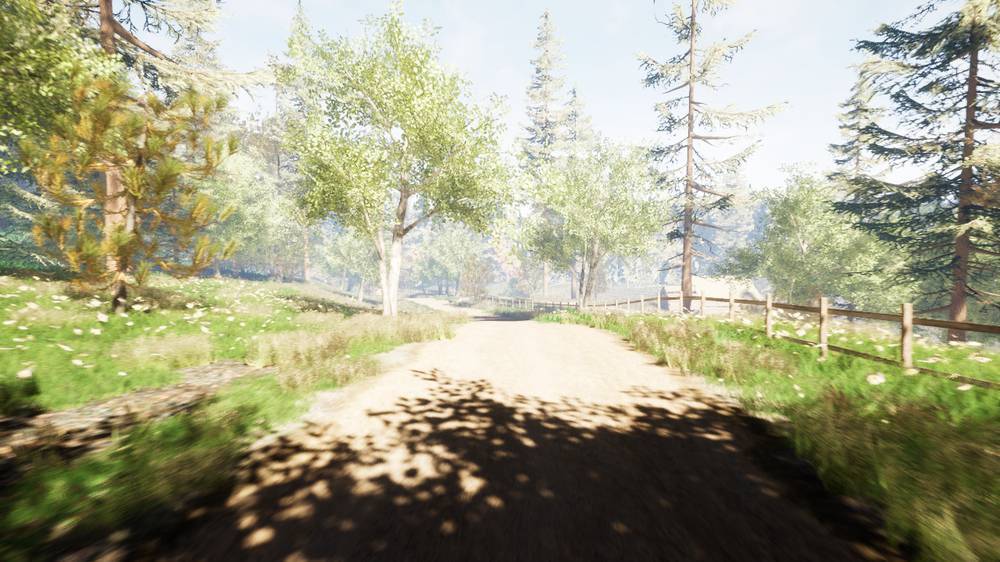} &
		\includegraphics[width=0.3\linewidth]{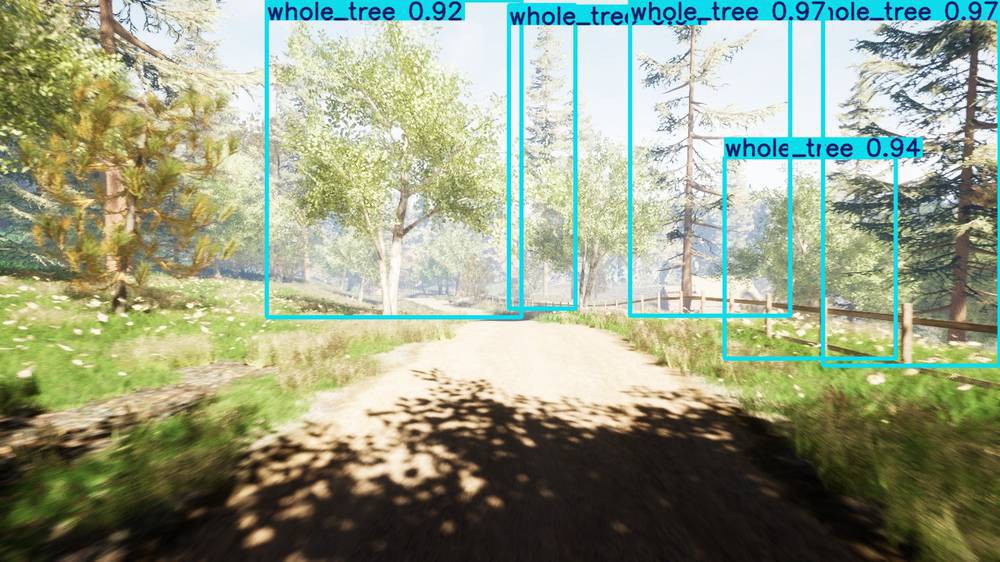} &
		\includegraphics[width=0.3\linewidth]{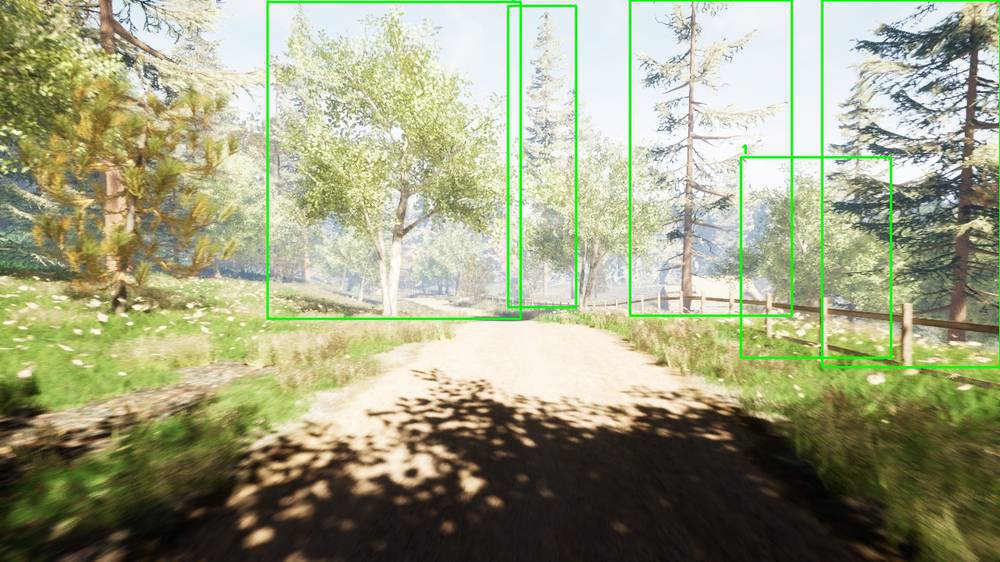} \\
		
		\bottomrule
	\end{tabular}
	
	\label{fig:yolo_phase_1_whole_tree}
\end{figure}

We show in Figure \ref{fig:yolo_phase_1_tree_trunk}, \ref{fig:yolo_phase_1_whole_tree} and Table \ref{tab:yolo_val_metrics_OD} both the qualitative and quantitative results of model training (object detection) on the simulated datasets. Quantitatively, the YOLO models achieve very strong validation performance on the synthetic domain, with tree trunks consistently outperforming whole trees. In particular, YOLO11m reaches the highest trunk accuracy (mAP@50:95 = 0.813), while YOLOv8m provides the strongest whole-tree performance (mAP@50:95 = 0.698). These results hypothesize that trunk detection is a ``comparatively simpler" task due to the vertical, elongated structure of stems, while whole-tree detection, which must account for irregular crowns and occlusions, introduces more ambiguity and results in lower precision.

The qualitative results in Figure \ref{fig:yolo_phase_1_tree_trunk}\&  \ref{fig:yolo_phase_1_whole_tree} further support these findings. On the simulated trunk val set, YOLO11m produces dense and consistent detections across diverse environmental conditions, including shadowed forest roads, high-contrast sunlit scenes, and cluttered backgrounds. The predictions align closely with the ground truth annotations, capturing both near- and far-field trees. Occasional errors occur in cases of severe occlusion or trees with very thin stems, where detections are sometimes fragmented or missed. Nevertheless, the overall high quality of detections visually mirrors the high quantitative scores (as seen in Table \ref{tab:yolo_val_metrics_OD}).
Altogether, the results demonstrate that synthetic training provides robust and reliable detectors within the simulation domain, setting a strong upper bound for downstream phases. However, as shown later in the real transfer analysis, this high simulated accuracy does not directly translate to real-world performance, emphasizing the importance of cross-domain adaptation strategies.

\begin{table*}[htp!]
\centering
\caption{\textbf{Phase 1:} Validation metrics for object detection on simulated tree trunk and whole tree datasets. Models are ordered by variant size (n, s, m, l).}
\resizebox{\textwidth}{!}{
\begin{tabular}{l l c c c c c c c c c c c c c c}
\toprule
Task & Model & Epoch & train/box\_loss & train/cls\_loss & train/dfl\_loss & Precision & Recall & mAP@50 & mAP@50:95 & val/box\_loss & val/cls\_loss & val/dfl\_loss & lr/pg0 & lr/pg1 & lr/pg2 \\
\midrule
Tree Trunk & YOLOv8n   & 100 & 0.64198 & 0.51119 & 0.85470 & 0.82755 & 0.87099 & 0.92217 & 0.77399 & 0.64272 & 0.53417 & 0.85107 & 0.000199 & 0.000199 & 0.000199 \\
           & YOLOv8s   & 100 & 0.55581 & 0.42323 & 0.83554 & 0.83738 & 0.87747 & 0.92875 & 0.79685 & 0.59484 & 0.49071 & 0.84336 & 0.000199 & 0.000199 & 0.000199 \\
           & YOLOv8m   & 100 & 0.52055 & 0.39357 & 0.83282 & 0.83614 & 0.88321 & 0.93024 & 0.80409 & 0.57936 & 0.47443 & 0.84955 & 0.000199 & 0.000199 & 0.000199 \\
           & YOLOv8l   & 100 & 0.50205 & 0.36515 & 0.82760 & 0.84004 & 0.88467 & 0.93074 & 0.80960 & 0.56565 & 0.46984 & 0.85554 & 0.000199 & 0.000199 & 0.000199 \\
           & YOLOv11n  & 100 & 0.62017 & 0.50105 & 0.84677 & 0.83320 & 0.87276 & 0.92494 & 0.78447 & 0.61998 & 0.52759 & 0.84075 & 0.000199 & 0.000199 & 0.000199 \\
           & YOLOv11s  & 100 & 0.54407 & 0.42742 & 0.83043 & 0.83856 & 0.88160 & 0.93020 & 0.80234 & 0.58274 & 0.48856 & 0.83403 & 0.000199 & 0.000199 & 0.000199 \\
           & YOLOv11m  & 100 & 0.51786 & 0.39687 & 0.82988 & 0.84114 & 0.88638 & 0.93341 & 0.81335 & 0.56064 & 0.46746 & 0.84689 & 0.000199 & 0.000199 & 0.000199 \\
           & YOLOv11l  & 100 & 0.51565 & 0.38905 & 0.82589 & 0.83684 & 0.89109 & 0.93232 & 0.81203 & 0.55931 & 0.45899 & 0.85141 & 0.000199 & 0.000199 & 0.000199 \\
\midrule
Whole Tree & YOLOv8n   & 100 & 0.35592 & 0.33256 & 0.84311 & 0.64212 & 0.86482 & 0.69876 & 0.66019 & 0.28603 & 2291.9 & 0.58077 & 0.000199 & 0.000199 & 0.000199 \\
           & YOLOv8s   & 100 & 0.30053 & 0.27154 & 0.82319 & 0.64408 & 0.88037 & 0.71664 & 0.68769 & 0.25435 & 1490.71 & 0.56835 & 0.000199 & 0.000199 & 0.000199 \\
           & YOLOv8m   & 100 & 0.27476 & 0.23765 & 0.82145 & 0.65013 & 0.88354 & 0.72400 & 0.69803 & 0.24033 & 793.191 & 0.58496 & 0.000199 & 0.000199 & 0.000199 \\
           & YOLOv8l   & 100 & 0.30578 & 0.27204 & 0.86636 & 0.64819 & 0.88396 & 0.72061 & 0.69591 & 0.23601 & 609.86 & 0.62346 & 0.000199 & 0.000199 & 0.000199 \\
           & YOLOv11n  & 100 & 0.42346 & 0.41189 & 0.88489 & 0.64137 & 0.87025 & 0.70673 & 0.66988 & 0.28085 & 1988.48 & 0.57785 & 0.000199 & 0.000199 & 0.000199 \\
           & YOLOv11s  & 100 & 0.30189 & 0.27667 & 0.82364 & 0.65025 & 0.87879 & 0.72431 & 0.69573 & 0.25015 & 1156.96 & 0.56736 & 0.000199 & 0.000199 & 0.000199 \\
           & YOLOv11m  & 100 & 0.29040 & 0.26305 & 0.83015 & 0.64740 & 0.88966 & 0.72153 & 0.69503 & 0.24031 & 632.75 & 0.58016 & 0.000199 & 0.000199 & 0.000199 \\
           & YOLOv11l  & 100 & 0.29188 & 0.26549 & 0.84434 & 0.64457 & 0.89342 & 0.71787 & 0.69204 & 0.23835 & 493.656 & 0.63417 & 0.000199 & 0.000199 & 0.000199 \\
\bottomrule
\end{tabular}
}

\label{tab:yolo_val_metrics_OD}
\end{table*}

\begin{table*}[!htbp]
\centering
\caption{\textbf{Phase 1:} Validation metrics for object detection on the validation set (tree trunk vs whole tree) using RT-DETR + RegNet.}
\resizebox{\linewidth}{!}{
\begin{tabular}{l l c c c c c c c c}
\toprule
\textbf{Type (Tree trunk / Whole tree)} & \textbf{Model} & \textbf{mAP@0.5:0.95} & \textbf{mAP@0.50} & \textbf{AP50} & \textbf{AP75} & \textbf{APs} & \textbf{APm} & \textbf{API} & \textbf{mAR@100} \\
\midrule
Whole tree & RT-DETR + RegNet & 0.920 & 0.962 & 0.962 & 0.957 & 0.817 & 0.938 & 0.917 & 0.974 \\
Tree trunk & RT-DETR + RegNet & 0.825 & 0.947 & 0.947 & 0.928 & 0.797 & 0.861 & 0.899 & 0.973 \\
\bottomrule
\end{tabular}}

\label{tab:val_rtdetr}
\vspace{-1.25em} 
\end{table*}

\clearpage
\subsection{Phase 2: Domain gap}

Out of the various model trained during phase 1, we take the best tree trunk (YOLOv11m) and whole tree (YOLOv8m) model and directly test on the real world tree images. Figures \ref{fig:yolo_tt_real}, \ref{fig:yolo_wt_real} and Table \ref{tab:yolo_domain_gap}, \ref{tab:sim2real_rtdetr} further provide a detailed view of the domain gap. On the synthetic validation set, both detectors achieve strong in-domain results (as seen from table \ref{tab:yolo_val_metrics_OD}). However, when evaluated zero-shot on real forest imagery, both models show considerable degradation, albeit with different severity. The trunk-trained YOLOv11m exhibits a steep performance decline, with accuracy falling by an absolute gap of 0.53 (corresponding to a 56.8\% relative drop). This sharp loss indicates that the fine-grained representation of tree trunks, while effective in simulation, does not generalize well to the varied textures, occlusions, and illumination conditions present in real forests. In contrast, the whole-tree trained YOLOv8m model shows a smaller but still significant absolute drop of 0.24 (33\% relative decline), suggesting that the coarser structural representation (crowns and stems combined) is somewhat ``a little more" robust to domain shift.

\begin{figure}[ht]
\centering
\includegraphics[width=0.7\linewidth]{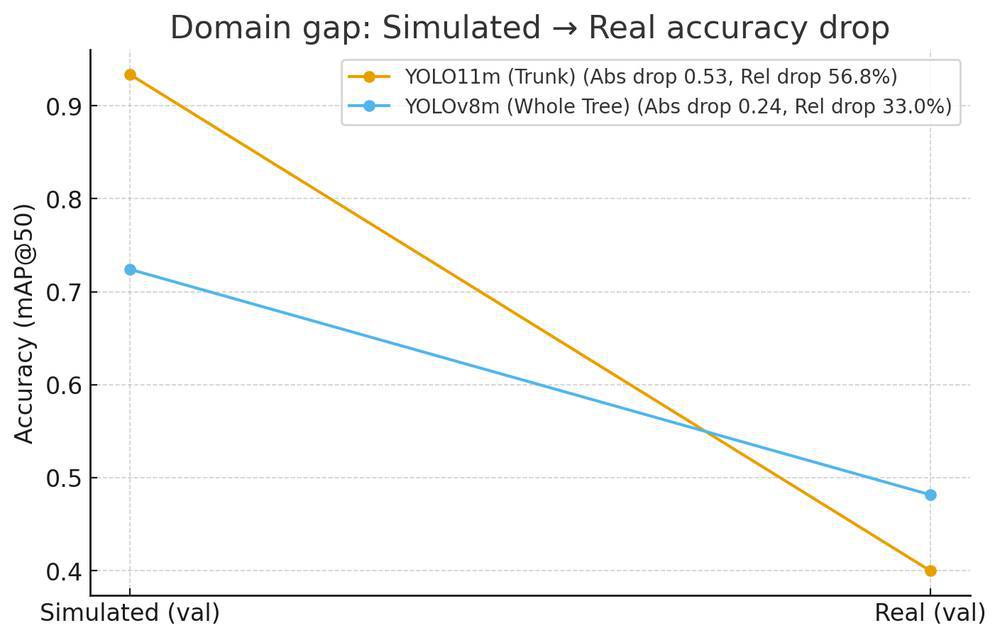} 
\caption{\textbf{Phase 2:} Domain gap when the best simulated (tree trunk and whole tree model) was directly tested on real tree images.}
\label{fig:yolo_domain_result}
\end{figure}

\begin{table}[htp!]
\vspace{-0.75em} 
\centering
\caption{Qualitative analysis of YOLOv11m (trained on simulated tree trunks) on the real trees test set.}
\setlength{\tabcolsep}{1pt} 
\renewcommand{\arraystretch}{0.9} 
\begin{tabular}{@{}ccc@{}}
\toprule
\textbf{RGB Image} & \textbf{Prediction} & \textbf{Ground Truth} \\
\midrule
\includegraphics[width=0.25\linewidth, angle=180, origin=c]{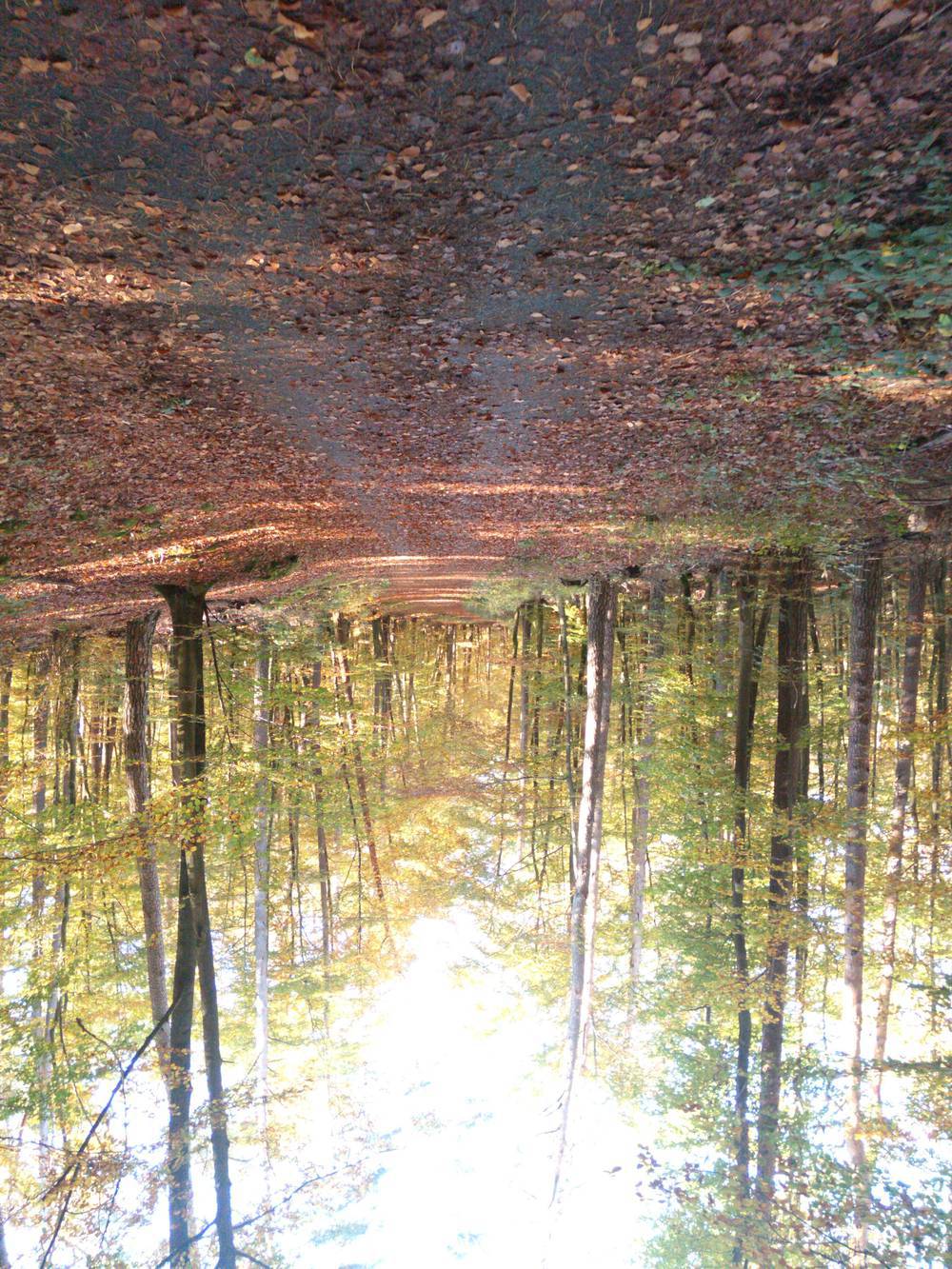} & 
\includegraphics[width=0.25\linewidth]{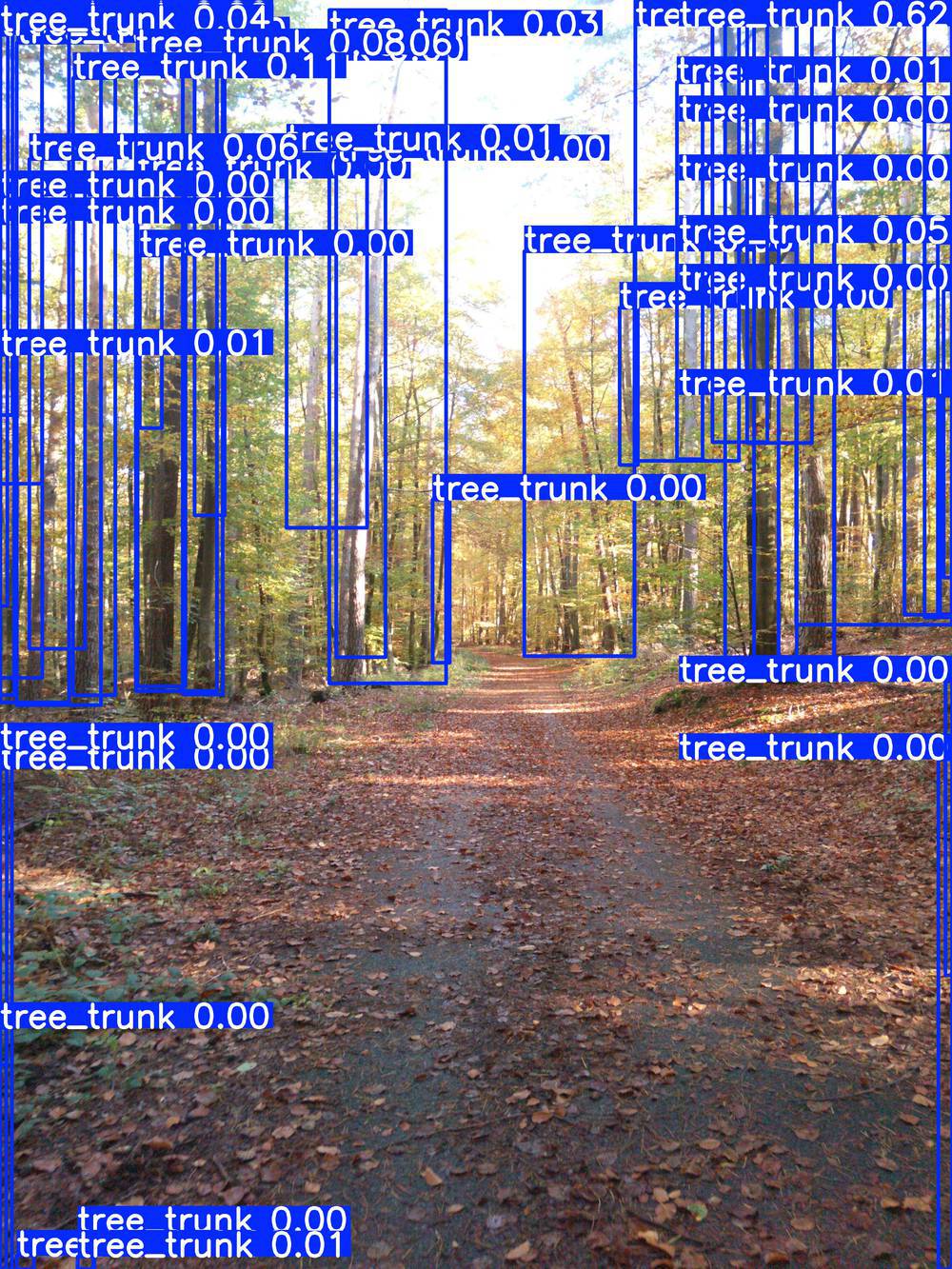} & 
\includegraphics[width=0.25\linewidth]{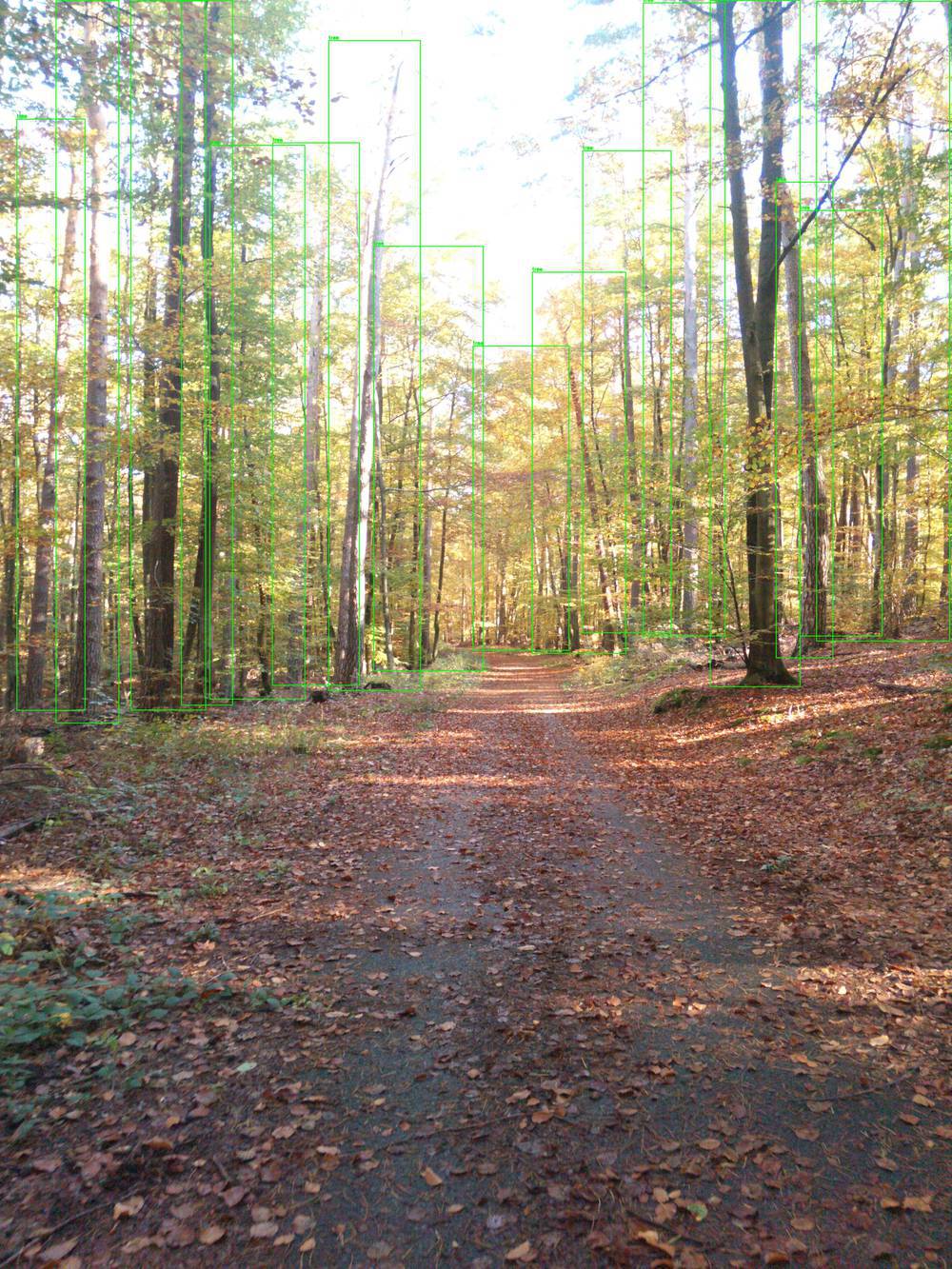} \\
[-2pt]
\includegraphics[width=0.25\linewidth, angle=180, origin=c]{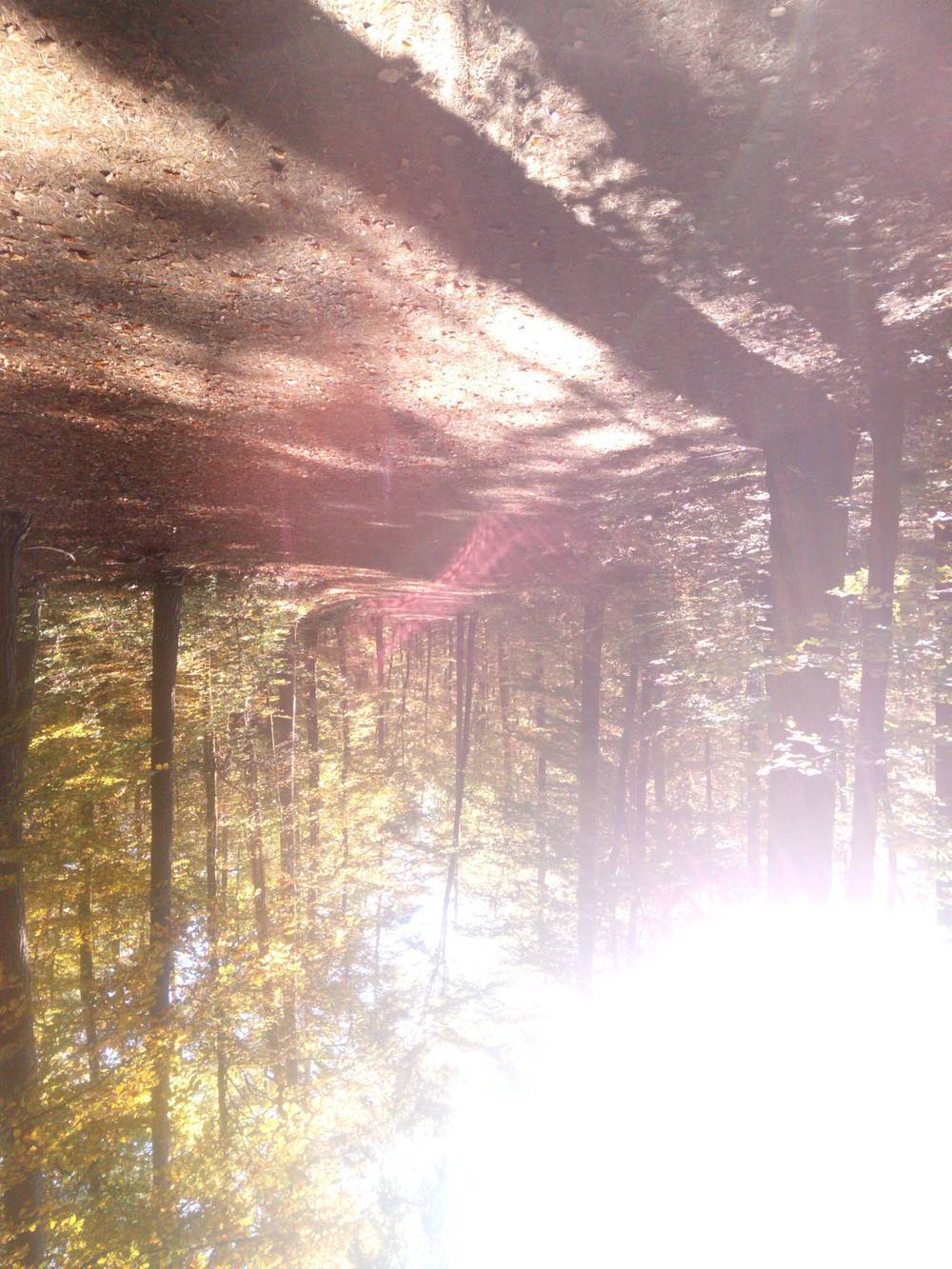} & 
\includegraphics[width=0.25\linewidth]{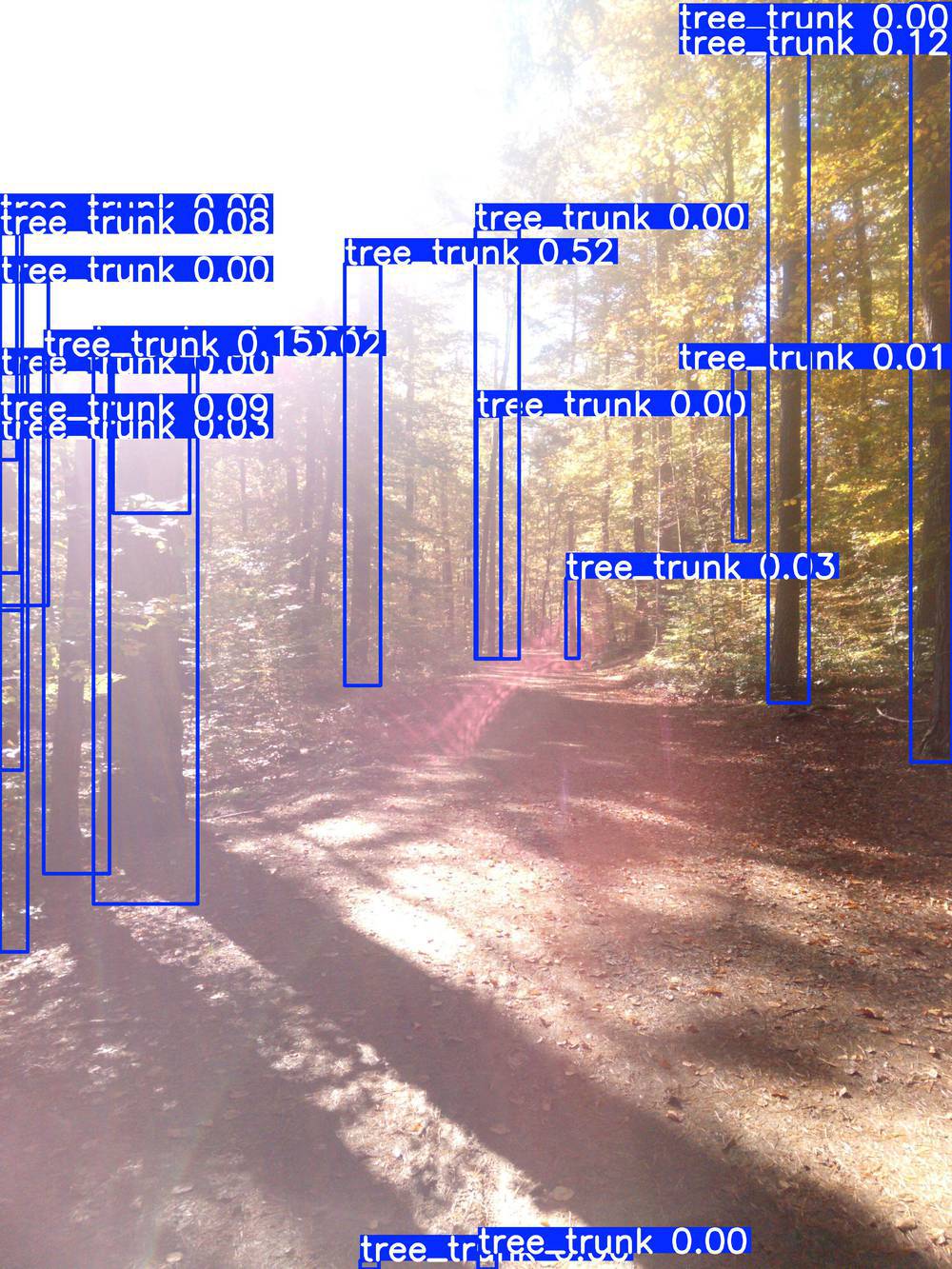} & 
\includegraphics[width=0.25\linewidth]{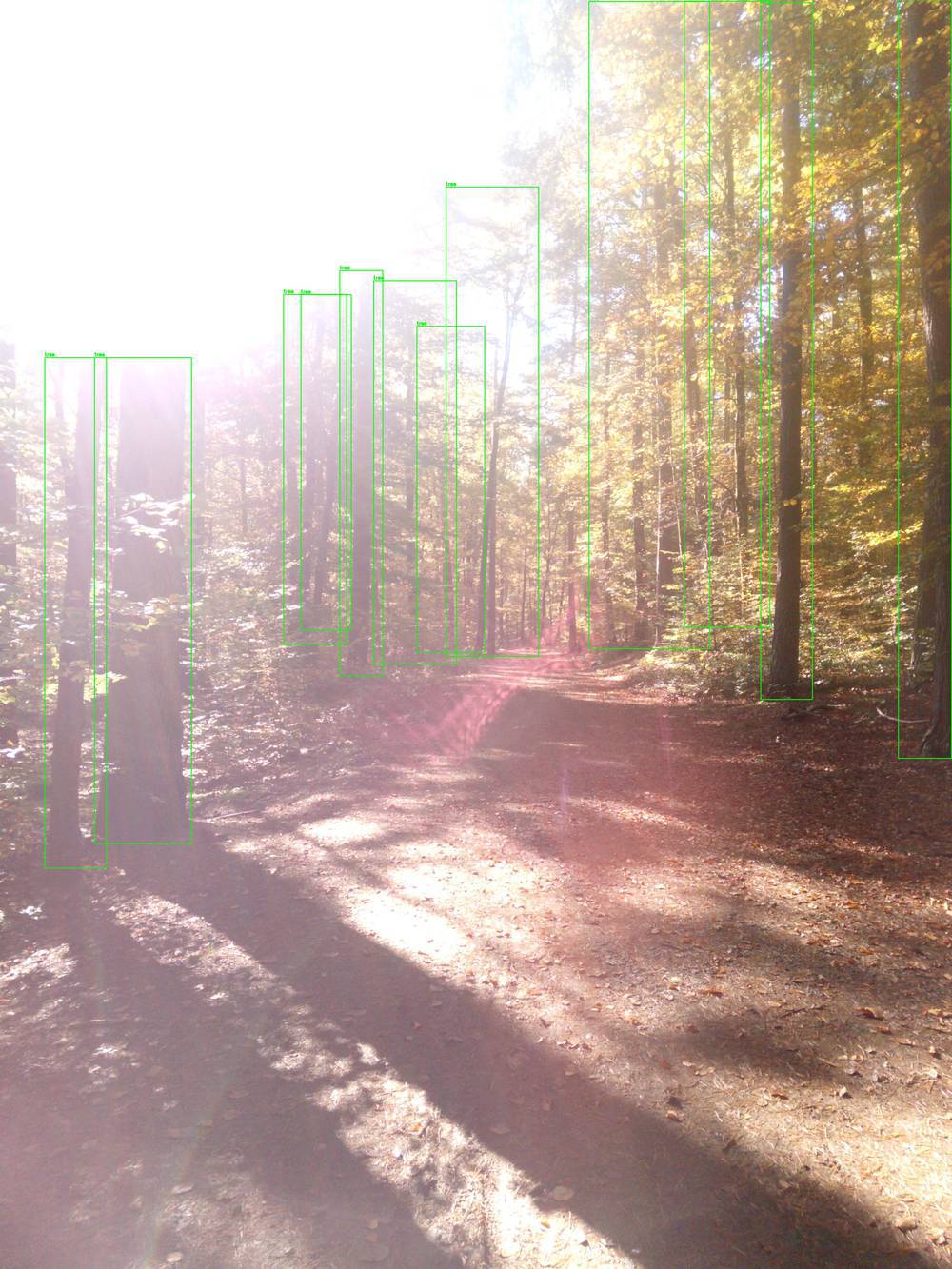} \\
[-2pt]
\includegraphics[width=0.25\linewidth, angle=180, origin=c]{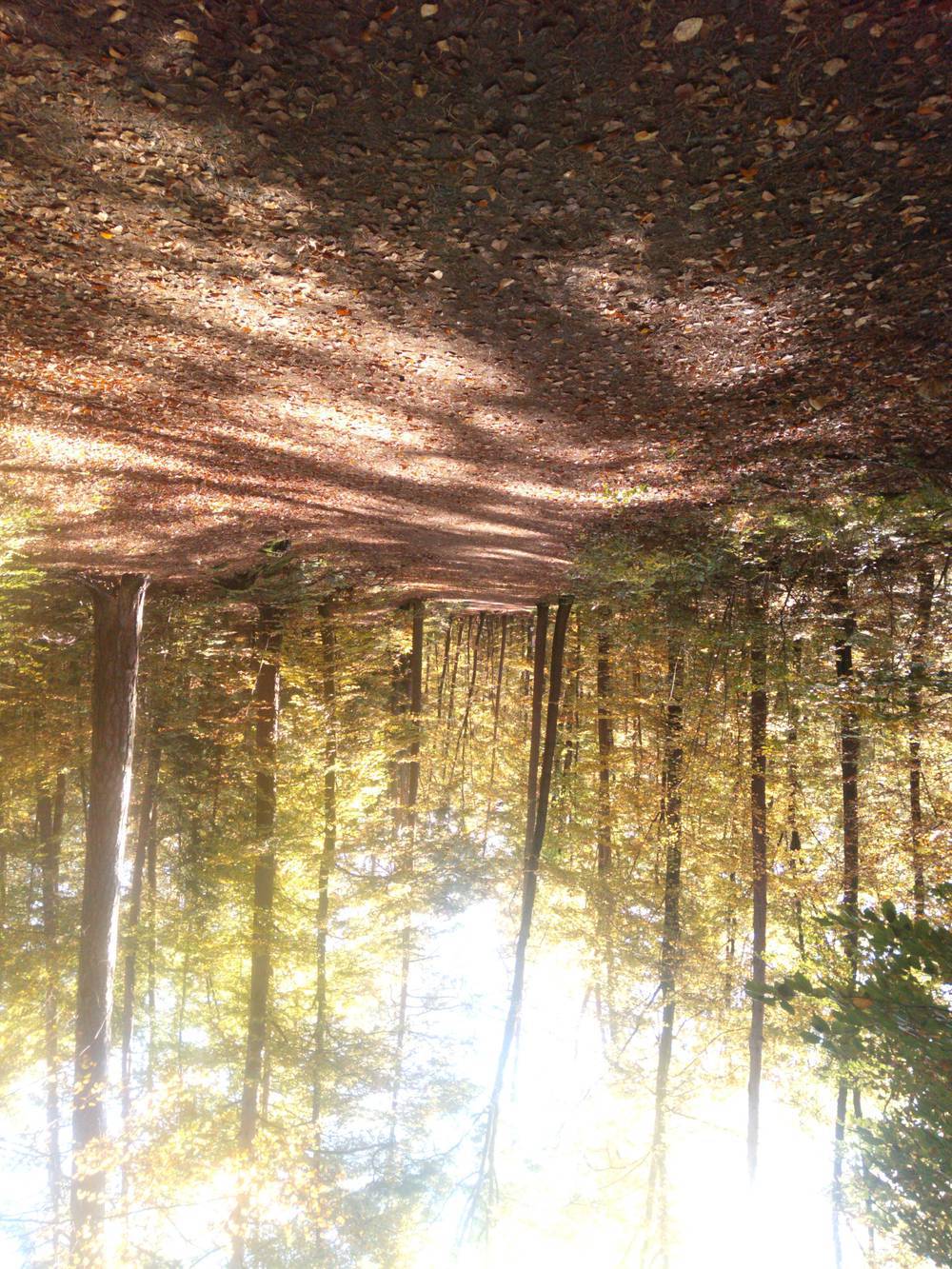} & 
\includegraphics[width=0.25\linewidth]{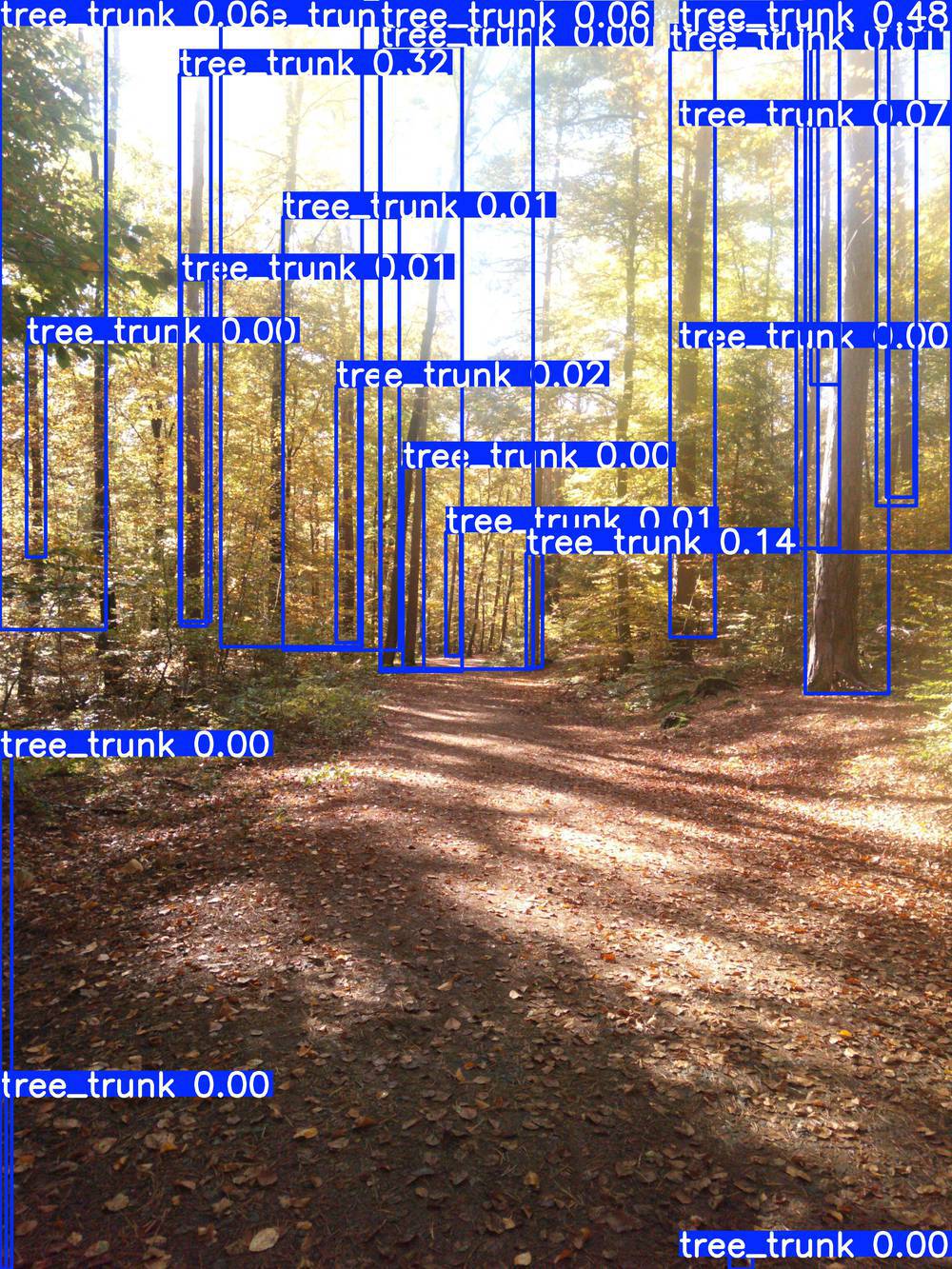} & 
\includegraphics[width=0.25\linewidth]{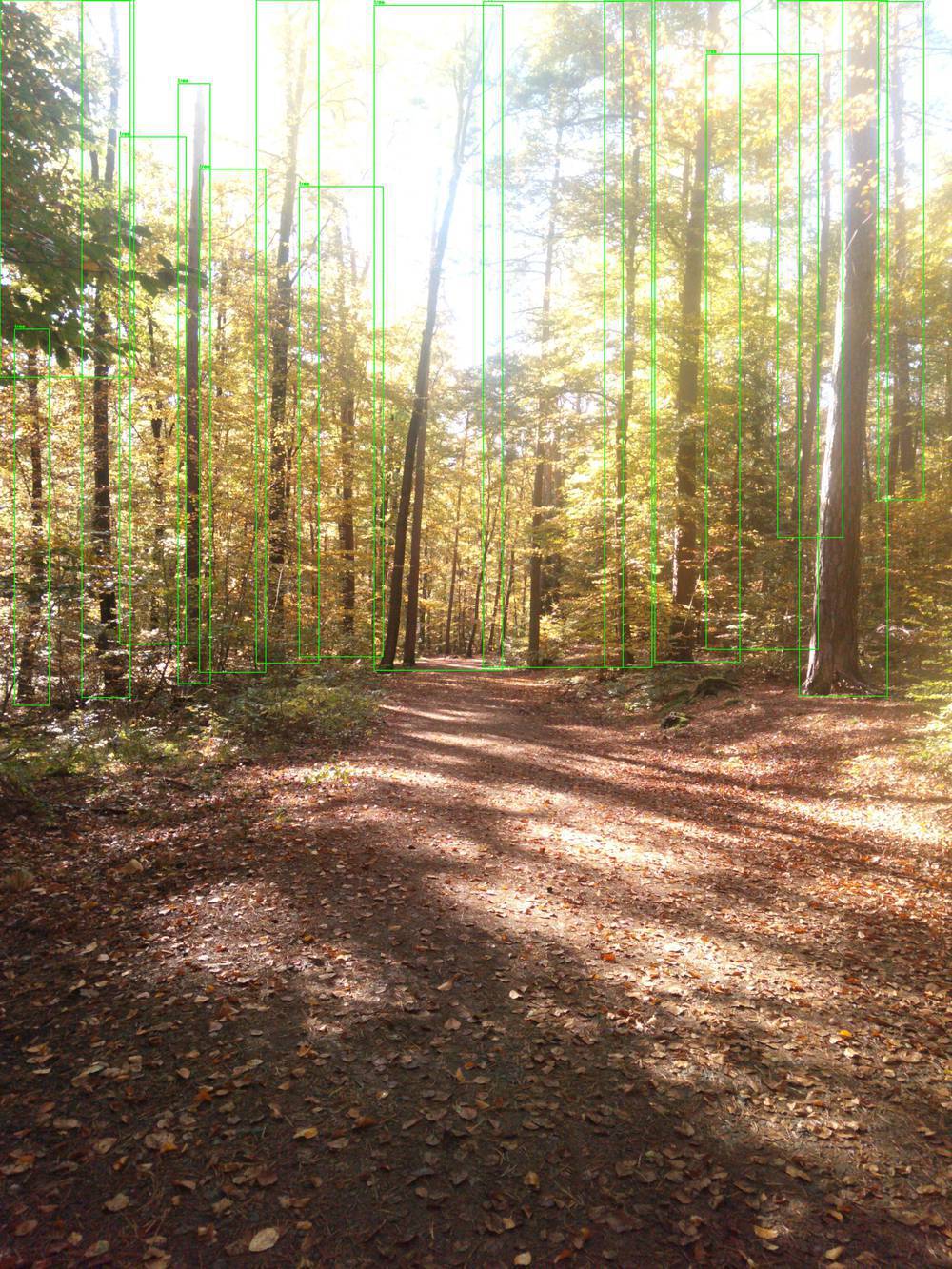} \\
\bottomrule
\end{tabular}
\label{fig:yolo_tt_real}
\end{table}

\begin{table}[ht]
\centering
\caption{Qualitative analysis of YOLOv8m (trained on simulated whole trees) on the real trees test set.}
\setlength{\tabcolsep}{1pt} 
\renewcommand{\arraystretch}{0.9} 
\begin{tabular}{@{}ccc@{}}
\toprule
\textbf{RGB Image} & \textbf{Prediction} & \textbf{Ground Truth} \\
\midrule
\includegraphics[width=0.25\linewidth, angle=180, origin=c]{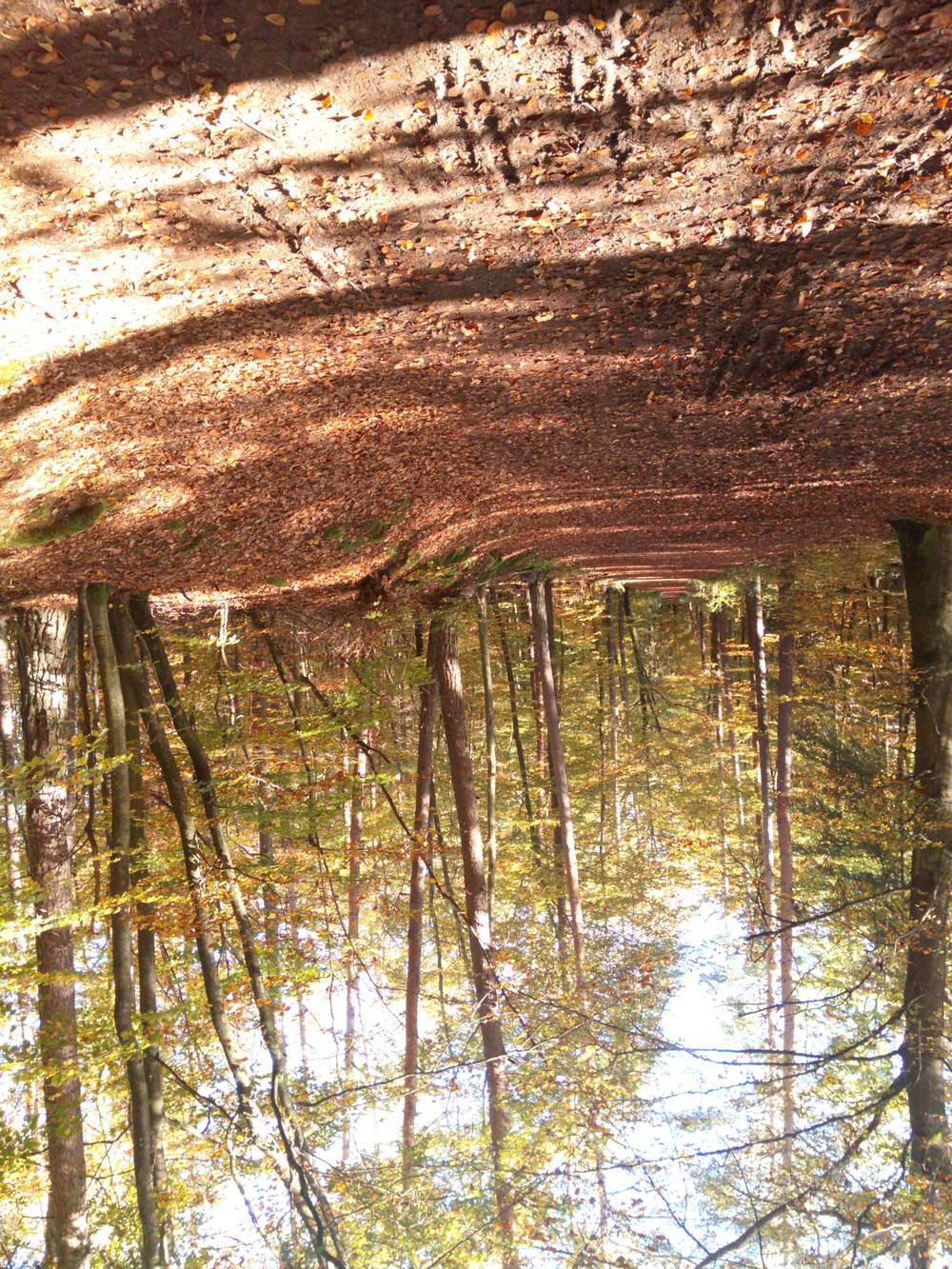} & 
\includegraphics[width=0.25\linewidth]{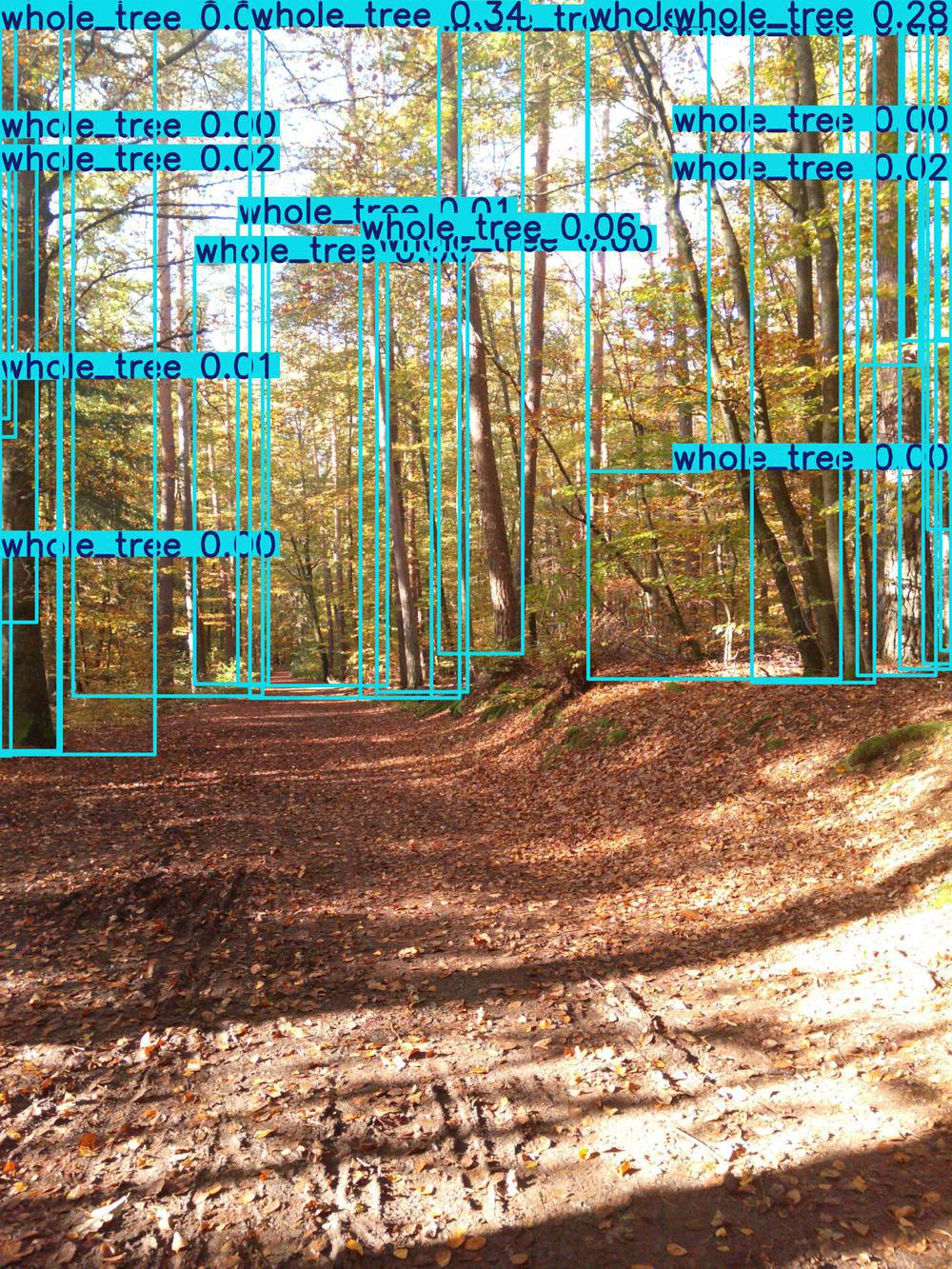} & 
\includegraphics[width=0.25\linewidth]{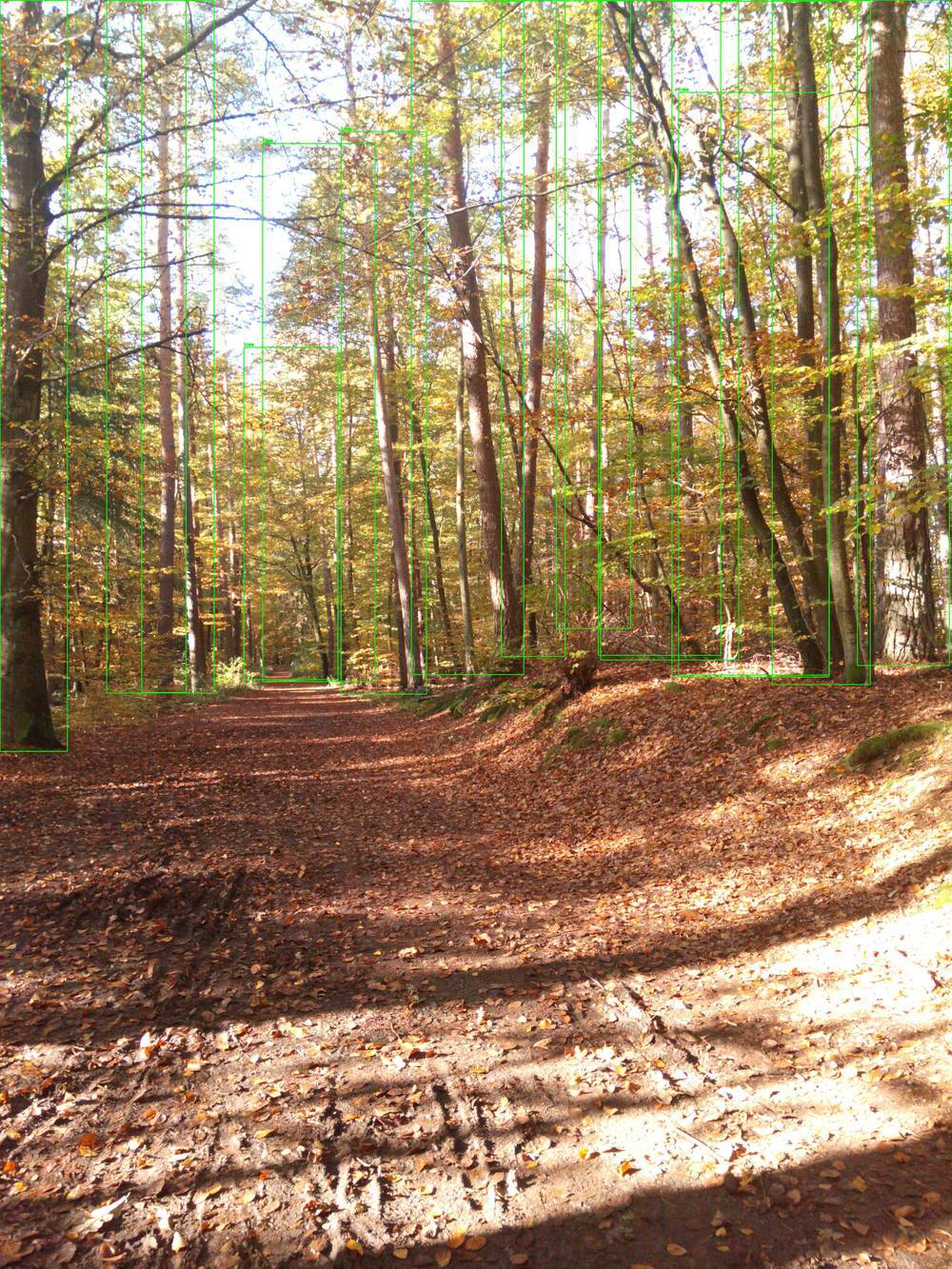} \\
[-2pt]
\includegraphics[width=0.25\linewidth, angle=180, origin=c]{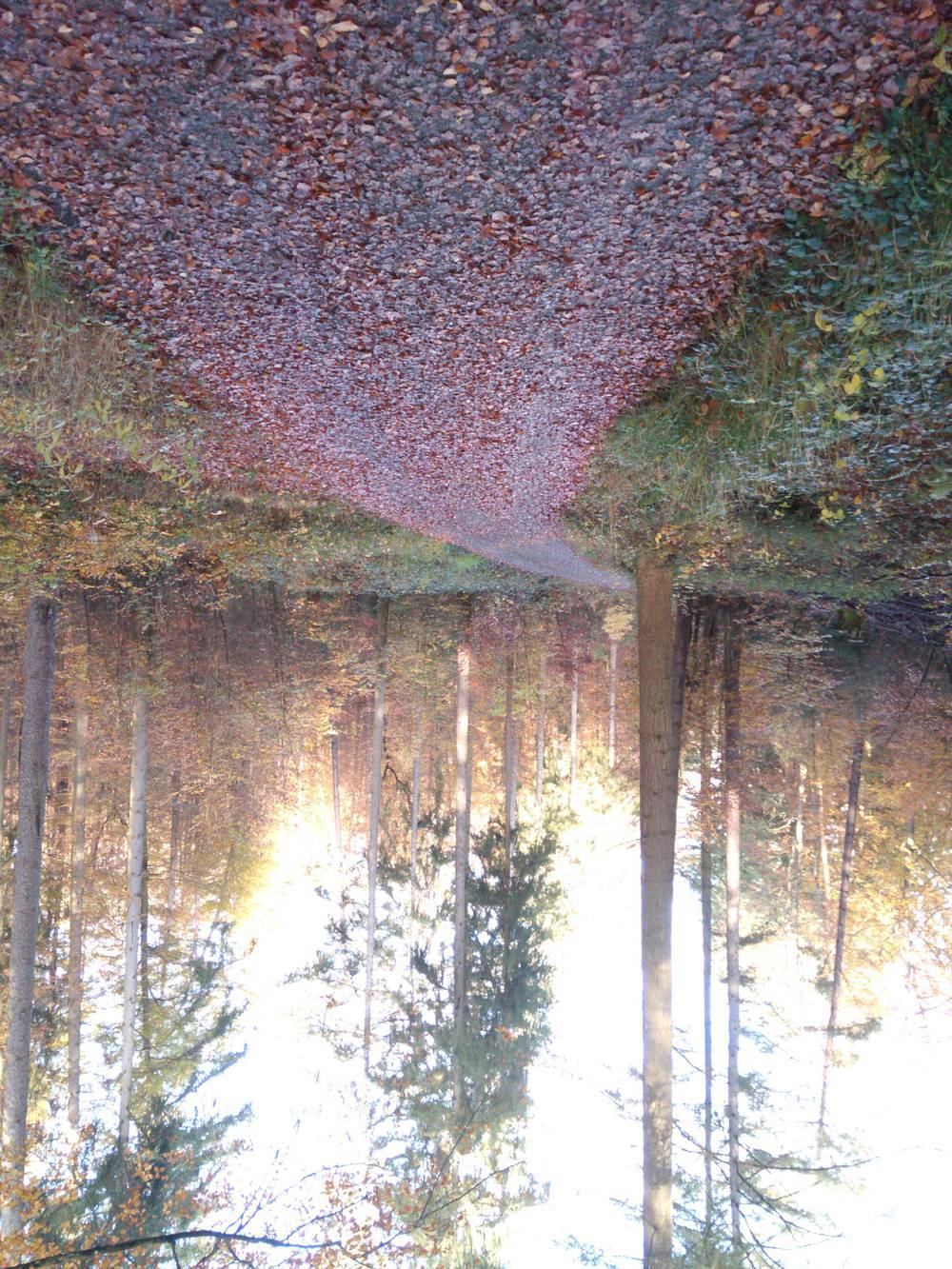} & 
\includegraphics[width=0.25\linewidth]{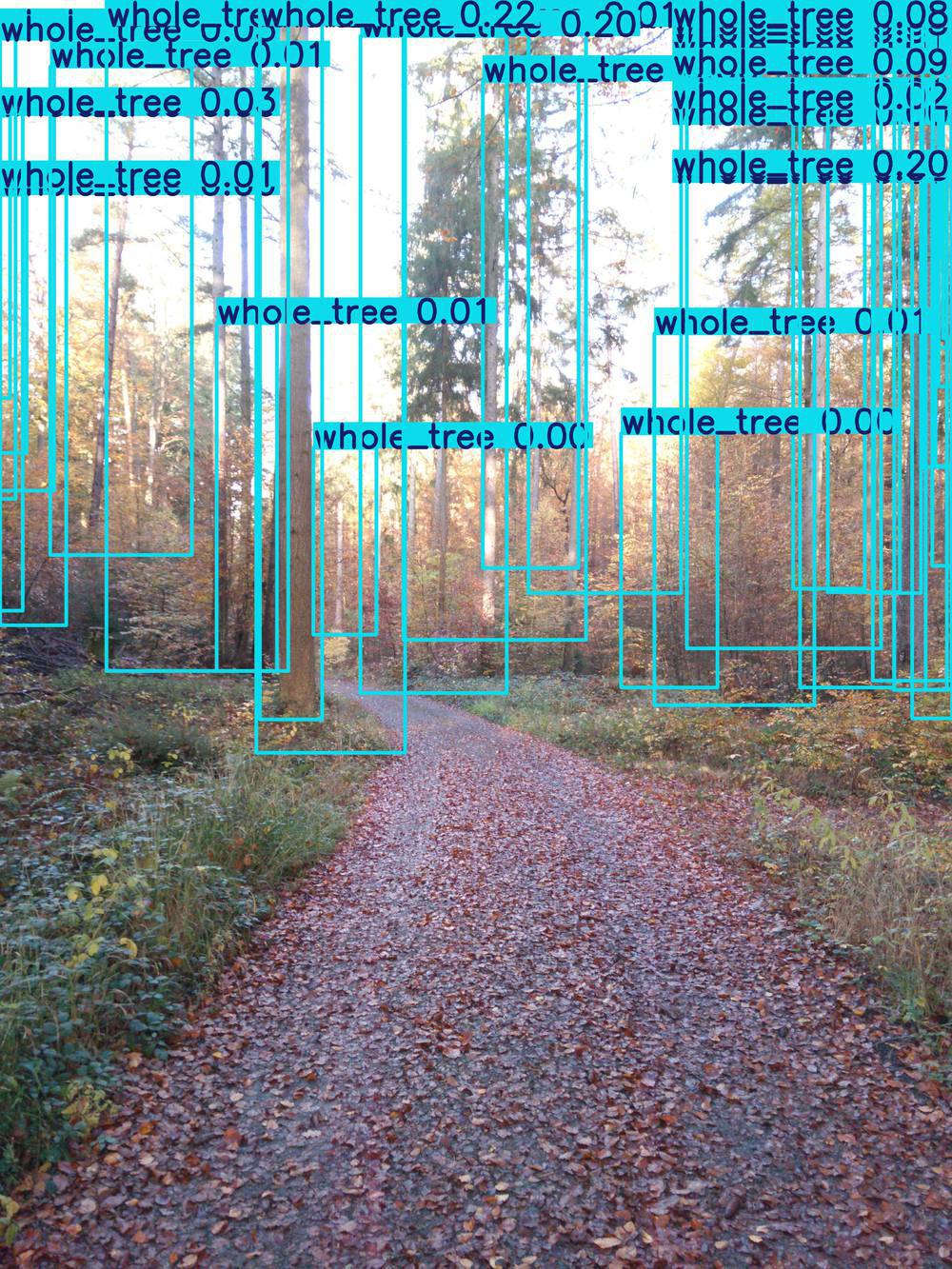} & 
\includegraphics[width=0.25\linewidth]{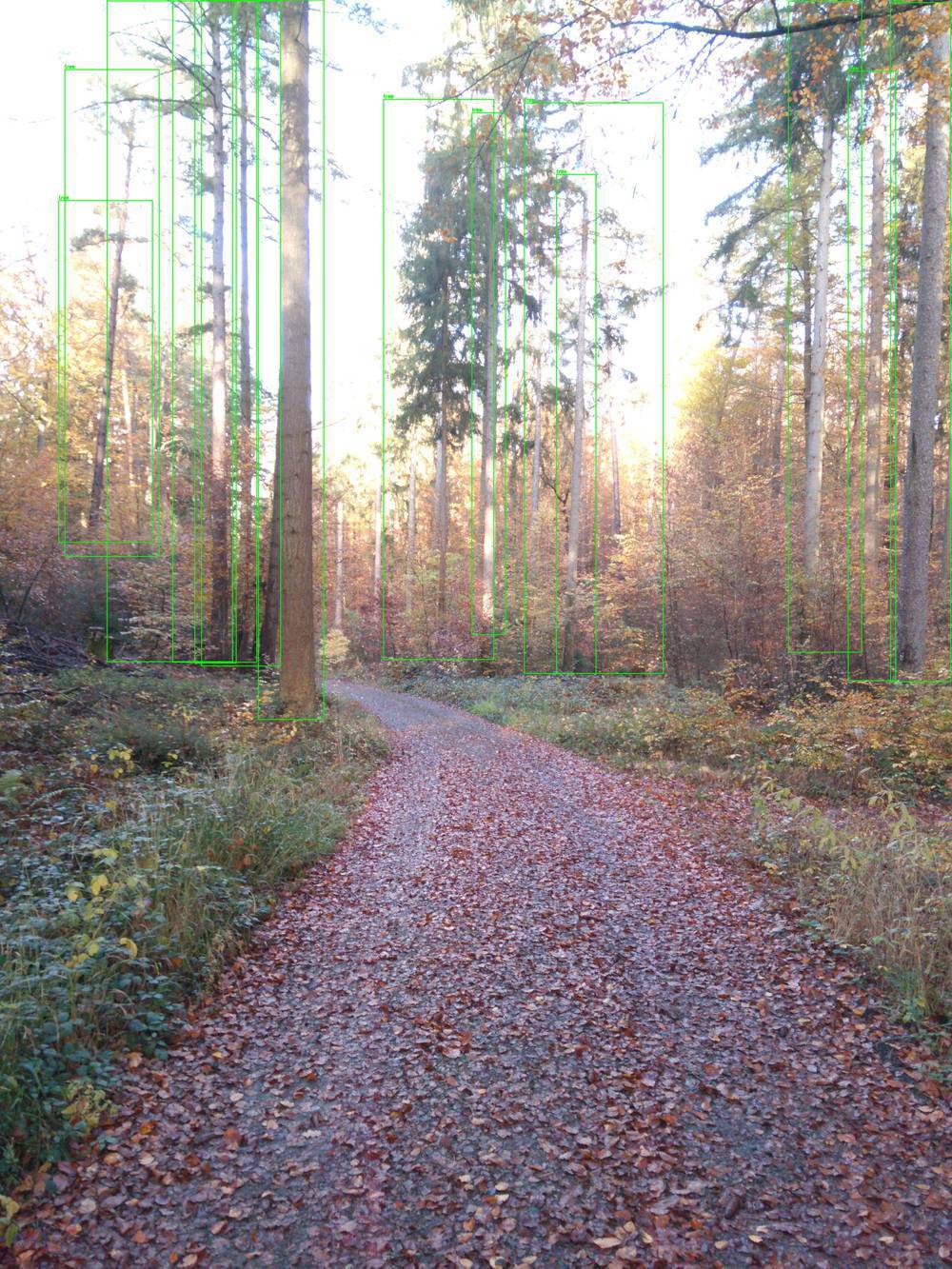} \\ 
[-2pt]
\includegraphics[width=0.25\linewidth, angle=180, origin=c]{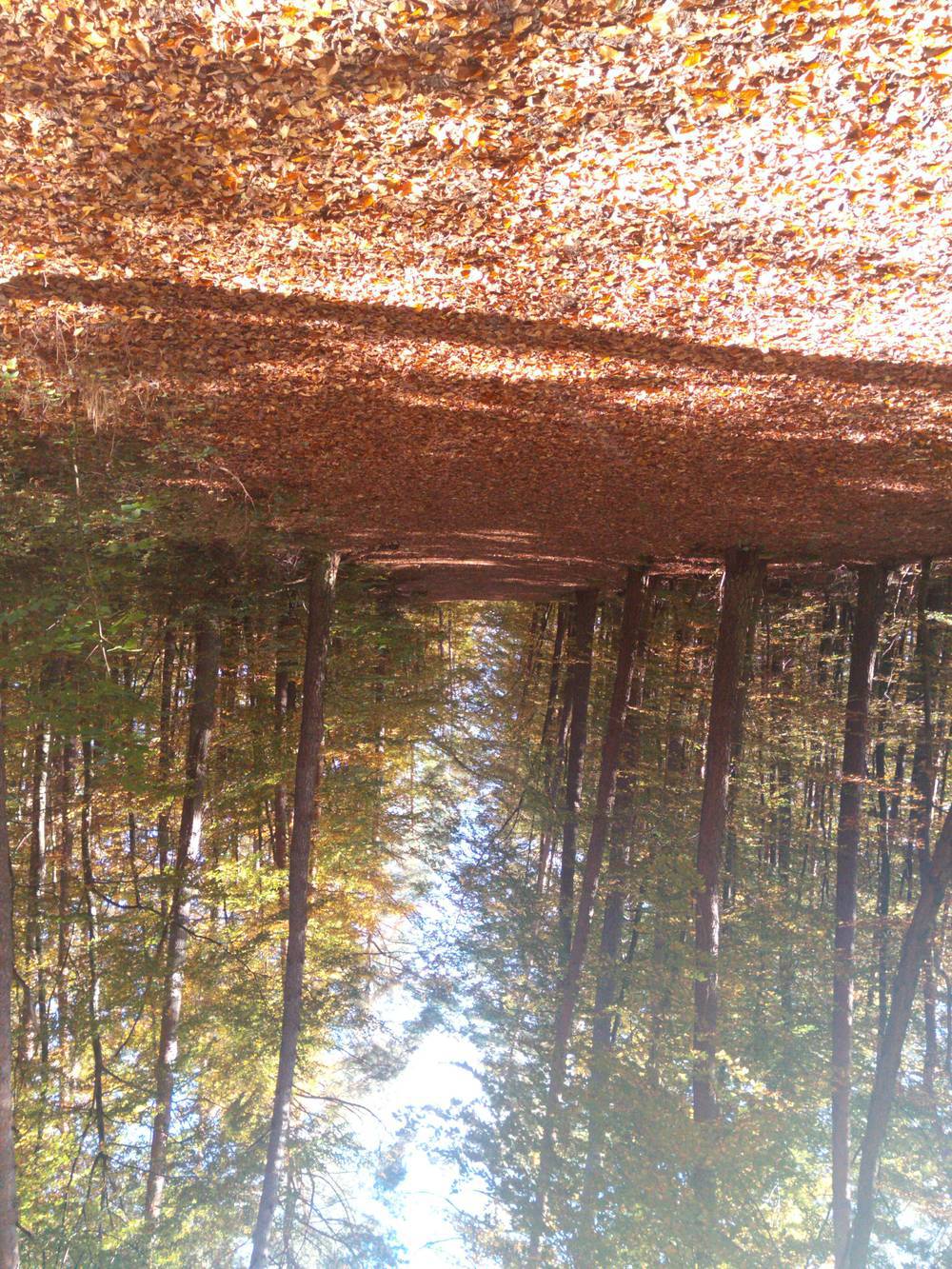} &
\includegraphics[width=0.25\linewidth]{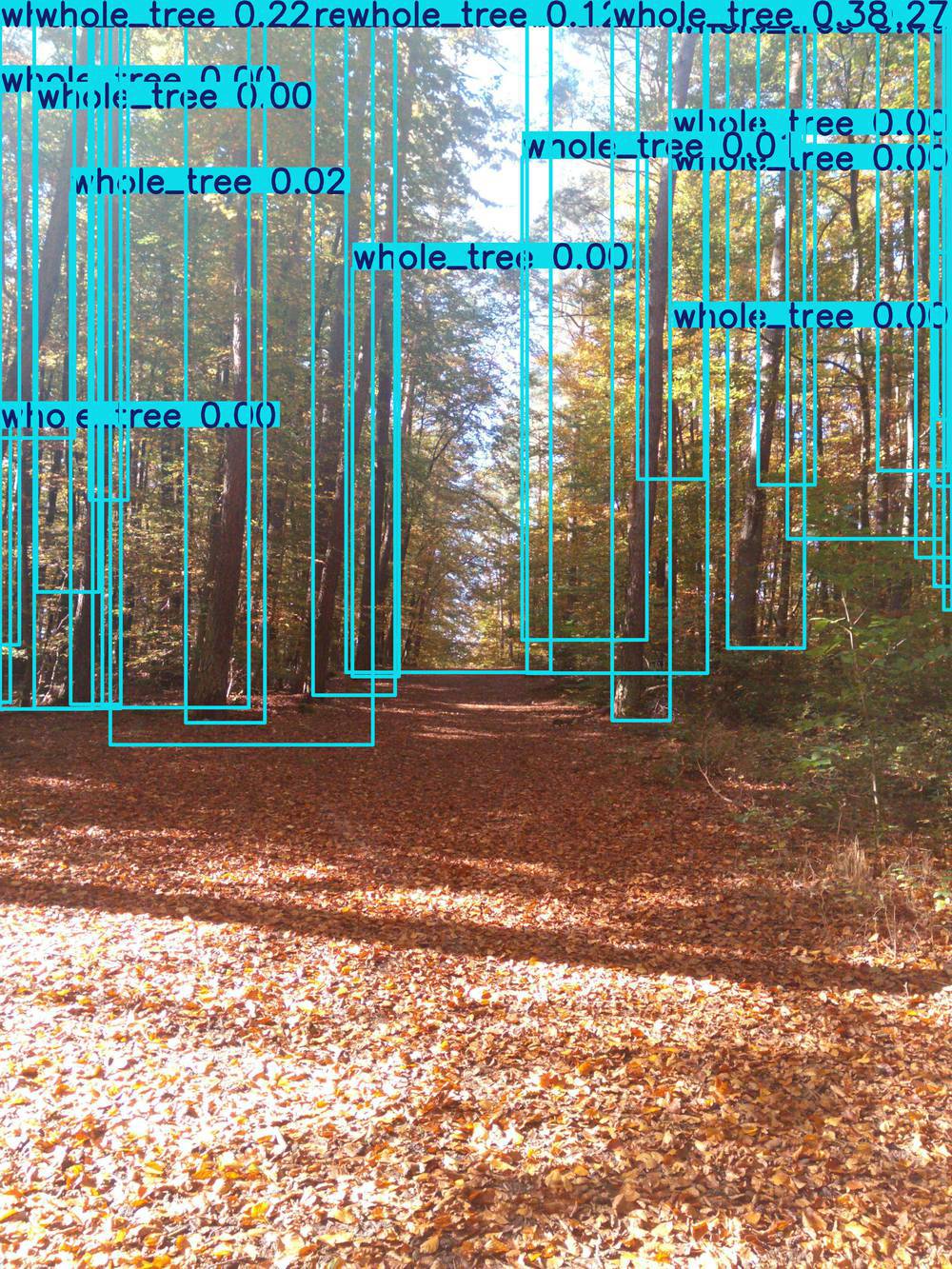} & 
\includegraphics[width=0.25\linewidth]{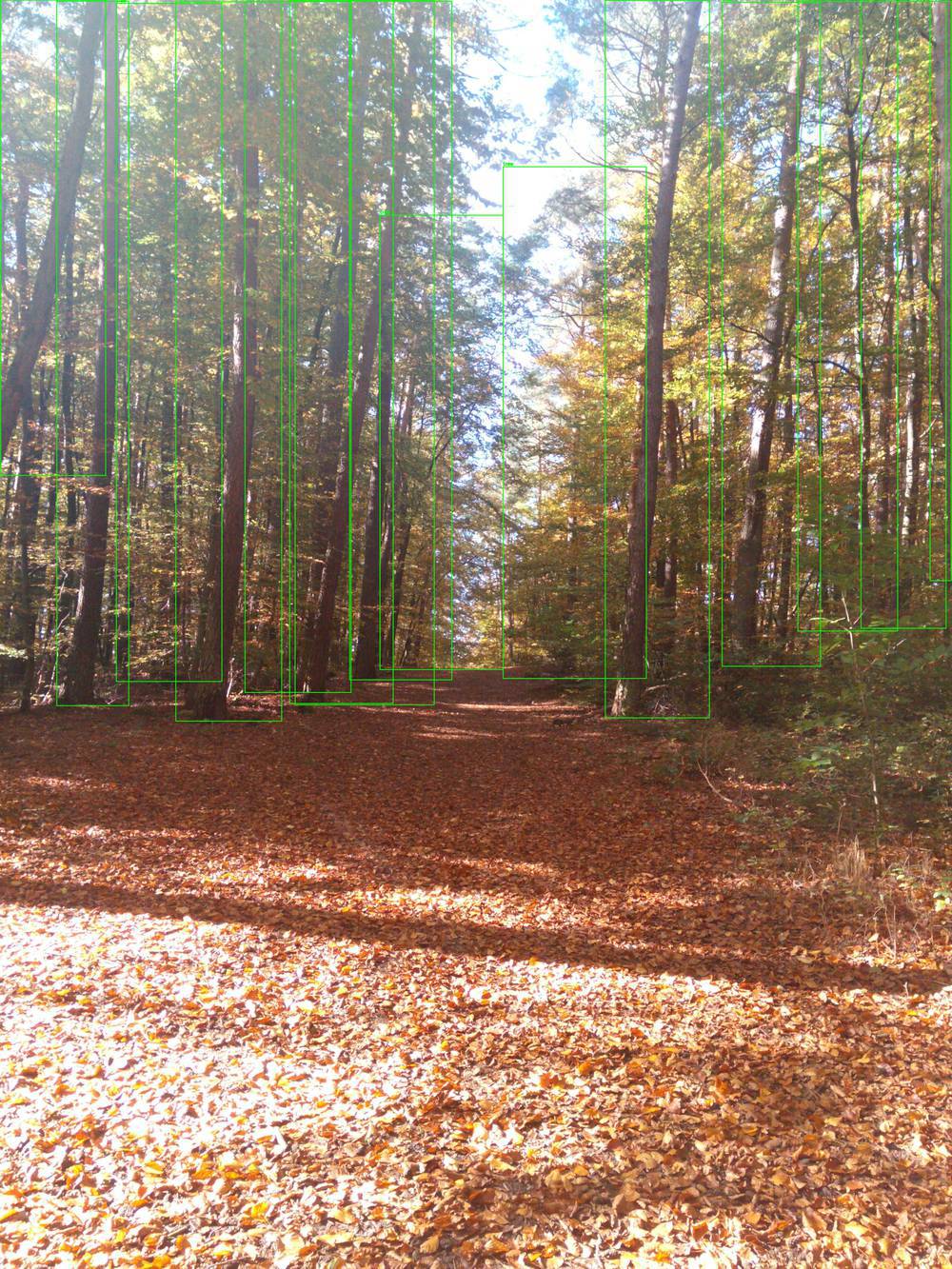} \\
\bottomrule
\end{tabular}
\label{fig:yolo_wt_real}
\end{table}

\begin{table*}[htp!]
\centering
\caption{\textbf{Phase 2 :} Domain gap analysis between simulated and real validation performance. YOLO series}
\begin{tabular}{l l l c c c c}
\toprule
\textbf{Model} & \textbf{Type} & \textbf{Task} & \textbf{Simulated accuracy (val set)} & \textbf{Sim→Real acc.} & \textbf{Absolute Gap} & \textbf{Relative Drop} \\
\midrule
yolo11m & Tree trunk & BBox & 0.93341 & 0.40   & 0.53 & 56.8\% \\
yolov8m & Whole tree & BBox & 0.724   & 0.4817 & 0.24 & 33\%   \\
\bottomrule
\end{tabular}

\label{tab:yolo_domain_gap}
\end{table*}

\begin{table*}[t]
\centering
\caption{\textbf{Phase 2: }Domain gap analysis between simulated and real validation performance. (RT-DETR + RegNet).}
\resizebox{\linewidth}{!}{
\begin{tabular}{l l c c c c c c c c}
\toprule
\textbf{Task} & \textbf{Model} & \textbf{mAP@0.5:0.95} & \textbf{mAP@0.50} & \textbf{AP50} & \textbf{AP75} & \textbf{APs} & \textbf{APm} & \textbf{API} & \textbf{Pretrained on} \\
\midrule
Object detection & RT-DETR + RegNet & 0.232 & 0.438 & 0.438 & 0.220 & -1 & -1 & 0.232 & Tree trunk sim \\
Object detection & RT-DETR + RegNet & 0.062 & 0.182 & 0.182 & 0.036 & -1 & -1 & 0.062 & Whole tree sim \\
\bottomrule
\end{tabular}}
\label{tab:sim2real_rtdetr}
\end{table*}

Qualitative examples further highlight these quantitative differences. Predictions on real images reveal that the trunk-trained model often detects only a limited subset of trees, focusing on highly salient or well-lit stems while missing thinner or background trunks, particularly under heavy glare or strong shadows.

\clearpage
\subsection{Phase 3: Training on real trees}

The validation metrics of YOLO (v8 \& v11) models on real trees for epochs can be seen from table \ref{tab:yolo_real_trees}. A consistent trend is observed where larger models (m/l/x) generally achieve slightly higher mAP scores compared to their smaller counterparts, however the improvements are marginal relative to the increase in model size. For example, YOLOv8x achieves the best overall performance among the v8 family with an mAP@50:95 of 0.660, compared to 0.616 for YOLOv8n, while precision and recall remain balanced across variants ($\sim$0.82). The YOLOv11 family, despite being trained with slightly different learning rates, shows competitive but slightly lower performance than YOLOv8 i.e. YOLOv11m and YOLOv11l reach mAP@50:95 scores of 0.599 and 0.616, respectively, but do not surpass the YOLOv8 counterparts. Interestingly, all YOLOv11 models exhibit lower precision values (0.784-0.802) compared to YOLOv8, while recall is comparable, indicating a tendency toward over-prediction. In terms of losses, both families show the expected pattern of lower training losses with larger model capacity, though validation losses remain relatively close across all variants.

\begin{table*}
\centering
\resizebox{\textwidth}{!}{
\begin{tabular}{l l c c c c c c c c c c c c}
\toprule
\textbf{Task} & \textbf{Model} & \textbf{Epoch} & \textbf{train/box\_loss} & \textbf{train/cls\_loss} & \textbf{train/dfl\_loss} & 
\textbf{Precision} & \textbf{Recall} & \textbf{mAP@50} & \textbf{mAP@50:95} & 
\textbf{val/box\_loss} & \textbf{val/cls\_loss} & \textbf{val/dfl\_loss} & \textbf{lr} \\
\midrule
Object Detection & YOLOv8n  & 50 & 0.9830 & 0.8459 & 1.0656 & 0.821 & 0.805 & 0.879 & 0.616 & 0.928 & 0.837 & 1.058 & 9.92e-05 \\
                 & YOLOv8s  & 50 & 0.8869 & 0.7426 & 1.0244 & 0.816 & 0.826 & 0.886 & 0.637 & 0.928 & 0.772 & 1.058 & 9.92e-05 \\
                 & YOLOv8m  & 50 & 0.8173 & 0.6936 & 1.0125 & 0.821 & 0.825 & 0.886 & 0.649 & 0.898 & 0.720 & 1.120 & 9.92e-05 \\
                 & YOLOv8l  & 50 & 0.8129 & 0.6565 & 1.0225 & 0.828 & 0.823 & 0.890 & 0.656 & 0.882 & 0.705 & 1.126 & 9.92e-05 \\
                 & YOLOv8x  & 50 & 0.7900 & 0.6260 & 1.0095 & 0.818 & 0.835 & 0.892 & 0.660 & 0.883 & 0.690 & 1.140 & 9.92e-05 \\
                 & YOLOv11s & 50 & 0.9979 & 0.8583 & 1.0650 & 0.787 & 0.807 & 0.863 & 0.590 & 0.961 & 0.814 & 1.102 & 5.96e-05 \\
                 & YOLOv11m & 50 & 0.9208 & 0.7813 & 1.0352 & 0.787 & 0.803 & 0.863 & 0.599 & 0.986 & 0.849 & 1.045 & 5.96e-05 \\
                 & YOLOv11l & 50 & 0.8773 & 0.7376 & 1.0365 & 0.802 & 0.803 & 0.871 & 0.616 & 0.954 & 0.811 & 1.102 & 5.96e-05 \\
                 & YOLOv11x & 50 & 0.8588 & 0.7084 & 1.0410 & 0.784 & 0.815 & 0.869 & 0.622 & 0.984 & 0.784 & 1.126 & 5.96e-05 \\
\bottomrule
\end{tabular}
}
\caption{\textbf{Phase 3:} Validation metrics for object detection on real trees across YOLOv8 and YOLOv11 variants. All models were trained for 50 epochs. (YOLO series)}
\label{tab:yolo_real_trees}
\end{table*}

\begin{table*}
\centering
\resizebox{\linewidth}{!}{
\begin{tabular}{l l c c c c c c c}
\toprule
\textbf{Task} & \textbf{Model} & \textbf{mAP@0.5:0.95} & \textbf{mAP@0.50} & \textbf{AP50} & \textbf{AP75} & \textbf{APs} & \textbf{APm} & \textbf{API} \\
\midrule
Object detection & RT-DETR + RegNet & 34.7 & 68.6 & 68.6 & 30.3 & -1 & -1 & 34.7 \\
\bottomrule
\end{tabular}}
\caption{\textbf{Phase 3: }Validation metrics for RT-DETR + RegNet trained on real trees.}
\label{tab:real_rtdetr}
\end{table*}

\clearpage
\subsection{Oriented object detection}

The results in Table \ref{tab:yolo_val_obb} highlight the performance of the YOLOv11 series for oriented bounding box detection on simulated tree trunk datasets. Across all variants, the models demonstrate strong localization accuracy, with mAP@50 values consistently above 0.90 and recall scores close to 0.85, indicating reliable detection of both near and far-field trunks. The larger models (YOLOv11l and YOLOv11x) achieve the highest precision and overall stability, reflecting their increased capacity to model complex structural cues. However, even the lightweight variants (YOLOv11n and YOLOv11s) maintain competitive performance, suggesting that efficient models are sufficient for this task when operating in a controlled simulation environment.
\begin{center}
\begin{minipage}{\textwidth}  
\centering
\resizebox{\textwidth}{!}{%
\begin{tabular}{l l c c c c c c c c c c c c c c}
\toprule
Task & Model & Epoch & train/box\_loss & train/cls\_loss & train/dfl\_loss & 
Precision (B) & Recall & mAP@50 (B) & mAP@50:95 (B) & 
val/box\_loss & val/cls\_loss & val/dfl\_loss & lr/p1 & lr/p2 \\
\midrule
Tree Trunk & YOLOv11n  & 100 & 0.71772 & 0.46044 & 1.52811 & 0.8287 & 0.85229 & 0.90939 & 0.74942 & 0.71198 & 0.57461 & 1.56044 & 0.000199 & 0.000199 \\
           & YOLOv11s  & 100 & 0.59922 & 0.51077 & 1.57878 & 0.8367 & 0.85275 & 0.91263 & 0.76162 & 0.58111 & 0.54875 & 1.59985 & 0.000199 & 0.000199 \\
           & YOLOv11m  & 100 & 0.55158 & 0.51558 & 1.55648 & 0.8568 & 0.85223 & 0.91623 & 0.79179 & 0.58911 & 0.52871 & 1.62083 & 0.000199 & 0.000199 \\
           & YOLOv11l  & 100 & 0.59349 & 0.51227 & 1.51467 & 0.8418 & 0.86181 & 0.92514 & 0.78889 & 0.60711 & 0.54427 & 1.51987 & 0.000199 & 0.000199 \\
           & YOLOv11x  & 100 & 0.51664 & 0.48910 & 1.50887 & 0.8493 & 0.87151 & 0.92512 & 0.79249 & 0.54909 & 0.53134 & 1.50861 & 0.000199 & 0.000199 \\
\bottomrule
\end{tabular}
}
\vspace{0.3em}
\captionof{table}{Validation metrics for Oriented Bounding Box Detection (YOLOv11 series) on simulated tree datasets.}
\label{tab:yolo_val_obb}
\end{minipage}
\end{center}


%
%
%
\bibliographystyle{splncs04}
\bibliography{samplepaper}
%

\end{document}